# Genetic Algorithms for Multiple-Choice Optimisation Problems

by

Uwe Aickelin

(Dipl Kfm, EMBSc)

School of Computer Science
University of Nottingham
NG8 1BB UK
uxa@cs.nott.ac.uk

Thesis submitted to the University of Wales
In candidature for the
Degree of Doctor of Philosophy

European Business Management School
University of Wales Swansea

September 1999

**Declaration**

This work has not previously been accepted in substance for any degree and is not concurrently being submitted in candidature for any degree.

Signed ………………………………. (Candidate)
Date ……………………………….

**Statement 1**

This thesis is the result of my own investigations, except where otherwise stated. Other sources are acknowledged by endnotes giving explicit reference. A bibliography is appended.

Signed ………………………………. (Candidate)
Date ……………………………….

**Statement 2**

I hereby give consent for my thesis, if accepted, to be available for photocopying and for inter-library loan, and for the title and summary to be made available to outside organisations.

Signed ………………………………. (Candidate)
Date ……………………………….

# Acknowledgements

Everybody has someone or something to thank for their success. Primarily I want to use this opportunity to thank my supervisor Dr Kathryn Dowsland for her guidance and support throughout my work. No doubt without the many of our discussions both this thesis and my research would not have been possible. Additionally, I am very grateful for the advice and support of Dr Bill Dowsland, the computer support staff and other people of the European Business Management School. Furthermore, special thanks go to Dr Jonathan Thompson for his early work on the nurse scheduling problem. I would also like to thank my family who in their own ways always supported me. Finally, thank you to Sonya for motivation, support and distraction at the appropriate times over the past years.

# Summary

This thesis investigates the use of problem-specific knowledge to enhance a genetic algorithm approach to multiple-choice optimisation problems. It shows that such information can significantly enhance performance, but that the choice of information and the way it is included are important factors for success. Two multiple-choice problems are considered. The first is constructing a feasible nurse roster that considers as many requests as possible. In the second problem, shops are allocated to locations in a mall subject to constraints and maximising the overall income. Genetic algorithms are chosen for their well-known robustness and ability to solve large and complex discrete optimisation problems. However, a survey of the literature reveals room for further research into generic ways to include constraints into a genetic algorithm framework. Hence, the main theme of this work is to balance feasibility and cost of solutions. In particular, co-operative co-evolution with hierarchical sub-populations, problem structure exploiting repair schemes and indirect genetic algorithms with self-adjusting decoder functions are identified as promising approaches. The research starts by applying standard genetic algorithms to the problems and explaining the failure of such approaches due to epistasis. To overcome this, problem-specific information is added in a variety of ways, some of which are designed to increase the number of feasible solutions found whilst others are intended to improve the quality of such solutions. As well as a theoretical discussion as to the underlying reasons for using each operator, extensive computational experiments are carried out on a variety of data. These show that the indirect approach relies less on problem structure and hence is easier to implement and superior in solution quality. The most successful variant of our algorithm has a more than 99% chance of finding a feasible solution which is either optimal or within a few percent of optimality.

# Contents

# List of Figures

# List of Tables

# 1 Introduction

## 1.1 The Nature of the Problem

Multiple-choice problems come in many varieties: Choosing one's lottery numbers, deciding what to wear in the morning or assigning which shift pattern a nurse should work. What all these problems have in common is that for each decision, be it a lottery number, piece of clothing or a worker, there is only one object we can assign to it. Hence, these are known as multiple-choice problems. Also, there are usually a number of hard and soft constraints of different importance guiding our decisions. For instance, it is only allowed to choose a particular lottery number once (hard constraint) or one would like to wear clothes that match each other (soft constraint). This thesis uses genetic algorithms to optimise multiple-choice problems, with the emphasis on balancing those soft and hard constraints.

In this research, we will concentrate on two multiple-choice problems with covering constraints. The research was motivated by the first problem to be tackled, which is to find work schedules for nurses in a major UK hospital. The hospital operates 24 hours per day using three shifts and nurses are graded into three bands. The required schedules have to be calculated weekly and must consider various hard and soft constraints. For instance, schedules must be perceived fair by staff and thus the preferences of nurses, their past working history and other objectives have to be taken into consideration. Furthermore, for each grade band and shift, strict covering requirements are set which must be met.

This problem was chosen for a variety of reasons. It is a linear problem but difficult to solve due to the multiple-choice component. Thus, although computationally expensive, optimal solutions can be obtained for a comparison of results. Furthermore, some insight into the problem structure existed from the tabu search approach reported in Dowsland [55]. Additionally, the existence of a large number of real-life data sets allowed for a realistic optimisation situation. The purpose of this research is not to find optimal solutions but to develop and compare new methods of handling constraints, which can then be applied to more complex problems.

After developing suitable solution methods for this problem, we turn our attention to mall layout and tenant selection. Although the problem and data in this case are artificial, it is modelled closely after similar problems in real-life. This problem was chosen as it is similar to yet more complex than the nurse scheduling problem. In particular, the objective is non-linear. The aim is to place shops into locations of the mall such that the overall revenue is maximised. This in turn maximises the rent generated by the shops, the actual objective, as it is largely proportional to the revenue. Constraints that have to be taken into account include upper and lower bounds on the number of shops of one type and restrictions on the number of shops of a certain size class. Soft constraints include some shops creating more revenue in certain areas of the mall, synergy effects between similar shops and efficiency savings of larger shops.

## 1.2 The Solution Method and Results

The methods chosen to solve these multiple-choice problems are genetic algorithms. They are inspired by evolution in nature and have the 'survival of the fittest' idea at their heart. In contrast to other solution methods, they work with a population of solutions in parallel and use stochastic crossover and mutation operators similar to those found in nature. Recently, a lot of interest has been shown in using genetic algorithms to solve real-life problems because of their flexibility and robustness. However, canonical genetic algorithms are not function optimisers and in particular, there is no explicit or generic way to include constraints. Thus, the purpose of this research is to add to the knowledge in the area of constraint handling in a genetic algorithm framework.

In a pilot study to this thesis (Aickelin [4]), it was shown that the nurse scheduling example is both a difficult and interesting problem to apply genetic algorithms to. Results for a very limited number of data sets were promising, although the main stumbling block was as anticipated the constraints. It was concluded that once this was overcome, the genetic algorithm would provide a robust and flexible solution method for this problem. One of the first tasks of this research was to investigate if the results found during the pilot study carry over for the much more extensive real-life data sets. For further details of the pilot study, see the full summary, which is contained in Appendix B.

Since the nurse scheduling problem is linear, optimal solutions for all data files are known. This allows for a thorough comparison of the results of our various genetic algorithm approaches. Additionally, Dowsland [55] solved the same problem with tabu search. Both methods will be compared in terms of solution quality, robustness and ease of including possible future expansions of the problem. Once successful genetic algorithms are established, they are tested on the non-linear mall layout problem. This allows for more general conclusions to be drawn about the suitability of our ideas for other scheduling and related problems.

Due to the nature of genetic algorithms, the focus of this thesis is on finding suitable ways of striking a balance between the soft and hard constraints of the problem. Our task is to find the best possible solution in terms of the soft constraints without violating any of the hard constraints. Over the years, many ways of doing this within a genetic algorithm framework have been suggested: Penalty functions, repair algorithms, special genetic operators and decoders to name but a few. A comprehensive literature review of these methods is provided.

In the course of the research, many of these traditional methods are applied to the nurse scheduling and mall layout problems. In particular, two avenues of research are followed: Direct genetic algorithms, which solve the actual problem themselves and indirect genetic algorithms, which solve the problem in combination with external decoder functions. Although seemingly making things more complicated, the latter approach is in fact shown to be less complex. This is because it lends itself better to the inclusion of problem-specific knowledge, which is the key to overcoming the problems, caused by the constraints.

## 1.3 The Structure of the Thesis

An introduction to genetic algorithms, their operators and the theory behind this type of meta-heuristic can be found in Appendix A. A summary of the pilot study for the nurse scheduling problem, assessing its difficulty and suitability to the proposed optimisation approach, is given in Appendix B. The rest of this work is structured in the following way.

Chapter 2 introduces the nurse scheduling problem in hand and a corresponding integer program is set up. The remainder of the chapter looks at various solution methods to nurse scheduling problems, with particular emphasis on linear programming and meta-heuristic approaches. The chapter concludes that the approaches described in the literature are not sufficient to solve the problem.

An overview of current genetic algorithm literature is given in chapter 3. The emphasis of the chapter is on the treatment of constraints, as the pilot study found that the issue of implementing constraints into a genetic algorithm framework is the most critical area of the research. A detailed review of various methods, including penalty functions, repair, decoders, special operators and others is given.

Chapter 4 details the encoding of our problem and presents the standard direct genetic algorithm approach. After experimenting with various parameter and strategy settings, the best values are retained. Then the first enhancement of the direct approach is presented in the form of adaptive penalty parameters that follow the development of the population. The chapter concludes with a summary of results and reasons for failure of the methods used so far.

After discussing the issue of epistasis and its relevance to our research, chapter 5 presents co-operative and hierarchical sub-populations in an attempt to overcome the problems encountered. Together with a special crossover operator and migration between the sub-populations, this method is shown to be very effective at solving the nurse scheduling problem. To improve upon the results, various further enhancements, namely delta coding, swaps and a local hillclimber, are introduced next. The chapter ends with a comparison of all direct genetic algorithm approaches used so far.

Chapter 6 is concerned with the indirect approach to the problem. After explaining the idea of an indirect genetic algorithm, permutation based genetic operators, made necessary by this type of genetic algorithm, are introduced. Then, some possible decoders are detailed and shortened parameter tests are performed. Further enhancements of the decoders and a new crossover operator are presented. Finally, the original nurse scheduling problem is extended and it is shown that genetic algorithms are flexible and robust enough to deal with this. The chapter concludes with a summary and comparison of all nurse scheduling results.

To validate the results found so far, all previous methods are applied to the mall layout problem in chapter 7. The superiority of the indirect over the direct approach is confirmed and possible problems with the co-operative co-evolutionary approach are discovered. Further enhancements of the indirect genetic algorithm are presented, which after proving to be successful, are applied to the nurse scheduling problem as well. A final comparison of results concludes the chapter.

The final chapter of this thesis summarises the findings of this project and makes various recommendations as to the best way of solving multiple-choice problems with genetic algorithms. This work is then put into the context of more general scheduling and areas for future investigation are identified.

# 2 Introduction to Nurse Scheduling

## 2.1 Problem Formulation

### 2.1.1 General Introduction

This chapter gives details of the nurse scheduling problem tackled in this research and presents an equivalent integer programming formulation. Following on from this, the many ways in which similar problems have been solved by other researchers are reviewed. A wide array of different methods has been proposed, all with their own strengths and weaknesses. However, due to the nature of nurse scheduling problems, most approaches described rely heavily on the particular problem structure, making their use for our problem impossible.

Our task is to create weekly schedules for wards of up to 30 nurses at a major UK hospital. These schedules have to satisfy working contracts and meet the demand for a given number of nurses of different grades on each shift, while seen to be fair by the staff concerned. The latter objective is achieved by meeting as many of the nurses' requests as possible and considering historical information to ensure that unsatisfied requests and unpopular shifts are evenly distributed. This will be further detailed in section 2.1.3. For additional details, refer to Dowsland [55] and for example data to Appendix C.

For scheduling purposes, the day at the hospital is partitioned into three shifts: Two day shifts known as 'earlies' and 'lates', and a longer night shift. Note that until the final scheduling stage, 'earlies' and 'lates' are merged into day shifts. Due to hospital policy, a nurse would normally work either days or nights in a given week, and because of the difference in shift length, a full week's work would normally include more days than nights. For example, a full time nurse works five days or four nights, whereas typical part time contracts are for four days or three nights, three days or three nights and three days or two nights. However, exceptions are possible and some nurses specifically must work both day- and night-shifts in one week.

As described in Dowsland [55] the problem can be decomposed into the three independent stages set out below. This thesis deals with the highly constrained second step.

1. Ensuring via a knapsack model that enough nurses are on the ward to cover the demand, otherwise introducing dummy and / or bank nurses to even out the cover.
2. Scheduling the days and nights on and off for a nurse.
3. Splitting the day shifts into early and late shifts using a network flow model.

### 2.1.2 The Three Solution Steps

In the following chapters, the nurse scheduling problem is often referred to as being particularly 'tight'. In order to understand this 'tightness' of the problem, one has to know that all data is pre-processed by a knapsack routine to smooth out over- and under-staffing. A knapsack is necessary due to the day / night shift imbalance, i.e. nurses usually working more day shifts than night shifts. Hence, the knapsack determines the maximum number of day shifts available subject to the night shifts being covered. The consequence of this is that there is no slackness in most of the covering constraints.

More precisely, the hospital requested that if any over-cover occurred it should be spread out over day shifts only such that weekdays are covered first and weekends last. To achieve this, additional dummy nurses are introduced who work as follows: Weekend dummies can only work day patterns not including any weekdays. Weekday dummies can only work day patterns not including any weekend days. They work as many shifts as necessary to complement the over-cover to five days.

An example of the knapsack's operation is as follows. Assume that there are 10 nurses required on each day. Thus, 70 'nurse shifts' are needed in total. Furthermore, assume that the knapsack has found that 15 full time nurses are available to work days with the remaining nurse required on nights. Since each full time nurse works five shifts, this gives a total of 75 nurse shifts available on days. So overall, there is a surplus of five nurse shifts. As the hospital requires the overstaffing to be spread over non-weekend days first, the demand for Monday to Friday is raised by one shift each. This eliminates the surplus and the knapsack routine is finished.

Now assume that one of the 15 nurses would only work four shifts in this week due to a day off work. As before the demand for Monday to Friday will be increased by one shift. In this case, this leads to an artificial 'shortage' of one shift. This is met with the introduction of a weekday dummy nurse who works one day shift. Again the overcover is smoothed out and the knapsack routine is finished. Situations with other combinations of over- and under-staffing are met in a similar way. The following list describes all possibilities grouped by the amount of over staffing, where a 'unit' refers to one single nurse shift:

- More than seven units: The demand is raised by one for all seven days until the over-staffing is by seven units or less.
- Seven units: The demand is raised by one for all seven days.
- Six units: The demand is increased by one for all seven days and additionally a weekend dummy nurse is introduced.
- Five units: The demand is raised by one unit for weekdays only.
- Less than five units: The demand is raised by one unit for weekdays only and additionally a weekday dummy nurse is introduced.
- If the knapsack determines that there are not enough nurses to cover the demand, then as many bank nurses as necessary are introduced. Bank nurses can only work one day shift each.

As 'earlies' and 'lates' are not yet merged at this stage, the second step of the problem can be modelled as follows. Each possible shift pattern worked by a given nurse can be represented as a zero-one vector with 14 elements, where the first seven elements represent the seven days of the week and the last seven the corresponding nights. A 1 in the vector denotes a scheduled day or night on and a 0 a day or night off. These vectors will be referred to as shift patterns and examples can be found in appendix C.4.

Depending on the working hours of a nurse there are a limited number of shift patterns available to her or him. For instance, a full time nurse working either 5 days or 4 nights has a total of 21 (i.e. $\binom{7}{5}$) feasible day shift patterns and 35 (i.e. $\binom{7}{4}$) feasible night shift patterns. Typically, a nurse has around 40 possible shift patterns available to her / him. Other data dimensions are between 20 and 30 nurses per ward, three grade-bands, nine part time options and 411 different shift patterns.

The third step, the network flow algorithm, is of little interest to us. It splits the day shifts into early and late shifts. The algorithm is always exact and takes little time to find an optimal solution. Again, more details can be found in Dowsland [55].

### 2.1.3 Setting Up of the Nurse Shift Pattern Cost $p_{ij}$

Before setting up this problem, the cost $p_{ij}$ of nurse $i$ working shift pattern $j$ has to be determined. This is done in the following way after close consultation with the hospital. For each nurse, the number of day and night shifts she must work is given. Each of these shift patterns is then assigned a ‘cost’ $p_{ij}$ according to their suitability. For sample data of these costs and the factors that contribute to them, refer to Appendix C. More precisely, the cost of a shift pattern is the sum of the following factors:

(1) Each shift pattern has been given a basic cost between one (no problems with pattern) and four (very unattractive). This cost generally depends on whether the pattern means that a nurse will have her days off together or separate. Note that some patterns in which a nurse will work both days and nights are given a cost of 18. This cost is imposed if the shift pattern means that a nurse will work some night shifts, then a day shift and finally, further night shifts.

(2) Nurses may prefer to work days or they may prefer to work nights. A nurse may be classified as one of the following, depending on her or his contract and general preferences:

- Days Only – in which case night shifts are not considered.
- Nights Only – in which case day shifts are not considered.
- Days Important – a cost of 12 is added to all night shifts.
- Nights Important – a cost of 12 is added to all day shifts.
- Days Preferred – a cost of 3 is added to all night shifts.
- Nights Preferred – a cost of 3 is added to all day shifts.

(3) A nurse may request not to work certain shifts and all shift patterns, which do not satisfy these requests, are given a cost. Thus, if a shift pattern means that *n* requests are not satisfied, *n* costs are added. These requests are graded from one (relatively unimportant) to five (binding). Note that if a nurse requests not to work an early but does not mind working a late, no cost is imposed as the nurse can be allocated a late shift in the subsequent network flow phase. Likewise, for a nurse who requests not to work a late but does not mind working an early. The costs added are:

- Grade 1 request – 3
- Grade 2 request – 8
- Grade 3 request – 12
- Grade 4 request – 18
- Grade 5 request – 90

(4) Nurses should not work more than seven days in a row. The cost added is equal to the number of days above seven that they would have to work in a row.

(5) The shift pattern costs do not include continuity problems with previous schedules. Thus, if the pattern a nurse worked last week finished 01, i.e. day off, day on; then any shift pattern which begins with a day off is penalised by three. Likewise, a cost of three is added if a nurse finished last week working 0 and starts this week working 10.

(6) Nights must be rotated: If a nurse worked nights last week, all night shift patterns for this week have an additional cost of ten. If a nurse has worked nights the week before, a cost of five is added.

(7) To rotate weekend work the following cost is added. If a nurse worked Saturday and Sunday last week, a cost of one is added to each pattern that involves working Saturday or Sunday this week.

(8) Before the problem is solved, one is deducted from the cost of each shift pattern for each nurse. Thus, perfect shift patterns have a cost of zero. However, if the cost of a shift pattern is above 89, it is set to 100. Costs for dummy and bank nurses are always set to zero.

(9) Finally, the working history of a nurse is taken into account. If a nurse had a cost for the shift pattern that she worked last week, then this is added to all non-zero cost shift patterns (but not above 100).

### 2.1.4 Integer Programming Formulation

The problem can now be formulated as an integer linear program as follows.

Indices:

$i$ = 1...$n$ nurse index.

$j$ = 1...$m$ shift pattern index.

$k$ = 1...14 day and night index (1...7 are days and 8...14 are nights).

$s$ = 1...$p$ grade index.

Decision variables:

$$x_{ij} = \begin{cases} 1 & \text{nurse } i \text{ works shift pattern } j \\ 0 & \text{else} \end{cases}$$

Parameter:

$n$ = number of nurses.

$m$ = number of shift patterns.

$p$ = number of grades.

$$a_{jk} = \begin{cases} 1 & \text{shift pattern } j \text{ covers day / night } k \\ 0 & \text{else} \end{cases}$$

$$q_{is} = \begin{cases} 1 & \text{nurse } i \text{ is of grade } s \text{ or higher} \\ 0 & \text{else} \end{cases}$$

$p_{ij}$ = Penalty cost of nurse $i$ working shift pattern $j$.

$F(i)$ = Set of feasible shift patterns for nurse $i$.

$N_i$ = Working shifts per week of nurse $i$ if night shifts are worked.

$D_i$ = Working shifts per week of nurse $i$ if day shifts are worked.

$B_i$ = Working shifts per week of nurse $i$ if both day and night shifts are worked.

$R_{ks}$ = Demand of nurses with grade $s$ on day respectively night $k$.

Target function:

$$\sum_{i=1}^{n} \sum_{j \in F(i)}^{m} p_{ij} x_{ij} \quad \rightarrow \quad \text{min!}$$

Subject to:

1. Every nurse works exactly one shift pattern:

$$\sum_{j \in F(i)} x_{ij} = 1 \qquad \forall i \qquad (1)$$

2. The shift pattern corresponds to the number of weekly working shifts of the nurse:

$$F(i) = \begin{cases} \sum_{k=1}^{7} a_{jk} = D_i & \forall j \in \textit{day shifts} \\ or & \\ \sum_{k=8}^{14} a_{jk} = N_i & \forall j \in \textit{night shifts} \\ or & \\ \sum_{k=1}^{14} a_{jk} = B_i & \forall j \in \textit{combined shifts} \end{cases} \quad \forall i \qquad (2)$$

3. The demand for nurses is fulfilled for every grade on every day and night:

$$\sum_{j \in F(i)} \sum_{i=1}^{n} q_{is} a_{jk} x_{ij} \geq R_{ks} \qquad \forall k, s \qquad (3)$$

Constraint sets (1) and (2) ensure that every nurse works exactly one shift pattern from her feasible set, and constraint set (3) ensures that the demand for nurses is covered for every grade on every day and night. Note that the definition of $q_{is}$ is such that higher graded nurses can substitute those at lower grades if necessary. This problem can be regarded as a multiple-choice covering problem. The sets are given by the shift pattern vectors and the objective is to minimise the cost of the sets needed to provide sufficient cover for each shift at each grade. The multiple-choice aspect derives from constraint set (1), which enforces the choice of exactly one pattern (or set) from the alternatives available for each nurse. Although this problem looks similar to the generalised assignment problem, it is different due to the additional shift pattern level, i.e. nurses are assigned to shift patterns, but days and nights must be covered.

## 2.2 Introduction to Nurse Scheduling

There are many different ways of solving manpower scheduling and in particular nurse scheduling problems. However, almost all solve a simplified version or are otherwise very problem-specific, for example no grades are taken into account, all nurses are assumed to be full time, or under- and over-staffing is allowed. During the course of this research, it therefore became clear, that these methods could not be used to solve our particular problem.

For the purpose of this chapter, nurse scheduling solution methods are classified into four types following the recommendation of Bradley and Martin [27]: Exact cyclical, heuristic cyclical, exact non-cyclical and heuristic non-cyclical. The problem in hand is of a non-cyclical nature, because the hospital wants high flexibility to allow nurses their requested days off and requests vary from week to week. Therefore, cyclical scheduling approaches are only discussed briefly in section 2.3. The remainder of chapter 2 deals with non-cyclical algorithms.

Exact non-cyclical solution methods, i.e. linear, integer and constraint programming, are presented in section 2.4. The most commonly used algorithms are heuristic (section 2.5) and meta-heuristic (section 2.6). Although heuristic algorithms do not guarantee to find the optimal solution, they tend to find very good solutions in a short time. The term meta-heuristic refers to algorithms that contain a number of simpler heuristics. This gives them the ability to be used for various problems with only slight modifications. Examples of these methods are tabu search, simulated annealing and genetic algorithms. In contrast, the methods of the heuristic section tend to be very problem-specific and not suitable for other problems.

Many of the early approaches to manpower scheduling were of a manual nature. This usually meant following a set of greedy rules when constructing a suitable roster. Due to their limitations, they are not reported here in detail. The interested reader is referred to Tien and Kamiyama [165] who give a good summary and comparison of many such approaches and Fries [71] who provides an extensive bibliography.

Bibliographies of more recent staff scheduling algorithms that concentrate on hospital nurse scheduling are given by Hung [97], Sitompul and Randhawa [151] and Bradley and Martin [27]. Additionally to the methods reviewed in this thesis, the authors mention self-scheduling, that is the (manual) scheduling by the nurses themselves. Self-scheduling is not further referred to in this thesis. The authors of the bibliographies conclude that most methods surveyed are limited and that decision support systems or the use of artificial intelligence might be possible future avenues for research.

## 2.3 Cyclic Nurse Scheduling

Cyclic nurse scheduling, as presented by Rosenbloom and Goertzen [141], is a common way of solving the nurse scheduling problem. It first generates all possible basic work patterns, usually on a weekly basis, by taking into account a variety of labour constraints (work stretches, weekends off, no isolated days off or on etc). Furthermore, only patterns that can be part of a larger schedule are allowed. For instance, if basic patterns are for one week then for work-stretch constraints reasons, certain patterns cannot be combined with others.

Once all feasible pattern pairs are determined, a linear programming optimisation decides how often each pattern pair is used. Finally, the patterns are assigned to the nurses who usually move onto a different pattern in every new planning horizon. Hence, the name cyclic scheduling as nurses cycle through all patterns. For a practical application of cyclic nurse scheduling, see Ahuja and Sheppard [3].

The nature of this approach is to regard all nurses as identical and hence no personal preferences can be taken into account. At best, nurses can choose from the set of optimal cyclic patterns. Our problem is not cyclic, because nurses' preferences, which will differ from week to week, must be taken into account. Thus, cyclic approaches cannot be used.

## 2.4 Linear, Integer, Constraint and Goal Programming

This section presents classical approaches that guarantee to find the optimal solution to the non-cyclical problem. However, the drawback of this is the often prohibitively long execution time. Thus, even though the problem formulation is still aimed at an optimal solution, the actual execution is often of a heuristic nature, for example cyclic descent or rounding of fractional variables rather than a full branch and bound approach.

One of the earliest examples of modelling the nurse scheduling problem as a mathematical program can be found in Warner and Prawda [169]. The authors formulate a linear program with the decision variables as the number of nurses of one grade working a particular shift on a specific day. Should a solution become fractional, a simple heuristic is used to correct it. To facilitate finding a solution, some substitution between nurses of different grades is allowed. Furthermore, only an absolute lower limit on the nurses required per shift is set. The target function is then to minimise any staffing below the required level. No preferences or working constraints are taken into account and the authors do not explain how to assign the shifts required in a solution to the actual nurses.

Warner [170] presents an extension of the above. This time the problem is formulated as an integer program, with the decision variables being the fortnightly shift patterns worked by a nurse. As in our approach, each shift pattern is given a penalty cost. However, Warner only bases this cost on work-stretch and isolated day on or off preferences of the nurses. The number of possible shift patterns for each nurse is kept small by having a fixed day and night rotation, alternate weekends off and further restrictions. Limited under-covering of shifts is also allowed. The target function is to minimise the sum of the costs of shift patterns for all nurses. The problem is solved via a block pivoting heuristic. The solution found is then manually improved as far as possible to form the final schedule.

Miller et al. [117] use the same problem formulation as Warner [170]. However, rather than penalising patterns they restrict the number of shift patterns available to a nurse by setting constraints for maximum work-stretches and not allowing any isolated days on. Nurses may request a particular day off which reduces the number of patterns further. Moreover, they only consider full time nurses working ten days per fortnight and do not distinguish between early, late and night shifts. The objective function is to minimise under-staffing. The problem is solved with a cyclic descent algorithm.

A constraint programming approach to the nurse scheduling problem is given by Weil et al. [171]. Constraint programming is similar to linear programming. However, rather than a 'blind' branch-and-bound on the full domain of the decision variables, the domains are dynamically reduced via the constraints in accordance with variables already fixed. The problem formulation differs from ours, as nurses working a particular shift on a specific day are the decision variables. The hard constraints are the same as in our problem. However, Weil et al. considerably reduce complexity by only scheduling full time nurses and not considering grades. Their objective is to minimise the violation of soft constraints regarding isolated days on or off and work-stretches. No individual preferences are considered. The authors are able to solve problems of similar sizes to ours on a workstation within seconds. No solution quality is reported.

Cheng and Yeung [37] present a hybrid expert system combined with a linear zero-one goal programming method to schedule full time nurses of one grade. The scheduling of days on and off is done by the goal programming module. The goals are to satisfy minimum staff levels, to minimise overtime, to grant requested days off, to limit work-stretches to maximal six consecutive working days and to prevent off/on/off patterns. For each goal, an aspiration level is set, for example the minimum required staff level on a particular day. Each goal also has a priority level assigned to resolve conflicts.

The actual allocation to early, late and night shifts is done by the expert system component, taking requests and required staff levels and other fairness measures into account. An expert system consists of a set of rules of 'if ... then ... else' format. These rules are gained by questioning experts, hence the name. The resulting hybrid system is able to solve nurse scheduling problems ten times faster than the head nurse, whilst halving constraint violations. However, this approach would be unsuitable for us because all goals have 'soft' aspiration levels, which is in conflict with our problem.

A similar goal programming approach is taken by Musa and Saxena [122]. In contrast to Cheng and Yeung they include three grades and allow for various part time options. However, they fail to include any preferences apart from the nurses choosing which one out of two alternative weekends to be off work. The most similar goal programming approach to our problem is given by Ozkarahan [124]. His problem is almost as complex as ours, apart from only using two grades of nurses and not allowing any substitution between the grades.

Arthur and Ravindran [7] present another two-phase goal programming heuristic to solve their nurse scheduling problem. Only full time nurses are considered and the three grades of nurses are scheduled independently. Since every other weekend is strictly scheduled to be off, only five shift patterns are available to each nurse. The goals are to meet the minimal staffing requirements and the individual preferences of the nurses. Although the authors propose to extend their model to allow part time nurses and to schedule all grades at the same time, it is not clear from the paper how they will achieve this. Furthermore, there is no limit on the work-stretch length and the model seems to rely on the use of an even number of nurses to function properly.

The nurse scheduling problem tackled in this thesis is also solved by Fuller [72] with XPRESS MP, a commercial integer programming software package. When solved as an integer program, as set up in section 2.1.4, optimisation times can be up to overnight. Using different branching rules and a different formulation with additional variables and constraints, computation time was reduced such that all files were solved within a reasonable time frame. Her full results are reported and compared to our genetic algorithm solutions in appendix D.2. These results show that in principle the problem is solvable with branch and bound methods. However, sophisticated extensions and software are necessary to do so.

## 2.5 Heuristic Scheduling

Gierl et al. [76] present a knowledge-based heuristic for scheduling physicians. Due to the nature of their problem, no grades are taken into account. In addition, no personal preferences are considered. Instead, an overall fairness measure is calculated. This is based on the working history of each physician and aims at spreading out undesired shifts and overtime. The algorithm then continuously cycles through all physicians, assigning shifts to maximise the overall fairness.

A simple staff scheduling heuristic for full time nurses of one grade only is presented by Anzai and Miura [6]. Cyclic descent and 2-opt heuristics are used to optimise the schedule. Schedules of reasonable quality are found after some 90 seconds on an IBM PC. However, the authors conclude that their model was too simplified which is to be addressed in a yet unpublished future paper.

Kostreva and Jennings [104] solve the nurse scheduling problem in two phases. In the first phase, groups of feasible schedules are computed. Each group fulfils the minimum staffing requirements and each individual schedule all major working constraints. Then in a second stage, the best possible aversion score is calculated for each group of schedules. The aversion score is based on the preferences of each nurse and corresponds to the $p_{ij}$ values as described in section 2.1. The group of schedules with the lowest score is chosen. In contrast to our model, Kostreva and Jennings schedule all grades independently from each other. Solution times are reported as approximately ten minutes to generate one schedule on a Macintosh PC.

Blau and Sear [25] solve the problem using full time nurses of three grades, where higher grades may substitute lower grades. In a first step they generate all possible shift patterns over a two week period and evaluate them for all nurses based on their preferences. The best 60 patterns for each nurse are kept and in a second step a cyclic descent heuristic is used to find an optimal overall schedule taking both the nurses' preferences and over- or under-staffing into account.

Decision support systems to solve the nurse scheduling problem are offered by Randhawa and Sitompol [132] and by Smith et al. [153]. Both work in a similar way: The usual constraints as well as nurses' preferences are taken into account. The user is asked to provide weights for various objectives. The algorithm then solves the problem greedily. The focus of these decision support systems is on the interaction between the user and the software, providing a what-if analysis for various sets of weights, rather than optimal solutions.

## 2.6 Meta-Heuristic Scheduling

This section looks at examples of the use of meta-heuristics to solve the nurse scheduling problem. The term meta-heuristic derives from the fact that these algorithms contain many smaller heuristics inside them. This makes meta-heuristics very generic in nature and they can often be used for various problems with only slight modifications. The three meta-heuristics presented here are simulated annealing, tabu search and genetic algorithms. For a concise summary and comparison of these three approaches see Glover and Greenberg [77].

Simulated annealing is a neighbourhood search method where downhill moves are always accepted and uphill moves are allowed under certain conditions to avoid being trapped in local optima. The probability of an uphill move being accepted depends on the change in the objective function value and on the temperature parameter. This parameter controls the search and generally starts out high and then cools down according to a cooling scheme. A higher temperature makes an uphill move more likely.

Isken and Hancock [98] use simulated annealing to solve a variant of the nurse scheduling problem. Their problem is more complex than the one tackled in this research since they have to deal with flexible starting times, instead of three fixed daily shifts. On the other hand, complexity is reduced by the fact that they only schedule nurses of one grade and that under- and over-staffing is penalised but allowed. The problem is modelled as an integer program and then solved with a simulated annealing heuristic. The authors found solutions within 25% of the optimal linear programming solution in less than 15 minutes on a 386/25MHz personal computer.

Another popular meta-heuristic is tabu search. Tabu search is a neighbourhood search method that usually accepts the best possible move. This can include uphill moves if no downhill moves are available. To avoid cycling, a tabu list of the last few moves is introduced. With every new move, the list is updated and moves currently on the list must not be made. The main control parameter of tabu search is the length of the tabu list.

Berrada et al. [21] formulate the nurse scheduling problem as a multi-objective optimisation problem. The authors decompose the problem such that they schedule early, late and night shifts separately and do not consider grades. The resulting problem is modelled with covering constraints and nurses working their contracted number of days as hard constraints and all other constraints (work-stretch, off/on/off patterns, preferences) as soft constraints. The authors use both tabu search and standard mathematical programming techniques to find pareto optimal solutions with regard to the soft constraints. The results presented show that tabu search is capable of solving the problem to the same quality as a commercial software packet (CPLEX), although tabu search was much slower.

Burke et al. [32] also use tabu search on their nurse scheduling problem. However, as their problem is of a very high complexity (planning horizon four weeks with up to 15 possible duties per day), they need to hybridise it with local search heuristics. The results are of better quality than manual solutions and are usually found within minutes.

The same nurse scheduling problem as discussed in this thesis is also solved by Dowsland [55]. Her tabu search algorithm uses a combination of different neighbourhood search strategies and strategic oscillation between finding a feasible solution and improving it in terms of preference cost. Furthermore, a succession of problem-specific special neighbourhood moves is used to improve upon solutions found. The final results match the quality of solutions produced by a human expert. Her results and findings will be compared to ours throughout this thesis and full results are reported in appendix D.2.

The final meta-heuristics presented in this section are genetic algorithms. As they are our chosen method of solving the nurse scheduling problem they are explained in detail in chapter 3 and in Appendix A. In a nutshell, genetic algorithms mimic the evolutionary process and the idea of the survival of the fittest. Starting with a population of randomly created solutions, better ones are more likely to be chosen for recombination into new solutions. In addition to recombining solutions, new solutions may be formed through mutating, i.e. randomly changing old solutions. Some of the best solutions of each generation are kept whilst the others are replaced by the newly formed solutions. The process is repeated until stopping criteria are met.

Easton and Mansour [56] use an enhanced genetic algorithm to solve an employee staffing and scheduling problem. Their approach includes penalty functions to cope with constraints (refer to sections 3.4 and 4.4), local hill climbing to improve solutions (refer to sections 3.5 and 5.4), rank-based selection (refer to section 4.3.4) and sub-populations (refer to section 5.2). The authors compare their results with those of various other heuristics and manage to improve on the best results found so far on a set of 36 test problems. No direct comparison of their tour scheduling problem to our rostering problem is possible as their emphasis is on minimising the number of employees needed to fulfil the schedule and does not take personal preferences into account.

Tanomaru [164] uses a genetic algorithm based heuristic for a staff scheduling problem of similar complexity to ours. He also has a weekly planning horizon and employees of distinct grades. However, rather than using a three shifts approach (early, late and night), employees can start at any time on the hour. Thus, solutions are represented by a list of seven pairs of integers, each pair indicating the start and stop times for each day.

In contrast to our problem, the number of employees is not fixed and overtime is allowed. Hence, the author's objective is to minimise total wage cost. His algorithm also makes use of penalty functions to cope with constraints such as total workforce requirements and maximum individual working shifts.

To reduce the number of infeasible solutions, crossover is only allowed such that 'whole employees' are exchanged. The major optimisation work is then done by a set of nine different heuristic operators. They act as a very sophisticated mutation operator on a single employee basis. Solutions for moderate sized problems obtained after ten minutes on a workstation were of similar quality as those of a human expert. Again, Tanomaru shows the capabilities of genetic algorithms to solve highly complex problems. However, he fails to report to what extent the nine heuristics used are responsible for his results, which makes a comparison to our findings difficult. He also concludes that for real-life problems, his heuristic mutation operators might be too time consuming and suggests a parallel implementation for speed-up.

## 2.7 Conclusions

As the literature review shows, a lot of interest has been paid to the area of nurse scheduling. This indicates that the problem is both interesting and difficult to solve. However, due to the nature of nurse scheduling problems, problem-specific knowledge was required in most cases to achieve good results. This makes it difficult to impossible to include any specific ideas into our model. For instance, cyclic models cannot be used due the importance of the nurses' preferences in our example. Nevertheless, it has been shown that heuristic approaches and in particular genetic algorithms have been successful at solving similar problems.

Two methods have been suggested to solve the same nurse scheduling problem as is tackled in this thesis. Tabu search by Dowsland [55] and integer programming by Fuller [72]. As has been shown in section 2.1.4, the nurse scheduling problem shares some similarities with set covering and generalised assignment problems. Fuller takes advantage of this and uses an advanced integer programming approach to solve our problem. However, this relies on having access to a sophisticated software package and can involve up to overnight computer runs. Nevertheless, the results found by Fuller allow for a thorough comparison and assessment of our results.

Results found by Dowsland are also of excellent quality. However, her algorithm is domain dependent due to the special moves employed. For instance, some moves take advantage of the fact that if one shift pattern containing a particular day is unfavourable so are all others containing this day. This reduces the robustness of her algorithm. In section 6.6, it is shown that this leads to poorer solution quality for more random data.

This leaves room for improvement for the genetic algorithm to capitalise on. As mentioned earlier, genetic algorithms are well known to be very robust for a variety of problems and data. In particular, as the section on meta-heuristic approaches has shown, genetic algorithms have been successful in solving similar manpower problems. Moreover, an earlier pilot study by Aickelin [4] had shown that using genetic algorithms is a challenging but promising approach for this particular problem. The pilot study concluded that the focus point of any future research into solving the nurse scheduling problem with genetic algorithms has to be the handling of the problem's constraints. The next chapter will outline current research into genetic algorithms and then concentrate on this particular aspect detailing various approaches.

# 3 Genetic Algorithms for Constrained Optimisation

## 3.1 Genetic Algorithm Introduction

Due to the increasing popularity of genetic algorithms, a vast amount of research has been published in this area. Thus, no literature review could possibly contain all the information available. Moreover, as mentioned earlier, it was established in the pilot study that the focus of future research into solving nurse scheduling problems with genetic algorithms has to be the successful treatment of constraints. Therefore, this literature review will concentrate on this area. Throughout this chapter examples of related problems, such as scheduling, set covering and generalised assignment problems, are used wherever possible.

However, beforehand this section will introduce the current state of research into genetic algorithms for optimisation purposes. Note that there will not be an extensive explanation of their actual workings. A more precise genetic algorithm tutorial based on Davis [48] and Whitley [174] can be found in Appendix A. Good textbooks on the topic are Goldberg [81] for earlier work up to 1989 and Michalewicz [115], Mitchell [119] and Bäck [9] for more recent research. After a quick summary of the main features of genetic algorithms, this section will go on to discuss recent research about the merits of using them for function optimisation. The remainder of this chapter will review the relationship between genetic algorithms and constraints.

Genetic algorithms are generally attributed to John Holland [96] and his students in the 1970s, although evolutionary computation dates back further (refer to Fogel [68] for an extensive review of early approaches). Genetic algorithms are stochastic meta-heuristics that mimic some features of natural evolution. Canonical genetic algorithms were not intended for function optimisation, as discussed by De Jong [51]. However, slightly modified versions proved very successful. For an introduction to genetic algorithms for function optimisation, see Deb [52]. Many examples of successful implementations can be found in Bäck [8], Chaiyaratana and Zalzala [35], Hedberg [91] and Ross and Corne [142].

To optimise a function, possible solutions are first encoded into chromosome-like strings, in order that the genetic operators can be applied to them. Genetic algorithms start with a population of usually randomly generated solutions. The two main genetic operators are crossover and mutation, both loosely based on their natural counterparts. The crossover operator takes (usually) two solutions, the so-called parents, and recombines them to form one or more new solutions, the so-called children. Parents are chosen from amongst all the solutions of the current population. However, the selection is stochastically biased towards solutions with better objective function values. These are also known as solutions with a higher fitness in evolutionary terms. Therefore, genetic algorithms follow Darwin's theory of 'survival of the fittest'.

Mutation takes one solution and modifies it slightly to form a new solution. After performing a certain number of crossovers and mutations, some of the solutions in the old population are replaced by new solutions and this concludes one generation of the algorithm. These generations are then repeated until a stopping criterion is met. Many additional features are usually necessary in order to optimise real-life problems: Elitism, i.e. the automatic survival of the $x\%$ best solutions, is used to preserve the best solution throughout generations. Additionally, some form of fitness scaling or ranking is often necessary for a robust performance. However, one major problem remains. How does one optimise constrained functions with genetic algorithms, which were originally intended for unconstrained problems? This issue will be discussed in more detail in section 3.2.

The inner workings of a genetic algorithm are often described in terms of the building block hypothesis. The hypothesis says that short low-order solution sub-strings, also known as schema, with higher than average fitness will be reproduced exponentially and spread throughout the population. This is because of the Darwinian selection of parents. The crossover operator then combines such schema or building blocks to form good full solutions.

Two articles discussing the merits of genetic algorithms for operational researchers are Dowsland [54] and Reeves [135]. Dowsland shows how researchers and practitioners were at first reluctant to use genetic algorithms. She argues that this was due to the lack of comparisons of results with those of other methods. Further problems mentioned are the impression that it is difficult to get started with genetic algorithms because of the biological background and terminology and the problems of dealing with constraints. Dowsland continues to point out that more and more of these obstacles are overcome and as this happens, the interest in genetic algorithms is growing. At the time of publication in 1996, she concluded that it was yet to be determined whether genetic algorithms will become a useful part of the operational researcher's toolbox.

As the overview of Reeves [135] from 1997 shows, genetic algorithms have become increasingly popular, especially in solving hard combinatorial optimisation problems. He summarises their essential attractions as:

- *Generality*: Only the encoding and the fitness function need to be changed from one problem to another.
- *Non-linearity*: No assumptions of linearity, convexity or differentiability of the problem are necessary.
- *Robustness*: A wide range of parameter settings will work well.
- *Ease of modification*: Unlike most other heuristics, variations of the original problem are modelled quickly.
- *Parallel nature*: There is a great potential for parallel implementation.

One of the most recent discussions surrounding genetic algorithms is the Free Lunch Theorem, which was originally presented by Wolpert and Macready [181] for non-revisiting algorithms. Non-revisiting algorithms are defined as not visiting the same point in the solution space twice during the course of the optimisation. Wolpert and Macready argue that the performance of all search algorithms on average over all functions is the same, i.e. there is no such thing as a 'best' meta-heuristic or a 'best' encoding for all problems. Therefore, choosing a specific heuristic due to its past performance on other functions may be misleading and it would be better to model the search algorithm after the actual function that needs to be optimised.

The theorem is extended to cover evolutionary algorithms by Radcliffe and Surry [130]. They argue in similar fashion to Wolpert and Macready that the role of the problem representation is central and point to the importance of incorporating problem-specific knowledge into representation and operators. The authors conclude that a much better understanding is still needed to establish a methodology and an underpinning theory.

Finally, a related and interesting observation is made by Ross et al. [144]. After experimenting with evolutionary algorithms to solve timetabling problems, they found a niche in the solution space in which these algorithms outperform hillclimbers. This situation occurred when there was a ‘medium number’ of constraints. If the problem was too tight, the algorithm had problems escaping local optima, whilst if the problem had only few constraints there would be many flat and unfriendly plateaux.

## 3.2 Constrained Optimisation with Genetic Algorithms

As seen from the descriptions in the previous section, there is no pre-defined way of including constraints into an optimisation using genetic algorithms. This is probably one of their biggest drawbacks, as it does not make them readily amenable to most real world optimisation problems. To solve this dilemma, many ideas have been proposed. These form the remainder of chapter 3. A good overview of most of the techniques presented in this chapter can be found in Michalewicz [115] and more concisely in Michalewicz [114].

Many genetic algorithms, including ours, often use a combination of the strategies described in the following. However, for easier understanding, all methods are explained independently and presented in the following order. The first method, that of including constraints in the encoding, is detailed in Section 3.3. However, as will be seen, the number and types of constraints that possibly can be implemented in this way is limited. Two widely applicable methods are presented in the two following sections, namely penalising in section 3.4, and repairing in section 3.5. Penalising tries to avoid infeasible solutions by steering the algorithm away from them, whilst repairing tries to 'fix' such solutions to become feasible.

The methods in section 3.6 are based on the standard genetic operators of mutation and crossover. In contrast to repairing, their main use is to preserve rather than to create feasibility. An indirect genetic algorithm approach incorporating a decoder is presented in section 3.7. Finally, section 3.8 gives an overview of other, problem-specific or otherwise less widely applicable, methods. For the purpose of simplicity, all examples in this chapter, unless otherwise stated, will be for the minimisation of the target function.

## 3.3 Implementing Constraints into the Encoding

Following the argument that binary alphabets offer a maximum number of schema that will be simultaneously sampled by a single individual (refer to implicit parallelism in appendix A.3), binary coding used to be the most common way of encoding strings. Even with binary coding, some simple constraints can be implicitly implemented. Take the following example: $f(x) = x^3 - x^2 + x$ is to be minimised for $0 \leq x \leq 15$ and $x$ integer. If $x$ is encoded as a binary string of length four, then the constraints are implicitly fulfilled by the encoding because $x$ will automatically be between 0 and 15 and no genetic operator can disturb this.

In the case of binary encodings, it is very difficult to include all constraints in such a way, as slightly different constraints like $0 \leq x \leq 20$ would already make this impossible. In this particular example, it would become easier if higher alphabets were used. Both Antonisse [4] and Radcliffe [131] argue that non-binary encodings have advantages over a binary encoding. Antonisse points out that although individuals encoded with a higher alphabet contain fewer schemata, each of these schemata is of a higher 'power' as wildcards now stand for more than just two possibilities. Radcliffe further argues that a binary coding is often inappropriate since similar solutions do not share many schemata, for example x = 7, i.e. (0111) and x = 8, i.e. (1000).

Nowadays, more and more researchers use non-binary alphabets, which allow a more 'natural' expression of their problem. We also chose a non-binary encoding which allowed us to include constraints (1) and (2) of section 2.1 implicitly. Our solutions are encoded as a string of shift patterns, such that the *ith* pattern is worked by nurse *i*. For more details of the nurse scheduling encoding refer to section 4.1.

## 3.4 Penalty Functions

Because of its relative straightforwardness, the penalty function approach is the most common way of dealing with constraints in a genetic algorithm context. The approach works by 'measuring' the infeasibility of a solution with a suitable penalty function. For instance, this could be the number of violated constraints, the magnitude of violation of constraints or any mixture thereof. This penalty is added to the target function value of the solution, usually after being scaled with a penalty factor. In genetic algorithm terms, the idea of this approach is to have infeasible solutions in a population, but to penalise them such that feasible solutions are fitter.

One advantage of penalty functions is that they can accommodate any number and type of constraints with relative ease. The standard way of doing this is to transform all constraints into inequalities of the form $f(x) \leq 0$. All violations, i.e. any positive left-hand sides, are then summed up and added as a penalty to the target function. Another advantage of penalty functions is that infeasible solutions are allowed in a population. Thus, the information stored in them is accessible to the genetic algorithm.

For example, one can imagine many cases where the successful combination of two infeasible but otherwise highly fit individuals leads to one feasible and highly fit solution. In addition, for many real-life optimisation problems, like our nurse scheduling, the problem is so tight that finding a feasible solution in itself is a very difficult task. Therefore, there simply must be a facility to evaluate and deal with infeasible solutions, rather than just discarding them.

However, the penalty function approach has some major drawbacks. For many problems simply adding up the violations of all constraints and adding this as a penalty to the target function is not successful as the search is not guided adequately. It is often necessary to develop more sophisticated penalty functions. For instance, one could rate the violation of certain constraints differently to the violation of others or introduce a general penalty weight with which the total violation is multiplied before it is added to the objective function. Unfortunately, it is highly experimental and thus time consuming to find good penalty functions and weights for one's problem as only general guidelines exist.

One unresolvable dilemma is that even with a suitable penalty function, finding a feasible solution is never guaranteed. It is always possible for a slightly infeasible but otherwise highly fit solution to dominate the search and for the whole search not to produce a single feasible solution, particularly if the problem's constraints are tight. In many practical cases, this might not matter, but for the nurse scheduling problem a solution has to be feasible. As the remainder of the thesis will show, this will be one of our main concerns. On the whole, penalty functions are a useful tool to assign a fitness value to infeasible solutions, but they are unlikely to dispense fully with the problems created by the constraints.

Other researchers have identified some further peculiarities of penalty functions. Hadj-Alouane and Bean [87] prove that for the Lagrangian relaxation of the general multiple-choice integer program there is no set of weights that guarantees to find the optimal solution with a linear penalty function. This is caused by dual degeneracy due to multiple optimal solutions, which is often the case in practical problems. The authors then prove that there must be a set of weights that finds the optimal solution with a quadratic penalty function. Another interesting observation is made by Johnson et al. [99] for a graph colouring problem. They prove that for their particular penalty function there is a constant set of penalty weights that guarantees to find a feasible solution over all problem instances.

Before presenting more details and examples of penalty functions, we cite Richardson et al. [138], who were one of the first to set up some general guidelines for the use of penalty functions. They concluded that penalties should be close to the expected completion cost and therefore, measuring the distance from feasibility is better than merely counting the number of violated constraints.

Richardson et al. define completion cost as the cost of transforming an infeasible and hence incomplete solution into a feasible solution. However, some difficulties arise from this. Not only are completion costs difficult to calculate, but many infeasible solutions can also be completed into different feasible solutions at various completion costs. Even if a satisfactory completion cost function for a problem can be found, it will often be the case that various stages of the search will need different levels of pressure.

Therefore, to simplify matters the penalty cost is often set equal to the ‘distance’ of an individual from the feasible region. For instance, this could be measured as the number of violated constraints or the sum of all constraint violations. Examples of this can be found in Easton [56], Li et al. [109] and Ross [145] who all use the sum of constraint violations as the penalty cost. We also use this type of penalty measure in section 4.1.

A survey of penalty functions and their performance on test problems is carried out by Gen and Cheng [74] and Michalewicz [116]. Amongst them they present and compare:

- Penalty functions with variable weights for each constraint.

- Penalty functions with weights depending on the number of iterations.
- Penalty functions with weights dynamically scaled depending on the best solution found so far.
- Penalty functions using levels of violations rather than the actual constraint violation.
- Penalty functions that use active and inactive constraints where only active constraints contribute to the penalty. A separate function decides in every generation if a constraint is declared active or not.
- Penalty functions that use near-feasibility thresholds, i.e. a certain amount of infeasibility is not penalised.

The authors conclude that a comparison of methods is difficult and the choice is highly problem-specific. It seems that adaptive penalties outperform static ones, however the more sophisticated the method the more parameters need to be set which is a difficult problem in itself. Further information about dynamic penalty functions and their merits for this research is presented in section 4.4.

## 3.5 Repair

As mentioned before, many of the methods reviewed in this chapter are used in combination with one another. This is particularly true for the use of penalty functions and repair algorithms. The reason for this is that the main weakness of penalty functions, i.e. not guaranteeing to find a feasible solution, can often be remedied by a suitable repair function. However, due to their very nature, repair functions are even more problem-specific than penalty functions so not even general guidelines exist. Therefore, this section will present various examples of repair algorithms rather than rules and guidelines.

Often repair functions are not only used to make an infeasible solution feasible, but also to improve the fitness of solutions. A common way of achieving this is to choose the best instance a solution can be repaired into from amongst all feasible possibilities. Similarly, one could apply such a repair function to already feasible solutions simply to improve upon their quality. When used in this way repair functions become local hill-climbers. This is often done to counter the 'lack of killer instinct' of genetic algorithms (De Jong [51]), which refers to their inability to make small changes. However, towards the end of the search small moves are often required to improve results further and the mutation and crossover operators are too disruptive to provide these.

When using repair algorithms, the user has a choice: Should the repaired version substitute the original individual, known as Lamarckian Evolution, or should only the fitness of the original string be changed, known as the Baldwin Effect? Baldwinian style repair operators are sometimes also referred to as memetic algorithms. Whitley et al. [172] investigate this question on some simple functions. Contrary to intuition they show that functions exist where following the Baldwinian strategy is superior to using Lamarckian Evolution. However, the authors also point out that a Baldwinian search is much slower than a Lamarckian search. Whitley et al. conclude that it was too early for definite results as to decide which strategy was superior and more research using more complex functions needs to be carried out first. Note that for the remainder of this thesis unless otherwise stated, 'repair' will refer to Lamarckian Evolution.

Two examples of a successful use of repair functions to find feasible solutions and improve upon them are Bäck et al. [12] and Beasley and Chu [18]. Bäck et al. compare the use of penalty functions with that of a repair scheme on a set covering problem. They use a graded penalty function, which by itself produces feasible but not high quality solutions. By repairing solutions, following a minimum cost principle, the authors manage to improve upon the quality of results significantly. The authors conclude that incorporating a simple repair method was far superior than using a penalty function on its own. They also point out that more research into repair functions needs to be done, particularly into frequency of repair.

Beasley and Chu [18] also solve the set covering problem using a genetic algorithm with a repair heuristic. They do not need a penalty function as their repair method guarantees feasibility. It works by greedily covering uncovered columns following a simple rule. The rule works by adding missing columns ordered by the cost of a column divided by the number of uncovered rows that the column would cover. They also have a local hill-climbing element in their repair algorithm. After achieving feasibility through adding columns, now redundant columns are dropped, with higher cost columns being dropped first. Using this algorithm Beasley and Chu were able to generate optimal solutions for small-size problems and high-quality solutions for large-size problems.

A slightly different use of the repair operators is made by Burke et al. [33] who apply genetic algorithms to highly constrained examination timetabling problems. The penalty function used is a weighted sum of timetable length and the number of conflicts between exams in adjacent periods. Rather than having a separate repair heuristic, they combine it with their uniform crossover operator. Instead of choosing exams at random to go from parents to children as in standard uniform crossover, heuristic rules are followed. For example, exams are selected according to how many conflicting exams they have in common with those already placed in the child. Another rule is to select exams such that the number of conflicts with exams in the previously scheduled period is minimised. Results are reported to be of good quality.

Further examples of repair algorithms are: Eiben et al. [60] and Eiben et al. [61], who solve constraint satisfaction problems, like the Zebra and N-Queens problems, successfully with a combination of genetic algorithms and heuristic-based repair strategies; Herbert and Dowsland [93], who solve the pallet loading problem using a genetic algorithm combined with a graph-theoretic based repair heuristic; and Miller et al. [118], who show that the addition of the right type of local improvement operator allows a genetic algorithm to solve the NP-hard problem of multiple fault diagnosis to within 99% of optimality.

## 3.6 Special Operators

Although very similar to the repair strategies detailed in section 3.5, the operators presented here are aimed more at preserving feasibility than at creating it. Moreover, they are usually enhancements of standard genetic algorithm operators like mutation and crossover rather than 'stand-alone' routines. Again, very little is published about the general case of genetic operators and constraints. Reid [137] presents a theoretical investigation of the behaviour of two-point crossover on solutions to a constrained integer optimisation problem. He concludes that more specialised crossover and mutation operators, which preserve feasibility, are often superior for loosely restricted problems. However, for highly constrained problems this would be too limiting, as finding (different) feasible solutions is a difficult task in itself. Thus, there is a danger of premature convergence in these cases.

Examples of special crossover and mutation operators that preserve feasibility can be found in the papers of Bilchev and Parmee [23], Levine [108], Tanomaru [164] and Ross et al. [143]. Bilchev and Parmee solve a loosely constrained fault coverage code generation problem. After creating all legal templates of possible combinations of test code, they apply crossover such that the template itself always stays intact.

Levine [108] solves a set partitioning problem with the help of a special 'block' crossover. Before using the special crossover, the set partitioning matrix is brought into block 'staircase' form. Block $B_i$ is then defined as the set of columns, which have their first one in row $i$. In any feasible solution, at most one column of each block may be present. The idea of the block crossover is to help preserve feasibility by setting the crossover column always to the first column of some block.

Tanomaru [164] uses a genetic algorithm for a staff scheduling problem. He defines a working shift as the list of shifts a particular employee works within the planning period. Thus, a chromosome or solution in his case is the list of working shifts for all employees. A special two-point crossover is then used, such that only whole working schedules for employees are exchanged. Within an employee's schedule, heuristic operators are applied.

To improve timetabling solutions, Ross et al. [143] use a violation directed mutation. The key aspect of this new kind of mutation is that both the position of the gene to be mutated and its new value are chosen such that constraint violations are reduced. In order to do this, the authors calculate a 'violation score' for each gene, which is equal to the sum of the violated constraints the gene takes part in. Experiments show that this type of mutation is highly successful in finding good solutions, in particular when the selection of the gene and its new value is stochastically biased. For instance, tournament selection is more successful than basing the mutation on the highest violation score or on the best improvement.

## 3.7 Decoders

Decoders offer an interesting option of solving constrained optimisation problems with a genetic algorithm. Instead of being a 'direct' solution to the problem as with standard genetic algorithms, a chromosome is now a set of instructions of how to build a solution by using another routine, namely the decoder. For instance, the genotype of an individual could represent a permutation of the items to be processed, for example in the job shop scheduling case a permutation of the list of jobs to be scheduled. After the usual genetic algorithm operations, the decoder then picks up the items in the order given by the permutation and builds a feasible solution from it, following specific rules. This final solution, i.e. the phenotype, is then assessed and its fitness score is given to the original permutation based individual.

One idea behind this approach is that 'difficult' variables will be moved to the front of the string and their values decided whilst there is still a high degree of freedom. 'Easier' variables follow later when some constraints are already tight. Another type of indirect genetic algorithm finds optimal settings of parameters or rules for heuristics that then solve the actual problem. For instance, these could be priority weights for various scheduling rules or a list of which sub-heuristics to use.

An advantage of these indirect approaches is that the genetic algorithm can be left almost unchanged from the canonical version presented in appendix A.2. In particular, there is no need for a penalty function as all solutions are decoded in such a way that they are feasible. The only difference to a standard implementation is that since the new chromosomes will be permutations, both crossover and mutation have to take this into account. Hence, appropriate operators, such as partially mapped crossover and swap mutation, must be used. For examples of these see appendix A.2.5 and section 6.2. All the problem-specific knowledge is built into the decoder, thus no further separate repair or hill-climbing heuristic is needed.

Note that the original intention when using a decoder was to ensure that solutions are guaranteed to be feasible once decoded. This follows the work and ideas of Davis [47] and Davis [49] who was one of the first to use decoders. This was achieved by making important constraints soft. In fact, the 'constraints' were often part of the objective in the first place. For example, in the exam scheduling case there is no fixed limit on the number of slots available. Although solutions are penalised for using more than a specific number of slots there is no upper limit to the length of the schedule, i.e. exams can always be placed at the end of the queue.

This is different from our approaches presented in chapters 6 and 7, where solutions are not guaranteed to be feasible and hence a penalty function approach is still needed to take this into consideration. In contrast, the traditional use of the decoder follows Palmer and Kershenbaum [125] who set out the following rules between the relationship of solutions in the original and in the decoder's space:

- For each solution in the original space, there is a solution in the encoded space.
- Each encoded solution corresponds to one feasible solution in the original space.

- All solutions in the original space should be represented by the same number of encoded solutions.
- The transformation between solutions should be computationally fast.
- Small changes in the encoded solution should result in small changes in the solution itself.

Details of how these rules conflict with the indirect genetic algorithm approach to the nurse scheduling problem and the nature of our decoder will be shown in section 6.3.3. However, some general points about using an indirect genetic algorithm with a greedy decoder can be made. Consider a minimisation problem. The above rules state that all solutions should be evenly represented in the encoded space. Often with a greedy decoder, this is not the case, as decoding is biased towards low-cost areas. Thus, large unattractive parts of the original solution space are cut out speeding up the search. Similarly, some decoded solutions will be represented more often than others, because the decoder is biased towards low-cost areas. As long as there is some variety in the initial population, this should not cause any problems, because children from two different permutations mapping to the same solution are unlikely to do so as well.

All other rules have some merit. In particular, it is important that an encoded solution always maps to the same decoded solution. Otherwise, there is no one-to-one relationship between fitness value, encoded and decoded solution. This could mislead the genetic algorithm. Unfortunately, the presence of hard constraints in our problems will prevent some decoded solutions from being feasible. In contrast to this, the remainder of this section will present examples of using the decoder in the traditional way, i.e. mainly following the above rules and guaranteeing feasibility by relaxing or not having any important constraints. Hence, the difference is that in our case infeasible solutions are not only penalised but are also unacceptable as final solutions.

One area where such decoders have proved popular is that of job-shop scheduling. Here it is difficult to maintain a feasible schedule using direct genetic algorithms. This was first noted by Davis [49]. He explains that using a direct representation, i.e. work station $w$ performs operation $o$ on object $x$ at time $t$, would lead to severe problems of upholding feasibility with standard crossover operators. Thus, he suggests an indirect representation, which for each work station is a preference list of the order of operations including the options ‘idle’ and ‘wait’.

A decoding routine then picks up the first legal action from the preference list of each work station. Crossover works by exchanging preference lists between work stations and mutation scrambles the members of a preference list. The evaluation function measures the cost of the whole schedule adding a large penalty if a certain time limit is not kept to. Using this approach the author finds good solutions and suggests using the combination of a ‘random’ genetic algorithm with a ‘deterministic’ decoder for other problems.

Other examples of using decoders for job-shop scheduling problems are described in Bagchi et al. [13] and in Fang et al. [64]. Bagchi et al. use an indirect encoding representing order priorities. However, they then point out that such a simple encoding would not be sufficient for their problem. This is because it restricts the search to the space of all permutations of job orders and does not take into account additional information. For instance, this additional information could be the possibility that some orders can be produced with different resources on different machines.

They argue that leaving these decisions to the decoder would be unsatisfactory, as it would be difficult to implement rules and slow down the search greatly. Therefore, they add allocated resources to each order in the priority list, making sure the random initialisation is feasible with respect to resources and orders. They then use a combined crossover operator, i.e. a partially-mapped crossover for the priority list and a uniform crossover for the resources. Similarly, two mutation operators are used. Experiments show that the more extensive encoding is vastly superior to the one based on simple order priorities.

Also using an indirect encoding representing order priorities, Fang et al. [64] investigate the convergence speed of different parts of the string. The authors discover that earlier parts of the string converge much faster than later parts. This is an expected effect due to the variation of 'significance' across the string. To exploit this and further improve their results they examine dynamic sampling of the convergence rates of different parts of the string and then use this information to target positions for crossover-points and mutation. Hence, high variance sections of the string are more likely to be chosen for crossover and low variance sections for mutation. Results using these enhancements are significantly better than those found using standard operators.

Two examples of the use of decoders for timetabling are given by Podgorelec and Kokol [127] and Corne and Odgen [42]. Podgorelec and Kokol use a genetic algorithm to optimise the ordering of events, in their case patients who have to attend therapies. Their objectives include minimising overall duration of all therapies, minimising maximum and average individual waiting times and idle times of devices. The ordering of events is decided by the genetic algorithm with a decoder constructing a timetable by giving each event the earliest possible slot. A weighted fitness function is used to measure the quality of a solution against all goals. Reported results are good and are further improved by dynamically adjusting parameter rates following some simple machine learning rules.

Corne and Odgen compare the effectiveness of hill-climbing, simulated annealing and genetic algorithms, all with direct and indirect representations, for optimising preaching timetables. In the direct representation, the first preacher is used for the first sermon, the second for sermon 2 etc. In the indirect case, the string is a permutation of all sermons and is then 'filled' by a decoding timetable builder following problem-specific heuristics.

They find that for all three methods the indirect approach works better than the direct approach. However, one could say that the timetable builder was at an advantage using more problem-specific knowledge. The authors agree, but argue that it would have been very time consuming in both implementation and use to employ 'smart' features in the direct case. However, in their experiments the genetic algorithm is out-performed by the hill-climber and by simulated annealing, which the authors attribute to the structure of the solution landscape being multimodal with steep-sided optima.

Further examples of interest are: Blanton and Wainwright [24], who solve a multiple vehicle routing problem with time and capacity constraints using an indirect encoding. They are successful with a problem-specific merge crossover operator that is based on a global precedence matrix of all customers to be visited. Jones and Beltramo [100], who use a greedy adding heuristic as the decoder for a partitioning problem that has been encoded as permutations of the objects to be partitioned. Reeves [136], who solves bin-packing problems by hybridising an indirect genetic algorithm with a heuristic best-fit decoder. Finally, a minimum span frequency assignment is dealt with by Valenzuela et al. [168] in a similar fashion to Reeves.

## 3.8 Miscellaneous Methods

This section concludes chapter 3 with a further selection of constraint handling methods that do not easily fit into any of the former categories, although some methods share similarities or are derivations of methods presented earlier.

Both Barnier and Brisset [14] and Bruns [29] combine a genetic algorithm with constraint logic programming techniques. Constraint logic programming is an advancement over linear programming making active use of constraints whilst conducting an exhaustive branch-and-bound search of the solution space. When one variable value is chosen, all other variables' domains are dynamically adjusted via the mutual constraints. This reduces the search space and speeds up the search.

The authors argue that constraint programming is suitable for highly constrained problems. However, due to its exhaustiveness, it is confined to relatively small search spaces. Genetic algorithms on the other hand, due to their fast sampling are suitable for large search domains, but have difficulties dealing with constraints. Hence, Barnier and Brisset and Bruns suggest hybridising both methods such that the genetic algorithm searches the whole problem space for promising sub-spaces and then using constraint programming techniques to optimise within these promising regions.

Following this approach, Barnier and Brisset successfully solve vehicle routing and radio frequency assignment problems, outperforming both genetic algorithms and constraint programming on their own. Bruns solves a job shop scheduling problem to similar quality as more problem-specific heuristics. However, he also points out that by using this combined approach, the constraint programming step might fail to find feasible solutions if the domains created by the genetic algorithm were empty with respect to feasible solutions. This could be the case for harder problems, although in his case due to the 'softness' of the main constraint it was not. He therefore suggests adding either a backtracking or otherwise suitable problem-specific operator to prevent this from happening.

Kowalczyk [105] also combines genetic algorithms and constraint programming. However, in his paper constraint programming techniques are applied during uniform and n-point crossover and also to initialise all solutions as feasible. Then, in the uniform crossover case, after randomly choosing which parent the first gene of the child comes from, a matrix for all other genes is constructed. This is done via constraint consistency checking and shows from which parent the other genes might come from to preserve feasibility. If the matrix contains only one parent then its gene is chosen automatically, otherwise it is chosen at random. After every gene, the matrix is updated again.

However, there are two drawbacks to Kowalczyk's method: By initialising all individuals via constraint programming, the variety of values for some genes might be restricted. This can lead to premature convergence. Furthermore, the crossover operator could lead to a situation where neither gene from either parent will lead to a feasible child. In those cases, the author suggests backtracking which again might lead to premature convergence.

A different hybridisation approach is presented by Cheung et al. [39]. They combine their genetic algorithm with a grid search method. The grid search heuristic works by performing a pattern search around solutions found by the genetic algorithm. Results are given for the mixed integer non-linear programming problems of the development of oil fields and the optimisation of a multi-product batch plant. The authors conclude that their hybrid method successfully handles problematic non-convex constraints but still needs further refinement.

Chu and Beasley [40] and Chu and Beasley [41] present a variation on penalty functions. They solve set partitioning and generalised assignment problems with a genetic algorithm using separate fitness and unfitness measures of individuals. Fitness is defined as in the standard genetic algorithm, i.e. it is equal to the objective function value of an individual. Unfitness measures the degree of infeasibility (in relative terms), i.e. for the set partitioning problem the number of rows covered more than once.

The authors then use the fitness and unfitness scores to divide the population into four distinct sub-populations. According to Chu and Beasley, these sub-populations allow for a much better replacement strategy than straightforward strategies, for instance replacing the solution with the worst fitness or the worst unfitness. Separation of fitness and unfitness also helps in terms of parent selection. Utilising these properties, they use a so-called matching selection method, which results in better quality offspring by choosing compatible parents. Overall, they achieve much higher feasibility amongst their solutions than with standard methods.

Wilson [180] also solves the generalised assignment problem. In a first step, the author finds the optimal solution by relaxing the capacity constraints. This solution is then used to seed the genetic algorithm population. This seeding works by randomly changing the covering row for some columns. The row is only changed if the corresponding capacity constraint is violated. The seeding results in a population of near-feasible solutions. The genetic algorithm then runs either for a pre-defined number of generations or until a feasible solution has been found. The final solution is then further improved with a 2-opt local search method. Wilson concludes that the solutions found to reasonably sized problems were of near optimal quality, but the genetic algorithm was outperformed by an older problem-specific heuristic approach.

Le Riche et al. [106] have two populations, each with its own penalty parameter in a 'segregated' genetic algorithm. One parameter is chosen to be large whilst the other is chosen to be small. In an inter-breeding phase in every generation, solutions from both populations generate children in addition to the standard inter-population crossover. The authors argue that this method allows them to approach the feasibility border, which usually holds the optimal solutions, from both directions, i.e. from inside and outside. However, they leave the question unanswered how to ensure that the algorithm converges to the area of the feasibility border containing the global optimum.

Paredis [126] experiments with a different representation of individuals. Each gene is allowed to take an additional '?' value on top of values from its domain. A '?' represents a choice that is still open. Initially, many '?' are present in the population. During the run of the genetic algorithm, more and more '?' are filled in by using constraint programming techniques. Thus, individuals are promising search states in the beginning and solutions later. The fitness of such partially defined individuals is defined as the objective function value of the best possible completed solution. Using this method, Paredis solves N-Queens and job shop scheduling problems successfully.

Dechter [53] and Kennedy [101] suggest backtracking. They initialise their populations with feasible solutions only. Should crossover produce infeasible children, the backtracking algorithm makes them increasingly similar to one parent until they are feasible again. Whilst this approach might work if the problem is not very tight or there is just one feasible region, for other problems this might lead to premature convergence.

If the solution space is convex and all variables are of numerical type, special geometric crossover operators and other geometric heuristics can be used to maintain feasibility. This can be seen in Michalewicz and Attia [112], Michalewicz and Janikov [113], Schoenauer and Michalewicz [148] and Schoenauer and Michalewicz [149]. As our problems are of a discrete nature, these methods cannot be used.

Schoenauer and Xanthakis [150] suggest cycling through the constraints, i.e. only one constraint is active at any time. Whenever a certain percentage of all solutions fulfil this constraint, the algorithm moves on to the next one and so on. The authors use this on a truss structure optimisation problem with hierarchical constraints where good solutions are found. Although Schoenauer et al. conclude that their method is problem independent, it seems fair to assume that it works best with problems that have only few constraints or are of the hierarchical constraints type.

## 3.9 Conclusions

As this chapter has shown, genetic algorithms have been successfully used to optimise a variety of problems. Moreover, there are various ways of dealing with constraints. Some of these are more general, whilst others are problem-specific. In particular, we have seen many applications to problems closely related to the nurse scheduling problem. With direct genetic algorithms these methods include implementing constraints into the encoding, penalty functions, repair algorithms and problem-specific crossover and mutation operators. Alternatively, indirect genetic algorithms have been used. There the problem-specific information is contained with a separate decoding heuristic.

None of these methods can guarantee feasibility in general and as a result there is no definitive way to handle constraints. Thus, the choice of constraint handling method is very problem dependent. This indicates that there is room for improvement and a more generic constraint handling technique would be of great benefit. In the following chapters we will look at most of the methods suggested in this chapter and show their merits and drawbacks for our particular problems. From these observations, new operators will be developed to handle the constraints present successfully. Reference to other relevant research will be made as appropriate.

# 4 A Direct Genetic Algorithm Approach for Nurse Scheduling

## 4.1 Encoding of the Problem

The previous chapter indicated that there are two main avenues for solving problems such as the nurse scheduling example with genetic algorithms: A direct and an indirect approach. This chapter concentrates on a straightforward direct encoding of the problem and experiments with genetic algorithm parameters and different penalty weights and functions. A more sophisticated direct approach to the problem is presented in the next chapter and an indirect approach is shown in chapter 6. Parts of this chapter are based on the results found during the pilot study (Aickelin [4]), which is summarised in Appendix B. However, due to the pilot study's limitations (one encoding, limited types of nurses and only six data sets) further work is necessary. First, consideration needs to be given to different direct encodings.

There are many different possibilities for encoding this problem. In the pilot study, good results were found with the encoding being the list of shift patterns worked by the nurses. As outlined in the remainder of this section, because of its superiority the same encoding is again used for this research. However, before doing so, the merits of using a different encoding have to be assessed. Generally, all good encodings should have the following qualities:

- Incorporate as many constraints into the encoding as possible (refer to chapter 3.3 for details).
- Allow for a fast evaluation of the fitness of the string.
- Offer some degree of inheritance by fitness between parents and children.
- Make sensible crossover and mutation operators possible.
- Be relatively short so there is less conflict with the building block hypothesis.

One possible direct encoding is to have a string of length equal to the total number of nurse shifts required in a particular week. Each shift on each day could then be represented by a sub-string of length equal to the number of nurses required at that time. Thus, the string would consist of 14 sub-strings, first the seven days then the seven nights. The values of each gene in the sub-string would then represent the index of the nurse working that shift. For example, the following would be the beginning of a string if five nurses were required on for each day shift: (1,13,14,21,5; 2,8,7,3,10…). In this example, nurses 1, 5, 13, 14 and 21 would work the day shift on Monday, nurse 2, 3, 7, 8 and 10 the day shift on Tuesday etc.

If this encoding was changed slightly, the three grade bands could also be taken into account. In order to do this, there have to be three sub-strings for each day and night. The length of each sub-string is then equal to the number of nurses required of grade one, two and three respectively. For example (1; 13,21; 5,14…) would be a partial solution where one nurse of grade one, two nurses of grade two and two nurses of grade three are required on the first day shift. The next step is then to ensure that only nurses of the required grade or higher are allowed to fill a 'slot'. Thus, the whole of covering constraint set (3) of the integer programming formulation in section 2.1 is implicitly fulfilled by this encoding.

However, there are some drawbacks with this encoding. Firstly, the strings would be of considerable length. For an average data set, they would consist of some 100 genes. This is calculated as seven days multiplied by the average number of nurses needed on a day shift, plus seven nights multiplied by the average number of nurses needed per night shift. Another problem is that the actual shift pattern worked by each nurse needs to be extracted first, which means that the whole string needs to be scanned for each nurse to arrive at her day on / off pattern. This is computationally expensive but necessary to include the 'cost' of shift patterns, as given by the $p_{ij}$ values defined in section 2.1.

Furthermore, even if initialised correctly, standard crossover and mutation operators are very likely to disturb each nurse working the correct number of shifts per week as required by her contract. Moreover, it could happen that a nurse was scheduled to work both day and night shifts in one week, although her contract does not allow this. It could even lead to situations where a nurse possibly works two shifts on the same day or night.

For instance, consider the following simplified example for two day and two night shifts, where three nurses are required on each. Two feasible solutions for this example are (1,2,3; 1,2,3; 4,5,6; 4,5,6;) and (4,5,6; 4,5,6; 1,2,3; 1,2,3;). In the first solution nurses 1, 2 and 3 work the two days and nurses 4, 5 and 6 the two nights. In the second solution, the situation is reversed. Any standard k-point or uniform crossover with these two solutions as parents will yield an infeasible solution where at least one nurse will work both a day and a night shift on the same day.

This disadvantage in particular renders this type of direct encoding awkward for the problem. Although special operators, similar to those presented in chapter 3.6, might be able to improve the situation, they will be difficult and time consuming to construct. Therefore, we chose not to use this type of encoding in the first instance.

A second possible encoding would be a string of as many binary vectors as nurses. Each vector would be of length fourteen and represent the seven days and nights of the week. As with the shift patterns, a 1 would represent a day (or night) worked and a 0 a day (or night) off. This encoding has the advantage that the children produced by crossover are of greater variety than those formed by the encoding used in the pilot study. In encodings with shift patterns, the child can only take the pattern of either parent. With this encoding there are many more possibilities if crossover points are allowed within the vector of a nurse.

However, there is also a downside to this when used with standard crossover operators. In almost all cases, this will result in infeasible vectors for a nurse, for instance a nurse working too many days or working both days and nights in a week. Additionally, solutions can be infeasible in respect to the covering constraint. Thus, no constraints are implicitly incorporated. Another disadvantage of this encoding is the considerable length of the strings. The first problem in particular led us to think that this type of encoding is not very practicable for the nurse scheduling problem.

After considering the above encodings, we chose to use the same encoding as in the pilot study. This encoding follows directly from the integer program formulation in section 2.1, i.e. an individual is a concatenated string of each nurses' shift pattern worked. Each individual thus represents a full one-week schedule and is a string of $n$ elements with $n$ being equal to the number of nurses. The *ith* element of the string is the index of the shift pattern worked by nurse $i$.

Such an encoding means that any standard crossover allocates some nurses to the shift patterns worked in one parent and the remainder to those worked in the other parent. A bit mutation operator changes the shift pattern of just one nurse. A possible individual for five nurses would therefore look like (12, 1, 128, 218, 24), indicating that nurse 1 works shift pattern 12, nurse 2 shift pattern 1 etc. For details about the transformation of a shift pattern into the actual days and nights on or off, see appendix C.4.

This encoding was chosen for the following reasons. Firstly, it builds on the previously established shift patterns and penalty costs $p_{ij}$ (for a definition of the $p_{ij}$ values, see section 2.1.3). Hence, with this encoding there is a nice one-to-one relationship between the penalty costs and the genes and therefore also with the building blocks. Secondly, it is equivalent to the one successfully used in the pilot study and by Hadj-Alouane and Bean [87] for general multiple-choice integer programs. Another advantage of this encoding is its compactness, as the length of the string is equal to the number of nurses. This again should help the formation of short highly fit schemata, as stipulated by the building block hypothesis.

Finally, as outlined in section 3.3, a good encoding should incorporate some of the problem's constraints. By using the above encoding, as each nurse corresponds to one position in the string, it is guaranteed that all child solutions obey the constraints that each nurse works exactly one shift pattern. This is the multiple-choice constraint set (1) of the integer programming formulation. As long as the initial values and those allowed for mutation of each gene are selected from the values of feasible shift patterns for each nurse, then constraint set (2) is automatically taken care of by the encoding, too. This leaves only constraint set (3) to be dealt with further. From our point of view, this last encoding is superior, as it is easier to ‘fix’ the covering constraint set (3) than constraint sets (1) and (2). This view is based on attempts to correct such infeasible solutions by hand.

The final constraint set (3) can then be incorporated via a penalty function approach as described in section 3.4. The resulting target function will be used to measure an individual’s raw fitness and is as follows. Note that only undercovering is penalised not overcovering, hence the use of the *max* function.

$$\sum_{i=1}^{n}\sum_{j=1}^{m} p_{ij}x_{ij} + w_{demand}\sum_{k=1}^{14}\sum_{s=1}^{p}\max\left[R_{ks} - \sum_{i=1}^{n}\sum_{j=1}^{m} q_{is}a_{jk}x_{ij};\, 0\right] \rightarrow \text{min!}$$

The parameter $w_{demand}$ is referred to as the penalty weight in the following. It is used to adjust the penalty that a solution has added to its raw fitness. Note that a fitter solution has a lower raw fitness as we are minimising the objective function. For each unit of undercover, a penalty of the size of $w_{demand}$ is added to an individual’s raw fitness. For instance, consider a solution with an objective function value of 22 that undercovers the Monday day shift by two shifts and the Wednesday night shift by one shift. If the penalty weight was set to ten, the raw fitness of this solution would be 22 + (2+1)*10 = 52.

We are now ready to apply a canonical genetic algorithm to the problem. This will be started in the section 4.3 along with an investigation into the effects of parameter settings and different genetic algorithm strategies. Beforehand, the next section describes the data and computer equipment used for the experiments and explains how the results were evaluated.

## 4.2 Description of Experiments

For all experiments, 52 real data sets as given to us by the hospital are available. The data was collected from three wards over a period of a few months. Unless otherwise stated, to obtain statistically sound results all experiments were conducted as twenty runs over all 52 data sets. All experiments were started with the same set of random seeds, i.e. with the same initial populations. The platform for experiments was a Pentium 200 MMX based IBM compatible PC, run in DOS 7.0. All algorithms are coded in Turbo Pascal for Dos 7.01. This programming language was chosen for its modularity and number crunching speed. At the time of starting this research, Turbo Pascal proved faster than the then current versions of Fortran and Delphi, which were the other alternatives available.

Note that in this chapter we do not look at the results to single data sets as we are searching for overall good strategies and parameter values. To make future reference easier, the following definitions apply for the measures used in the remainder of the thesis: 'Feasibility' refers to the mean probability of finding a feasible solution averaged over all data sets and runs. 'Cost' refers to the sum of the best feasible solution (or censored cost as described below) for all data sets averaged over the number of data sets with at least one feasible solution. Thus, cost measures the unmet requests of the nurses, i.e. the lower the cost the better the performance of the algorithm.

Should the algorithm fail to find a feasible solution for a particular data set over all twenty runs, a censored cost of 100 is assigned instead. The value of 100 was used as this is more than ten times the average optimal solution value and significantly larger than the optimal solution to any of the 52 data files. Thus, a particular algorithm that finds feasible solutions of 20 for 50 data sets and fails to solve the other two sets would have a cost of (20 * 50 + 2* 100) / 50 = 24.

The average of the best solution for each data set, rather than the average over all feasible solutions, was chosen to measure the performance of algorithms for the following reasons. As experiments showed, most of the time there was little difference which one was used for comparisons from the genetic algorithm's point of view. An example of this can be seen in Figure 4-1 in section 4.3.

However as the research progressed, situations where improved algorithms found more feasible solutions than before were encountered. Although, the best of these was as good or even better than before, the average was worse as some of the new feasible solutions were of below average quality. If the average over all feasible solutions was used in these cases, then it would be misleading. Note that using the average of the best of each run also allows for a more realistic comparison of our solutions to those of Dowsland [55], who also used the best result out of ten runs for each data set.

Furthermore, in a real-life situation we would always recommend executing the genetic algorithm several times over a specific data set. To allow for this the solution time is kept very fast by keeping time critical parameters small. It took on average less than ten seconds on a personal computer as specified above per optimisation run and data set. If for any reason multiple runs for a single data set are not wanted, we recommend compensating this by increasing the population size and extending the stopping criteria. This recommendation will become clear from the discussions in this chapter.

## 4.3 Parameter and Strategy Testing

### 4.3.1 General Introduction

The last step before running a canonical genetic algorithm is to set its various parameters and strategies. These are population size and type, initialisation method, selection strategy, crossover and mutation rate, crossover and mutation type, survival strategy and stopping criteria. The sheer number and range of these parameters show that this is by no means a trivial task. Moreover, as it has been mentioned earlier there is not one set of parameters that is superior for all types of problems (see Goldberg [81], Davis [48], Wolpert and Macready [181] and Radcliffe and Surry [130]).

Nevertheless, where appropriate, literature sources will be cited concerning the 'optimal' setting of parameters or the 'superiority' of certain strategies. For the purpose of this chapter, all parameters and strategies will be of a static nature, i.e. once chosen they will stay the same throughout the optimisation. The only exception of this will be experiments with a simple dynamic mutation rate. The topic of dynamic and truly adaptive parameters, as suggested for example by Yeralan and Lin [182], Tuson and Ross [166] and Tuson and Ross [167], will be further discussed in section 4.4 for the penalty weight and in section 7.5 for other parameters.

The experiments were conducted as follows. Initially, all parameters and strategies were set to the values found during earlier research based on a limited number of experiments and data sets (Aickelin [4]). These values can also be found in Table 4-1. Extensive experiments were then conducted for each parameter or strategy in turn. The parameter or strategy that gave the best results was then kept for all future experiments. Full results are not given in those cases where much worse results were encountered. Some of these can be found in Appendix D.

| Parameter / Strategy | Initial Setting |
|---|---|
| Population Size | **1000** |
| Population Type | **Generational** |
| Initialisation | **Random** |
| Selection | **Rank Based** |
| Uniform Crossover | **Non-parameterised** |
| Parents and Children per Crossover | **2** |
| Per String Mutation Probability | **5%** |
| Replacement Strategy | **Keep 20% Best** |
| Stopping Criteria | **No improvement for 20 generations** |
| Penalty Weight | **10** |

Table 4-1: Initial parameter settings for the direct genetic algorithm.

We are aware that the order of the experiments will influence the 'best' value for each parameter or strategy. Moreover, as we look at the best performance over all 52 data sets, we might not find the best parameters for all individual cases. Nevertheless, in view of the well-reported robustness of genetic algorithms regarding parameter settings, the parameters and strategies found this way will be amongst the best possible, without resorting to meta-level algorithms or other more complicated methods.

### 4.3.2 Population Type, Size and Initialisation

The first parameter tested was the population size. It is expected that the bigger the population the better the results found by the genetic algorithm because of the increased diversity and number of individuals processed. However, a bigger population also leads to a longer solution time due to the additional computation time per generation. Some theory about optimally setting the population size in the case of binary encodings can be found in Goldberg [82]. However, this is not directly relevant for our research, as we do not have the binary encodings assumed by Goldberg.

Before deciding whether to employ a generational or a steady-state population, some initial experiments with both types were carried out. An extensive discussion of this question can for example be found in De Jong and Sarma [50], Syswerda [160] and Goldberg and Deb [79]. In line with their findings, neither of the two methods proved consistently superior to the other. We decided to concentrate on one model to carry out all future experiments and the generational type population was chosen.

The individuals of the first population were seeded at random, in our case paying attention to assigning only feasible shift patterns to each nurse with respect to constraint set (2) of the integer program of section 2.1. However, Bramlette [79] suggests that by 'intelligently' seeding the initial population by choosing the best out of $n$ random individuals, better results can be obtained. We tried this for $n=5$, $n=10$ and $n=20$ and found that this in fact degraded solution feasibility and cost by up to 10%, producing worse results for higher $n$.

It is conjectured that this is due to the high number and tightness of the constraints in our problem. An observation of initial solutions showed that although their mean fitness value was almost twice as good as before, this was achieved by a strong bias towards cheaper shift patterns rather than by 'more feasible' solutions. This was because the seeding selection was based on the standard fitness function, which resulted in high cost, but potentially essential shift patterns, for example weekend and night patterns, being discriminated against. Thus in our case, it seems more important to have a good random spread rather than fitter but still infeasible solutions which lack certain 'unfavourable' values in the initial population. This is in contrast to Bramlette who solved unconstrained problems where a solution's fitness is a more direct representation of its quality.

This view is supported by results presented in Burke et al. [34]. The authors state that although seeding the population may result in better quality solutions by providing good starting points, diversity is also important. They conclude that if the seeding produces very similar individuals then the loss of genetic diversity might lead to worse final solutions in comparison to random initialisations.

Having decided on a random initialisation scheme, experiments regarding the best size of the population were carried out. Figure 4-1 shows how the average cost over all feasible solutions compares with the average cost over only the best feasible solution for each data set. As mentioned previously, both behave in a very similar manner and from now on, the term cost will only refer to the average cost of the best feasible solution for each data set. For a further explanation of these definitions refer to section 4.2.

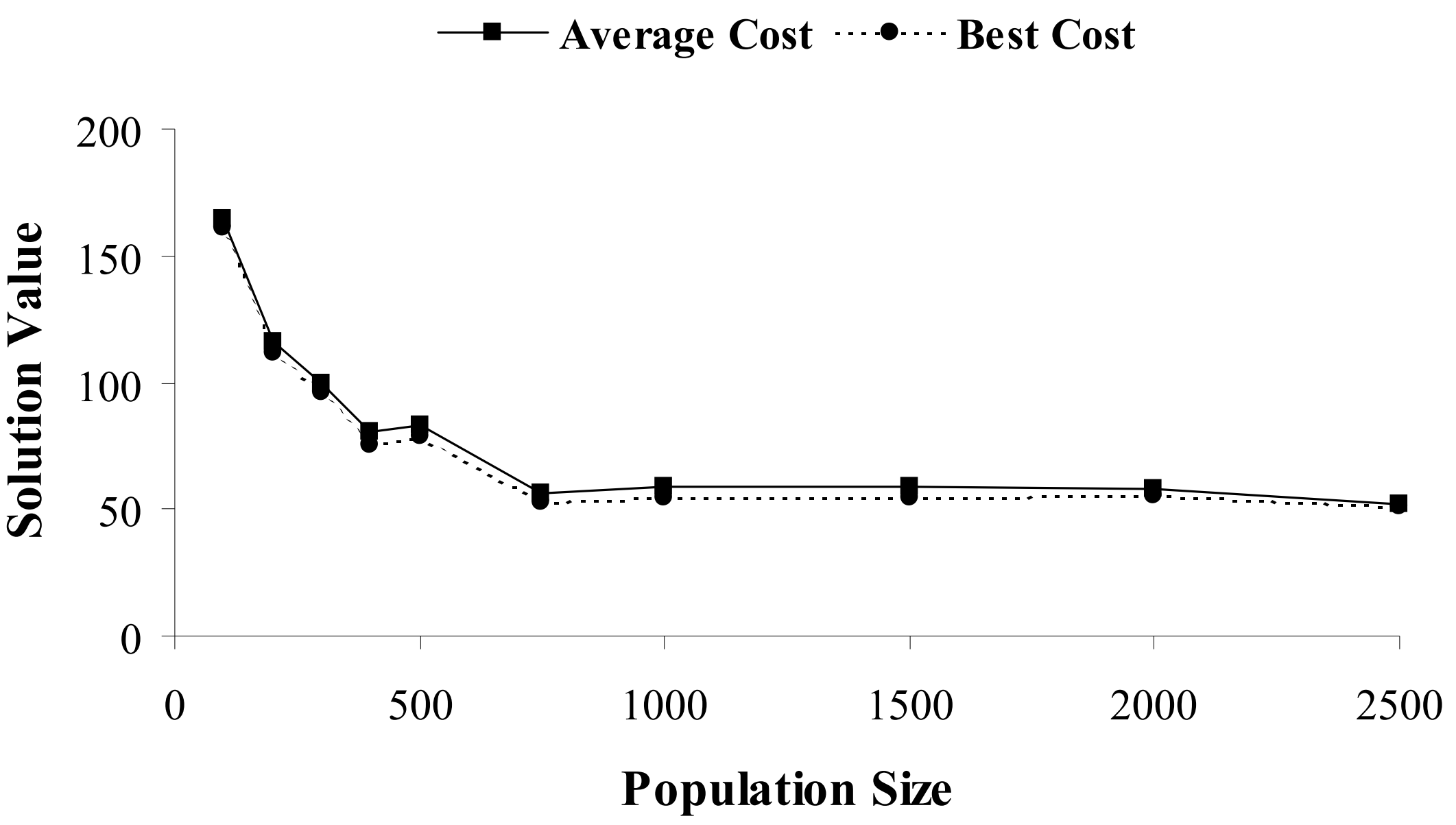

Figure 4-1: Population size versus average and best solution cost.

As seen from Figure 4-2, a bigger population leads to an increase in feasibility. However, the cost is only improved up to a population size of 750. Note that the choice of stopping criterion, i.e. no improvement for 20 generations, is the most likely reason for the improvement being capped here. If this was extended, larger populations would perform better because they had the necessary number of generations to converge further. For the current criterion, the chosen population size of 1000 is a good compromise between cost, feasibility and the necessary computation time. Figure 4-3 shows the average computation time for a single run of one data set. It grows almost linearly with the population size indicating an efficient algorithm. For a population size of 1000, an average solution time of less than 9 seconds is achieved.

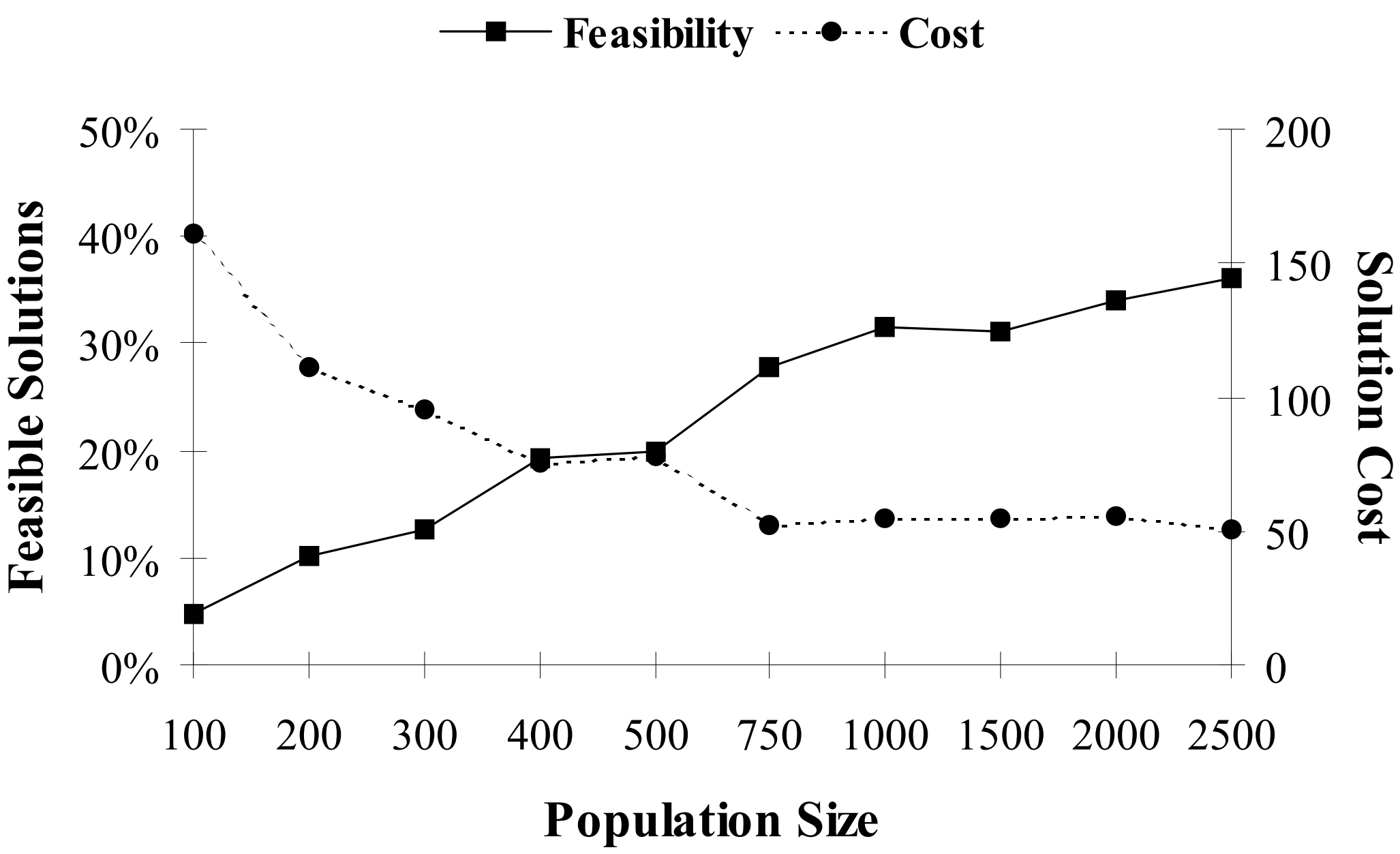

Figure 4-2: Population size versus feasibility and solution cost.

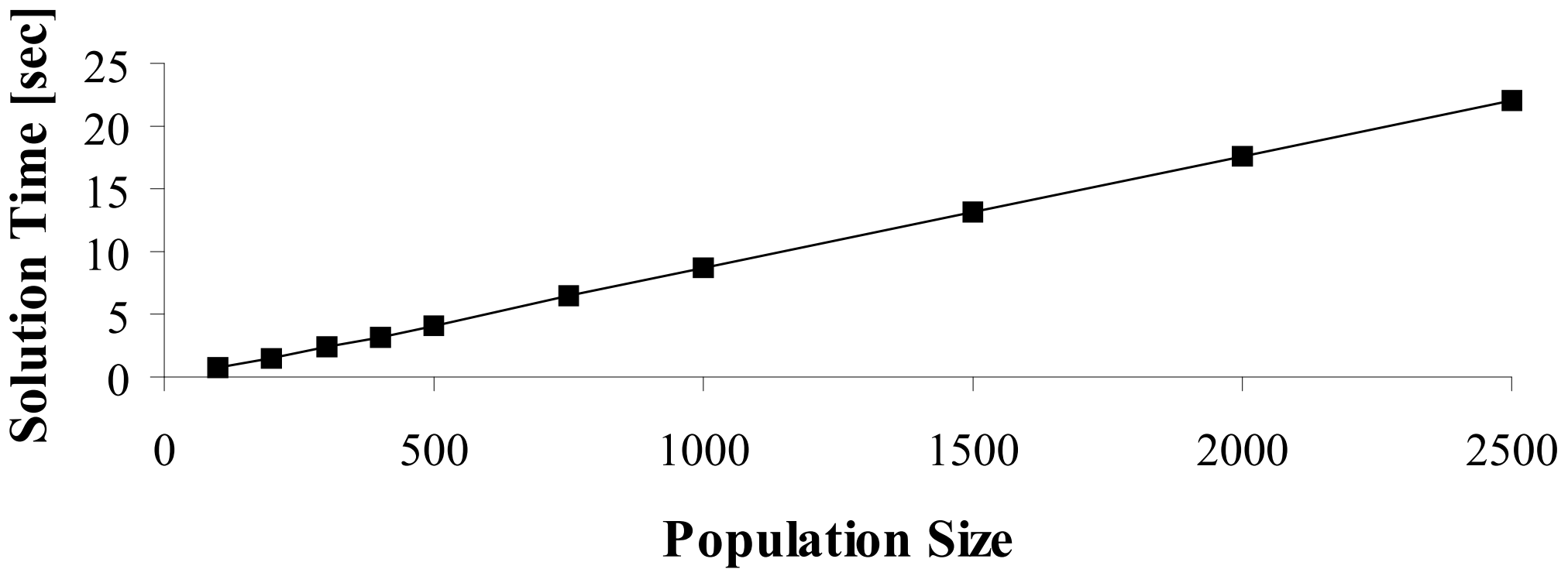

Figure 4-3: Population size versus solution time.

### 4.3.3 Penalty Weights

The second parameter to be tested is the penalty weight for uncovered demand, $w_{demand}$. Intuitively, one would think that the higher the penalty weight, the more likely it is that the solutions are feasible. Figure 4-4 shows the cost and feasibility for various penalty weights. However contrary to intuition, as shown in the figure, if the weight is set too high, feasibility is decreased drastically. A possible reason for this is the premature convergence to a low-cost and almost feasible solution not allowing enough time for a broader exploration of the search space. A weight in the region of 10 to 30 gives the best results. For all further experiments, a weight of 20 will be used.

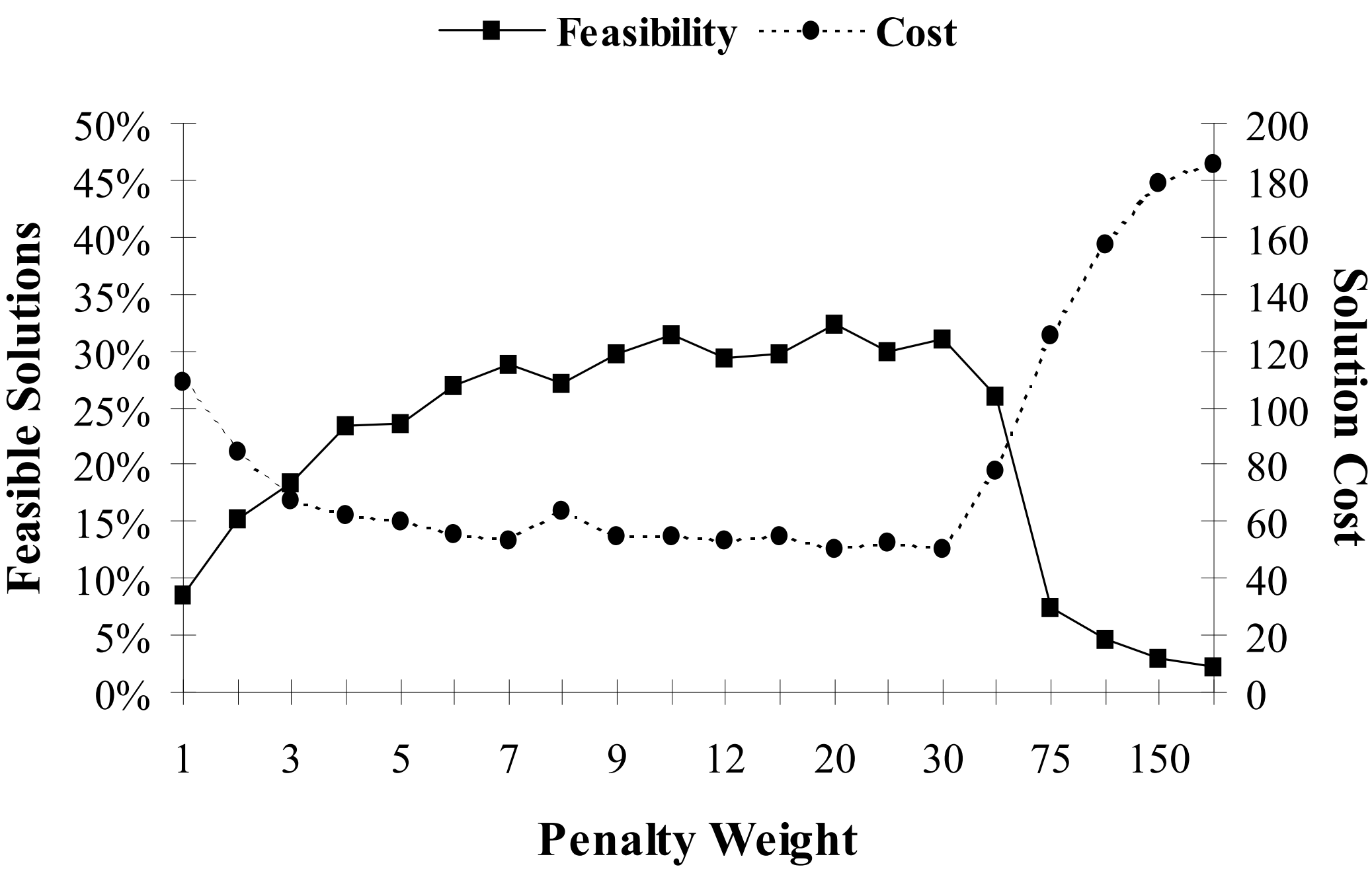

Figure 4-4: Penalty weight versus feasibility and solution cost.

In view of the results of Hadj-Alouane and Bean [87], concerning the theoretical impossibility of finding a set of perfect linear penalty weights, it is also worth noting that a variety of experiments with quadratic penalties failed to improve results. Details about Hadj-Alouane and Bean's results are described in chapter 3.4. The experiments using a quadratic penalty function were performed by squaring the undercover and using the same range of penalty weights as above. The results were worse than for the linear weight. Full results are pictured in appendix D.3. In hindsight, the likely reason for the failure of the quadratic penalties is that they are too high in the beginning and thus lead to premature convergence. For instance, for a typical number of uncovered shifts of 6 this results in a penalty of 36 times the penalty weight.

### 4.3.4 Selection

Following the genetic algorithm paradigm, the probability of selecting an individual for crossover is proportional to its fitness, i.e. the fitter an individual the more likely it is to be chosen. This is usually achieved with a roulette-wheel type selection (for further details see Goldberg [81]). However, if individuals are selected in proportion to their raw fitness (or in inverse proportion when minimising), then the problems of domination and lack of selective pressure can occur.

Domination can happen in the early stages of the search when following random initialisation there are a few individuals with much better fitness than the remainder of the population. Lack of selective pressure can happen towards the end of the search, when many individuals have similar fitness values. This could lead to a situation where the probability of better individuals being selected for crossover is almost the same as the probability for below average individuals. Further development would therefore be hampered.

To avoid these problems, some form of fitness scaling is necessary. Whitley [179] recommends rank-based selection. This avoids any 'ad hoc' scaling but still solves the problems of both domination and lack of selective pressure. Individuals are sorted according to their raw fitness, with the best receiving the highest rank equal to the number of individuals in the population. Selection is then made in proportion to these ranks, which gives the best individual an average of four children. For more details about this type of scaling and the average number of children produced by certain individuals see the examples in appendix A.2.

### 4.3.5 Crossover

In the preliminary study [4], only one-point and standard uniform crossover, as originally presented by Syswerda [162], were compared, with uniform crossover giving better results. Standard uniform crossover is defined as uniform crossover with $p = 50\%$ for each gene coming from either parent. In this more extensive comparison, two-point crossover and parameterised uniform crossover with various parameters are also included. Parameterised uniform crossover is explained in more detail in Spears and De Jong [154]. It works as standard uniform crossover, but $p$, representing the probability of genes coming form the first parent, can range from 0.5 to 1.

Figure 4-5 shows the results of this study of various crossover operators on the nurse scheduling problem. As has been observed by many other researchers tackling different problems (for instance Booker [26], Schaffer et al. [146] and others), two-point crossover (label '2p') gave superior results to one-point crossover (label '1p'). Two-point crossover also resulted in lower cost and roughly the same feasibility as standard uniform crossover, i.e. uniform crossover with $p = 50\%$.

However, the real winner is parameterised uniform crossover with $p = 80\%$. In the graph, the x-axis label indicates the value for $p$ as a percentage. Parameterised uniform crossover with this value is less disruptive to the strings than standard uniform crossover, whilst at the same time offering greater variety of possible children than two-point crossover. In view of the results in Figure 4-5, this combination seems most suited for our problem and hence this operator will be used for all future experiments.

Usually only two parents are used for crossover. Eiben et al. [59] conduct a study into using more than two parents for crossover operators. Their experiments show that for some problems using more parents per crossover can be beneficial. They conclude that it is worthwhile using up to four parents, depending on the problem at hand. We also tried this with $p = 80\%$ uniform crossover and two, three and four parents. Note that in the case of three and four parents, there is an 80% probability of a gene to come from the first parent and a 20% probability to come at random from any other parent.

And indeed four parents (label '4 par') gave slightly better results than two parents with both cost and feasibility being improved by some 5%. This could be due to two factors: The increased flexibility, as described by Eiben et al., or the relatively higher number of samplings better individuals received due to our rank-based selection. The best individual now participates in on average 8 crossovers (formerly 4), whilst an average individual takes part in on average 4 crossovers (formerly 2). However, one also has to take into account that for more than two parents each parent contributes fewer genes than before. For all further experiments, parameterised uniform crossover with $p = 80\%$ and four parents will be used.

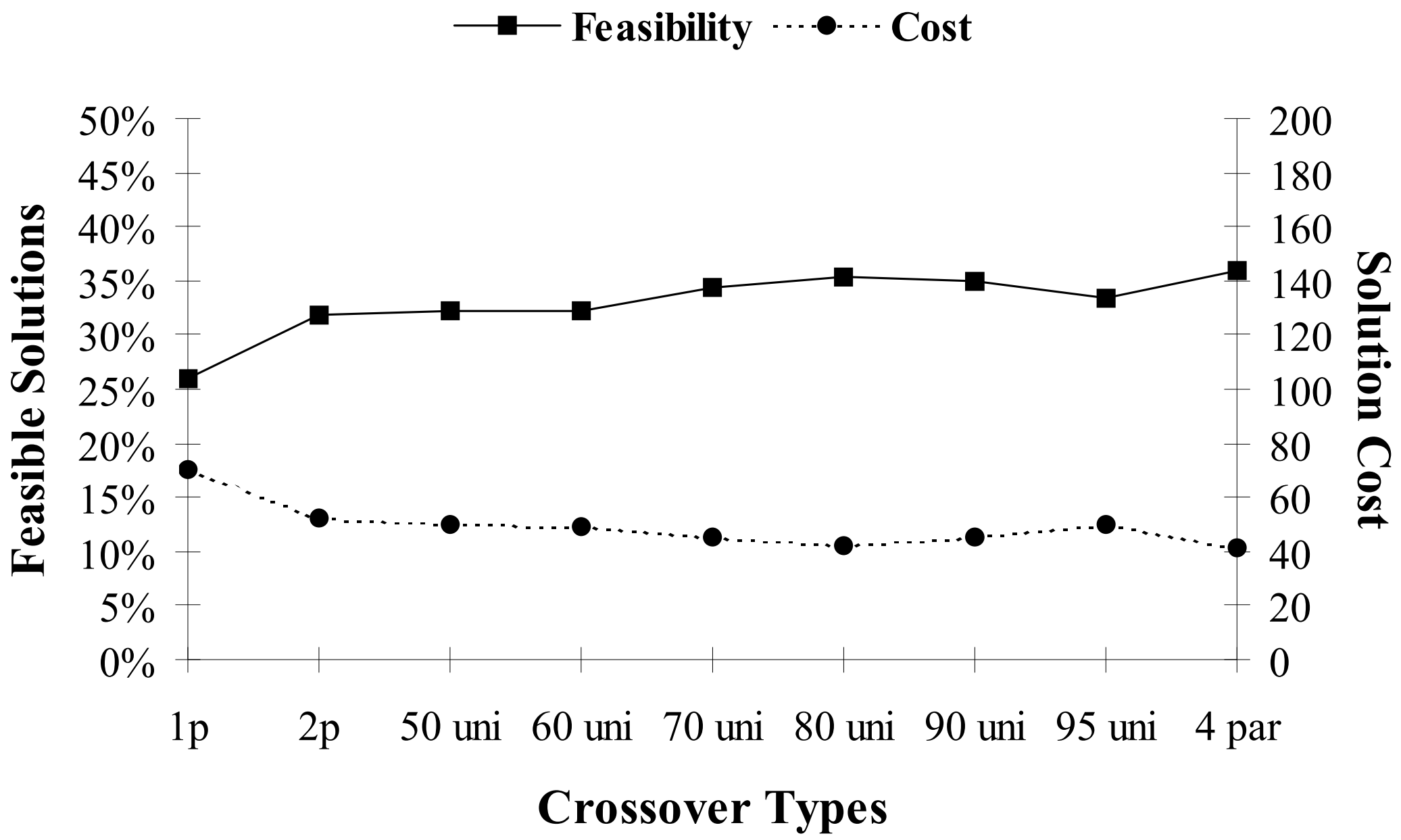

Figure 4-5: Performance of different types of crossover strategies.

### 4.3.6 Mutation

In the pilot study of Aickelin [4], a per individual mutation rate of 5% was used. If an individual was chosen for mutation, exactly one of the genes was changed at random. In contrast to this, all the genetic algorithms used in this research have a single bit mutation probability. This gives a higher flexibility, as there is a chance for more than one gene to be mutated per string. Bäck [8] suggests a single bit mutation probability of [1 / length of string], which in our case would be 3% to 4%. In view of the results of the experiments, Bäck's suggestion is a good guess. Figure 4-5 shows variations of the single bit mutation rate. Results are as expected for a genetic algorithm, with good values for under 5% single bit mutation rate. Henceforth, a single bit mutation rate of 1.5% will be used. This value, at the lower end of the good range, was chosen as it requires less mutation operations and hence is computationally faster.

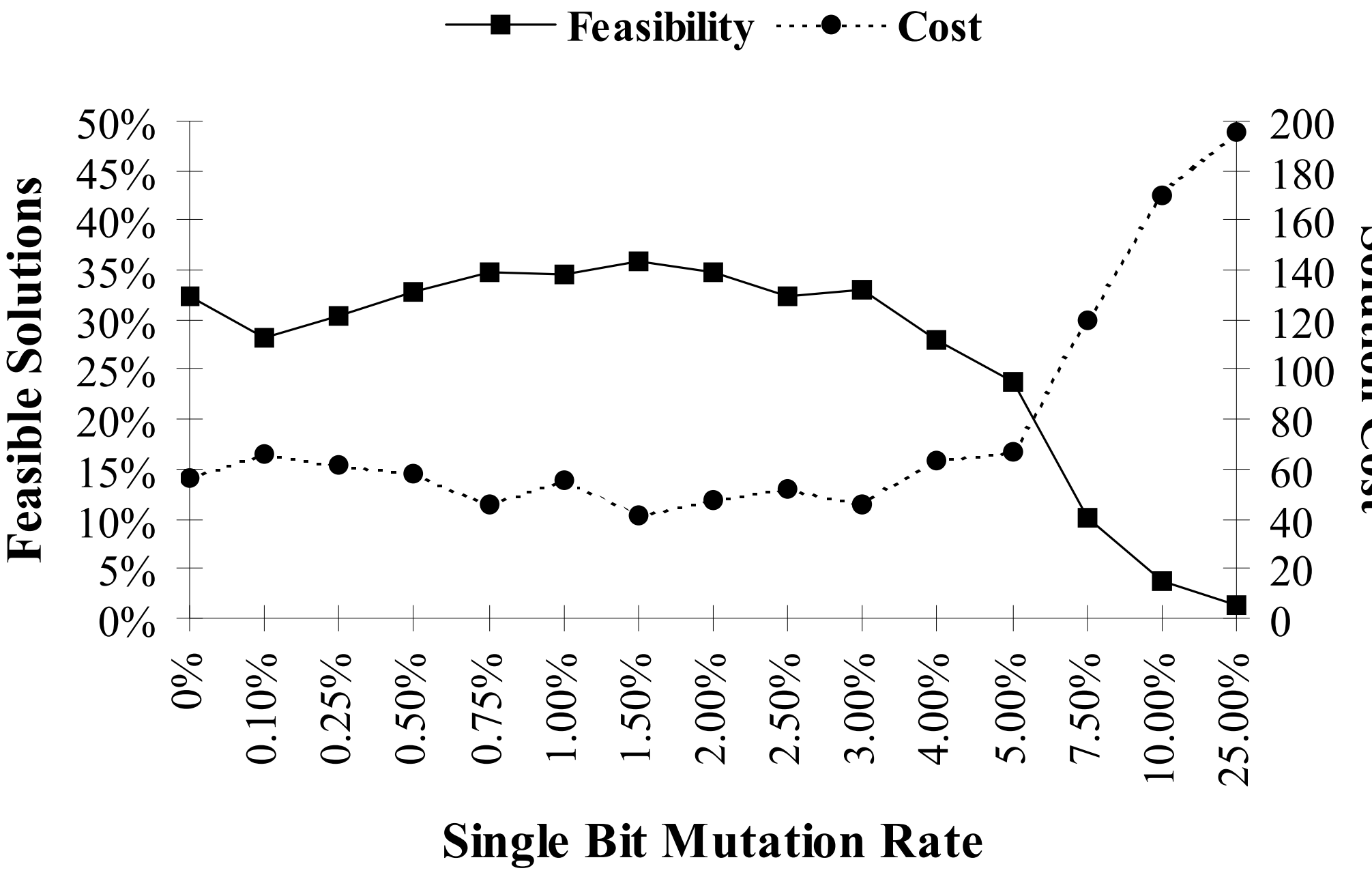

Figure 4-6: Varying the mutation rate versus feasibility and solution cost.

An alternative to mutation, as suggested by Ghannadian [75], was also tried. Rather than mutating a chosen solution, a completely new individual was introduced into the population as a substitute. However, this strategy produced worse results than using 1.5% single bit mutation and hence was not pursued further.

Fogarty [67] uses adaptive mutation rates. The general argument runs that in the early stages of the search, mutation is less important because there is still a good variety of solutions and big changes are needed which are more efficiently carried out with crossover. However, in the later stages of the search mutation might not only re-introduce lost values, but also help fight premature convergence by enabling the genetic algorithm to escape local optima. Hence, it is suggested that mutation be kept low in the beginning and made more prominent in the later stages of the search.

This idea was implemented following two different strategies: Both increased the mutation rate with every generation passed by multiplying it by a factor of 1.1. The difference between the two methods was whether or not the mutation rate was reset to its starting value (0.1%) if a better solution than the currently best was found. Neither of these strategies produced better results than a standard fixed mutation rate. This could be due to two reasons: Either our function to increase the mutation rate was not suitable or the idea does not work with our problem.

An example in support of the second argument is the problem of steel truss design studied by Lee and Takagi [107]. Lee and Takagi experimented with a genetic algorithm that had a dynamic population size and dynamic crossover and mutation rates but was otherwise canonical. The authors found that in their particular problem out of all parameter variation strategies, exponentially *decreasing* the mutation rate improved their results the most.

### 4.3.7 Replacement

So far, an elitist strategy with keeping the best 20% in each generation had been used as the replacement policy. Further experiments were carried out allowing different proportions to survive. Additionally, experiments with a tournament approach to survival were carried out. The results of these parameter tests are shown in Figure 4-7, where the label indicates the percentage of best solutions kept.

One can see that keeping the best 10% is a better strategy than keeping the best 20%. For the experiments shown in the graph, duplicates were eliminated from the population. This has been suggested, for example, by Ronald [140] and leads to slightly better results than keeping the duplicates. A tournament approach to survival (label 'tourn'), in which one parent competes with one child for the place in the next generation, gives good results. However, it is inferior to an elitist strategy keeping the best 10%, which is the one used hereafter.

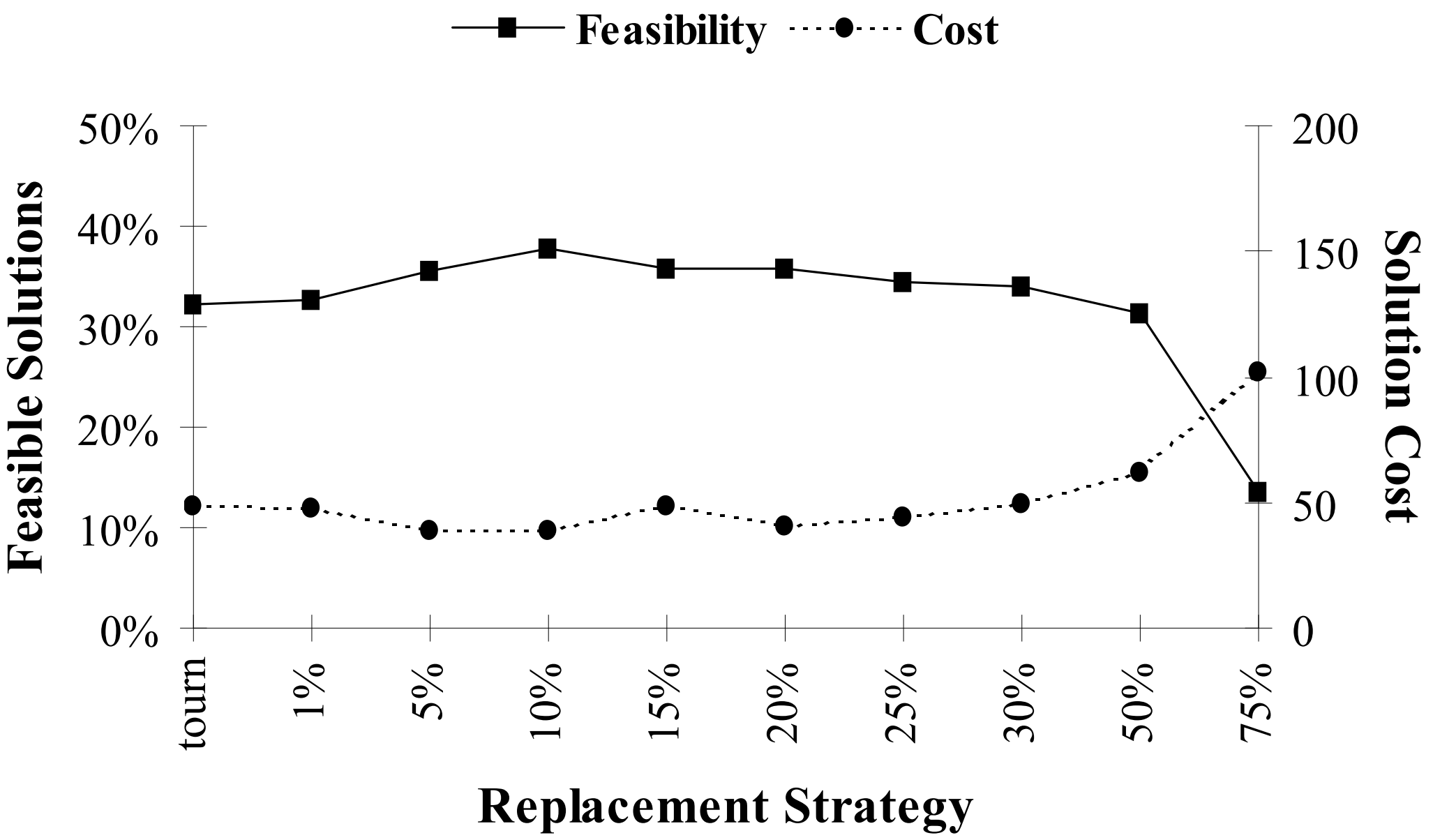

Figure 4-7: Comparison of different replacement strategies.

### 4.3.8 Stopping Criteria

The stopping criterion was kept as the number of generations without an improvement of the best solution found so far. Other criteria are suggested in the literature, for example Kim et al. [102] use a convergence criterion based on a percentage of all strings with the same value for certain genes. However, from a practical optimisation point of view, where solution speed is important, our criterion is simpler and hence computationally faster. Furthermore, it is anticipated that no improvement for a large enough number of generations is equivalent to the convergence of the genetic algorithm. Unfortunately, in circumstances where an almost optimal solution is present in the starting population, this might not hold. However, judging by the difficulty to create good solutions by random initialisation, this is a very unlikely situation.

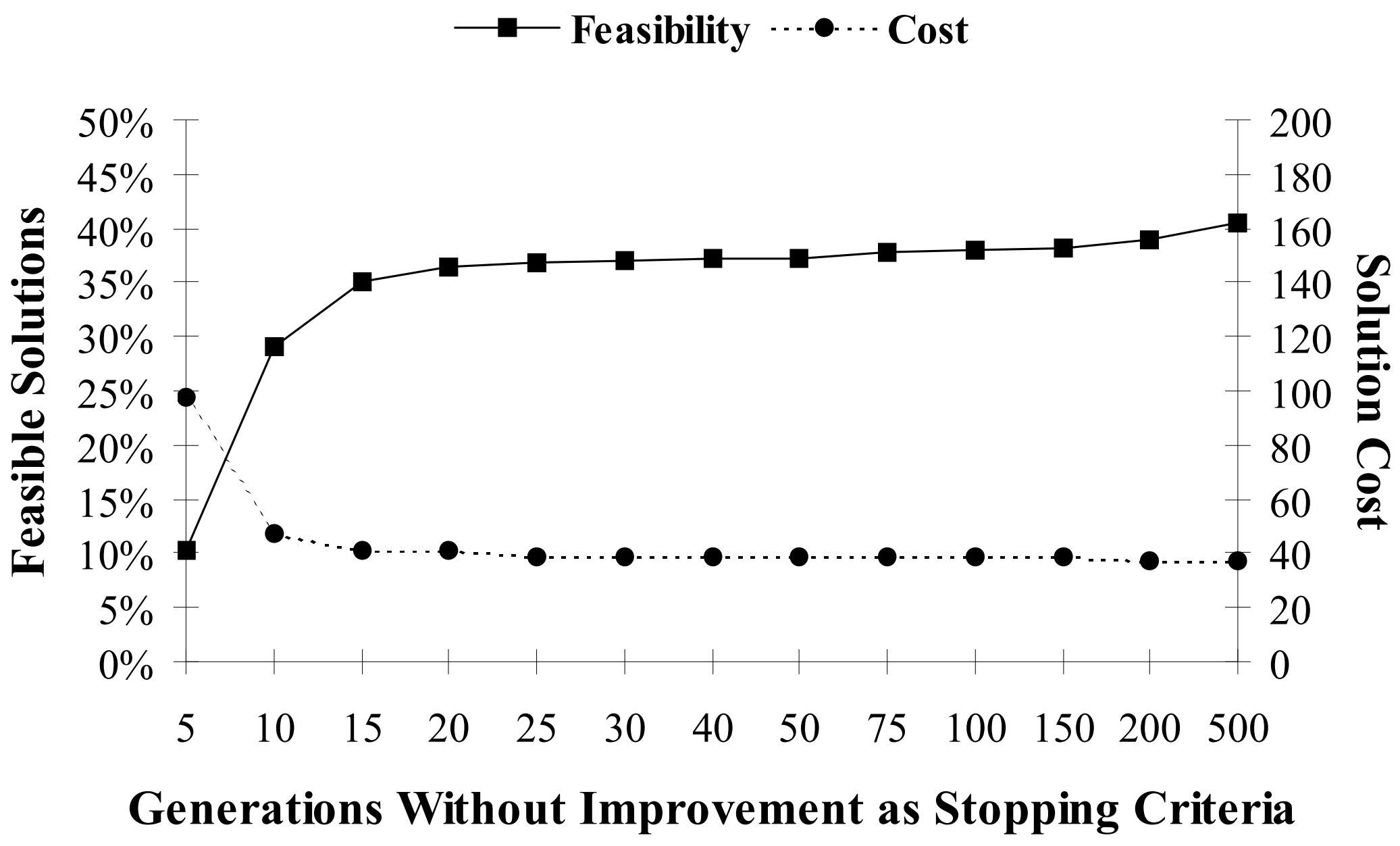

Figure 4-8: Stopping criteria versus feasibility and solution cost.

Experimenting with different values for the number of generations without improvement leads to a trade-off between solution time and solution quality. Full results are shown in Figure 4-8 and Figure 4-9. A good trade-off is achieved for a value of 25 to 30 generations without further improvement. Thus, 30 generations without improvement was the chosen stopping criterion for future experiments.

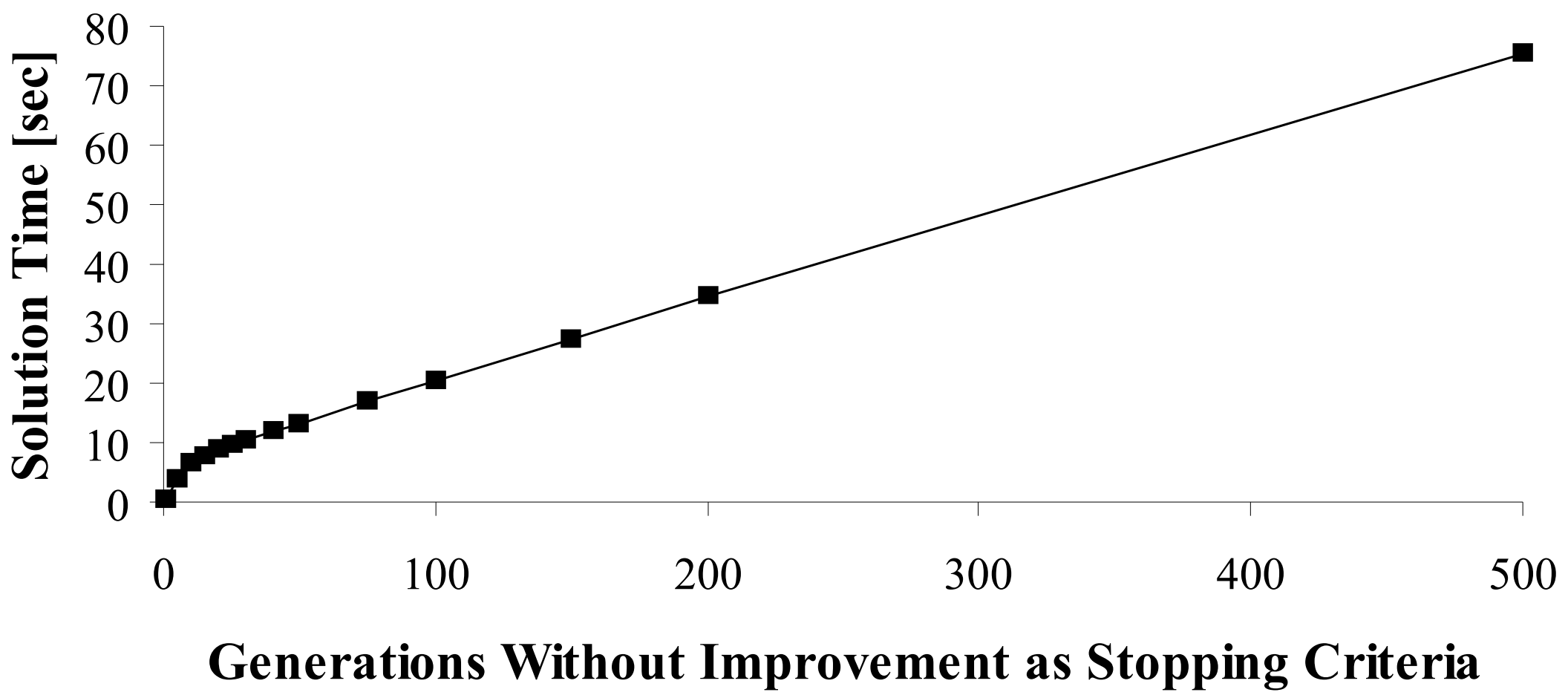

Figure 4-9: Stopping criteria versus solution time.

### 4.3.9 Summary of Parameter Tests

Table 4-2 shows a final summary of the values and parameters identified to work best with the nurse scheduling problem and used in the remainder of this thesis unless otherwise stated. Overall, feasibility was improved from 31.5% to 37.1% and cost was down from 54.8 to 38.9. The biggest improvements were made using parameterised uniform crossover and increasing in the number of generations without improvement for the stopping criterion. Whilst the latter is trivial, the former is more interesting and will be looked at more closely in section 5.2 when a new type of crossover is introduced. Most disappointing was the result of the penalty weight tuning. No major improvement was achieved. However, it was conjectured that more could be done about the penalty weights and the next section examines dynamic penalty weights.

| Parameter / Strategy | Optimised Setting |
|---|---|
| Population Size | **1000** |
| Population Type | **Generational** |
| Initialisation | **Random** |
| Selection | **Rank Based** |
| Uniform Crossover | **Parameterised with *p = 0.8*** |
| Parents and Children per Crossover | **4** |
| Per Bit Mutation Probability | **1.5%** |
| Replacement Strategy | **Keep 10% Best** |
| Stopping Criteria | **No improvement for 30 generations** |
| Penalty Weight | **20** |

Table 4-2: Final parameter values and strategies for the direct genetic algorithm.

## 4.4 Dynamic Penalty Weights

Clearly, the results found so far by the genetic algorithm are very poor. Particularly disappointing, with regards to increasing the feasibility of solutions, were the tests on varying the penalty parameter. Even with an 'optimal' penalty weight, more than two thirds of all solutions were infeasible. During the pilot study (Aickelin [4]) it was found that dynamic penalty weights, adjusting with the number of generations passed, improved results.

However, it was noted that some dynamic schemes proved better for certain data sets than others did. To overcome this problem of data specific performance, a more sophisticated dynamic penalty weight strategy is presented in this section. The penalty weight used here is truly dynamic or adaptive, as it adjusts itself in proportion to the actual development of the current population.

The use of such adaptive weights is not uncommon in the genetic algorithm literature. As pointed out by Reeves [135], several researchers have found that the use of adaptive or truly dynamic penalty weights can overcome the problems observed with fixed weights. Hadj-Alouane and Bean [87], who solve a general multiple-choice assignment problem, start with a small penalty weight and increase it when the best solution has been infeasible for several generations. They decrease it again when the best solution over several generations has been feasible. Thus, the effect is one of oscillating around the feasibility boundary. This effect is also referred to as strategic oscillation, a term derived from tabu search and described fully in Glover et al. [78].

Smith and Tate [152] suggest a penalty weight that is scaled according to the difference between the fitness of the best feasible solution and the overall fittest solution found so far. More precisely they use the following formula to determine the penalty $p_i$ of solution $i$, where $n_i$ is the number of constraints violated by solution $i$, $V_{feas}$ and $V_{all}$ are the values of the best feasible and best solution found so far and $k$ is a severity parameter:

$$p_i = \left(\frac{n_i}{2}\right)^k \left(V_{feas} - V_{all}\right)$$

The authors chose not to base their penalty on the degree of violation as suggested by Richardson et al. [138]. They argue that in their particular example, an area facility layout problem, the total amount of infeasibility is less important than the number of constraints violated. After extensive tests, Smith et al. conclude that the exact value of the severity parameter $k$ is not important to achieve significant improvements in comparison to a fixed penalty weight approach.

In the following, experiments with four different adaptive penalty weight strategies are carried out. In addition to one similar to the weight presented by Hadj-Alouane and Bean [87] (referred to as Hadj), a variant of Smith and Tate [152] (called Smith) is tested. Furthermore, a 'reverse' of the function used by Hadj-Alouane and Bean and a dual weight approach are tried. The variation from the method of Smith and Tate is that we measure the amount of constraint violation as opposed to the number of violated constraints.

This was done in agreement with the view of Richardson et al. [138] that this is a better measure of infeasibility. For instance, if a solution violates two constraints by four shifts each then this is far worse than three constraints being violated by one shift only. In the reverse method, rather than starting with a low weight and then increasing it until a feasible solution has been found (Hadj weight), we start high and decrease the weight in line with increased feasibility.

We conjectured that this 'reverse' approach is more suitable to our tight problem. Hence, the penalty weight is high at first, forcing the genetic algorithm into regions more likely to contain feasible solutions. When solutions improve, the weight is lowered, allowing for a full exploration of these areas. The fourth approach is a simplified version of the third method, called dual weight approach in the following. Whilst the best solution found so far is still infeasible a high penalty weight is used. Once that solution becomes feasible, the weight is dropped to a low value.

As Smith and Tate concluded that the precise setting of the severity parameter is unimportant, we follow their example and use $k = 1$ in the case of the Smith weight. Hence, the penalty weight becomes

$$w_{demand} = \begin{cases} \frac{(V_{feas} - V_{all})}{2} & for\ V_{feas} > V_{all} \\ v & else \end{cases}$$

This weight is then multiplied by the sum of all constraint violations, as described in section 4.1, and this penalty is then added to the objective function value. Should $w_{demand}$ become zero, it will be set equal to a small number $v$. This was done after observing problems in the behaviour of the genetic algorithm for a weight of zero. Trying various values for $v$ produced little difference in results, hence for our experiments $v = 5$ was used because $V_{feas}$ was usually at least ten greater than $V_{all}$.

In the case of the Hadj style weight and its 'reverse' the following formula to update the penalty weight is used:

$$w_{demand} = \begin{cases} \alpha \cdot q & for\ q > 0 \\ v & else \end{cases}$$

For the reverse type weight, $q$ is the sum of constraint violations by the best solution found so far. For the original style weight, $q$ is set to ten minus the sum of constraint violations by the best solution. Ten was chosen as it was larger than the average maximum violation score observed by the best solution of the first generation throughout a number of experiments. In other words, $q$ is usually in the range [0…10]. To adjust severity, $\alpha$ is used as a pre-set severity parameter.

As with the Smith weight, the penalty weight is set to a small number $v$ should the best solution be feasible and hence $q = 0$. Again, this was done to avoid problems with a zero penalty weight. Furthermore, this follows the definition of the Hadj weight, which decreases once a feasible solution has been found. In our experiments $v = \alpha / 2$ is used as this is roughly similar to the above for the tested range of $\alpha$.

For the computational experiments, the penalty weight was updated once every generation for all methods. Figure 4-10 shows the results for a fixed penalty weight, dual penalty, Smith style penalty and reverse style penalty for various $\alpha$. Note that for all severity parameters $\alpha$ the results for the Hadj type weight were far worse than for the fixed penalty. Hence, no results for this penalty weight are given.

Various combinations for the two values of the dual penalty approach were tried, with the combination 50 and 5 giving best results. These are pictured under the ‘hi/lo’ label and are worse than the results for the fixed weight. The best results for the ‘reverse’ approach were achieved for $\alpha = 8$. They are a good improvement over the fixed penalty weight results and notably the number of feasible solutions had been raised by almost a quarter. The result for the variant of the Smith and Tate type weight was also an improvement over the fixed penalty and gave the overall best result. Hence, for all future experiments the variant of the Smith and Tate dynamic penalty weight will be used.

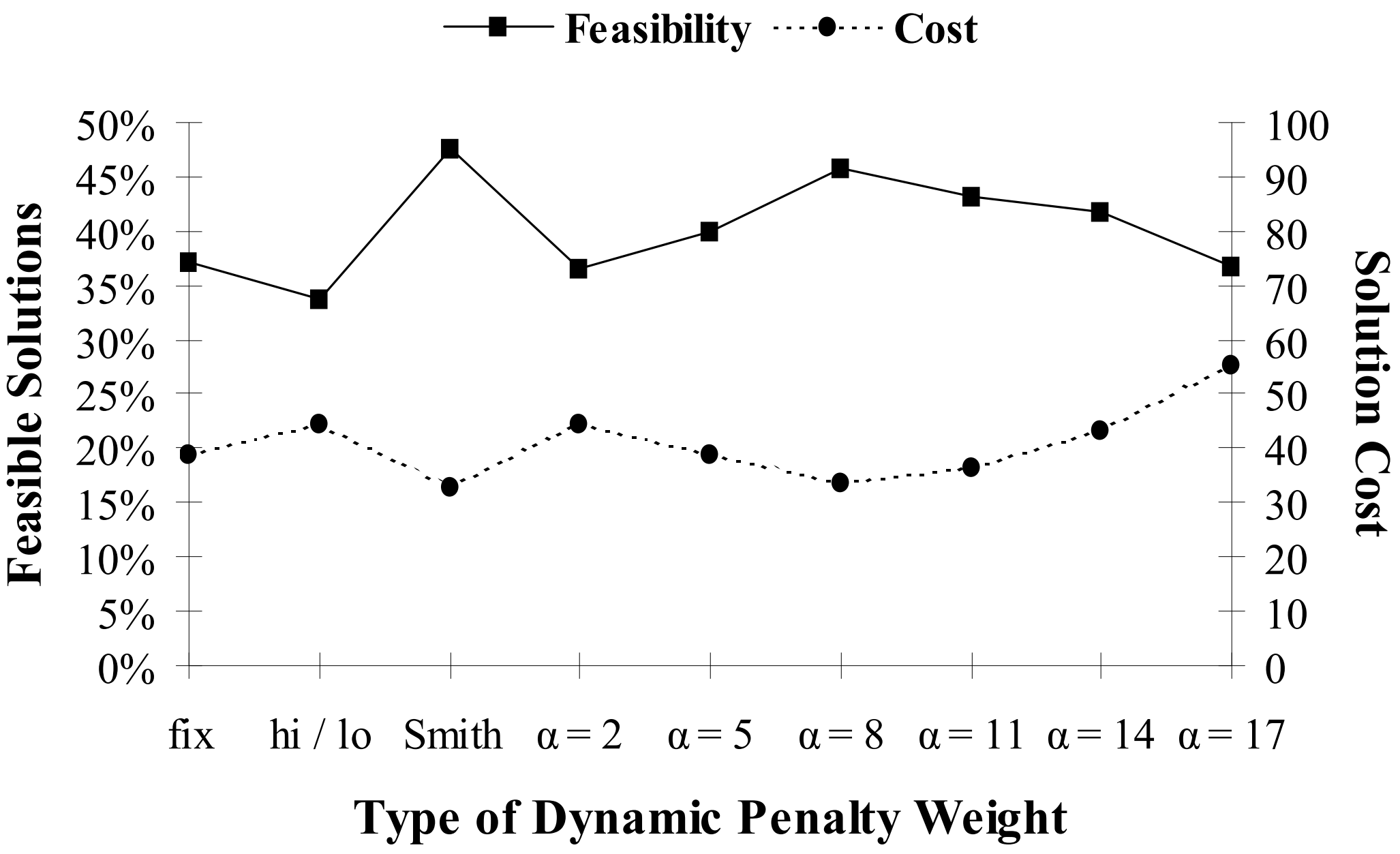

Figure 4-10: Comparison of various types of dynamic penalty weight strategies.

Figure 4-11 shows a closer comparison of how the penalty weights actually developed under the three strategies to try to explain the success or respectively failure of the methods. The experiment was made for a data set of average difficulty with $\alpha = 10$ and $v = 5$. Predictably, the Hadj style weight starts low and grows bigger with the other two methods starting high and gradually getting lower. The Hadj type weight fails to find a feasible solution at all, which would be indicated by the penalty weight reaching 90 and then dropping to 5. This is not surprising, as it is in line with our findings, that using high penalty weights hardly finds feasible solutions, which is demonstrated in Figure 4-4.

Both the variant of Smith and Tate and the ‘reverse’ method behave similarly, with some subtle differences. The ‘reverse’ method is rigid and hence occasionally tends to oscillate between two values. This happens in a situation where under a higher weight a particular solution with less constraint violations is better than a fitter solution with more constraint violations. Thus, the weight is decreased. Under the new weight, the situation is reversed and the weight is increased again. The variant of the Smith and Tate method makes more subtle changes and therefore avoids this cycling.

Another advantage of the Smith weight is that if no new and better feasible solution is found, the penalty weight gradually increases, as does the difference between the overall best and best feasible solution. Hence, more and more pressure is put on the algorithm to improve feasibility, without using too high weights as the Hadj style strategy does. This results in the typical behaviour shown in the graph: The weight slowly edges up when there is no better feasible solution and then suddenly drops when a new improved feasible solution has been found. Therefore, its better results are attributed to this overall greater responsiveness.

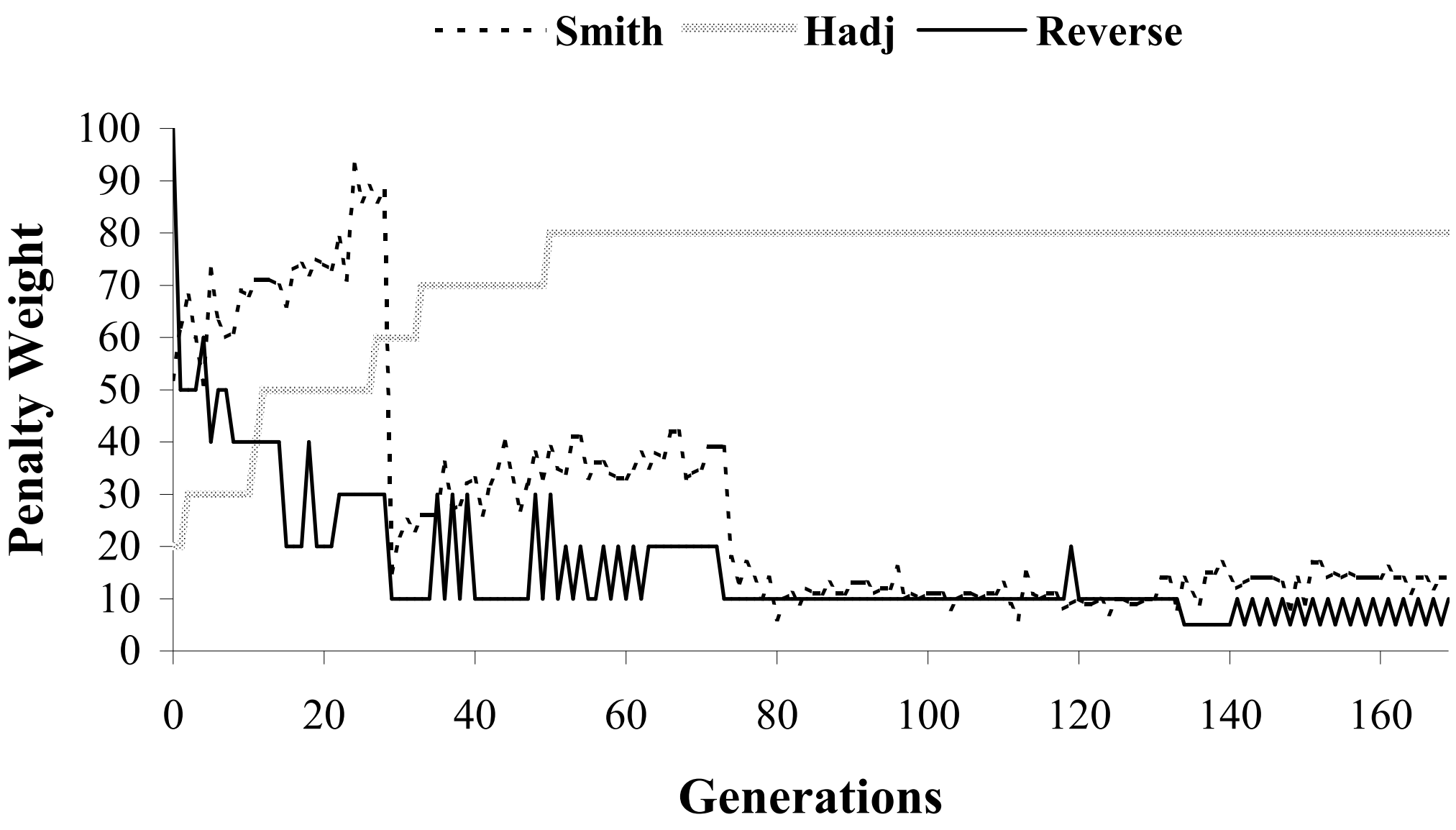

Figure 4-11: Development of dynamic penalty weights under three strategies.

## 4.5 Conclusions

Overall, the simple genetic algorithm fails to solve the nurse scheduling problem adequately, even with optimised parameters and sophisticated dynamic penalties.

Figure 4-12 gives a summary of results found so far. The graph shows the results for the genetic algorithm with (label 'Optimised') and without (label 'Basic') parameter optimisation and compares this to the dynamic penalty approach (label 'Dynamic'). For further comparison, a summary of the results found by tabu search (label 'Tabu') is also shown.

Although improvements were made both for feasibility and for cost of solutions in comparison to the basic approach, results are still far worse than those found for example by tabu search. Section 5.1 will examine the reasons for this failure, namely the breakdown of the building block hypothesis due to epistasis caused by the inclusion of the constraints via a penalty function approach.

The experiments in the area of penalty weights have been exhaustive and no further improvement is expected in this field. However, many of the other ideas presented in chapter 3 have not been implemented so far. These include using repair operators and other local hillclimbers, having more than one population in parallel, employing problem-specific crossover and other ways of including problem-specific information. The following chapter will look at these avenues.

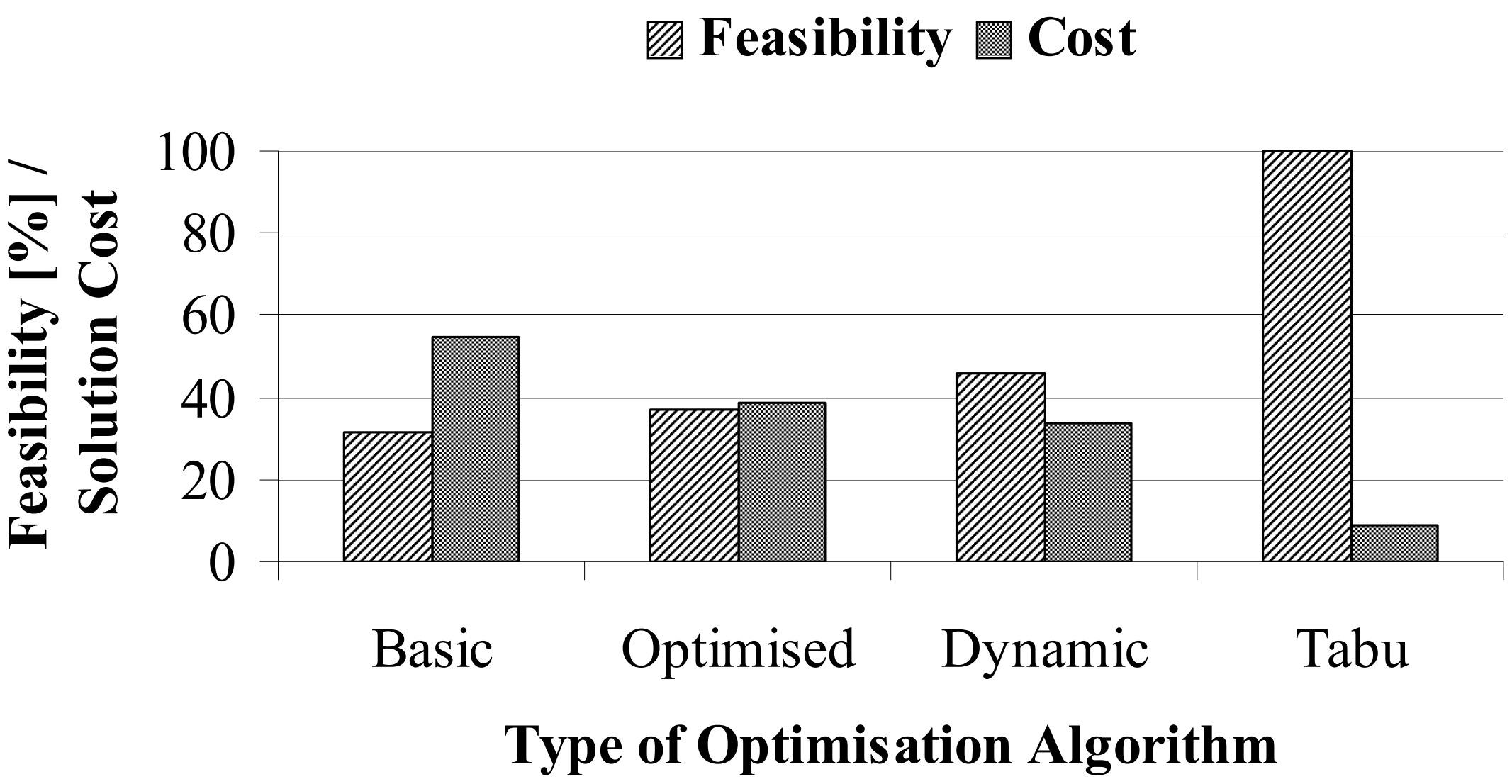

Figure 4-12: Comparison of simple direct genetic algorithms with tabu search for nurse scheduling.

# 5 An Enhanced Direct Genetic Algorithm Approach for Nurse Scheduling

## 5.1 Epistasis or why has the Genetic Algorithm failed so far?

The success of a genetic algorithm is usually attributed to the validity of the Building Block Hypothesis, which is fully explained in appendix A.3. In brief, the hypothesis stipulates that genetic algorithms rely on the crossover operator being able to combine good partial solutions, so-called building blocks, into complete good solutions. However, it is well recognised (see for example Davidor [44]) that for problems with a high degree of epistasis this is not guaranteed. Essentially, epistasis is present if the total fitness of an individual is not a linear combination of the fitness of its elements. This happens, for instance, if the contribution to fitness of single elements in the solution string depends on the values of other elements. Hence, epistasis is a measure of the 'non-linearity' of the relationship between the encoding and the fitness function.

For illustration purposes, consider the following example of binary strings of length 5, where the fitness is simply the sum of the genes. Two good building blocks are (###11) and (101##) where a # stands for either a 0 or 1. If these two are combined to (10111), a highly fit solution is formed. Because of the 'linear' relationship between encoding and fitness function, no epistasis is present and the genetic algorithm can successfully combine good building blocks to highly fit full solutions.

On the other hand consider the same problem, but with the additional constraints that no more than two adjacent genes can take a value of 1. Although the quality of the building blocks is still high, their combination is now infeasible. However, had the first building block contained the same number of ones but as (110##) then the combination would have been successful again. The genetic algorithm will find this new problem harder to solve, due to the epistasis created by the additional constraint.

To arrive at a better understanding of epistasis, Davidor [44] suggests measuring it as the difference between the sum of the fitness scores of the single genes and the total fitness of the string. He defines the fitness score of a single gene by averaging the fitness of all individuals containing this particular gene and comparing this to the average fitness of the whole population.

Reeves and Wright [134] note that Davidor's definition of epistasis is flawed because positive and negative interactions within strings are allowed to cancel each other out. Reeves and Wright state that more information is needed, for instance the sign and magnitude of each interaction, which in practice is hardly obtainable. However, they also conclude that Davidor's measure and idea of epistasis still gives some guidance as to the likely difficulty of a problem with regards to its degree of epistasis.

Obviously, the quality of such a measure also depends on the size of the alphabet. In the binary case, on average half the population has the same value for a particular gene. This allows for statistically sound results for the mean fitness score for each single gene, as the sample size is large. For our nurse scheduling encoding the size of the alphabet is the number of possible shift patterns for a particular nurse. As explained previously, this can be as high as 75 and averages at around 40 depending on the data. Consequently, on average only 2.5% of the population carry the same value for a particular gene, which makes for a sample size that is too small to perform a meaningful analysis. Therefore, we were not able to use Davidor's epistasis measure.

However, we can say that it is the penalty function approach that renders the nurse scheduling problem with its particular encoding highly epistatic. If the fitness was just the sum of the nurses' $p_{ij}$ values, then no epistasis would be present. In fact, the epistasis created by the covering constraints is twofold. The main effect derives from the fact that the contribution to cover by a nurse working a particular shift depends on the patterns worked by the other nurses. However, because higher graded nurses are allowed to cover for nurses of lower grades there is a second dimension to this interdependency. It is this second aspect of epistasis that the following modifications are intended to reduce.

Our approach follows an idea by Beasley et al. [16] and Beasley et al. [17] who suggest that epistasis can be reduced by decomposing an epistatic problem into several sub-problems. They then overlay these sub-solutions within an individual and eventually merge them to form a normal solution. This seems counter-intuitive at first as it makes the search space larger. Nevertheless, the problem is made simpler because the epistasis between the elements, i.e. sub-problems rather than single genes, is reduced. Beasley et al. find good solutions with this method for a circuit layout problem. However, for a second problem, a Walsh transform computation, the technique is less successful. The authors suspect this to be due to the merging algorithm becoming too complex making an exhaustive search as used for the first problem infeasible.

In this thesis, a similar approach is taken, but instead of overlaying sub-solutions, sub-populations are built. If the problem could be decomposed into three separate problems, one for each grade, then this type of epistasis would disappear. However, because of the interdependencies such a straightforward tactic is not possible. Instead, the ideas underpinning a parallel genetic algorithm (see appendix A.4) are adopted and we attempt to breed sub-populations that are highly fit with respect to nurses within specific grades. These are then combined strategically using a special fixed-point crossover operator.

This approach was motivated by an observation made during parameter testing. There it was noted that one-point crossover sometimes gave better results than uniform crossover. This led us to conjecture that in some cases combining large building blocks, made up of good partial solutions for one or more grades, aided the solution process. This hypothesis was further supported by the improvements in the results obtained by parameterised uniform crossover for parameters significantly larger than 0.5 (see Figure 4-5 for details). The following summarises this new approach which will be detailed in the next section:

1. Sort the solution strings according to nurses' grades (from high to low).
2. Introduce a new type of crossover: a fixed-point crossover on grade-boundaries.
3. Split the population into several sub-populations based on grades.

4. Introduce new (sub-)fitness functions based on a pseudo measure of under-covering for each grade.
5. Produce some of the children by applying the crossover operator to individuals from complementary sub-populations.
6. Allow individuals to migrate between sub-populations to overcome limitations.

## 5.2 Co-Operative Co-Evolution

### 5.2.1 Introduction

These sections contain a detailed description of the new crossover operator and the new fitness measure and sub-population structure surrounding it. First, the grade-based crossover and its potential for helping to solve the problem are explained in more detail. Then the need for a different fitness measure to supply the grade-based crossover with more relevant information is justified. In turn, this new 'pseudo fitness' makes the creation of sub-populations, each based on one of these new fitness scores, beneficial. Finally, to overcome a potential drawback of using sub-populations a migration operator is introduced, which allows individuals to swap sub-populations.

Before the new crossover operator can be applied, the strings need to be sorted in respect to the nurses' grades. These had actually been sorted before, as the hospital supplied the data sorted by grades. Although not required by the genetic algorithm so far, without it the observation that one-point crossover sometimes outperformed other crossover operators would probably never have been made. Now, however, it becomes mandatory, because only sorted strings can be regarded as three substrings - one for each grade.

The new crossover is defined as a one- or two-point crossover on these grade boundary points. The idea behind this 'grade-based' crossover is to force the genetic algorithm to keep good sub-solutions or 'building blocks', based on the grades, together and therefore to build up good partial solutions. To ensure a continued variety of different individuals in the population only some of the crossover operations are performed in this new way. The rest are done with the parameterised uniform crossover operator as before.

Similar approaches can be found in Adachi and Yoshida [1] and Hatta et al. [90]. Adachi and Yoshida temporarily protect parts of the strings from crossover and mutation. They also use sub-populations, with each sub-population 'protecting' some promising parts of its individuals. However, their sole aim is to speed up solution time, in their case the solving of small to medium sized travelling salesman problems. Hence, they protect those parts of strings, which have values that appear most frequently in the sub-populations. The effect of this is to accelerate convergence which is reported as being twice as fast as before. From our point of view, particularly solving hard problems, this approach would lead to low quality solutions due to the forced premature convergence.

Hatta et al. [90] use both uniform and two-point crossover. To decide which operator to apply to two individuals, they introduce a new measure called the 'elite degree'. This elite degree of an individual is measured by the number of recent ancestors with a high fitness value. If the sum of the elite degrees of both parents is over a certain threshold value, two-point crossover is applied, otherwise uniform crossover.

Hatta et al. report results on a number of test functions which are solved faster than with either operator on its own. However, it remains unclear why the authors do not base the choice of crossover operator directly on the fitness scores of the parents and compare it to the average fitness of the population. We conjecture that the authors wanted to keep inherited features that remained fit over a number of generations. Furthermore, it is possible that some part of the performance gain is due to the fact that two different crossover operators are used at the same time rather than just one operator on its own.

## 5.2.2 Pseudo Constraints and Pseudo Fitness

To make full use of our new crossover operator and to dispense with the effects of epistasis, the selection of parents can no longer depend on the fitness of the whole string. Thus, additional sub-fitness scores according to the 'building blocks' have to be introduced. This is achieved by partitioning the covering constraints into three independent groups in order to define a pseudo measure of under-covering. To do this we define new constants $r_{is}$ = 1 if nurse $i$ is of grade $s$, $r_{is} = 0$ otherwise, and $S_{ks}$ as the demand on day (respectively night) $k$ for grade $s$ only, excluding nurses required at lower grades.

We then rewrite the constraint set
$$\sum_{j=1}^{m}\sum_{i=1}^{n} q_{is} a_{jk} x_{ij} \geq R_{ks} \qquad \forall k,s \qquad (3)$$

as:
$$\sum_{j=1}^{m}\sum_{i=1}^{n} r_{i1} a_{jk} x_{ij} \geq S_{k1} \qquad \forall k \qquad (3.1)$$

$$\sum_{j=1}^{m}\sum_{i=1}^{n} r_{i2} a_{jk} x_{ij} \geq S_{k2} \qquad \forall k \qquad (3.2)$$

$$\sum_{j=1}^{m}\sum_{i=1}^{n} r_{i3} a_{jk} x_{ij} \geq S_{k3} \qquad \forall k \qquad (3.3)$$

With
$$S_{ks} = \begin{cases} R_{ks} & s = 1 \\ R_{ks} - R_{k(s-1)} & s = 2,3 \end{cases}$$

Note that (3.1), (3.2) and (3.3) only match (3) if the covering constraints are tight at each grade. Otherwise, (3) allows higher graded nurses to cover requirements at lower grades. For example, if there are more than enough grade 1 and grade 2 nurses and the overall cover is tight, constraints (3.1) and (3.2) will be slack, but constraints (3.3) will never be satisfied. Thus, these new pseudo constraints are not considered binding, but are merely included to guide the search.

Now the pseudo fitness of an individual for a particular grade of nurses can be defined. It is the sum of the nurses' $p_{ij}$ values penalised by any violation of the appropriate pseudo constraint. Thus, solutions that are close to providing the required cover at a particular grade will have a higher pseudo fitness than those that are not.

### 5.2.3 Sub-Populations and Co-Evolution

However, the introduction of these new pseudo fitness scores also creates problems. To use them in combination with the grade-based crossover, individuals now need to be given various pseudo fitness values to express how well they cover each of the grades separately. Thus, since selection is based on rank, multiple rankings have to be done which is computationally expensive. Furthermore, there is the question as to on which pseudo fitness score, respectively rank, should selection and replacement be based? These issues can be resolved if the population is split up into a number of sub-populations each optimising a different part of the problem based on the nurses' grades.

However, these sub-populations will have to be different from the standard 'competing' schemes as described in appendix A.4 and used for instance in Spears [156] or Mühlenbein et al. [121]. There, sub-populations are mainly introduced to speed up the solution process. Mating only takes place within a sub-population and all sub-populations follow the same fitness function, hence they are 'competing' for the best solution.

More advanced are the sub-population strategies of Gordon and Whitley [83] (island models), Starkweather et al. [158] and Whitley and Starkweather [173] (both distributed genetic algorithms). Although they still only breed within identical sub-populations, they now allow the exchange of information by letting individuals migrate from one population into another. Herrera et al. [95] take these ideas one step further: Each sub-population solves the original problem but to preserve diversity uses different selection, crossover and mutation strategies. We will take a closer look at the benefits of migration towards the end of this section. However, none of these schemes is sufficient for our endeavours, as we want each sub-population to follow a different fitness function based only on a part of the problem.

The sub-populations envisaged for our problem will be more similar to the co-operative co-evolution idea as presented by Handa et al. [88] or Potter and De Jong [129]. Handa et al. have a two level system of populations to solve constraint satisfaction problems. The higher level population works like a standard genetic algorithm, directly optimising the phenotype level of the problem. The lower level population however is different. It tries to find good schemata rather than solutions. The fitness of schema is measured by superimposing them onto full solutions and calculating the resulting difference in fitness. Good schemata are regularly communicated to the upper level population by a transmitting operator. Computational results are presented showing the effectiveness of this approach on graph colouring problems.

Potter and De Jong [129] present a general co-operative co-evolutionary approach to function optimisation. They summarise their ideas as follows:

- Sub-components of solutions are produced in sub-populations called species.
- Complete solutions are formed by assembling appropriate sub-components.
- The fitness of sub-solutions is defined as the fitness of the complete solution it takes part in.
- Each species evolves as a standard genetic algorithm.

Hernandez and Corne [94] follow the above rules to set up a ‘divide and conquer’ genetic algorithm approach for a set-covering problem where 200 rows had to be covered by 1000 columns. In their encoding, each gene represents a row and the allele is the index of a column covering it. The string is then split up into *k* chunks of even size. In their experiments, *k* is in the range between 5 and 50. Subsequently they use *k* sub-populations, each optimising one chunk only.

The authors calculate the fitness of each chunk by combining it several times with random chunks from all other sub-populations to form a full solution. In their experiments, the average of only five such complete solutions is used to keep computation time down. To cut computation time even further, the fitness of a chunk is only updated every five generations. Although one can argue that fitness calculations of Hernandez and Corne are prone to sampling errors, the computational overhead created is still immense. The best results are reported to be within 2% of optimality with better results for higher values of *k*. However, the average run time of the algorithm is in excess of one hour.

### 5.2.4 Details of Sub-Populations

To overcome the vast computational overheads as reported above and to avoid sampling errors, a different approach is proposed to arrive at the fitness of a sub-solution. Rather than calculating a ‘compatibility’ score as proposed by Potter and De Jong and used by Hernandez and Corne, we directly assign a sub-fitness via the above pseudo covering constraints to each sub-solution. Thus, the fitness of a sub-solution will be the sum of the $p_{ij}$ values of the nurses in it plus a possible penalty for a violation of the relevant pseudo covering constraints.

As a further difference from the model of Potter and De Jong, different levels of sub-populations forming a hierarchical structure are used. Low level sub-populations that optimise one grade only, medium level sub-populations that optimise a combination of grades and high level sub-populations that solve the original problem. Solution parts ‘flow’ from lower levels to higher level sub-populations due to appropriate crossover combinations. Another way of looking at this is to describe the lower level sub-populations as stepping stones towards fuller solutions produced in higher levels.

The following describes our approach in detail. Firstly, the new pseudo covering constraints are used to guide the search in seven sub-populations defined as follows:

- Individuals in sub-populations 1, 2 and 3 have their fitness based on cover and requests only for grade 1, 2 and 3 respectively.
- Individuals in sub-populations 4, 5, 6 and 7 have their fitness based on cover and requests for grades 1+2, 1+3, 2+3 and 1+2+3 respectively.

For example, the fitness functions for populations 1 and 4 become:

$$\sum_{i \in grade1} \sum_{j=1}^{m} p_{ij} x_{ij} + w_{demand}(1) \sum_{k=1}^{14} \max \left[ \left( S_{k1} - \sum_{i \in grade1}^{n} \sum_{j=1}^{m} r_{i1} a_{jk} x_{ij} \right) ; 0 \right]$$

$$\sum_{i \in grade1,2} \sum_{j=1}^{m} p_{ij} x_{ij} + w_{demand}(4) \sum_{k=1}^{14} \left\{ \begin{array}{l} \max \left[ \left( S_{k1} - \sum_{i \in grade1}^{n} \sum_{j=1}^{m} r_{i1} a_{jk} x_{ij} \right) ; 0 \right] \\ + \max \left[ \left( S_{k2} - \sum_{i \in grade2}^{n} \sum_{j=1}^{m} r_{i2} a_{jk} x_{ij} \right) ; 0 \right] \end{array} \right\}$$

As dynamic weights following the rules of Smith and Tate [152] are used, it is necessary to decide if $w_{demand}(B)$ should be based on the best individual overall or whether it should be calculated separately for each sub-population *B*. The pseudo covering constraints will be slacker for some sub-populations than for others, as for instance there are more grade one nurses than strictly required. Therefore, a different penalty weight for each sub-population seems appropriate and the latter option was adopted. Note that this can result in different values for $w_{demand}(B)$ in each sub-population *B* at any point in time.

Note also that sub-population 7 does not represent the original optimisation problem as it is only concerned with the total cover and not with the cover at grades 1 and 2 independently. It is therefore necessary to maintain a 'main' sub-population - sub-population 8 - whose fitness is the original fitness as given in section 4.1. This main sub-population is the most important, as it is the only sub-population that aims to find solutions to the original problem. This will be reflected when sizing the sub-populations in terms of the number of individuals in them.

In order to solve the original problem, we also need to strike a balance between producing highly fit individuals with respect to sub-strings in the sub-populations and allowing individuals from complementary sub-populations to combine. Additionally, the total number of individuals in all sub-populations has to be the same as before, i.e. 1000, for a fair comparison of results. These issues are achieved by the following rules:

1. Sub-populations 1-3, each with 100 individuals, evolve from parameterised uniform crossover within themselves for maximum diversity.
2. Sub-populations 4-7, each with 100 individuals, evolve from 50% parameterised uniform and 50% grade-based crossover. In the case of grade-based crossover the parents are selected by rank from sub-populations 1-3 as appropriate and combined accordingly, for example grade-based children in sub-population 4 would have parents from sub-populations 1 and 2.
3. Main sub-population 8, with 300 individuals, evolves from 50% parameterised uniform and 50% grade-based crossover. In the case of grade-based crossover the parents are picked from sub-populations 1-7 and combined accordingly. For instance, possible grade-based crossover combinations are individuals from sub-populations 1+4, 5+2, 1+2+3 or 7 combined with any other.

### 5.2.5 Migration

A well-known drawback with the use of sub-populations is the loss of information due to the limited choice, in comparison to one big population. To help overcome this problem individuals are allowed to swap sub-populations. This operation is called migration (see Starkweather et al. [158], Tanese [163], Whitley and Starkweather [173] and Whitley [175]). The migration operator keeps the number of individuals in each sub-population constant whilst spreading the information they contain across others. An additional benefit in our case is that through successful migration an individual may become 'fit' for more than one sub-objective.

Two different kinds of migration with various parameter settings were tried. The first is random migration in which there is a small chance $p_M$ *[1%...10%]* in every generation for any individual to move into another randomly chosen sub-population. The second is a migration of the best five individuals of each sub-population to a random sub-population every *g [1...20]* generations. The swap partner is determined randomly for all migrations. The difference in results for various parameter settings was only small, and the best results were achieved for $p_M = 5\%$, and $g = 5$. Full results are reported in appendix D.3. For all future experiments random migration with $p_M = 5\%$ was used.

### 5.2.6 Results

An overall summary of all results can be seen in Figure 5-1. Clearly the introduction of the sub-populations based on the nurses' grades and the use of the special fixed point crossover have improved both feasibility and cost of solutions dramatically over the original results. As an additional benefit, the average solution time has been reduced from 12 seconds to less than 10 seconds for a single run. A further good improvement is made when random migration is added. However, migration of the best does not improve results as much as random migration. This is thought to be because of the 'specialisation' of each sub-population. Whilst a random migration might introduce some interesting schemata, the migration of schemata that are already optimised for some criteria, proves less helpful.

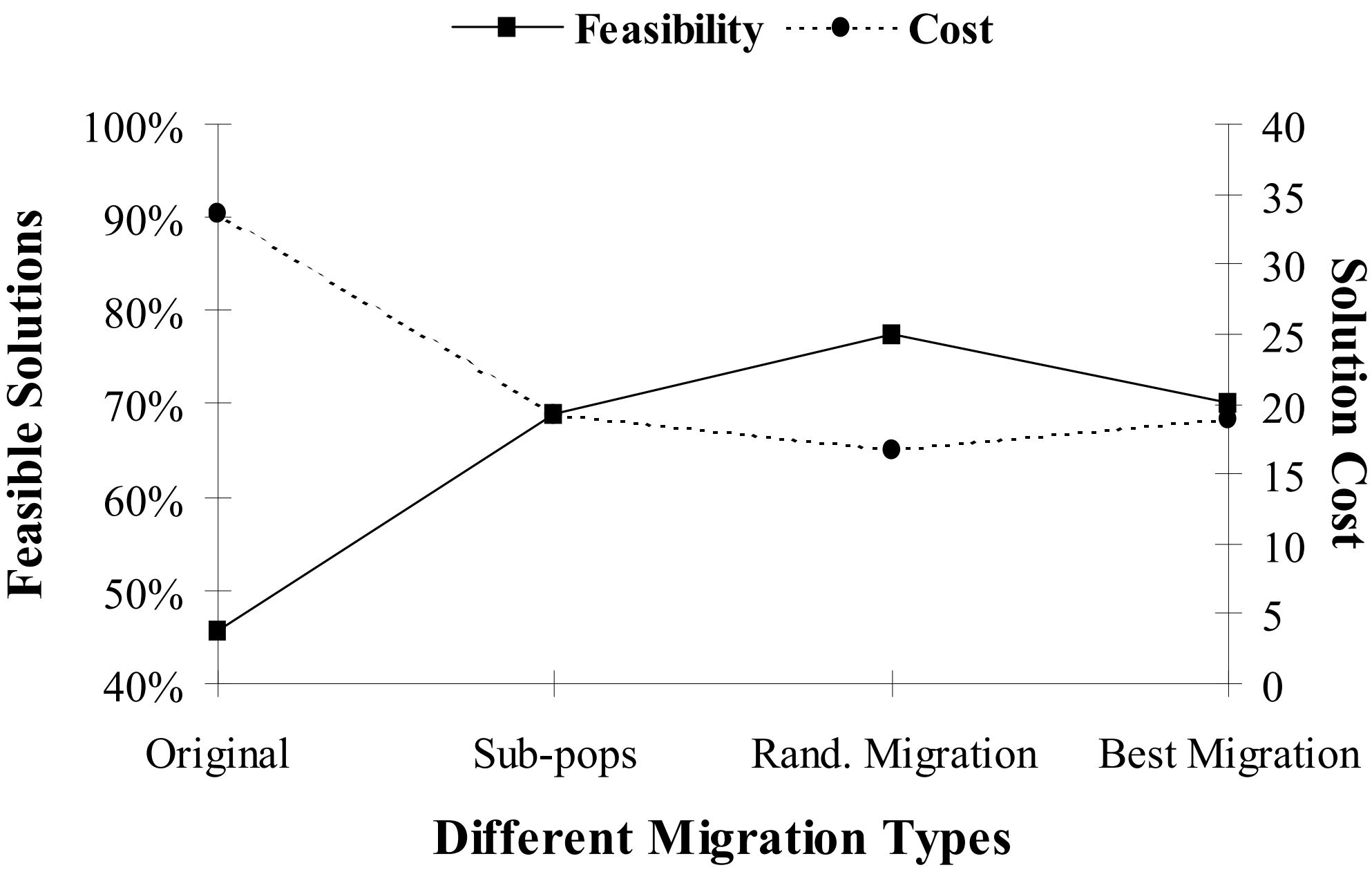

Figure 5-1: Comparison of different migration types with no migration and no sub-populations.

## 5.3 Swaps and Delta Coding

### 5.3.1 Swaps

The results of the co-operative co-evolutionary approach are an immense improvement over the original results found with a standard genetic algorithm in chapter 4. However, they still fall short in comparison to the quality of solutions found by tabu search or integer programming software. This section will portray two further enhancements of our genetic algorithm, namely swaps and Delta Coding. Both ideas try to improve solutions further by making small changes not usually achieved by the genetic algorithm itself.

Together with the approaches portrayed in section 5.4, these are examples of using problem-specific information to enhance the genetic algorithm. This differs slightly from the co-evolutionary scheme, where it was more the problem structure that was exploited. Adding problem-specific knowledge is a well-known strategy to improve the performance of genetic algorithms, as suggested for instance by Grefenstette [84], Suh and Gucht [159] and others.

Swaps – a simple form of repair - try to improve the fitness of individuals by allowing nurses to swap their shift patterns worked. Note that throughout this section the word 'swap' is used loosely for cyclic exchanges rather than in its original meaning of a two-way exchange. Delta Coding follows the ideas presented in Whitley et al. [178]. The search starts with a population based on a previous good solution rather than initialising the algorithm randomly. However, as this section will show, both enhancements are limited in their ability to improve solution quality further.

Three types of swaps were tried: Chain swaps, special swaps and adjacent swaps. In chain swaps, up to four nurses can swap their shift patterns amongst themselves. A swap is allowed if the sum of the nurse-shift pattern costs $\Sigma p_{ij}$ after the swap is smaller than before. Note that for a single nurse this can lead to a worse shift pattern than before because it is not necessarily a pareto optimal swap. Only nurses of the same grade are allowed to take part in the swap. This is done so that the cover provided is not influenced by the swap. The limit of four was set because of the computationally extensive calculations involved. However, as nurses have to be of the same grade (and logically of the same number of working days and nights), this is hardly restricting, as there are rarely more than four 'identical' nurses.

Special swaps were introduced after it was noticed that for certain data sets 'impossible' solutions were found. This occurred when there were nurses present that could either work $k$ days or $k$ nights and nurses that could work either $k$ days or $(k-1)$ nights. This could lead to a situation where there is an overall shortage by one shift. A situation like this could arise when a nurse of the former type works on days whilst a nurse of the latter type is on nights.

Normally, this should be sorted out by the genetic algorithm. However, if there are unfavourable $p_{ij}$ values, resulting in a large penalty on the former nurse working nights and on the latter working days, then the genetic algorithm might find it very hard to do so. Special swaps detect such a situation and then force two nurses of the above types to swap working days and nights irrespective of the $p_{ij}$ values. The swap takes place by moving the nurse working $k$ days or $k$ nights onto the night shift of the second nurse and choosing a random day shift for the first nurse.

Adjacent swaps are designed to improve feasibility of solutions without increasing the penalty too much. Adjacency is defined as two shift patterns that differ only by one working day (night) being moved to another position. Initially, a matrix of adjacency is calculated. In it are all adjacent patterns for each shift pattern. For example, (0111110 0000000) and (1111100 0000000) are adjacent.

The swaps then work as follows: First, all days and nights with surpluses and shortages are determined. Then for each nurse working on a day (night) with a surplus and not on a day (night) with a shortage, all her adjacent shift patterns are checked. If one of them would rectify the situation, then a swap takes place. Only adjacent patterns are allowed, because they are easily identifiable and usually only involve a small change in the $p_{ij}$ values. This is due to the way $p_{ij}$ values are set up as detailed in section 2.1.3.

A comparison of results with those achieved by co-operative co-evolution with random migration (label 'No Swaps') is shown in Figure 5-2. Because of the large amount of extra computation, all swaps are only applied to the top ten individuals in the main sub-population. This increases the average solution time from under 10 to about 13 seconds. Experiments applying the swaps to more individuals resulted in an even larger increase in solution time without giving significantly better results.

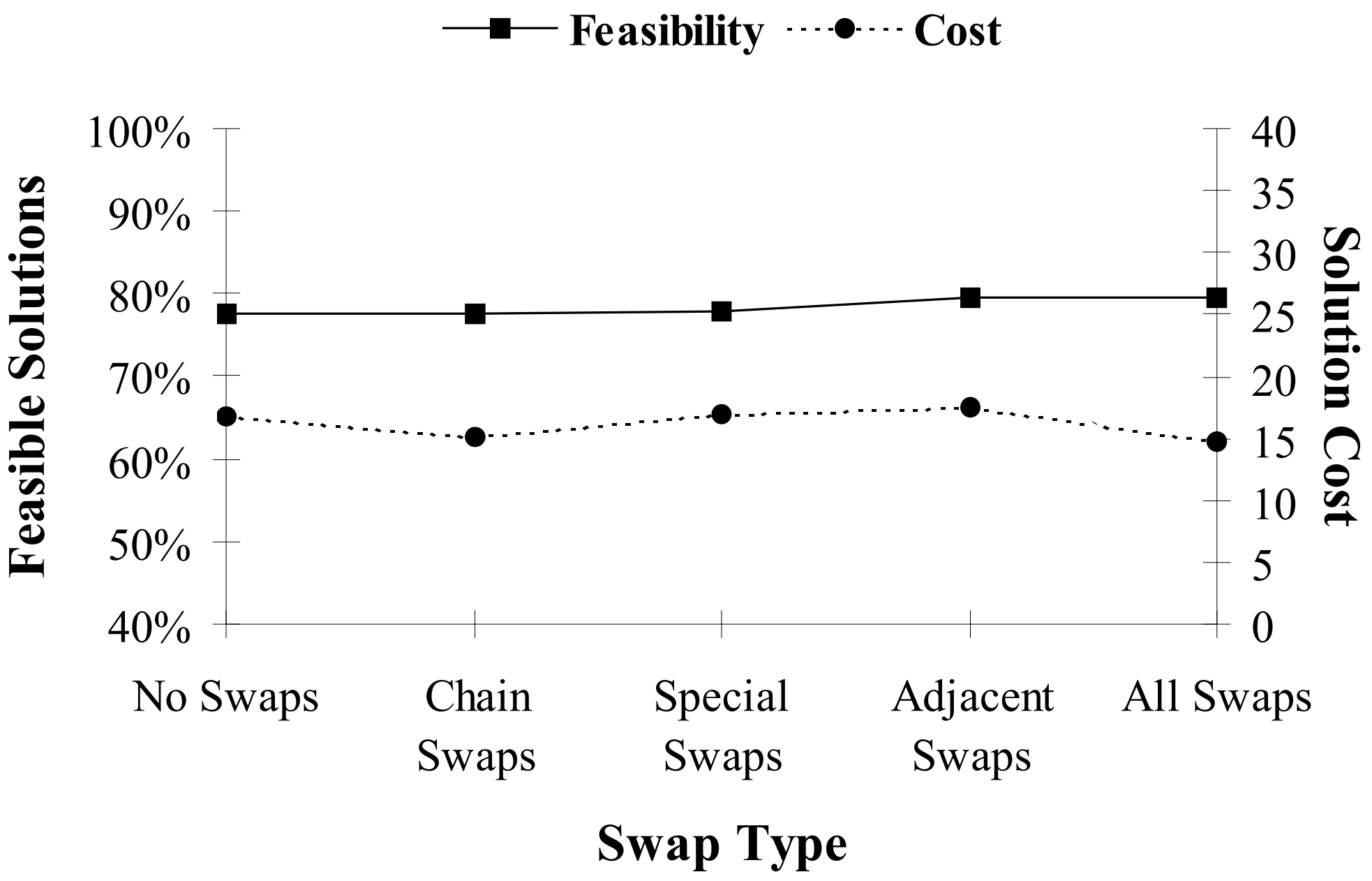

Figure 5-2: Results for various types of swapping.

In general, all three swaps behave as expected. Chain swaps improve the cost of solutions without affecting feasibility. Special swaps have little effect, just slightly improving feasibility by overcoming some of the problems of awkward data sets. Adjacent swaps improve feasibility slightly without making the penalty cost much worse. All three combined have a slightly lesser effect than the sum of the three single swaps. Overall, both feasibility and cost are improved slightly, but they are still nowhere near the quality of tabu search. Nevertheless, the benefit was considered sufficient to use all three types of swaps for all future experiments.

### 5.3.2 Delta Coding

The idea of Delta Coding is taken from Whitley et al. [178]. Delta Coding tries to strike a balance between diversity by initialising populations at random and preserving information by basing new populations on previous good solutions. This is done by initialising a new population at random but within a hypercube around a previously found good solution. The search starts by performing an initial run with a standard genetic algorithm. Once that search has terminated, a new genetic algorithm is started. However, the solution space is reduced to the hypercube around the best solution found in the initial run via the above mentioned initialisation scheme. This process is then repeated as often as required with the hypercube gradually becoming smaller.

Whitley et al. use a binary encoding to optimise problems with real variables. After the initial standard genetic algorithm, their string does not represent the actual parameter values but rather difference values (i.e. delta values and hence the name 'Delta' Coding). These differences are always measured from the previous best solution used for initialisation purposes. To reduce the solution space, in subsequent runs Whitley et al. allow less and less bits to encode these differences up to a certain lower limit. This results in smaller and smaller differences from the original solution. The authors then argue that due to this reduced solution space population sizes can be reduced without affecting solution quality.

In order to apply Delta Coding to our problem some slight changes from the original idea have to be made. In our encoding the genes are the shift patterns worked by the nurses. Thus, a way to measure differences or delta values between shift patterns has to be found. This can be achieved with the previously mentioned adjacency matrices of similar shift patterns. In order to use these matrices, the degree of adjacency of two shift patterns is defined as the number of working shifts that have to be moved to transform one shift pattern into the other one.

For instance, the degree of adjacency between (1100000 0000000) and (0001100 0000000) is two and between (0000000 0100111) and (0000000 0101110), it is one. For those nurses who work more day shifts than night shifts, the degree of adjacency between their day and night shift patterns cannot be calculated in this way. Instead, it is fixed as the number of working shifts in the day shift pattern.

This degree of adjacency is now used as a distance measure between shift patterns. Therefore, the next step to use delta coding for our problem is to calculate five adjacency matrices, with different degrees of adjacency. Thus for each shift pattern, all shift patterns that differ for up to five, four, three, two and one working day(s) or night(s) are found. To start the Delta Coding optimisation, a standard genetic algorithm run is performed which is initialised at random as before. Subsequently there are five delta coding runs, each being initialised using the appropriate adjacency matrix and based around the best solution found in the most previous run.

Hence, Delta Coding run 1 uses the adjacency matrix for up to five shifts, Delta Coding run 2 uses the adjacency matrix for up to four shifts and so on. Note that Delta Coding run 1 uses the same solution space as the standard genetic algorithm. However, due to the additional Delta Coding probability $p_{DC}$ introduced in the following, the initial population is less diverse as it is centred around the best previous solution.

Because of the discreteness of our variables, we have to use the concept of adjacency matrices. However, the difference between two directly 'adjacent' solutions is quite big, as all nurses will have to work a different shift pattern. Thus, for 25 nurses this can result in up to 50 single shifts being different. Although some of the changes will cancel each other out, we still felt that the difference created would be too big. Thus, to be able to fine tune the level of differences, one further parameter was introduced. It is called the Delta Coding probability $p_{DC}$ and defined as the probability that a gene is changed from the value of its counterpart in the previously best solution. So for instance, if $p_{DC} = 10\%$ then there is a 90% probability that a gene is initialised to the same value as in the previously best solution and a 10% probability that a value from the appropriate adjacency matrix is chosen.

Figure 5-3 summarises the results for various values for the Delta Coding probability $p_{DC}$ and compares this to a co-evolutionary genetic algorithm without Delta Coding (label 'w/o'). When Delta Coding was used, 24 runs were made per data set. These were four 'standard' genetic algorithm runs plus five Delta Coding runs for each standard run. The population size of the Delta Coding runs was not reduced, as suggested by Whitley et al., since this gave far worse results. This is not surprising, as unlike for the binary encoding example used by Whitley et al., there is no reduction in the size of our encoding, only the range of potential values is reduced.

As the graph shows, a co-evolutionary genetic algorithm with Delta Coding produced worse results for any setting of $p_{DC}$ in comparison to a co-evolutionary genetic algorithm without (label 'w/o'). The labels indicate the value of $p_{DC}$. The results tend to get worse the 'more' Delta Coding is used. For full Delta Coding, i.e. $p_{DC} = 100\%$ the solutions are far worse than before

The most likely reason for this failure is the inadequacy of the adjacency matrix scheme together with the discreteness of the shift patterns. However, it is difficult to see any other way to apply the idea of Delta Coding to our problem. Some further improvements could possibly be achieved by seeding each sub-population around the best individual of the corresponding sub-population, rather than basing all individuals on the best overall solution of the previous run. This is an interesting idea for future research.

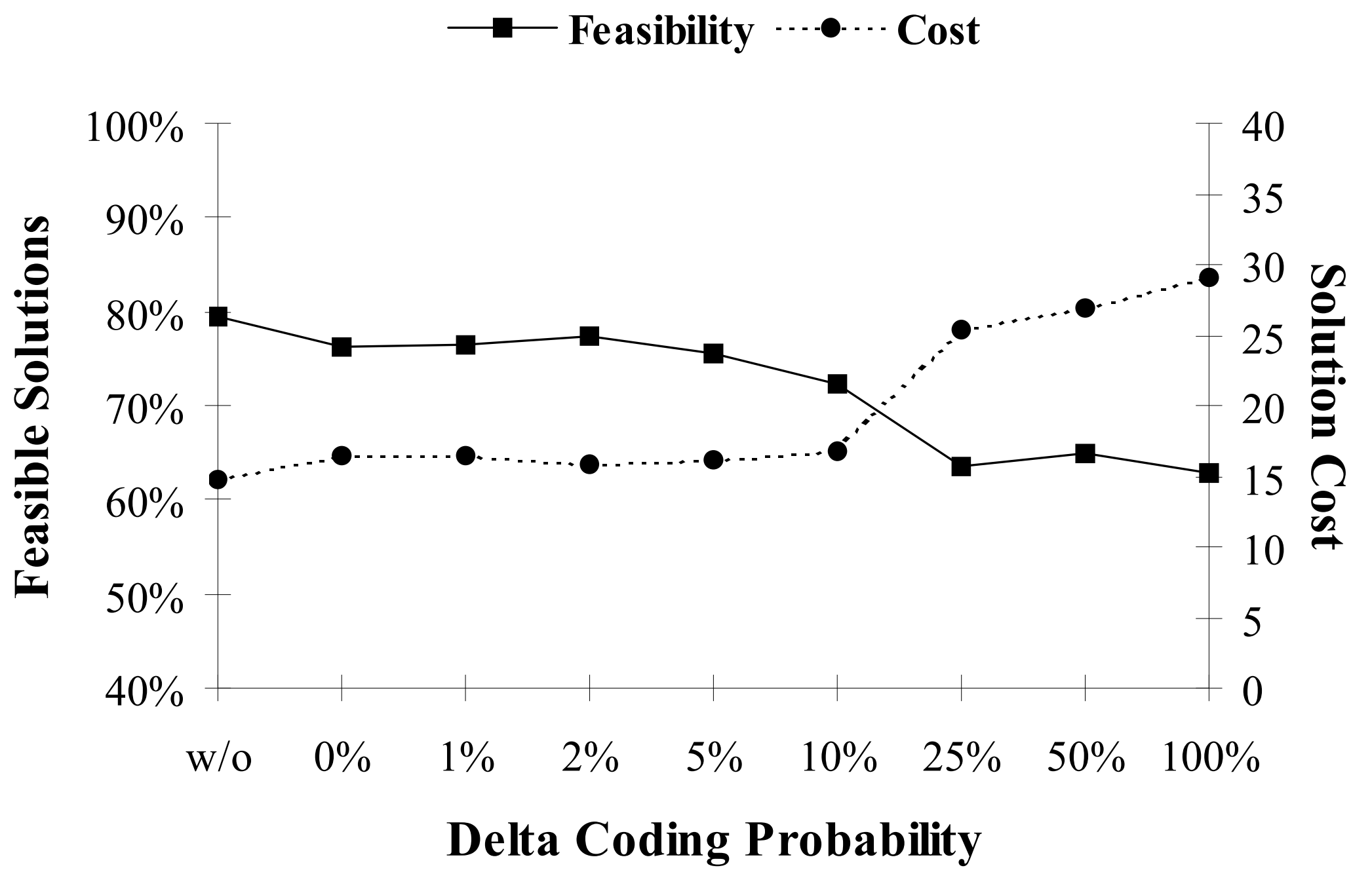

Figure 5-3: Comparison of solution quality for various Delta Coding Probabilities.

## 5.4 Hill-Climber, Repair and Incentives

### 5.4.1 Introduction and Definitions

Since the quality of solutions still lags behind the results found by Dowsland [55], we decided to have a closer look at the failed attempts (i.e. those runs that failed to find either good cost or feasible solutions). This showed that the remaining problems were due to the search converging towards solutions that satisfied most, but not all, of the covering constraints. These infeasible solutions could be further partitioned into two distinct types, which we shall refer to as *balanced* and *unbalanced* solutions. Additionally, there are *undecided* solutions, which yet have to converge to either type.

In *balanced* solutions, the unsatisfied constraints relate to either (but not both) days or nights, and there is sufficient over-covering on other days or nights respectively to possibly correct this. In the second type, *unbalanced* solutions, there is a day and night shift imbalance, for example one night shift missing and one day shift too many or vice versa. Both these problems highlight the difficulty of making small changes with our genetic algorithm implementation, due to the disruptiveness of the crossover operators and our encoding that only allows moving 'whole' nurses.

For the purpose of defining these three classes of infeasible solutions formally, the following four conditions C1 – C4 are set up:

- C1: There is at least one day shift with a shortage.
- C2: There is at least one day shift with a surplus.
- C3: There is at least one night shift with a shortage.
- C4: There is at least one night shift with a surplus.

*Balanced* solutions are those that satisfy both C1 and C2, but not either of C3 or C4. *Balanced* solutions are also those that satisfy both C3 and C4, but not either of C1 or C2. *Unbalanced* solutions are those satisfying either C1 but not C2 or C3; and furthermore those satisfying C3 but not C1 or C4. All other solutions are defined as being *undecided*.

Table 5-1 gives some examples for clarification. Mon – Sun refers to the seven day shifts and mon – sun to the seven night shifts. A positive number indicates a surplus of nurses and a negative number a shortage. The first *balanced* solutions shows a situation where all over- and under- cover is on day shifts. The second *balanced* solution shows a similar situation with all under- and over-coverings on nights. In the two following *unbalanced* solutions, the situation is different. Either the surplus is on days and the shortage on nights or vice versa.

The third *unbalanced* solution is a special case. No over-cover is present to compensate for the uncovered shift. This can happen if a nurse working an equal number of day and night shifts is working days and a nurse who works less night shifts than day shifts is on nights. A further discussion of this issue can be found in section 5.3.1 where a special swap operator was designed to remedy the situation.

*Undecided* solutions are characterised by usually allowing for the movement of 'whole' nurses, i.e. they have not yet converged to either situation and it is therefore possible that the genetic algorithm operators will still turn them into feasible solutions. However, the second *undecided* solution looks more likely to develop into an *unbalanced* solution, with a surplus on nights and a shortage on days.

| | Days | | | | | | | Nights | | | | | | |
|---|---|---|---|---|---|---|---|---|---|---|---|---|---|---|
| Type | Mon | Tue | Wed | Thu | Fri | Sat | Sun | mon | tue | wed | thu | fri | sat | sun |
| Balanced | -2 | 0 | 1 | 0 | 0 | 1 | 0 | 0 | 0 | 0 | 0 | 0 | 0 | 0 |
| Balanced | 0 | 0 | 0 | 0 | 0 | 0 | 0 | 0 | -1 | 0 | 0 | 0 | 1 | 0 |
| Unbalanced | 0 | 0 | -1 | 0 | 0 | 0 | 0 | 0 | 1 | 0 | 0 | 0 | 0 | 0 |
| Unbalanced | 0 | 0 | 0 | 0 | 0 | 2 | 0 | 0 | -1 | 0 | -1 | 0 | 0 | 0 |
| Unbalanced | 0 | 0 | -1 | 0 | 0 | 0 | 0 | 0 | 0 | 0 | 0 | 0 | 0 | 0 |
| Undecided | 0 | 0 | -1 | -1 | 0 | 2 | 0 | 0 | -1 | 1 | -1 | 0 | 1 | 0 |
| Undecided | 0 | -1 | -1 | 1 | 0 | 0 | -2 | 0 | 0 | 2 | 0 | 2 | 0 | -1 |

Table 5-1: Examples of balanced, unbalanced and undecided solutions.

### 5.4.2 Implementation

The problem of *balanced* solutions could usually be overcome if the algorithm was allowed to run for long enough, due to the use of mutation or the adjacency swap operator presented in section 5.3.1. This is because a *balanced* solution can be made feasible by moving a single nurse onto a different shift pattern. To achieve this, a nurse has to be moved from a shift pattern that includes some overcovered days to one covering some uncovered days. Standard mutation and the adjacency swap operator have a chance of doing this.

On the other hand, the second type of infeasibility (*unbalanced* solutions) remains a problem because of the hospital's policy that all but a few nurses on special contracts must work either days or nights in any week. Thus, once the 'wrong' nurses are on days respectively nights, the genetic algorithm cannot sort it out, because both crossover and mutation can only move 'whole' nurses in our encoding. Thus, unless a lucky swap with another nurse is performed, moving a 'whole' nurse is equivalent to moving up to five shifts which by far overcorrects most situations. Therefore, this would result in worse solutions, which are unlikely to take part in crossover or survive replacement in the following generation.

The solution to this dilemma is to look ahead and avoid these unfavourable situations altogether by rewarding solutions with a 'future' potential, i.e. *balanced* solutions, and penalising those that are *unbalanced*. For this purpose, further problem-specific knowledge needs to be incorporated into our algorithm.

In our case, this is achieved as follows: The fitness scores of *balanced* and *unbalanced* solutions are adjusted by adding an incentive or bonus to *balanced* solutions and a disincentive or negative bonus to *unbalanced* solutions. This results in *balanced*, but less fit, solutions ranking higher than *unbalanced* fitter solutions, which is the desired effect. Both the incentive and disincentive are based on the weights used for the constraint violation penalties, multiplied by a constant factor. Thus, they change dynamically at the same rate as the penalties and are measured in multiples of violated constraints.

Logically, the result of the above changes will result in more runs converging towards *balanced* solutions. As discussed above many of these eventually become feasible as a result of random mutation and / or swap operators. It makes sense to exploit this situation by introducing a more intelligent mutation operator that will attempt to repair these solutions directly. This will not only reduce the number of generations before such solutions become feasible, but will also counter the disruptiveness inherent in the crossover operator and ensure that such solutions reach their full potential before being destroyed by crossover.

In view of the Lamarckian and Baldwinian discussion, as described in section 3.5 and in Whitley et al. [172], we decided to use a Lamarckian strategy. Hence, our repair operator is a simple hill-climber and works as follows. A *balanced* solution is taken and subjected to an improvement heuristic that cycles through all shift patterns of each nurse, accepting a new pattern if it improves fitness. This results in the ability to make only slight changes in the days and nights worked by a single nurse. Note that this type of hillclimber is similar to the gradient-like bit-wise improvement operator introduced by Goldberg [81] and used in Chen and Chen [38] and elsewhere.

However, this repair routine is only applied to the top five *balanced* solutions of the main sub-population. This avoids potential problems of premature convergence resulting from using too much aggressive mutation and saves wasting computation time repairing and improving large numbers of *balanced* solutions in later generations. Note that in the absence of five *balanced* solutions, the hill-climber is also applied to feasible solutions. In this case, it aims to improve the penalty score.

Note also that due to the previously mentioned day / night shift peculiarity (see special swaps in section 5.3.1 for more details about this), a solution that is *balanced* but overall has a shortage of a shift, still gets the bonus and has a chance to be repaired. For instance, (0,-1,-1,1,0,0,0; all nights 0) is a solution of this type. After going through the repair operator, this solution will possibly be (0,0,-1,0,0,0,0; all nights 0). Thus, it has become *unbalanced* and will be dropped from the population quickly. This is the desired result, as solutions of this type are difficult to be repaired to feasibility, with only the special swap operator offering some chances of achieving this.

### 5.4.3 Results

Figure 5-4 shows the outcome of experiments with different magnitudes for incentives only (label 'i'), incentives combined with repair (label 'r') and disincentives only (label 'd'). The magnitude is measured in terms of violated constraints. For instance, an incentive of two (label '2 i'), nullifies two constraints violated by one unit or one constraint violated by two units. As before, the genetic algorithm is based on the co-operative co-evolutionary approach.

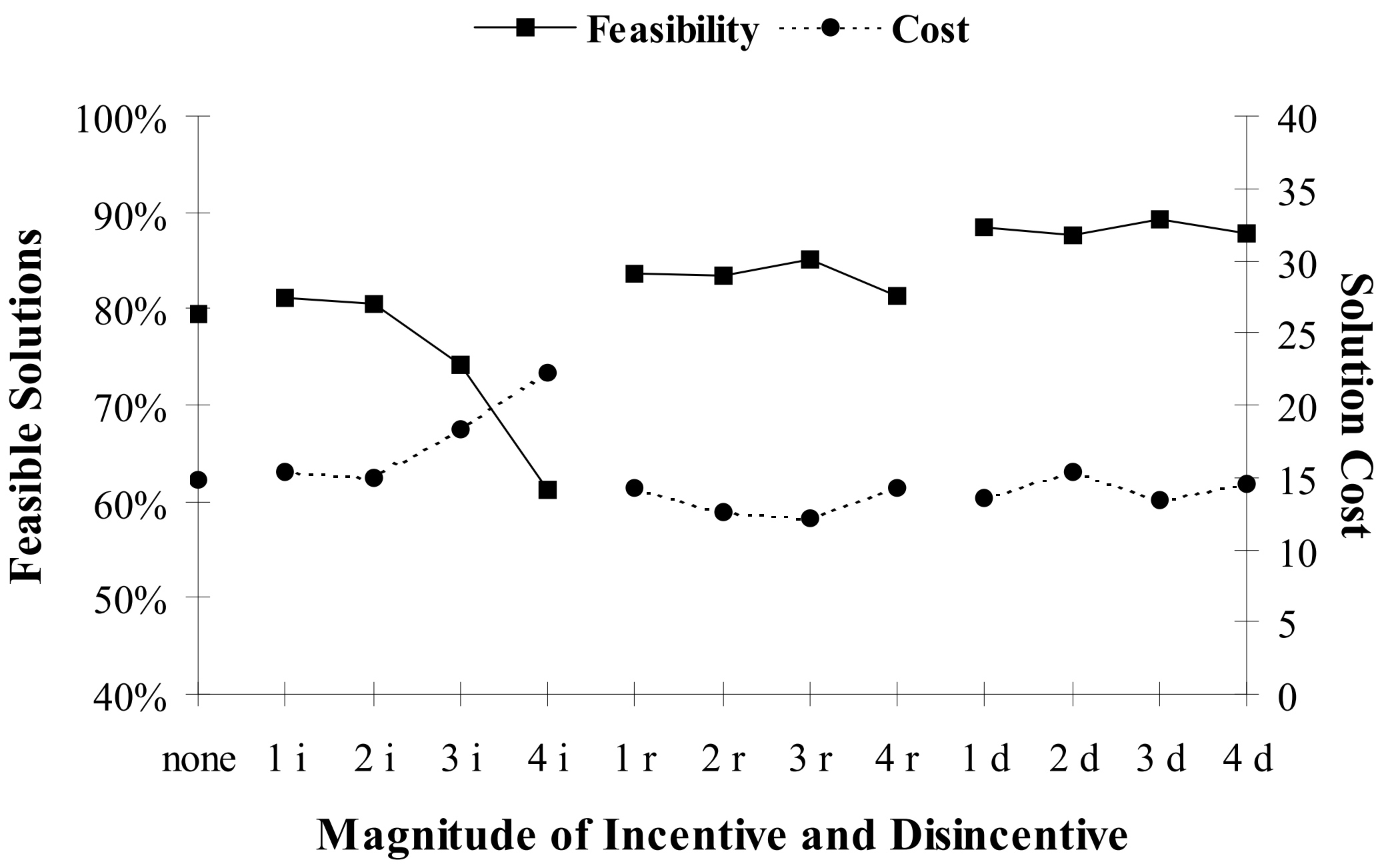

Figure 5-4: Different incentives (i), incentives and repair (r) and disincentives (d).

The graph shows that a small incentive improves results slightly whilst a bigger incentive biases the search in the wrong direction. However, this bias is exploited once the repair operator is active and now a bigger incentive gives better results. The disincentive works on its own to improve feasibility by penalising 'dead end' situations as described above. Within the range examined, the magnitude of the disincentive is largely unimportant. For all further experiments, both incentive and disincentive is set to three, i.e. equivalent to three constraints violated by one unit.

A final summary of results is shown in Figure 5-5. If used together, disincentives and incentives plus repair significantly improve solution quality both in terms of feasibility and in terms of cost. Solution time remained largely unchanged at on average 13 seconds.

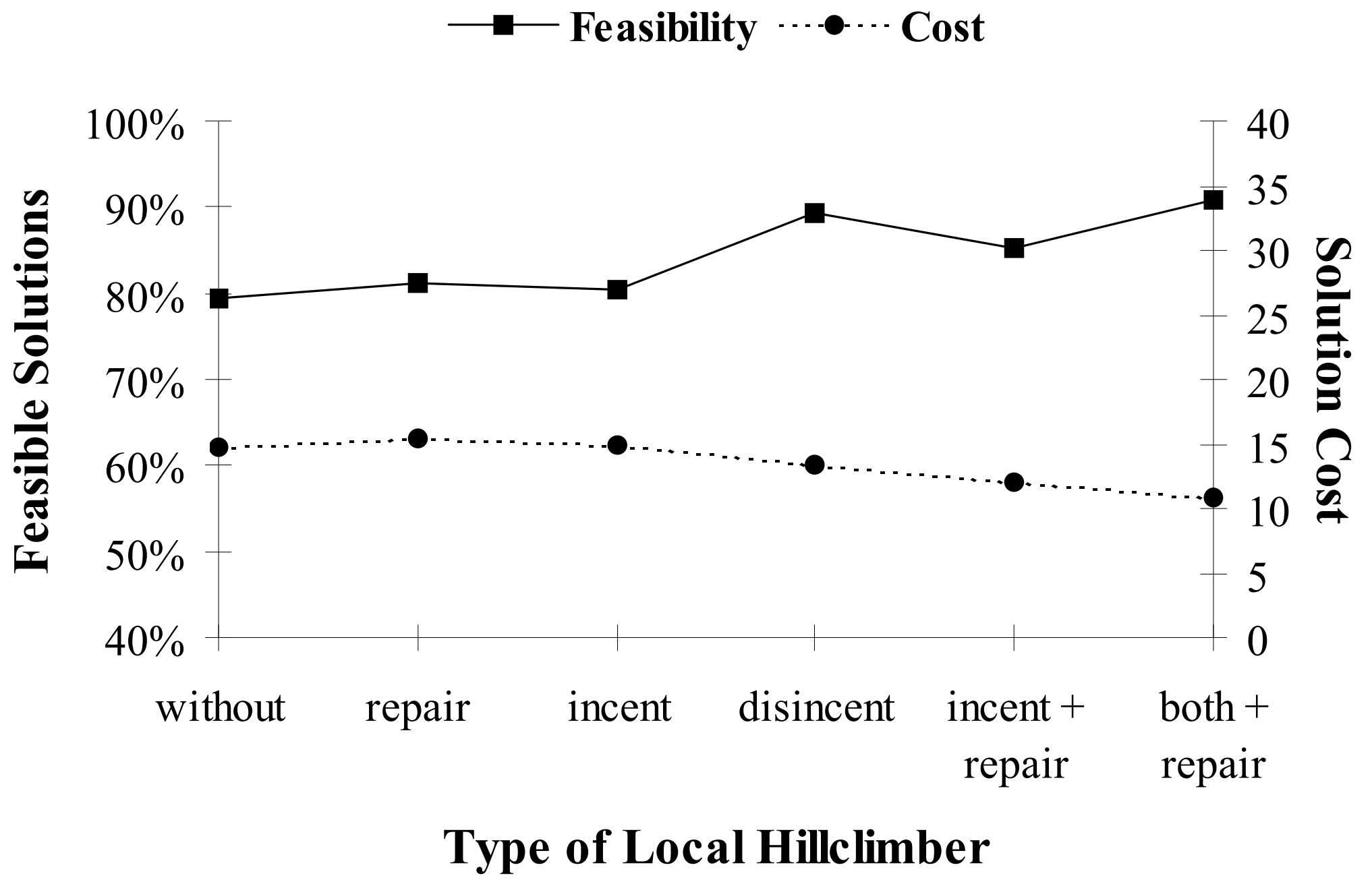

Figure 5-5: Various local hillclimbing strategies.

## 5.5 Conclusions

So far this thesis has shown how successive additions of problem-specific information were able to improve an at first unsuccessful genetic algorithm implementation for the nurse scheduling problem. This had been done to the point where the genetic algorithm was able to produce feasible solutions of good quality. A summary of the results is shown in Figure 5-6. Detailed results for all 52 data sets are reported in appendix D.2.

Given the high degree of epistasis introduced by penalising the covering constraints in the fitness function, the lack of success of the original implementation is not surprising (label 'Basic'). As mentioned previously, the genetic algorithm relies on assembling good building blocks to form good solutions. Unfortunately, because of the importance of the covering constraints, there is a strong interdependency between the shifts worked by the nurses. Therefore, 'good' shift patterns in one solution are less useful in another. This would probably not be such a key issue if the constraints were less tight as this would move the emphasis away from the covering constraints to the nurse-shift pattern costs $p_{ij}$. However, as explained in section 2.1.2, all data is pre-processed by a knapsack routine to remove slackness before it is optimised with the genetic algorithm.

The use of dynamic penalties was able to increase the number of runs terminating with feasible solutions, but these improvements were small (label 'Dynamic'). Again with hindsight, this is not surprising, as making the penalties more intelligent does not tackle the main issue of epistasis. Consequently, the biggest single success was obtained by exploiting the grade based structure of the problem, using sub-populations to provide large building blocks that were then combined using special crossover operators (label 'Sub-pops'). This was a way of circumventing the epistasis problem.

Although the epistasis present in the original problem cannot be reduced since it is inert in the chosen encoding and problem structure, it is lower in the newly created sub-problems. This is because the sub-problems only optimise the cover for parts of the problem by abstracting from some of the grade requirements. Since these could not be disposed of altogether, they were re-introduced step by step.

This was done via a hierarchical sub-population structure: lower level sub-populations optimised the cover for one grade, medium level sub-populations for two grades and the highest level for all grades. Finally, a main sub-population combined the information gathered in all other to solve the original problem. To improve results further, the exchange of information between sub-populations was encouraged by the introduction of a migration operator. This partially overcomes the drawback of using sub-populations in comparison to one big population, which is the limited choice of individuals in the selection phase.

Once the sub-populations with special crossovers and migration were in place, failures appeared to be due to the well-known problem that genetic algorithms 'lack the killer instinct' (De Jong [51]). This means that they converge towards good solutions, but are too disruptive to make the minor changes necessary to improve these. This problem is frequently overcome by adding problem-specific information, for instance in the form of a hill-climber to make small improvements to some or all of the population. Two such approaches, one based on 'swapping' shift patterns and the other on Delta Coding as presented by Whitley et al. [178], were not very successful at doing this.

However, we were finally successful with a more intelligent approach of including problem-specific knowledge by recognising solution attributes that were likely to lead to successful hill-climbing, and those that certainly would not. We then repaired and rewarded the former class and penalised the latter (label 'Repair'). Although those solutions with favourable attributes would eventually become good feasible solutions via mutation or the swap operators, to speed this up they are processed by a hill-climbing repair routine.

The result is a fast robust implementation, in which all the add-ons remain very much within the spirit of genetic algorithms and whose solutions are of only marginally worse quality than those found by tabu search (label 'Tabu'). A look at the detailed results in appendix D.2 confirms this view: 49 out of 52 data sets are solved to (or close to) optimality, with the remainder having at least on feasible solution.

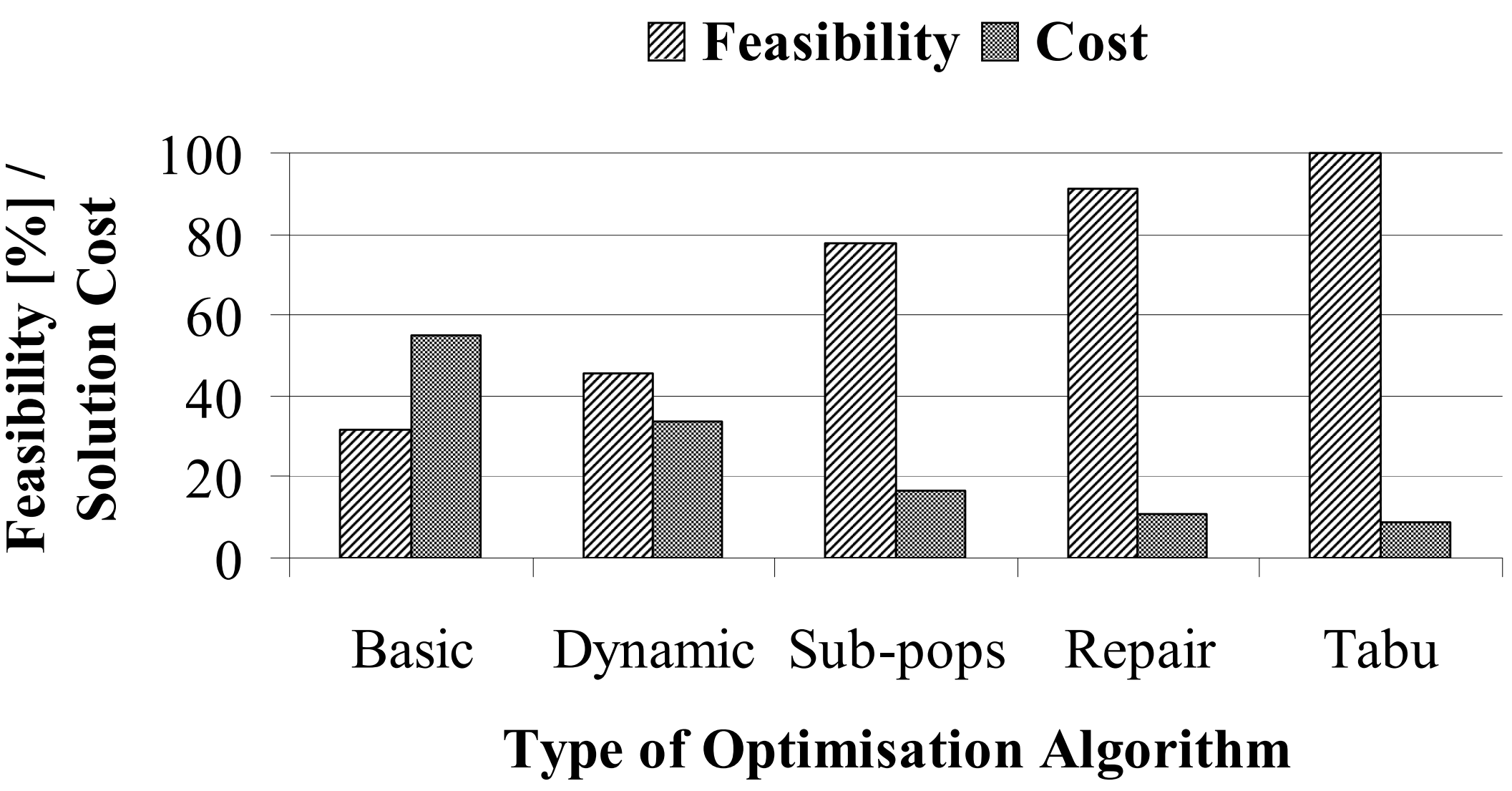

Figure 5-6: Comparison of nurse scheduling results for various direct genetic algorithm approaches and tabu search.

It is not claimed that the algorithm used so far is necessarily the best genetic algorithm approach to the problem. In particular the development effort was considerable. However, rather than adding even more to it, like for example developing a form of intelligent mutation or extending the number of dynamic parameters, we would like to take a different approach.

In an early paper, Davis [47] suggested using an indirect coding combined with a heuristic decoder as an effective means of overcoming epistasis. Since then, this type of genetic algorithm has been used successfully on a variety of problems. More details about decoders can be found in section 3.7. This approach has the advantage that all the problem-specific information is contained within the heuristic decoder, whilst a basic genetic algorithm works to optimise its parameters, for example, the order in which the nurses are to be processed. In the next chapter such an indirect approach is developed for the nurse scheduling problem and compared to the results found so far.

# 6 An Indirect Genetic Algorithm Approach for Nurse Scheduling

## 6.1 What is an 'Indirect' Approach?

This section starts with an introduction to indirect genetic algorithms. Then the merits of this approach for the nurse scheduling problem are investigated. Following on from this, the differences between our decoder and the decoders usually found in other indirect genetic algorithms, like those summarised in section 3.7, are discussed. Before the decoder used is presented in detail, a variety of order-based crossover and mutation operators are examined. These are made necessary by the order-based encoding of the indirect approach.

An indirect genetic algorithm is the hybridisation of a standard genetic algorithm with other heuristics. For this purpose, the genetic algorithm can be kept almost canonical. However, rather than representing the problem 'directly', the genotype of an individual is now fed into a decoding heuristic, which produces a phenotype and its fitness value.

This can be done in a number of ways. One possibility is for the genotype to be a set of parameters. The decoder then assembles the phenotype following a set of rules guided by these parameters. For instance, in the nurse scheduling problem, there could be a set of different scheduling rules, such as 'maximise preferences', 'spread cover out evenly', 'cover days with highest uncovered demand first' etc. The parameters of the genotype would then decide in which order to use these rules to form the phenotype.

Unfortunately, this type of encoding would require the decoder to decide who and when to schedule. This requires very sophisticated rules and leads away from the original idea of optimising a nurse scheduling problem with genetic algorithms. Hence, we will follow a different strategy. The genotype will be a permutation of the nurses whilst the decoder will take the form of a greedy heuristic, assigning shift patterns to the nurses in the order given by the genotype.

Such an indirect approach has the advantage that most of the problem-specific information will be contained within the decoder. Of course, a suitable decoder must be constructed first. This will be discussed in section 6.3. It is in the co-operation and counter-balancing between a problem-specific and greedy decoder and a generic but stochastic genetic algorithm that the success of an indirect approach lies. In addition to the general information on the use of this method, summarised in section 3.7, we refer to Eiben et al. [58] who focus on the deterministic and stochastic balancing discussion.

Eiben et al. solve constraint satisfaction problems, such as the N-Queens and graph colouring problems, with genetic algorithms and with deterministic constructive search algorithms. They argue that in general, the constructive heuristics work well, but sometimes are misled and fail to give good results. This can be overcome by diversifying the search, for instance by keeping multiple solutions in parallel or incorporating random elements into the construction mechanism.

Eiben et al. then note that these are in fact essential principles of genetic algorithms. Finally, they show that the combination of a genetic algorithm with a deterministic construction heuristic gives better results than either method on its own on a number of problems. The authors conclude that there is vast potential from the amalgamation of deterministic heuristics and evolutionary algorithms because of the way they complement each other. Similar conclusions have been found by Davis [48] and Reeves [136].

However, there is one distinct difference between the indirect approach used in this research and those usually reported elsewhere. All the examples in section 3.7, and to our knowledge most others found in literature, are ‘soft’ concerning one of the main constraints. In fact, often the problems are unconstrained and limiting a particular resource is part of the objective function. For instance, in a graph colouring problem, the objective is to minimise the number of colours to be used or there is the option of not colouring a node. Another example would be timetabling or job shop scheduling when there is no upper limit on the number of slots or days available. This makes finding a feasible solution very easy.

If not part of the objective originally, since they are still important, the number of colours, slots, days etc. are then optimised simultaneously with the other objectives. This is usually done by adding a weighted penalty to the fitness of each string depending on the amount used. Although this might look similar to a penalty function approach, there is a key difference. In these approaches, 'penalised' solutions are feasible. In our case, the number of nurses and the demand to be covered are strictly given. Thus, solutions that violate constraints are invalid because they cannot be used at all by the hospital.

Therefore, unless we can find a decoder that guarantees feasibility infeasible individuals will still be encountered. This is in contrast to the standard approaches outlined above where all solutions are feasible. To assign infeasible solutions a fitness score, the same penalty function approach as described in section 4.1 will be used. Thus, the fitness of a solution is calculated as

$$\sum_{i=1}^{n}\sum_{j=1}^{m} p_{ij}x_{ij} + w_{demand}\sum_{k=1}^{14}\sum_{s=1}^{p}\max\left[R_{ks} - \sum_{i=1}^{n}\sum_{j=1}^{m} q_{is}a_{jk}x_{ij};\ 0\right]$$

## 6.2 Permutation Crossover and Mutation

Before discussing how the decoding heuristic is constructed in the next section, permutation based genetic operators need to be introduced. Due to the difference between direct and indirect genetic algorithms described in the previous section, 'standard' crossover and mutation operators can no longer be used. This is because crossover operators such as one-point or uniform crossover would in most cases lead to duplicates in the list of permutations. An example of this is shown in appendix A.2.5 for one-point crossover.

Hence, new permutation based crossover operators have to be devised. Four of the most common of these operators are presented here: Order based crossover (similar to two-point crossover), C1 crossover (similar to one-point crossover), PMX (partially mapped crossover) and uniform order based crossover (similar to uniform crossover). After describing the crossover operators in detail, we will go on to compare their effects on a string of permutations.

Order based crossover is first presented in Davis [47]. It works with two parents and creates two children. Similar to two–point crossover, two crossover points are randomly chosen along the parent strings. The first child inherits the part of the string between the crossover points of the first parent and the second child the same part respectively from parent two. These sub-strings are kept in exactly the same position in the children as they were in the parents. The missing genes are now filled in from the other parent, that is parent two for child one and vice versa. The genes are taken in the order in which they appear in the appropriate parent and if not already present in the child, they are placed in the first free position of the child’s string.

Figure 6-1 is an example of an order based crossover. To create child 2 the values between the crossover points have been taken from parent 1. This results in the values 1, 2, 8 and 9 missing in child 2. In parent 2, they appear in the order of 1-8-9-2. Consequently, child 2 is completed by inserting the missing values in this order. Child 1 is formed by taking the values between the crossover points from parent 2. This leaves the values 2, 3, 4 and 5 out. As they appear in the order of 2-3-4-5 in the first parent, they are placed like this in the empty slots of child 1.

| | | | | | | | | | |
|---|---|---|---|---|---|---|---|---|---|
| 1 | 2 | 3 | 4 | 5 | 6 | 7 | 8 | 9 | Parent 1 |
| 3 | 4 | 7 | 1 | 6 | 8 | 9 | 2 | 5 | Parent 2 |
| | ^ ¦ ^ | | | | | ^ ¦ ^ | | | Crossover-Points |
| 1 | 8 | 3 | 4 | 5 | 6 | 7 | 9 | 2 | Child 1 |
| 2 | 3 | 7 | 1 | 6 | 8 | 9 | 4 | 5 | Child 2 |

Figure 6-1: Example of order based crossover.

A variant of order based crossover is C1 crossover as described by Reeves [136] and used by others without acquiring a definite name. C1 crossover works similarly to order based crossover but has only one crossover point. Children receive the part before the crossover point exactly as in one parent and the missing genes as ordered in the other parent. More details and an example of C1 crossover can be found in appendix A.2.5.

Partially mapped crossover (PMX) was invented by Goldberg and Lingle [80]. It derives its name from the fact that a portion of one string ordering is mapped onto a portion of another. Again, PMX uses two parents to create two children and two random cutting points are chosen. The sub-string between the cutting points is referred to as the mapping section. The children start as identical copies of their parents before a series of swaps is carried out. Then the two mapping sections are exchanged between the children. This will lead to most of the values in the mapping section being present twice in the children. To correct this a series of swaps is carried out. Each duplicated value is replaced by the value in the other parent's mapping section opposite the duplicate's counterpart in this parent's mapping section.

This operation is best explained with the example in Figure 6-2. As outlined above, child 1 starts as an identical duplicate of parent 1 and child 2 as a duplicate of parent 2. Then the mapping sections are defined, in this example they are 3-4-5 and 7-1-6. Next, the complete mapping sections are swapped between the children. In child 2, this results in the values of 3, 4 and 5 existing twice in the string. Therefore, they all have to be swapped in the manner described above. Firstly, 3 needs to be replaced. 7 is opposite 3 in the mapping section. Hence, the 3 outside the mapping section is replaced by 7. In the same manner, 4 is replaced by 1 and 5 is replaced by 6 in child 2. In child 1 the values 7, 1 and 6 are duplicated. Thus, the duplicated values outside the mapping section are replaced as follows: 7 by 3, 1 by 4 and 6 by 5.

| | | | | | | | | | |
|---|---|---|---|---|---|---|---|---|---|
| 1 | 2 | 3 | 4 | 5 | 6 | 7 | 8 | 9 | Parent 1 |
| 3 | 4 | 7 | 1 | 6 | 8 | 9 | 2 | 5 | Parent 2 |
| | ^ | ^ | | ^ | ^ | | | | Cutting Points |
| 4 | 2 | 7 | 1 | 6 | 5 | 3 | 8 | 9 | Child 1 |
| 7 | 1 | 3 | 4 | 5 | 8 | 9 | 2 | 6 | Child 2 |

Figure 6-2: Example of Partially Mapped Crossover (PMX).

The final operator presented here is uniform order based crossover, as described in Syswerda [161]. In line with the other two operators, two children are produced from two parents. However, the mechanism is quite different as it is a mixture of standard uniform crossover and order based crossover. No crossover points are chosen. Instead, a uniformly random binary template of the same length as the individuals is created. Child 1 takes the genes from parent 1 for every position where the binary template is one. Similarly, child 2 takes all genes from parent 2 when there is a zero in the template. The remainder of the children is filled in the same way as in order based crossover: Child 1 receives the missing genes in the order they appear in parent 2 and for child 2 it is vice versa. Figure 6-3 shows an example for clarification.

| | | | | | | | | | |
|---|---|---|---|---|---|---|---|---|---|
| 1 | 2 | 3 | 4 | 5 | 6 | 7 | 8 | 9 | Parent 1 |
| 3 | 4 | 7 | 1 | 6 | 8 | 9 | 2 | 5 | Parent 2 |
| 0 | 0 | 1 | 0 | 1 | 1 | 1 | 0 | 1 | Binary Template |
| 4 | 1 | 3 | 8 | 5 | 6 | 7 | 2 | 9 | Child 1 |
| 3 | 4 | 5 | 1 | 6 | 7 | 8 | 2 | 9 | Child 2 |

Figure 6-3: Example of order based uniform crossover.

A practical comparison of these and other permutation crossover operators is performed by Fox and McMahon [70] and Poon and Carter [128] for travelling salesman problems, by Adelsberger et al. [2] for a flow shop scheduling problem, by Bierwirth et al. [22] for job shop scheduling, and by Starkweather et al. [157] for a warehouse scheduling application. A more theoretical comparison using forma (i.e. schema in the permutation encoding case) analysis can be found in Cotta and Troya [43].

Most of the authors conclude that as in the case of standard crossover operators, there is no one ‘best’ crossover operator for all problems. The effectiveness of a crossover operator depends on the interaction between the problem-specific decoder and the operator’s disruptiveness of the following three types of orderings for genes: The relative order of the genes (e.g. A three positions before B), the absolute order of genes (e.g. A anywhere before B) or the absolute position of genes in the string (e.g. B in position 5).

Order based crossover preserves the order and position of genes of the first parent in the sub-sequence between the cutting points and the absolute order of the remaining genes from the other parent. On average one third of the first parent’s genes are kept unchanged both for position and order. A proof of this can be found in Cotta and Troya [43]. The remaining two thirds keep their absolute order as given in the second parent. For C1 crossover on average half of the first parent is kept unchanged and the other half of genes retain their absolute position from the second parent.

In the case of PMX, children are created that preserve the order and position of genes in the mapping sections from one parent. Again, this is on average roughly one third of the string. Another third of the string preserves the order and positions of genes from the other parent, as it remains untouched by the swapping mechanism. The final third will have different absolute positions and probably orderings depending on the actual mapping. One can anticipate that the proportion of genes in the last third becomes less and less as convergence sets in. This is because strings become increasingly similar and genes will therefore be more likely to be mapped onto themselves.

Under uniform order based crossover on average half of all genes from each parent retain their original position and hence order. The other half keeps the absolute order from the other parent. Although this sounds similar to the effect of C1 crossover, there is one major difference due to the nature of the binary template. With C1 crossover one large chunk of either parent is kept. Therefore, there is a strong bias towards keeping genes together that are close in the string. A similar situation arises when filling the second half of the string from the other parent. This bias does not exist for uniform order based crossover. Hence, uniform order based crossover is more flexible or disruptive depending on one’s point of view.

Thus, the operators sorted in increasing order of disruptiveness to the absolute string positions of the genes are: Partially mapped crossover, C1 crossover, uniform order based crossover and order based crossover. However, this does not take into account that with the use of PMX one third of the genes retain neither the absolute nor the relative position of either parent. With all other operators, at least the relative position is kept from either parent. Hence, one could also argue that PMX is the most disruptive crossover. In section 6.4 all four crossover operators are tested for the nurse scheduling problem and we try to find an answer for this question in our particular case.

Finally, the mutation operator must also be different from the ones used with canonical direct genetic algorithms. This again is because genes might otherwise become duplicated or omitted. Davis [48] suggests two operators. Firstly, a swap mutation, which is the most common in genetic algorithms with order based encodings. It operates by swapping the position of two elements. The second operator, called scramble sublist mutation, is more disruptive. It works by randomly choosing two points in the string and then randomly reordering the positions of symbols between these two points. Both strategies will be experimented with in section 6.4.

## 6.3 The Decoder Function

### 6.3.1 Encoding

In this section, we will first describe an indirect genetic algorithm approach whose main feature is a heuristic decoder that transforms the genotype of a string into its phenotype. After discussing which type of permutation to choose for our encoding, two different decoders will be presented and compared. Finally, it is investigated if the rules for efficient decoders set up in section 3.7 are fulfilled and explain possible conflicts between our decoders and the rules.

An indirect genetic algorithm is very similar to a canonical direct genetic algorithm as described in appendix A.2. In fact, it is identical apart from requiring different crossover and mutation operators. The need for different operators, because of the order-based encoding, is explained in detail in section 6.2. One additional step is also needed, since the genotype of an individual is no longer the same as the phenotype. Thus, a decoder is necessary to transform one into the other. One way of looking at this is as an extended fitness function calculation. Rather than directly deducting the fitness of a string, an intermediate step is necessary to transform the genotype of an individual into an actual solution to the problem.

The first decision, before using a decoder based genetic algorithm, has to be what the genotype of individuals should represent. Essentially, there are two possibilities in our case: The encoding can be either a permutation of the nurses to be scheduled or a permutation of the shifts to be covered. For comparison's sake, the encoding has to be similar to the one used in the direct genetic algorithm. The equivalent to that is the permutation of the nurses, as this would again satisfy the multiple-choice part of the constraint sets. However, before deciding, consideration needs to be given to the other possibility, as it might prove more effective.

The decision which of the two to choose, follows a similar pattern to that of deciding on the encoding for the direct genetic algorithm. In section 4.1, it was explained that using an encoding which is a list of the shift patterns worked by the nurses is superior to having an encoding which is a list of the nurses covering the shifts. This is because it is easier to fix possible violations of the demand constraints than fixing the constraints regarding the number of shifts nurses must work.

Here we argue in a similar fashion. If the string was a permutation of the nurses, then the decoder would have to assign single shifts or whole shift patterns to them. To keep the decoder simple, all shifts have to be assigned at once to a particular nurse. Because a nurse must work one of her shift patterns, assigning all her single shifts in one go will give the same results as assigning a shift pattern. Therefore, one can subsume assigning single shifts with the more general case of assigning whole shift patterns. This again has the advantage that the constraint sets (1) and (2) of the integer program formulation are implicitly fulfilled. As before, the disadvantages of this approach are that there is no guarantee that the demand is covered and that only whole shift patterns can be changed resulting in a loss of flexibility.

However, if strings as permutations of the shifts to be covered are used, then the decoder would have to assign nurses to them. Without any further enhancements, this would result in the decoder having to overcome the same difficulties as described in section 4.1 for the similar encoding. In other words, a simple decoder may assign a nurse to more than one shift per day / night or to too many shifts per week. This problem can be reduced with an intelligent decoder. For instance, a 'look-ahead' operator could address this issue. Such a sophisticated device is necessary, because simply assigning a nurse to a shift will often result in situations where for the last nurse no assignable shift leads to a feasible solution. However, this would make this type of decoder much more complicated than the decoder for the other encoding.

Another argument to consider in the decision is the following. Intuitively, the main effect of using a permutation based encoding and a decoder is that objects which are difficult to schedule eventually appear early in the string, when there is still a lot of flexibility left. This could, for instance, be nurses that can only work a limited number of shift patterns or shifts that can only be covered by a small number of nurses. In our problem, some nurses can only work a limited number of shift patterns. This can be due to shift patterns worked in previous weeks or because of requested days off. On the other hand, although some shifts are less popular than others, it is difficult to find shifts that can only be covered by a few nurses. This is because there is always a surplus of top graded nurses who by definition are allowed to cover all shifts.

The final argument to consider about which encoding to choose are possible conflicts with the building block hypothesis due to the length of the string. As mentioned previously, the longer the string, the more likely it is that successful building blocks are split up, as they tend to be longer, too. A permutation of shifts based string would comprise of 14 genes, calculated as seven days plus seven nights. An individual based on a permutation of the nurses has as many genes as there are nurses on the ward, i.e. it is approximately of length 25. Thus, there is little difference between the two encodings in this respect.

Considering all the points made above, there seems to be an advantage in using the permutation of nurses over the permutation of shifts. This requires a less sophisticated decoder and is more in tune with the idea of difficult items appearing early in the string. Additionally, it is similar to the encoding chosen for the direct genetic algorithm, which allows for results to be compared on a more equal footing. Thus, the permutation of nurses is chosen as an encoding for our indirect genetic algorithm.

### 6.3.2 Decoder Details

Now that we have decided that our genotypes are permutations of the nurses, a decoder that builds a schedule from this list has to be constructed. This 'schedule builder' needs to take into account those shifts that are still uncovered. Additional points to consider are the grades of nurses required, the types and qualifications of the nurses left to be scheduled and the cost $p_{ij}$ of a nurse working a particular shift pattern. Thus, a good schedule builder would construct feasible or near-feasible schedules, where most nurses work their preferred shift patterns.

First, two simple decoders are presented for this task. The first covers those days and nights with the highest number of uncovered shifts. The second decoder schedules shift patterns to maximise their overall contribution regarding the covering requirements of all three grades. A comparison with the decoder rules set up by Palmer and Kershenbaum [125] in section 3.7 follows and computational experiments (section 6.4) will show weaknesses of these decoders. To correct these, a third decoder that combines elements from the other two is presented in section 6.5.1 and shown to solve the nurse scheduling problem very successfully.

The first decoder, referred to as 'cover highest' in the future, is relatively simple to construct. Due to the nature of this approach, a nurse's requests or $p_{ij}$ costs are not taken into account by the decoder. However, they will influence decisions indirectly via the fitness function, which decides the rank of an individual and the subsequent rank-based parent selection. The level of influence will obviously depend on the setting of the penalty weight $w_{demand}$. The higher the weight, the less likely that this effect will filter through.

The decoder constructs solutions as follows. A nurse works *k* shifts per week. Usually these are either all day or all night shifts (standard type). In some special cases, they are a fixed mixture of day and night shifts (special type). Therefore, the first step in the cover highest decoder is to find the shift with the biggest lack of cover. This will decide whether the nurse will be scheduled on days or nights if she is of the standard type. We then proceed to find the *k* day or *k* night shifts with the highest undercover. If she is of the special type, we directly find the *k* shifts with the highest undercover, taking into account the number of day and night shifts worked by this particular nurse. The nurse is then scheduled to work the shift pattern that covers these *k* days or nights.

For nurses of grade *s*, only the shifts requiring grade *s* nurses are counted as long as there is a single uncovered shift for this grade. If all these are covered and $s < 3$, uncovered grade $(s+1)$ shifts are taken into account. Once those are filled and $s < 2$, grade $(s+2)$ shifts are considered. This operation is necessary, otherwise higher graded nurses might fill lower graded demand, whilst higher graded demand might not be met at all. Note that if there is more than one day or night with the same number of uncovered shifts, then the first one is chosen. For this purpose, the days are searched in Sunday to Saturday order.

The second decoder, called the 'overall contribution' decoder, works differently. It goes through all feasible shift patterns of a nurse and assigns each one a score. The one with the highest score is chosen. If there is more than one shift pattern with the best score, the first such shift pattern is chosen. For this purpose, shift patterns are searched in increasing index order as sorted in appendix C.4, with the search starting at the first possible day shift pattern of a nurse. Unfortunately, there is a potential flaw in this simple search order. The first nurses to be scheduled will always be assigned the first of their shift patterns with the lowest $p_{ij}$ value. This seriously reduces the diversity of solutions created by the decoder. To counter this, other search orders are investigated in section 6.4.3.

The score of a shift pattern is then calculated as the weighted sum of the nurse's $p_{ij}$ value for that particular shift pattern and its overall contribution to the cover of all three grades. The latter is measured as a weighted sum of grade one, two and three uncovered shifts that would be covered if the nurse worked this shift pattern, i.e. the reduction in shortfall. Obviously, nurses can only contribute to uncovered demand of their own grade or below.

More precisely, the score $s_{ij}$ of shift pattern $j$ for nurse $i$ is calculated with the following parameters:

- $d_{ks} = 1$ if there are still nurses needed on day $k$ of grade $s$ otherwise $d_{ks} = 0$.
- $a_{jk} = 1$ if shift pattern $j$ covers day $k$ otherwise $a_{jk} = 0$.
- $q_{is} = 1$ if nurse $i$ is of grade $s$ or higher otherwise $q_{is} = 0$.
- $w_s$ is the weight of covering an uncovered shift of grade $s$.
- $w_p$ is the weight of the nurse's $p_{ij}$ value for the shift pattern.

Finally, $(100 - p_{ij})$ must be used in the score, as higher $p_{ij}$ values are worse and the maximum for $p_{ij}$ is 100. Note that $(-w_p p_{ij})$ could also have been used, but would have led to some scores being negative. Thus, the scores are calculated as follows:

$$s_{ij} = w_p(100 - p_{ij}) + \sum_{s=1}^{3} w_s q_{is} \left( \sum_{k=1}^{14} a_{jk} d_{ks} \right)$$

The ‘overall contribution’ decoder is more complex than the ‘cover highest’ decoder, but has certain advantages. In contrast to the simpler previous decoder, it considers the preferences of the nurses and also tries to look ahead a little. An example of the second advantage is a situation where there are more grade two shifts uncovered than could be covered by grade two nurses yet to be scheduled. Also, suppose that there are still enough grade one and two nurses left to cover the demand for grade one and two shifts together. The cover highest decoder would first cover all grade one demand. If that was awkwardly spread, some grade one cover might be wasted on days where it is not required (and where neither grade two cover is required). This could happen because ‘whole’ nurses are scheduled at a time, i.e. up to five shifts at once. Only once all grade one demand is fulfilled would it consider uncovered grade two shifts.

The overall contribution decoder on the other hand looks ahead and might cover the grade two shifts early enough with grade one nurses. However, there is a drawback of this look-ahead behaviour: The decoder might waste higher graded nurses too early on shifts where they are not strictly required to work. This could be due to the $p_{ij}$ costs or due to the weights associated with the uncovered shifts of the three grades. In section 6.4, we will have a closer look at this by experimenting with various weight settings.

### 6.3.3 Decoder Rules and Counterexample

To judge the efficiency of the two decoders presented above, let us recapitulate the rules set up for efficient decoders by Palmer and Kershenbaum [125] and summarised in section 3.7 as:

1) For each solution in the original space, there is a solution in the encoded space.
2) Each encoded solution corresponds to one feasible solution in the original space.
3) All solutions in the original space should be represented by the same number of encoded solutions.
4) The transformation between solutions should be computationally fast.
5) Small changes in the encoded solution should result in small changes in the solution itself.

It can be proven that the first rule is violated by both decoders. For the cover highest decoder, imagine that there are only two nurses present and only days have to be considered. Furthermore, assume that nurse 1 works four days and nurse 2 three days. If one shift has to be covered on every day, then nurse 1 working (1000111) and nurse 2 working (0111000) would be a feasible solution. However, due to the fixed search order of shift patterns, the decoder would never construct this solution, nor would it find many others. The only two solutions it is capable of finding are the following two. If nurse 1 is scheduled first, she will be given the pattern (1111000), whilst nurse 2 receives (0000111). If nurse 2 is scheduled first, she will work (1110000) and nurse 1 will be scheduled as (0001111).

For the overall contribution decoder suppose that the demand to be covered is the same for all days. Thus, the first nurse to be scheduled will end up working the shift pattern with the lowest $p_{ij}$ value. This is true for any nurse that would be scheduled first and hence there is always at least one nurse in any decoded solution that works her cheapest shift pattern. Thus, a solution where no nurse works any of her shift pattern with the lowest $p_{ij}$ value would never be constructed.

To visualise this, consider the example above with two nurses and one shift to be covered per day. Furthermore, assume that both nurses can only work two patterns. Nurse 1 can work (1111000) with $p_{11} = 10$ or (0001111) with $p_{12} = 1$. Nurse 2 can work (0000111) with $p_{21} = 10$ or (0111000) with $p_{22} = 1$. Thus, if nurse 1 is scheduled first she will be given her second pattern because it has a lower $p_{ij}$ value whilst covering the same number of uncovered shifts as her other pattern. Unfortunately, neither of the two patterns of nurse 2 will now make the solution feasible. The same is true if nurse 2 is scheduled first. She will receive her second pattern and again, neither of the patterns of nurse 1 can complete the solution to feasibility. The optimal solution, i.e. both nurses working their first pattern, cannot be constructed by the decoder.

The violation of rule 1) by both decoders cannot be taken lightly as this possibly means that the optimal solution is excluded from the solution space. However, the larger the number of nurses to be scheduled, the more flexibility is introduced to schedule nurses onto different patterns. In our problems, there are usually between 20 and 30 nurses. Although it is still possible to create theoretical counterexamples, practically they are very unlikely to contain the optimal solution. For instance, for the 'overall contribution' decoder, this would mean that none of the 20-30 nurses works any of her shift patterns with the lowest $p_{ij}$ value. This is extremely unlikely, as the objective is to minimise the sum of the $p_{ij}$ values. Practically, this situation does not occur in any of the 52 data sets. However, it cannot be ruled out in general and will remain a permanent disadvantage of our decoders.

Feasibility as in rule 2) can no longer be guaranteed either, as tight constraints have to be observed. Otherwise, an unlimited supply of nurses, respectively overtime, would be necessary which is strictly forbidden. This is a problem-specific issue and cannot be changed. Therefore, a penalty function approach is still necessary. The penalty function used is the same as before and outlined in section 4.1. Nevertheless, because all decoder operations are deterministic, each permutation still decodes to only one solution in the original space as required by the rule. This second aspect is important, as it ensures a one-to-one relationship between the encoded solution, the decoded solution and the fitness function value. Without this, the genetic algorithm might run into difficulties, as the basic principle of allocating more reproductive trials to fitter individuals would be violated.

Again, rule 3) is not fulfilled by either decoder. Both methods are designed to aim for feasible solutions, with an additional bias towards cheaper solutions for the second decoder. Thus, those parts of the solution space containing high-cost or highly infeasible solutions are unlikely to be reached by our decoders. As outlined in section 3.7, this is not a problem, as those parts are not of interest to us, since we are indeed looking for low-cost feasible solutions.

Transformations between decoded and encoded solutions are relatively fast, as requested by 4). However, particularly for the second more complicated decoder, the time spent on them is non-negligible. Thus, to keep the optimisation time in the same range as before, the population sizes will have to be reduced in comparison to the direct genetic algorithm. Finally, rule 5) is not observed as small changes in the permutation might lead to big changes in the actual solution. Only if two identical nurses swap their positions in the string, will there be no changes to the solution. However, if they are of a different grade or have different preferences, then there will be a 'domino effect' to the shift patterns worked by all nurses after the first swapping position.

Overall, it seems that most of the rules are not observed by the decoders. However, as detailed in section 3.7, this does not render them useless. Since they are purposely biased to promising regions of the solution space, the exclusion of valuable solutions is unlikely. Moreover, this can possibly be further corrected with an appropriate choice for the parameters and weights. This and further issues arising from the decoders, along with results to the nurse scheduling problem will be looked at in the next section.

## 6.4 Parameter Testing

### 6.4.1 General Introduction

For these experiments the replacement and stopping criteria, mutation probability and penalty weight were maintained as those derived in section 4.3. These parameters still worked well and it was thought that no significant improvements could be achieved by further tweaking. One reason for this is that both the string length and fitness function are identical to the ones used for the direct genetic algorithm. However, there are some major differences to the direct algorithm.

Firstly, permutation based crossover and mutation operators must be used as explained in section 6.2. Secondly, the size of the solution space is different. For *n* nurses each working one of 40 possible shift patterns, the solution space size is $40^n$ in the direct and $n!$ in the indirect genetic algorithm. Typically, there are 25 nurses to be scheduled, which means that the direct solution space is of size $40^{25} \approx 10^{40}$ and the indirect of size $25! \approx 2*10^{25}$. Additionally, as explained in the previous section, it is in the nature of the decoder to be biased towards certain regions of the solution space. Thus, the size of the actual solution space sampled by the indirect genetic algorithm is many magnitudes smaller than for the direct genetic algorithm.

Therefore, smaller population sizes can be used without jeopardising the quality of the results. On the other hand, there is non-negligible computational effort necessary to decode solutions. To arrive at comparable run times of around 15 seconds per optimisation run, the population size has to be reduced from 1000 to 100. In view of the reduced size of the solution space, this seemed a reasonable setting and was consequently adapted for all experiments. Hence, the following parameters and strategies were used for the remainder of this chapter:

| Parameter / Strategy | Setting |
|---|---|
| Population Size | **100** |
| Population Type | **Generational** |
| Initialisation | **Random** |
| Selection | **Rank Based** |
| Crossover | **Order-Based** |
| Swap Mutation Probability | **1.5%** |
| Replacement Strategy | **Keep 10% Best** |
| Stopping Criteria | **No improvement for 30 generations** |
| Penalty Weight | **20** |

Table 6-1: Parameters and strategies used for the indirect genetic algorithm and nurse scheduling.

### 6.4.2 Cover Highest Decoder

The first set of experiments is designed to investigate the cover highest decoder. Three types of experiments were carried out. Firstly, to establish the extent of work done by the decoder itself, it was tested on its own without a genetic algorithm. Then the decoder is combined with a genetic algorithm and experiments with different permutation based crossover operators are carried out. Finally, the benefits of using a special form of seeding are investigated. The results of all tests are shown in Figure 6-4. The average solution time for all experiments was some 15 seconds per single run.

The results under the 'no GA' label are for the decoder on its own processing 10000 random solutions. This is equivalent to approximately 100 generations with a genetic algorithm, which is roughly the length of a typical genetic algorithm run. The results show that the decoder alone is not capable of solving the problem with less than 5% of runs finding a feasible solution.

Next, Figure 6-4 shows a comparison of the performance of the four different crossover operators described in section 6.2 in conjunction with the cover highest decoder. Ordered by the quality of results, starting with the best, they are PMX, uniform order based crossover, C1 and order based crossover. This indicates that the higher the percentage of genes staying in their absolute positions the better. Moreover, in the case of two operators with the same percentage, i.e. uniform order based and C1 crossover, the more disruptive and thus more flexible uniform crossover performs better.

These are interesting results and will lead us to develop a new crossover operator based on a combination of PMX and uniform order based crossover in section 6.5. Overall, the results are better than those found by the simple direct genetic algorithm (pictured in Figure 5-6) in terms of feasibility. However, the cost of solutions is much higher. As stated previously, the $p_{ij}$ values are not taken into account by this decoder, but it was thought that there might be an indirect effect via the fitness function. This has been proven wrong. In the next section, it will be investigated if the overall contribution decoder, which considers the $p_{ij}$ values directly, performs better.

Finally, under the 'Sorted' label, the strings are no longer initialised as a random permutation of the nurses. Instead, they are 'seeded' such that grade three nurses appear first in the string, then grade two nurses and finally grade one nurses. The idea behind this is that higher graded nurses are more 'flexible' than lower graded ones, because higher graded nurses can cover for lower grades. Thus, because they are more flexible they can be scheduled later and hence appear towards the end of the string. For this experiment, C1 crossover was used to maintain the grade order. Note that this operation is in fact more than mere 'seeding', as it restricts all future children to the same structure.

As the graph shows, this 'seeding' produces worse results. This might sound surprising at first, but biasing the string by seeding it in the above way seriously reduces diversity in the population. Moreover, as explained in section 4.3.2, this can reduce solution quality as had happened in this case. Furthermore, not all higher graded nurses are necessarily more flexible than lower graded nurses. For instance, some nurses, irrespective of their grade, can only work a very limited number of shift patterns due to days off or last week's schedule. To summarise, this shows that scheduling the nurses is more complex than just sorting them by 'flexibility' and then assigning shift patterns.

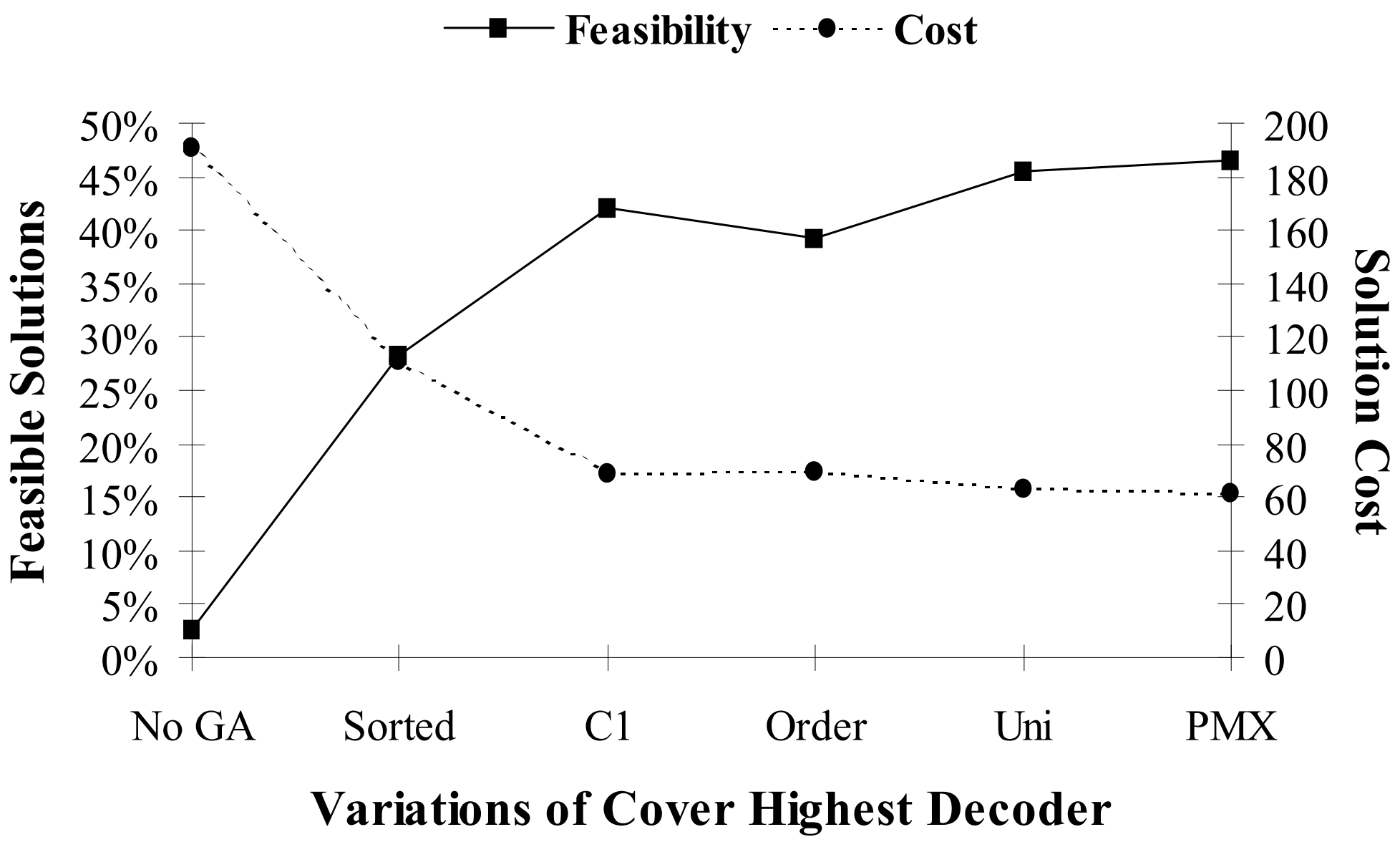

Figure 6-4: Crossover operators and other variations for the cover highest decoder.

### 6.4.3 Overall Contribution Decoder

The first tests carried out for this decoder were the same as in the previous section, i.e. four different crossover operators, a grade-seeded string and the decoder on its own without a genetic algorithm. Again, for the decoder without a genetic algorithm, 10000 randomly generated solutions were decoded in each optimisation run. Before conducting a set of experiments concerning the internal decoder weights, these weights for the contribution of a shift pattern to the four objectives, namely grade one, two and three cover and the preferences of the nurses were set intuitively to $w_1{:}w_2{:}w_3{:}w_P = 4{:}2{:}1{:}1$.

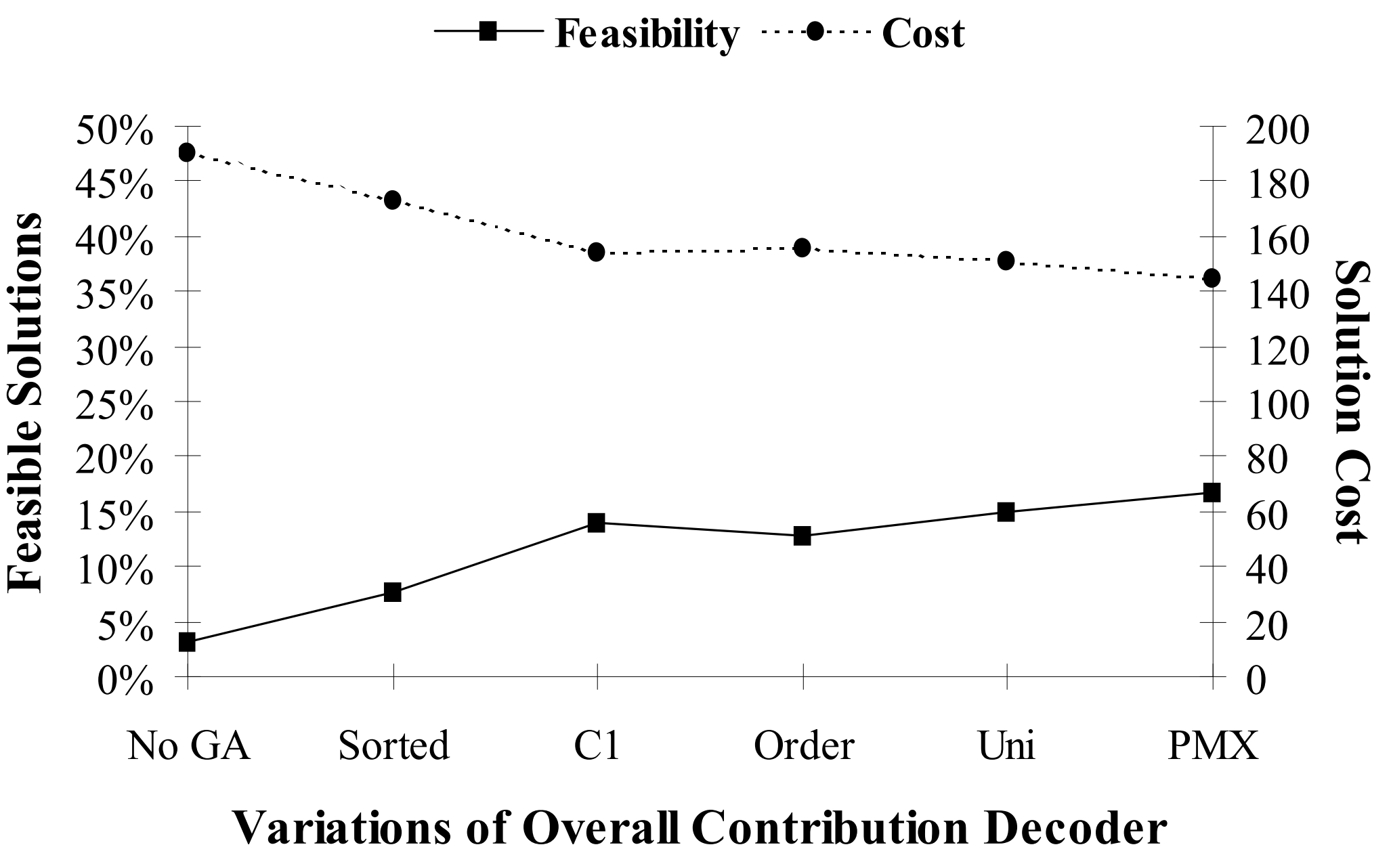

Figure 6-5: Crossover operators and other variations for the overall contribution decoder.

The results of the first set of experiments are shown in Figure 6-5. As the graph shows, the behaviour of this decoder is similar to the previous one. Again, using the decoder on its own or seeding the string sorted by grades produces far worse results than a normal indirect genetic algorithm approach. For the four crossover operators, the results are also similar to those in Figure 6-4: PMX performs best, followed by uniform, order-based and C1 crossover. However, the overall results are poorer than those found by the cover highest decoder. This could possibly be because of the arbitrary choice of decoder weights. The next set of experiments is designed to examine this issue.

In Figure 6-6 the results of experiments with different weights are shown. The labels give the weights used as $w_1$:$w_2$:$w_3$:$w_P$. Clearly the results are very poor for any weight setting, although the ratio *8:2:1:1* performed slightly better than the others and will be used for future experiments. As the results were poor for all weight combinations, the fault had to be elsewhere. Thus apart from the decoder operation itself, the most likely reason for the failure was the search order of shift patterns.

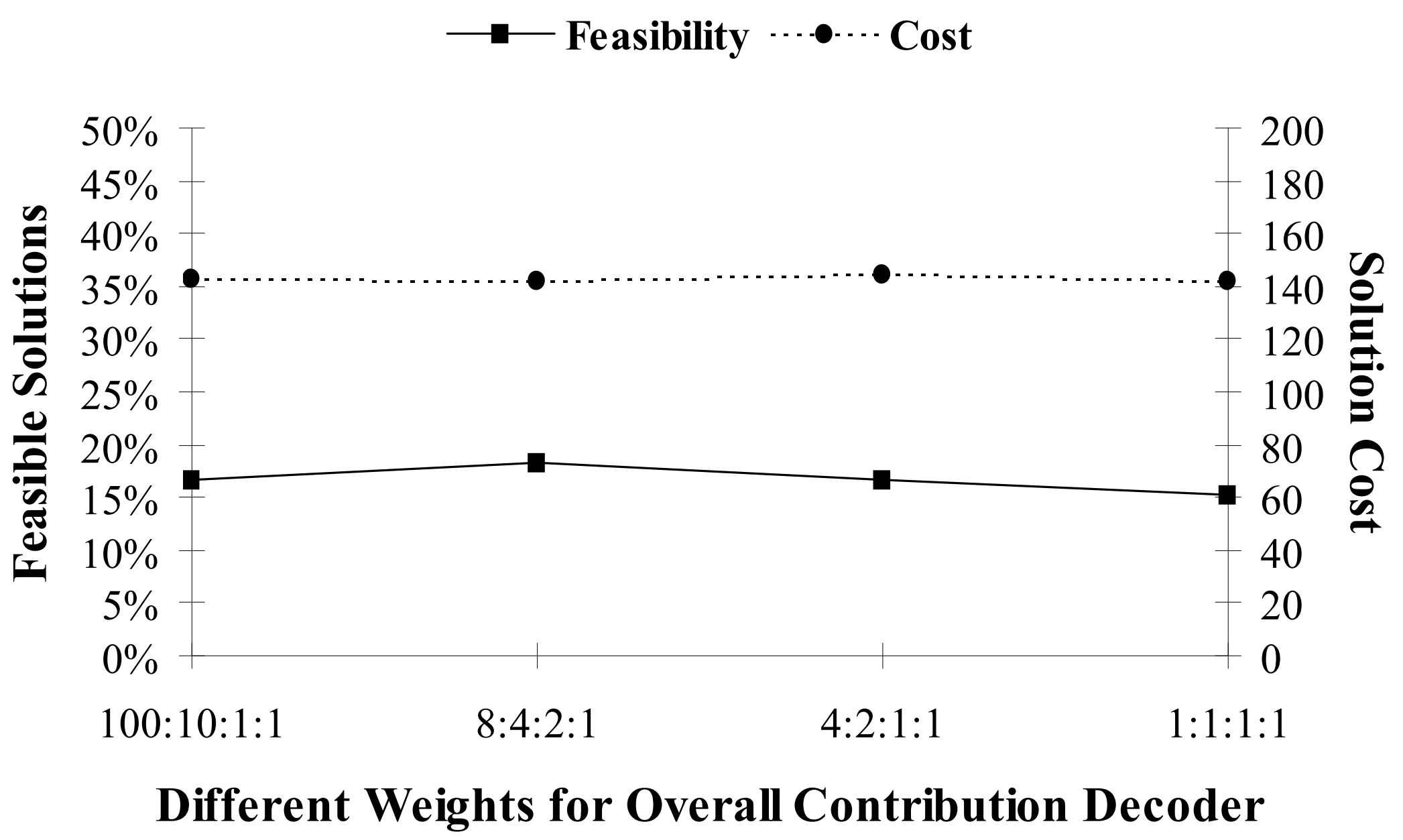

Figure 6-6: Weight ratios for the overall contribution decoder.

We recalled that by definition the decoder searches through the shift patterns of a nurse starting from her first possible day shift pattern and then follows the order of shift patterns as given in appendix C.4. One also needs to remember that if there are multiple shifts with the same score, the first one achieving such a score is picked. This typically happens for the first few nurses to be scheduled, as all shifts are uncovered at this stage. Therefore, until there is at least on fully covered shift, nurses are always given one particular low-cost shift pattern. Thus, the search is biased towards assigning patterns that include the start of the week. This greatly reduces diversity amongst the solutions and leads to very poor results as shown in Figure 6-6.

Thus, different search orders are introduced to correct this behaviour. Figure 6-7 shows the results of different types of search orders compared to the 'lowday' one used previously. In the following, 'random' search orders refer to fixed random orders, i.e. the shift patterns are shuffled (separately for days and nights) for each nurse before each optimisation run and then always searched in that order.

The simple random order (label 'Rand Order') starts randomly with the first day or night shift pattern of a nurse. It then follows the order as shuffled, continuing with the first night (day) shift pattern once the last day (night) pattern is reached. The biased search order (label 'Biased') does the same. However, it starts with a 75% probability with a day shift pattern. This has been done because approximately 75% of nurses work day shift patterns rather than night shift patterns, as the demand is much higher on days.

For the final two search orders, a nurse's shift patterns are ordered by increasing $p_{ij}$ values, with the lower index number of a shift pattern as tiebreaker. The cheapest order (label 'Cheapest') starts the search with the shift pattern with the lowest $p_{ij}$ value, whilst the random cost (label 'Rand Cost') search starts with a random shift pattern and then follows the cost order. Again, if the highest cost shift pattern is reached by the random cost ordering, the search continues with the lowest cost shift pattern until the search has come full circle.

The graph in Figure 6-7 shows that the random search orders achieve much better results than the 'deterministic' ones following the shift patterns as indexed. Of the three orderings with random starting points, the biased search does slightly better than the other two. Overall, random search orders significantly improve the results of the overall contribution decoder. However, in comparison to previous approaches, the results are only slightly better than those found by the simple direct genetic algorithm.

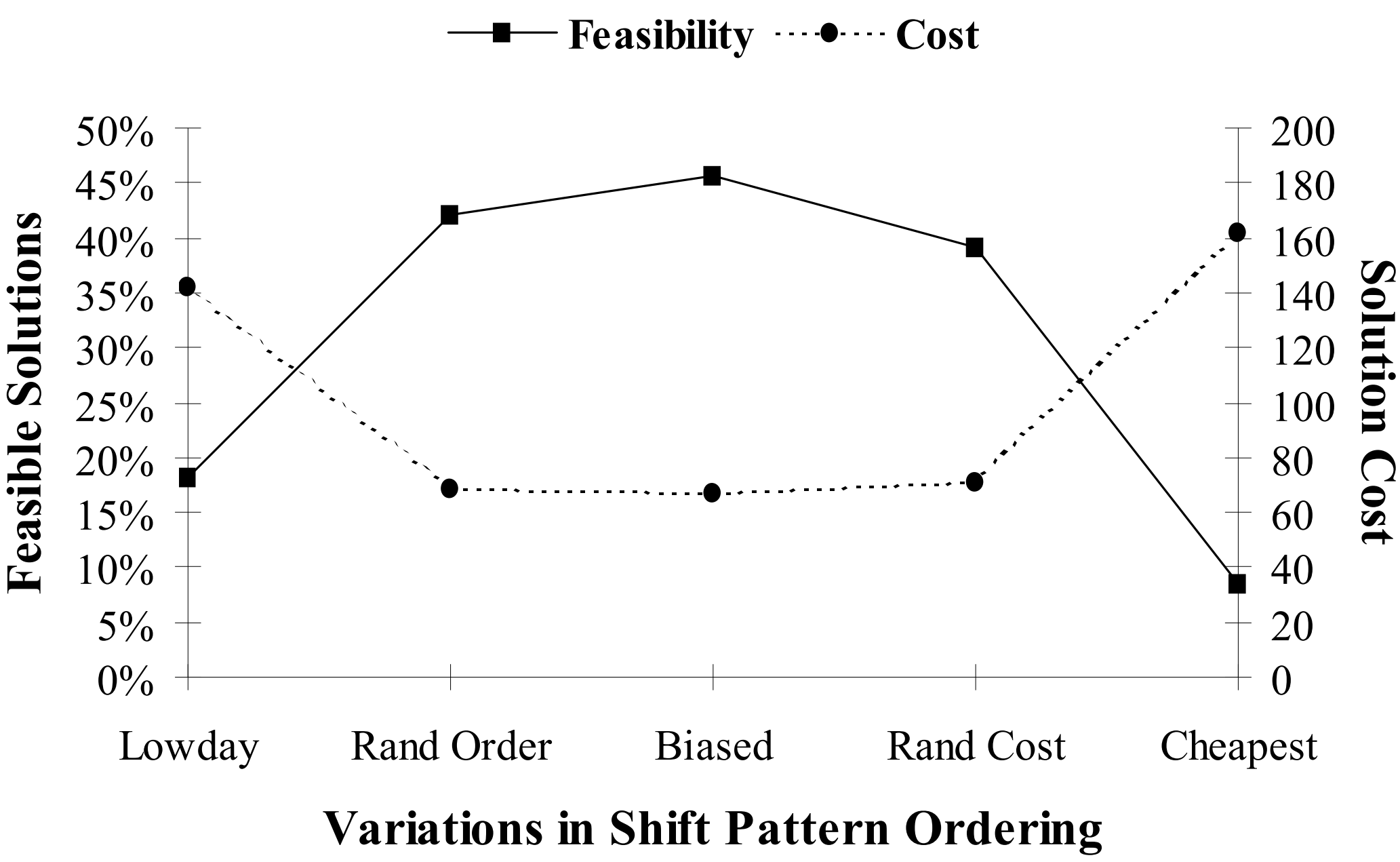

Figure 6-7: Different shift pattern orderings for the overall contribution decoder.

We conjecture that the problems of the two decoders, as described in section 6.3.2, prevented the results from becoming any better. For the cover highest decoder the problems were that is does not take the $p_{ij}$ values into account and that there is no looking ahead. The disadvantage of the overall contribution decoder is the possible wasting of higher graded nurses when they are not strictly required. Hence, in the next section a new decoder is developed to overcome these problems by combining the positive features of both the cover highest and overall contribution decoders.

## 6.5 Decoder Enhancements

### 6.5.1 Combined Decoder

Both decoders used so far only produced mediocre results because each one had some drawbacks. In this section, we will try to correct this by merging both attempts into a new decoder, which is called the combined decoder. This new decoder takes into account both the overall contribution of a shift pattern for all three grades, the preferences of the nurses and the days with the highest number of uncovered shifts.

The combined decoder also calculates a score $s_{ij}$ for each shift pattern and assigns the shift pattern with the highest score to the nurse, breaking ties by choosing the first such shift pattern. To avoid bias, a fixed random search order will be used. However, this time a shift pattern scores more points for covering a day or night that has a higher number of uncovered shifts. Hence, $d_{ks}$ is no longer binary but equal to the number of uncovered shifts of grade $s$ on day $k$. Otherwise using the same notation as in section 6.3.2, the score $s_{ij}$ for nurse $i$ and shift pattern $j$ is calculated as follows:

$$s_{ij} = w_p(100 - p_{ij}) + \sum_{s=1}^{3} w_s q_{is} \left( \sum_{k=1}^{14} a_{jk} d_{ks} \right)$$

As this decoder is very similar to the overall contribution decoder, no separate experiments concerning the crossover operator and search order were carried out. Due to its poor performance, grade-based seeding was not tried and as before the decoder on its own optimising 10000 random solutions performed very poorly. However, due to the different nature of the contribution elements, new experiments concerning the covering and preference weights were necessary. For these, the weights for contribution to cover were fixed at values for $w_1{:}w_2{:}w_3$ of $8{:}2{:}1$. Experiments were then carried out on a range of values for the preference weight $w_p$.

Figure 6-8 shows the outcome of a series of experiments for different values of $w_p$, indicated by the x-axis labels. In these experiments PMX crossover and a random biased search order of shift patterns to break ties were used. These values and strategies were chosen as they gave best results for the similar overall contribution decoder. As the graph shows, the results are of excellent quality. For any value of $w_p$ in the range

tested, cost and feasibility of solutions was better than for the cover highest (label 'High') and overall contribution (label 'Cont') decoders.

The behaviour for variations of $w_p$ is as expected. If it is set too low, then solutions are very likely to be feasible but are of high cost. If $w_p$ is set too high, solution quality rapidly drops due to the bias towards cheap but infeasible solutions. A value of $w_p = 0.5$ gives the best results, sacrificing only a small amount of feasibility for a good improvement in cost. With this setting, which is used for future experiments, the solutions are of higher feasibility than ever achieved with the direct genetic algorithm. However, the cost of solutions is slightly worse than for the best direct genetic approach. Thus, further improvements are necessary. These are considered in the following sections.

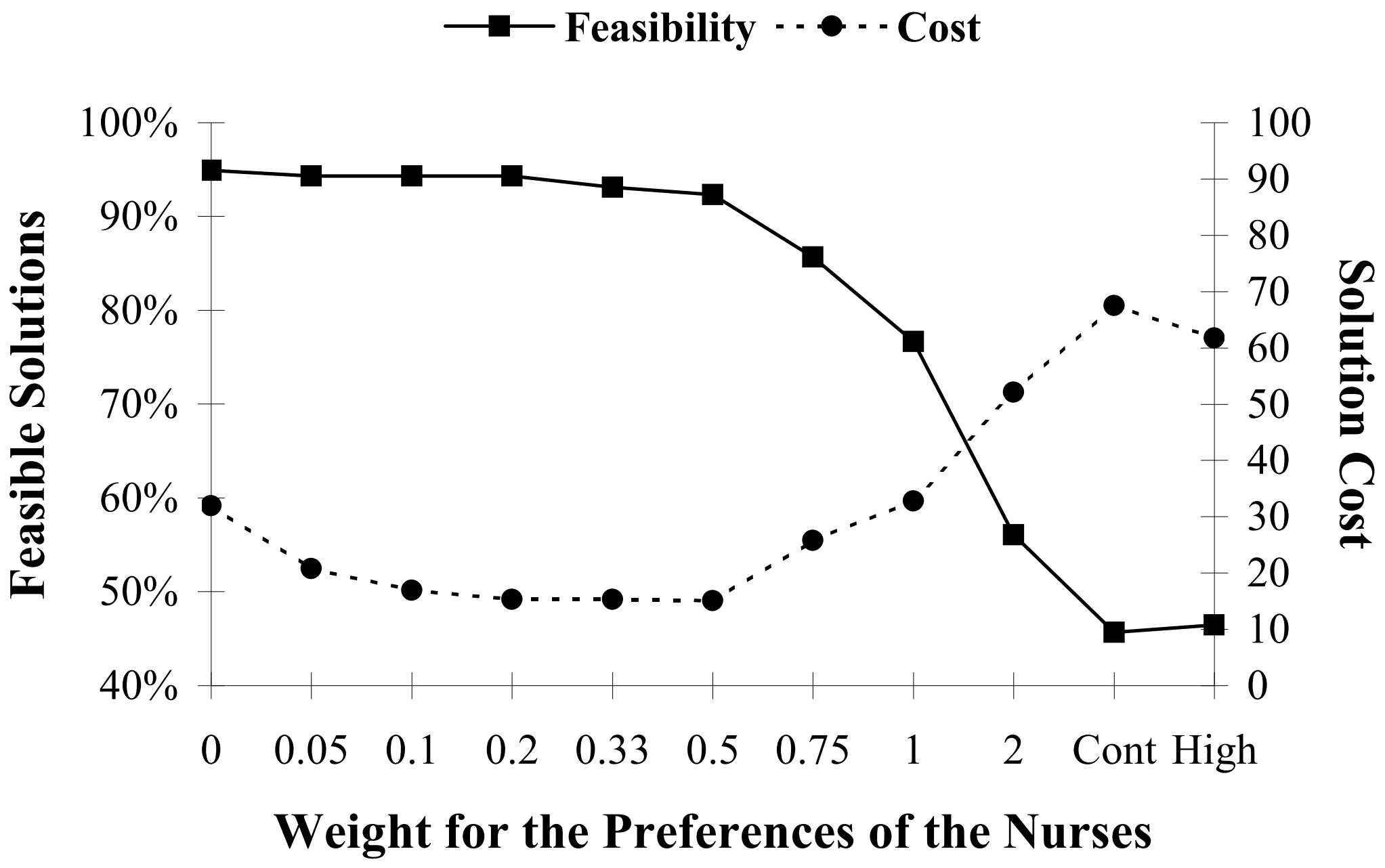

Figure 6-8: Combined decoder with different preference weights.

### 6.5.2 Parameterised Uniform Order Crossover (PUX)

This section is concerned with experimenting with a new type of crossover and the scramble mutation operator. This new crossover operator is inspired by the results found in sections 6.4.2 and 6.4.3 and displayed in Figure 6-4 and Figure 6-5. There it was shown that PMX outperformed the other operators. It was conjectured that this was because of the higher number of genes left in their absolute positions. Amongst the other operators, uniform order based crossover was best. This was attributed to its flexibility or disruptiveness.

To capitalise on this, a new crossover operator utilising both these advantages is introduced. This new operator will be called parameterised uniform order crossover or PUX for short. PUX works in a similar way to uniform order based crossover as described in section 6.2. However, when creating the binary template, the probability for a one will be equal to a parameter $p$, similar to standard parameterised uniform crossover. For instance, if $p = 0.66$, then there will be a 66% chance for a one and hence a 66% chance that a gene will be in the same absolute position as in one of the parents.

Thus, PUX with $p = 0.66$ has an equal probability of keeping a gene in the same absolute position as PMX. Moreover, PUX has an additional advantage. Whilst with PMX the remaining 33% of genes were positioned without any reference to the parents, PUX retains the absolute order of these as found in the second parent. However, in line with other uniform crossover operators, PUX is disruptive in a sense that it does not transmit large chunks of the parents.

We experimented with various values for $p$ using the new combined decoder with optimal weights, as described in the previous section. Results for this and a comparison to C1 and PMX crossover are shown in Figure 6-9. The value after each PUX label indicates the percentage used for $p$. As the graph shows, particularly for the cost of solutions, results are further improved with an appropriate choice for $p$. For instance, for $p = 0.66$ feasibility is as high as for PMX, but solution cost is significantly lower. This proves that our hypotheses about the necessary qualities of a successful crossover operator for our problem, as detailed in the first paragraph of this section, were right.

Another interesting observation is that the higher the value of *p* the lower the feasibility of solutions. This indicates that a more disruptive crossover, i.e. $p = 0.5$, is more flexible and has the power to create feasible solutions when the other operators fail. On the other hand, solution cost is best for a medium setting of *p*, i.e. $p = 0.66$ or $p = 0.8$. This shows that for best results a careful balance has to be struck between flexibility and a parameter setting, which allows larger chunks to be passed on more frequently. In our case a setting of $p = 0.66$ is ideal.

We also tried the scramble mutation operator, as opposed to the swap mutation operator used in all other experiments. The scramble mutation operator is described in section 6.2. Here it was tried with PUX 50 (label 'U Scram') and C1 crossover (label 'C Scram'). The corresponding results are illustrated in the graph and are of worse quality than using the swap mutation. This could be because the same mutation probability of 1.5% as with the swap mutation had been used. This is probably too high now, as the scramble mutation operator is far more disruptive. For all future experiments and unless otherwise stated, PUX 66 and swap mutation will be used.

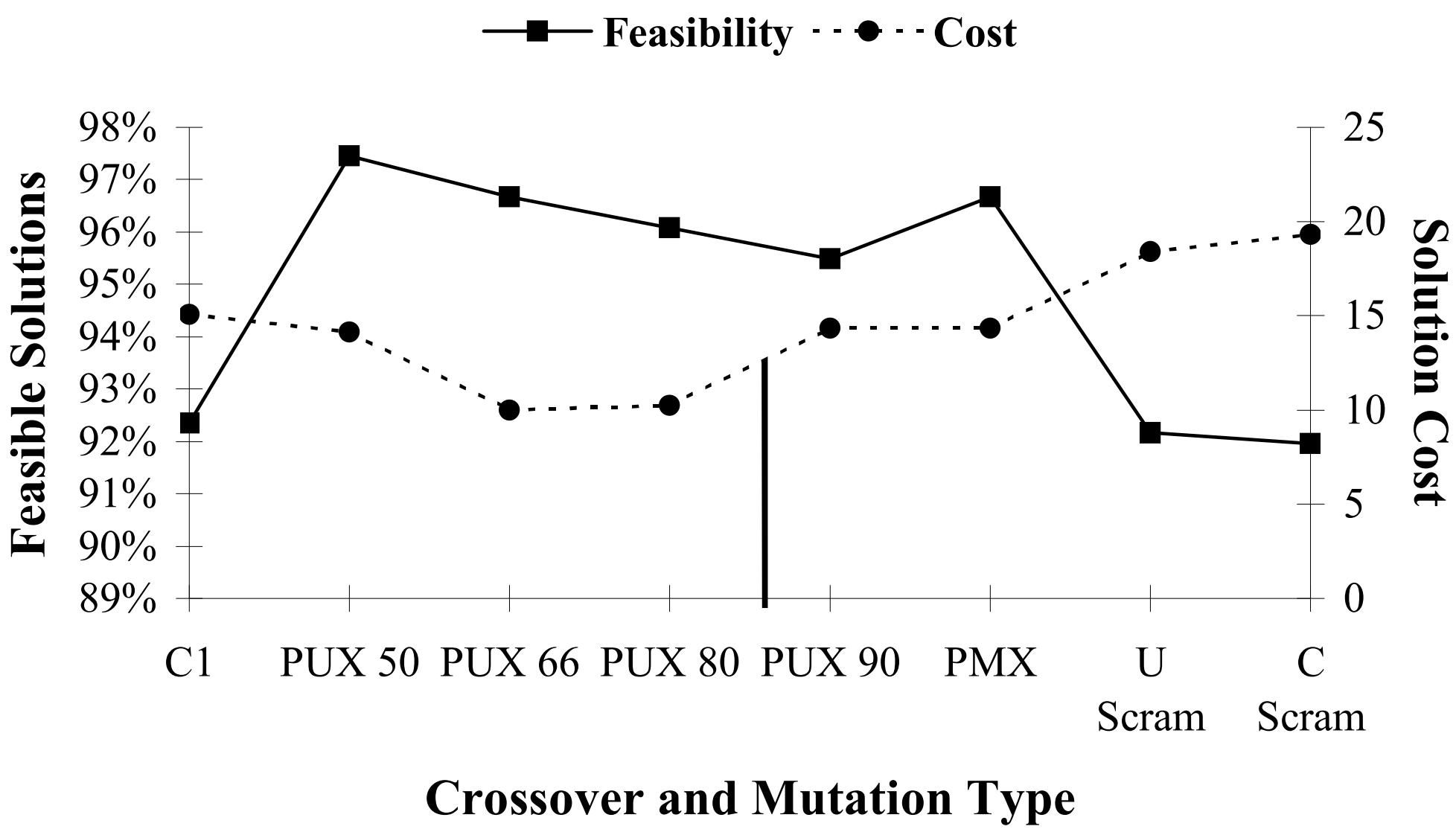

Figure 6-9: Different types of crossover and mutation.

### 6.5.3 Bounds

The final enhancements of the indirect genetic algorithm are based on the intelligent use of bounds. Three methods are proposed: Crossing over before bounds, mutating before bounds and using bounds when assigning a shift pattern with the schedule builder. The former two are based on ideas taken from Herbert [92] and Nagar et al. [123] and will be called boundary crossover and boundary mutation in this section. The latter, referred to as simple bound, is a similar idea to the above and will be discussed afterwards.

Herbert concludes in his research that bin packing results using a genetic algorithm and decoder can be improved with a new type of crossover operator. In his work, Herbert maximises value and uses a permutation based encoding and C1 crossover. He argues that once a lower bound for the solution has been found, only crossover points within that part of the string with a cumulative fitness of greater than the lower bound should be used. This is because the ‘mistakes’ in a particular solution must clearly have happened before such a boundary point.

When maximising value, the placement of the first $k$ pieces will define a sub-layout for any $k$. Subsequently certain areas, for example those surrounded by pieces, will be unusable. Thus at any point the remaining useable area can be determined. It is also known which pieces are left and their respective values. Therefore, one possible upper bound is the value of the pieces used so far plus the result of a knapsack calculation to maximise an area-based estimate of the possible future value.

Thus, the algorithm should concentrate on the area before and including the boundary point, rather than wasting time and optimising ‘tails’ of strings that later need to be revised anyway. A similar approach can be imagined using upper bounds when minimising waste. Note that it is essential for this approach that cumulative bound scores of partial strings can be defined and that C1 crossover is used. A similar argument is brought forward by Nagar et al. who mutate within the part of the string before the boundary point.

An example of these boundary operators is shown in Figure 6-10. Pictured is a sample individual (not from the nurse scheduling problem) encoded as a permutation of events.

Underneath, the cumulative fitness score of the string up to a particular gene is given. If we were to minimise the problem and knew that a solution of 18 existed, an upper bound of 18 could be set. This would give us the bound point as shown. Thus, C1 crossover would have its crossing point before the bound, that is between position one and six of the string. Similarly, swap mutation would have at least one swapping partner from before this boundary point.

| | | | | | | | | | |
|---|---|---|---|---|---|---|---|---|---|
| 5 | 2 | 9 | 4 | 1 | 7 | 6 | 8 | 3 | Individual |
| 3 | 3 | 7 | 9 | 14 | 14 | 21 | 23 | 29 | Cumulative Lower Bound Score |
| | | | | | ^ | ^ | | | Boundary Point |

Figure 6-10: Example of crossover and mutation before a boundary point.

Note that the boundary crossover and mutation operators are quite different from the gene variance based operators proposed by Fang et al. [64]. As outlined in section 3.7, Fang et al. concentrate crossover points in slow converging parts of the string, whilst mutation is more likely to take place in fast converging parts. In encodings with decreasing significance across the genotype, as in our encoding, gene variance based operators will typically result in more mutations towards the front of the strings and more crossovers towards the tails. However, boundary crossover and mutation will always tend towards the front of the strings following the idea of operating before the boundary point.

However, when using the boundary crossover and mutation with the nurse scheduling problem there is one dilemma. How can cumulative bound scores for parts of the string be calculated? Unless all nurses are scheduled, one does not know the quality of a solution. Hence, it makes little sense using a partial fitness based on the cover so far as a measure for the bound. Instead, we propose to base the bounds on the $p_{ij}$ values. These are easily summed up for partial strings and make perfect sense in that respect. The drawback of this is that we can hardly use the lowest cost value found throughout the population as an upper bound, as this is most probably found in an infeasible solution. Therefore, the cost of the best feasible solution is used as an upper bound, which means until we have found a feasible solution, normal unbounded operators will apply.

The simple bound is based on a similar idea as the boundary operators described above. When building a schedule from the genotypes it is easy to calculate the sum of the $p_{ij}$ costs so far. Once this sum exceeds the bound, set equal to the best feasible solution found so far, the schedule has 'gone wrong'. One could now employ backtracking to try to correct this. However, to guarantee that the solution is improved with the actual permutation of nurses at hand, a sophisticated algorithm of exponential time complexity would be necessary. This is outside the scope of this piece of research, but might be an idea for future work.

Instead, a simpler approach is proposed. Once a feasible solution of cost $C^*$ has been found, one knows that in the optimal solution no nurse $i$ can work a shift pattern $j$ with $p_{ij} > C^*$. This will be used as a rule when assigning shift patterns. Of course, particularly in the early stages of optimisation this simple bound is of little use, as $C^* >> p_{ij}$. However, towards the end of the search when good feasible solutions have been found, the simple bound should prevent wasting time on dead-end solutions by making sure shift patterns with $p_{ij} \leq C^*$ are assigned.

The results of various experiments with all three types of bounds are shown in Figure 6-11. The graph compares C1 crossover without bounds (C1) to the use of either boundary crossover (C1 Cross), boundary mutation (C1 Mutat) or both boundary operators (C1 All). The results using the boundary operators are clearly worse than not using them, which is mainly due to the boundary crossover.

The reason for the failure of the boundary operators is thought to be the inadequate measure of the upper bounds. The bounds were defined as the cumulative sum of the $p_{ij}$ values. How good or bad a (partial) solution provided cover was not taken into account. Hence, low-cost solution parts with a poor cover were kept in one piece by the boundary crossover. As the results show, this simplified approach is not suitable for the nurse scheduling problem. Thus, a more problem-specific method would be necessary.

However, this is against the idea of developing a generally applicable indirect approach that does not rely on the problem structure and hence this is not pursued further.

For both C1 and PUX crossover, the use of the simple bound slightly improved the cost of solutions whilst leaving the feasibility unchanged. The slight improvement in cost can be attributed to those runs where good solutions were found which then were further improved by forcing nurses on even cheaper shift patterns. An additional benefit of using the simple bound was that average solution time was accelerated from around 15 seconds per single run to less than 10 seconds.

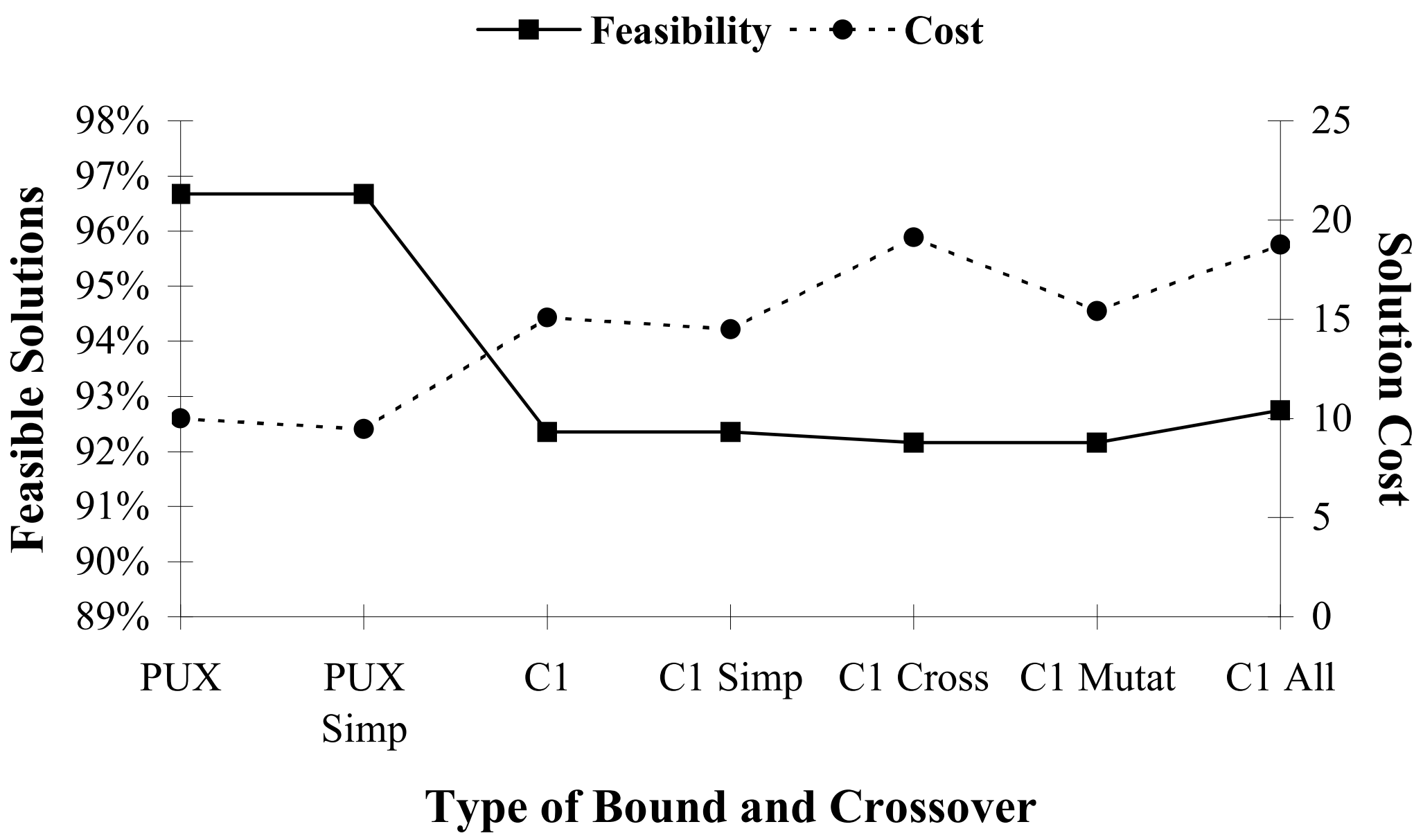

Figure 6-11: Different ways of using bounds.

The results found by the enhanced indirect genetic algorithm are of excellent quality. However, as mentioned in the introduction to this thesis, solution quality is only one aspect why genetic algorithms were chosen to optimise the nurse scheduling problem. Another reason was the well-known robustness of genetic algorithm if the problem changes slightly. In the next section, it is investigated how the algorithms performed when additional constraints and different data were introduced. The former followed requests by the hospital, which shows again how important these issues are for the practical optimisation of real problems.

## 6.6 Extensions of the Nurse Scheduling Problem

### 6.6.1 Head Nurses and Teams

In this section, two possible extensions of the original nurse scheduling problem as presented in section 2.1 are presented. Both extensions were suggested by the hospital and data was provided. However, no solutions of other methods are available to compare with our results. Nevertheless, we will present the genetic algorithm results, showing that our model is robust and comparatively easy to modify.

The first expansion concerns head nurses of which two are present on each ward. In terms of grade bands, they are treated like grade one nurses. However, their salary is different. In particular, the rate of pay for head nurses at the weekends is significantly higher than for grade one nurses. Thus, to reduce cost, the hospital imposed a new constraint that on any weekend day a maximum of one head nurse is allowed to work. This leads to the following new parameters and constraints in the integer programming model of section 2.1.4:

$$o_i = \begin{cases} 1 & \text{if nurse } i \text{ is a head nurse} \\ 0 & \text{else} \end{cases}$$

New constraint: Only one head nurse is allowed to work on Sunday ($k = 1, 8$) and Saturday ($k = 7, 14$):

$$\sum_{k=1,8}\sum_{i=1}^{n}\sum_{j=1}^{m} o_i x_{ij} a_{jk} \leq 1$$

$$\sum_{k=7,14}\sum_{i=1}^{n}\sum_{j=1}^{m} o_i x_{ij} a_{jk} \leq 1$$

Since these new constraints cannot be included implicitly in the encoding, a penalty function approach has to be taken. Note that it would be possible to include this in the decoder by adding a new component to the score of a shift pattern, but this would not guarantee feasibility either. Thus, the raw fitness function of section 4.1 must be expanded by the following term, where $w_{head}$ is the respective penalty weight:

$$w_{head}\left( max\left[\sum_{k=1,8}\sum_{i=1}^{n}\sum_{j=1}^{m} o_i a_{jk} x_{ij} - 1;0\right] + max\left[\sum_{k=7,14}\sum_{i=1}^{n}\sum_{j=1}^{m} o_i a_{jk} x_{ij} - 1;0\right]\right)$$

The second expansion concerns nurses working in teams. The hospital requested that at any one time at least one nurse of each team was present. This is to allow for the continuous care of patients, as each patient is treated by nurses of one team only. Therefore, only permanent members of staff and not bank or dummy nurses are in a team. Thus, two nurses of each team are required on days (one for the early shift and one for the late shift) and one on nights. At present, the nurses are split into two teams. To allow for teams, the integer programming model of section 2.1.4 has to be extended as follows:

$h$ = 1…t team index.

$$b_{ih} = \begin{cases} 1 & \text{if nurse } i \text{ is in team } h \\ 0 & \text{else} \end{cases}$$

New Constraints: At any one time, there must be at least one nurse of each team present:

$$\sum_{i=1}^{n}\sum_{j=1}^{m} a_{jk} b_{ih} x_{ij} \geq 2 \qquad \forall h; \quad k = 1,\ldots,7$$

$$\sum_{i=1}^{n}\sum_{j=1}^{m} a_{jk} b_{ih} x_{ij} \geq 1 \qquad \forall h; \quad k = 8,\ldots,14$$

Again, these new constraints cannot be included implicitly in the encoding. Similarly, including it in the decoder score would not guarantee feasibility either. Hence, the raw fitness function of section 4.1 must be expanded by the following additional penalty term, where $w_{team}$ is the penalty weight associated with it:

$$w_{team} \sum_{h} \left( \sum_{k=1}^{7} max\left[ \sum_{i=1}^{n}\sum_{j=1}^{m} a_{jk} b_{ih} x_{ij} - 2;0 \right] + \sum_{k=8}^{14} max\left[ \sum_{i=1}^{n}\sum_{j=1}^{m} a_{jk} b_{ih} x_{ij} - 1;0 \right] \right)$$

The results of using the indirect genetic algorithm with PUX and simple bounds for the extended nurse scheduling problem can be seen in Figure 6-12. In preliminary experiments, $w_{team}$ and $w_{head}$ were tested in the range between zero and twenty. Both weights gave best results for a value of five, which was used for the results pictured.

The cost of solutions of the extended nurse scheduling problem is higher than for the original problem. This is hardly surprising, as fewer nurses will be able to work their ideal shift patterns with additional constraints in place. Although the new optimal solutions are unknown, the increase in cost is reasonable. The number of feasible solutions is also reduced, more for the inclusion of the teams' constraints than for the head nurses' constraints. If both are included, it is further reduced. Again, this is no surprise, as these new constraints had to be included via penalty functions. Thus, solutions are not guaranteed to fulfil the additional constraints. However, the reduction in feasibility is small compared with results to the original problem, which shows the robustness of our genetic algorithm approach. As mentioned previously, results could possibly be further improved by including separate scores in the decoder.

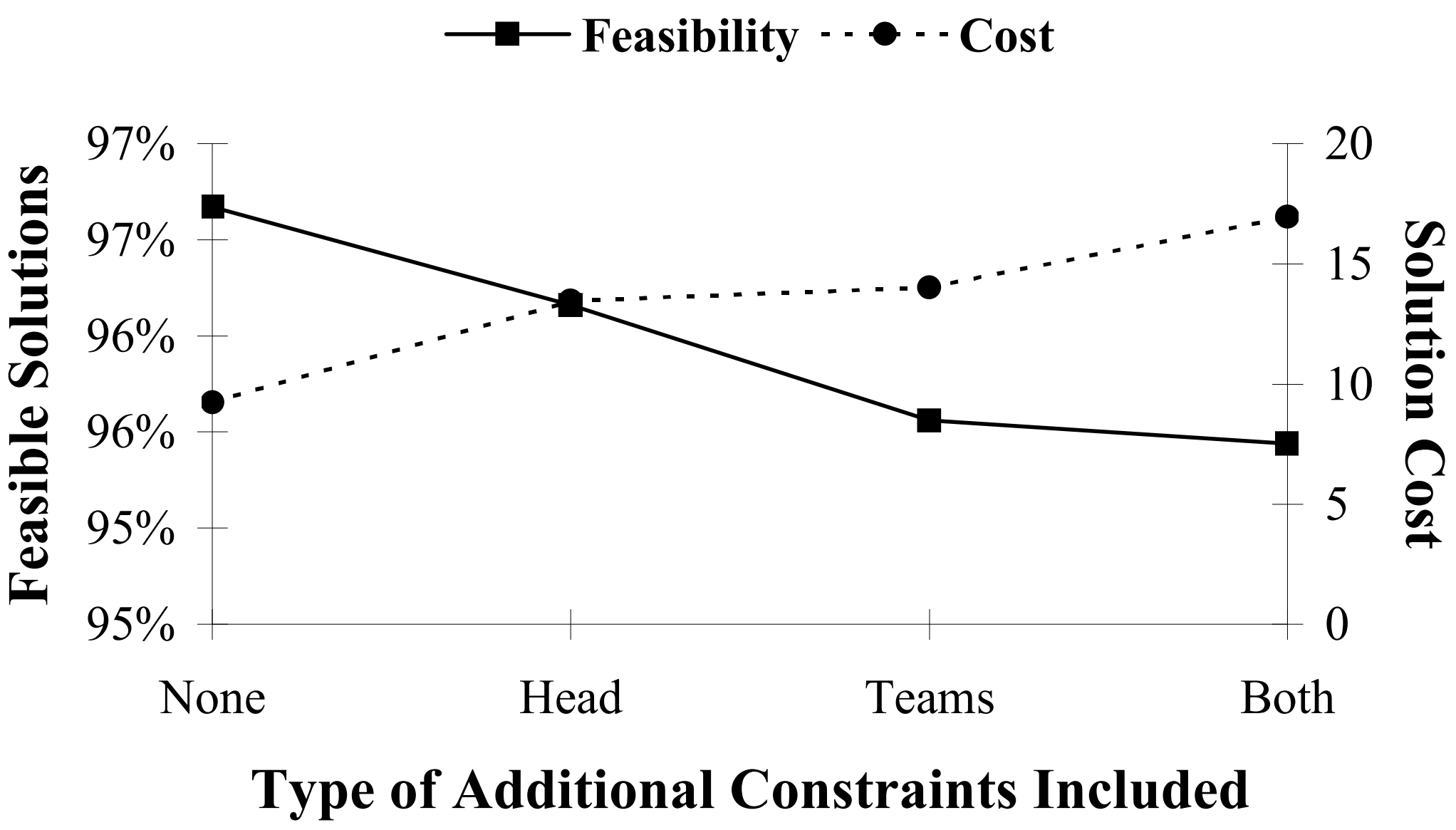

Figure 6-12: Results of the extended nurse scheduling problem.

### 6.6.2 Different Data

To test the robustness of both the direct and indirect genetic algorithms further, the original problem was solved with some different data. These results were then compared to those found by tabu search and XPRESS MP for the new data. In order to do this, two new sets of 52 weeks worth of data were generated. Both sets were identical to the original for the number of nurses, their qualifications and the demand to be covered. However, they differed in the way the $p_{ij}$ values were calculated.

In the first set, called random, the $p_{ij}$ values were set to a random integer between zero and one hundred for each nurse and shift pattern combination. For the second data set, referred to as high pattern cost, the rules outlined in section 2.1.3 were followed. However, one was subtracted from the basic shift pattern cost used in step (1) and the result was then multiplied with twenty. This leads to basic shift patterns costs of 0, 20, 40, 60 etc. Thus, a higher priority was put on the quality of the shift pattern in comparison to the requests of the nurses.

The results for the new and the original data sets are compared in Figure 6-13. The solutions under the IP label are those found with XPRESS MP and occasionally involved an overnight run. More details about this solution method can be found in Fuller [72]. The graph shows that for the high pattern cost data tabu search (label 'Tabu') and the indirect genetic algorithm (label 'GA (i)') gave solutions of similar quality and the direct algorithm (label 'GA (d)') performed slightly worse. However, for the random pattern cost data, the results of both genetic algorithm approaches were superior to those found with tabu search.

This is thought to be due to some specific neighbourhood moves used with tabu search. For instance, some of these moves rely on the fact that if one shift pattern including day *d* is of high cost for a nurse, so are all other shift patterns including day *d*. This is usually true for the original and for the high pattern cost data, but no longer holds for the random data. Full details of the special moves are reported in Dowsland [55].

Although the genetic algorithms proved to be more robust than tabu search for the random data set, all algorithms found it harder to solve than the original or high pattern cost sets. This can be seen from the fact that they produce worse results than those found by XPRESS MP. We conjecture the reason for this being the reduced number of optimal or nearly optimal solutions. Both in the original and high pattern cost sets nurses usually had some shift patterns with the same cost. This leads to plateau like areas in the solution space. With random shift pattern costs, this is no longer true.

However, it is thought that the indirect genetic algorithm could have found better results by using higher penalty weights in line with the on average higher random shift pattern costs. Note that the average shift pattern cost was around 30 for the original data compared to around 50 with the random data. Thus, to balance these higher $p_{ij}$ values against constraint violations, higher penalty weights seem justified. However, as we were testing our algorithm for robustness, penalty weights were not increased. On the other hand, an 'intelligent' penalty weight setting mechanism, for instance taking the average $p_{ij}$ value of a data set into account and setting the penalty weight accordingly, is an interesting idea for future research.

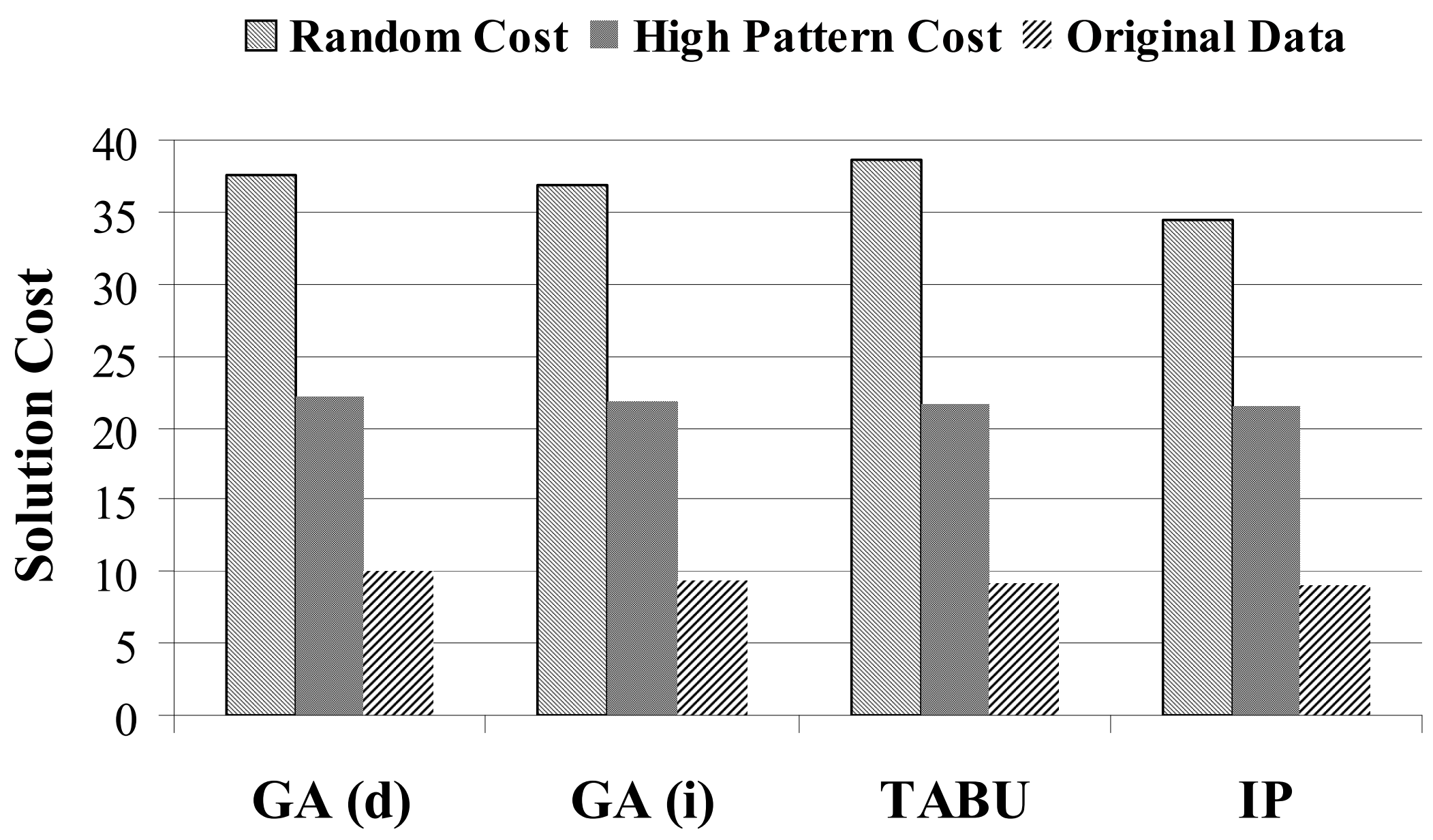

Figure 6-13: Comparison of results for different data sets between the direct and indirect genetic algorithm, tabu search and XPRESS MP.

## 6.7 Conclusions

Overall, excellent results have been achieved with the indirect genetic algorithm plus schedule builder. Furthermore, this was achieved in less development time than for the direct genetic algorithm approach. The main benefit of the indirect approach proved to be the fact that it is easier to adapt the decoder to problem-specific information than the whole of the genetic algorithm, as is necessary for the direct approach. A comparison of final results is shown in Figure 6-14 and detailed results are reported in appendix D.2.

Already the simple cover highest decoder (label 'Highest') and overall contribution decoder (label 'Overall') produced better results than the simple genetic algorithm. Once both decoders were combined (label 'Combo'), the results rivalled those found with the most sophisticated direct genetic algorithm including all enhancements (label 'Direct'). However, once the PUX operator and the simple bounds were employed as well (label 'PUX'), the results were better than for any direct approach and within 2% of optimality. A look at the detailed results in appendix D.2 shows that 51 out of 52 data sets are solved to or near to optimality. A feasible solution is found for the remaining data set.

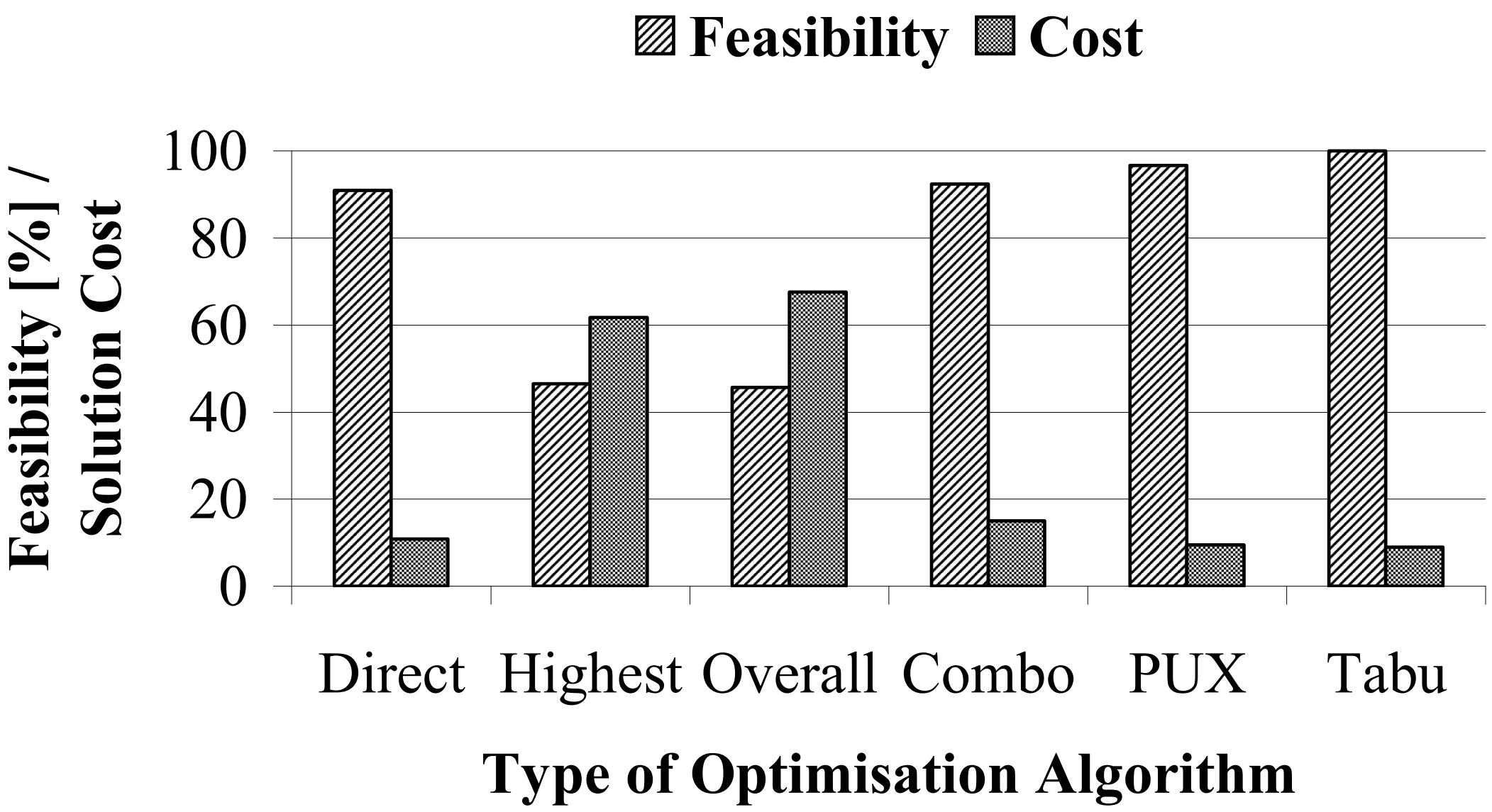

Figure 6-14: Comparison of genetic algorithm approaches with tabu search.

These excellent results together with the fact that the indirect genetic algorithm approach proved to be more flexible and robust than tabu search, makes the genetic algorithm a good choice to solve the nurse scheduling problem. However, it remains to be seen how both the direct and indirect approach fare on a different problem with a non-linear objective function and larger solution space. This will be investigated in chapter 7 with the mall layout and tenant selection problem.

# 7 The Mall Layout and Tenant Selection Problem

## 7.1 Introduction to the Problem

### 7.1.1 General Introduction

In this section, the Mall Layout and Tenant Selection Problem is presented, in future Mall Problem for short. After explaining the reasons why it was decided to tackle this new problem, its origins are discussed. Then the problem itself and the data used are described. A more mathematical description of the problem can be found in sections 7.1.3 and 7.1.4. The way our data was created is described in 7.1.5.

So far, we have presented results of the direct and indirect genetic algorithms and all their enhancements for the nurse scheduling problem only. The flexibility and robustness of the indirect approach was shown in section 6.6 with possible extensions and new data. In this chapter, this is further investigated for both the direct and indirect genetic algorithms by applying them to a different problem. This problem is of multiple-choice nature to allow for a like for like comparison, but also sufficiently different from the nurse scheduling problem to test the robustness of the algorithms.

The idea of using the Mall Problem stems from Hadj-Alouane and Bean [87], who optimise this multiple-choice problem with their genetic algorithm. More details about their approach can be found in sections 3.4 and 4.4. The Mall Problem they tackle is also described in more detail in Bean et al. [15]. However, we were unable to obtain the full details of their problem and the original data as it was the property of a private development company. Therefore, we decided to build our own model and new data was created accordingly. An example of a mall layout is shown in Figure 7-1, which is the floor plan of the Cribbs Causeway Mall near Bristol. This example is of similar size to the problem tackled in this thesis and shows the placement of the big ‘anchor’ stores (e.g. Marks & Spencer) and of other ‘normal’ shops.

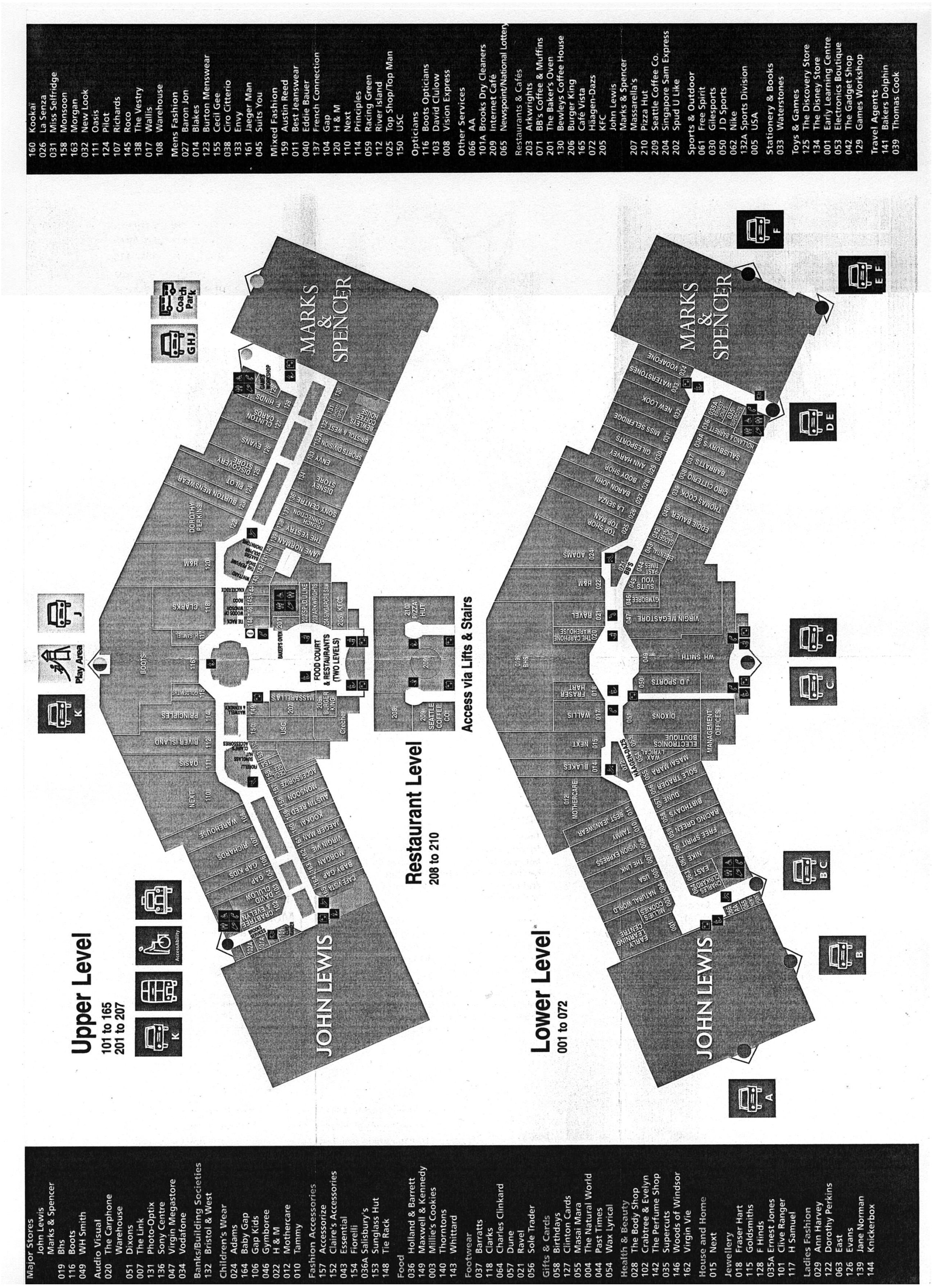

Figure 7-1: Floor plan of the Cribbs Causeway Mall near Bristol.

The Mall Problem differs from the nurse scheduling problem in that it has a non-linear objective function and that we constructed our own data. Thus, we were able to control all characteristics of the data. These two features make it a very interesting problem to tackle, as it is both difficult to solve and customisable to our needs. We will take advantage of the second fact by making the problem similar enough for comparison, but with a different focus than before. In the nurse scheduling case, achieving feasibility was the main concern. Here, finding a feasible solution will be easier, but the solution space will be larger and less uniform.

### 7.1.2 The Problem

The principle of the Mall Problem is to maximise the rent revenue of the mall. Although there is a small fixed rent per shop, a large part of a shop's rent depends on the sales revenue generated by it. Therefore, it is important to select the right number, size and type of tenants and also to place them into the right locations within the mall to maximise revenue. As outlined in Bean et al. [15], the rent of a shop will depend on the following factors:

- The attractiveness of the area in which the shop is located.
- The total number of shops of the same type in the mall.
- The size of the shop.
- Possible synergy effects with neighbouring shops (not used by Bean et al.).
- A fixed amount based on the type of the shop and the area in which it is located.

Before placing shops in the mall, the mall is divided into a discrete number of locations, each big enough to hold the smallest shop size. Larger sizes can be created by placing a shop of the same type in adjacent locations. The placement of 'normal', shops is done after placing some large 'anchor' stores, for instance big department stores. For the remainder of this thesis these anchor stores will be ignored, as in practice, they pay a predefined rent or even own the mall and their locations are fixed.

For each type of the other shops there will be a minimum, ideal and maximum number allowed in the mall, as consumers are drawn to a mall by a balance of variety and homogeneity of shops. For instance, if there are too many record shops in the mall, then competition will be too high amongst them. On the other hand, if there are too few, then they will not be able to hold a great variety of records. If they are close to the ideal number, then the mall will be known as a good place to buy records.

Some locations in the mall are more attractive than others. These could be locations close to the anchor stores, towards the centre of the mall or near parking facilities. To facilitate the handling of this, the mall is split into a number of distinct areas with each location belonging to one of them. For example, these areas could be the north, east, south, west and centre of the mall or close to an anchor store, close to parking, close to the centre etc.

Furthermore, some shop types will do better in certain locations than others. For instance, a newspaper kiosk might be better placed towards the exit of the mall than in its centre. However, this is less important than the overall attractiveness of an area. To tackle this, each location in the mall belongs to an area, and for each area there is an attractiveness rating. In addition to this and to consider the second point, the fixed amount of rent a shop has to pay will depend on the type of the shop and the area it is located in.

The revenue of a shop, and hence its rent, also depends on its actual size. The size of shops is determined by how many locations they occupy within the same area. We decided to group shops into three size classes, namely small, medium and large, occupying one, two and three locations in one area of the mall respectively. For instance, if there are two locations to be filled with the same shop type within one area, then this will be a shop of medium size. If there are five locations with the same shop type assigned in the same area, then they form one large and one medium shop etc.

Usually, there is a maximum total number of small, medium and large shops allowed in the mall.

To determine the rent it is argued that a medium sized shop is more efficient than two small shops and a large shop is more efficient than three small shops or one medium and one small shop. This is due to reduced overhead costs, more room because no partitioning walls are required and other reasons. Therefore, an efficiency rating depending on the size of the shops is introduced. Note that whenever we refer to a shop's rent in the remainder of this thesis, it denotes the rent of one 'unit' of the shop. Thus, a large shop has to pay 'three' rents, one for each of its small units.

To make the problem as realistic as possible, an additional twist is added by assigning most shop types one or more groups of which they are a member. For instance, childrenswear, menswear and womenswear are all part of the clothes shop group. Additionally, women's wear is part of the women's shop group to which, for example, cosmetic shops and jewellers might belong. To account for synergy effects between shops of one group, a revenue bonus is allocated to certain shops if they are not too far apart. This bonus is applicable to all shops of one group within one area of the mall, if at least one small shop of each group member is present within this area.

To test the robustness and performance of our algorithms thoroughly on this new problem, 70 data files were created. The data were grouped into seven sets with ten files each. The seven data sets differ in the dimension of the data, i.e. the number of locations in the mall, the number of areas the locations are grouped in, the number of shop types and the number of shop groups. A further difference between the sets is in the tightness of the constraints regarding the minimum and maximum number of shops of a certain type or size. Full details on how the data was created, its dimensions, the differences between the sets and the anticipated effects on our algorithms can be found in section 7.1.4.

### 7.1.3 Problem Formulation

In this section, a mathematical model of the Mall Problem is set up. This leads to a non-linear objective function and numerous constraints. Recall that the problem is that of assigning one shop to each location in the mall such that the rental revenue (basic rent plus the rent proportional to sales) is maximised. The following can then be used to set up a mathematical model:

Indices:

$i = 1...N$ location index.
$j = 1...S$ shop index.
$k = 1...A$ area index.
$l = 1...G$ group index.

Decision Variables:

$$x_{ij} = \begin{cases} 1 & \text{if a shop of type } j \text{ is in location } i \\ 0 & \text{else} \end{cases}$$

Parameters:

$a_j$ = minimum number of shops of type $j$ required.
$b_j$ = maximum number of shops of type $j$ allowed.
$c_j$ = ideal number of shops of type $j$ to maximise revenue, respectively rent.
$t_j$ = total number of shops of type $j$ in the mall.

$n_{jk}$ = number of shops of type $j$ in area $k$.
$n_{jS}$ = number of small shops of type $j$.
$n_{jM}$ = number of medium shops of type $j$.
$n_{jL}$ = number of large shops of type $j$.

$m_S$ = maximum number of small shops.

$m_M$ = maximum number of medium shops.

$m_L$ = maximum number of large shops.

$$d_{ik} = \begin{cases} 1 & \text{if location } i \text{ is in area } k \\ 0 & \text{else} \end{cases}$$

$$g_{jl} = \begin{cases} 1 & \text{if shop type } j \text{ is a member of group } l \\ 0 & \text{else} \end{cases}$$

$$B_{kl} = \begin{cases} 1 & \text{if group } l \text{ is complete in area } k \\ 0 & \text{else} \end{cases}$$

$r_{jk}$ = basic rent if shop $j$ is allocated in area $k$.

$r_k$ = attraction factor of area $k$.

$B(B_{kl})$ = Bonus factor depending whether a group is complete within an area or not.

$E(n_{jk})$ = Shop Size Efficiency factor of shops of type $j$ if they occupy $n_{jk}$ locations within area $k$.

$I_j(c_j,t_j)$ = Shop Count Efficiency factor of shops of type $j$, depending on $t_j$ and $c_j$.

Constraints:

1. The number of shops of type $j$ in area $k$ is calculated as:

$$n_{jk} = \sum_{i=1}^{N} d_{ik} x_{ij} \quad \forall j,k$$

2. The number of shops of each size:

$$n_{jL} = \sum_{k=1}^{A} \text{int}(n_{jk}/3) \qquad \forall j$$

$$n_{jM} = \sum_{k=1}^{A} \begin{cases} 1 & \text{if } (n_{jk} \bmod 3 = 2) \\ 0 & \text{else} \end{cases} \qquad \forall j$$

$$n_{jS} = \sum_{k=1}^{A} \begin{cases} 1 & \text{if } (n_{jk} \bmod 3 = 1) \\ 0 & \text{else} \end{cases} \qquad \forall j$$

3. The total number of small, medium and large shops in the mall must be less than the respective maximum number allowed:

$$N_L = \sum_{j=1}^{S} n_{jL} \leq m_L$$

$$N_M = \sum_{j=1}^{S} n_{jM} \leq m_M$$

$$N_S = \sum_{j=1}^{S} n_{jS} \leq m_S$$

4. The total number of shops of one type must be more than the minimum number and less than the maximum number allowed:

$$a_j \leq \sum_{i=1}^{N} x_{ij} \leq b_j \qquad \forall j$$

5. There must be exactly one shop in each location:

$$\sum_{j=1}^{S} x_{ij} = 1 \qquad \forall i$$

6. A group within one area is either complete or not:

$$B_{kl} = \begin{cases} 1 & \text{if } \prod_{j=1}^{S} n_{jk} g_{jl} > 0 \\ 0 & \text{else} \end{cases} \qquad \forall k,l$$

<u>Target Function:</u>

For one location and shop unit, the rent is: $[r_k \times B(B_{kl}) \times E(n_{jk}) \times I_j(c_j, t_j)] + r_{ik}$

Or in other words:

$$\begin{bmatrix} \text{Attraction Factor of Area} \times \text{Total Group Bonus of Shop} \\ \times \text{Shop Size Efficiency Factor} \times \text{Shop Count Efficiency Factor} \end{bmatrix} + \text{Fixed Rent Shop/Area}$$

Thus, the overall target is the sum of this function for all locations. As mentioned before, this objective function is non-linear. In the following section, the efficiency functions used are described and dimensions of the relevant parameters are given.

### 7.1.4 Efficiency Functions and Parameter Dimensions

From the information contained in Bean et al. [15], we know that the revenue dependent amount of rent is a multiple of the fixed amount of rent. In our data, the revenue depending rent is approximately ten times as big as the fixed rent. To achieve this, the following ranges for the parameters of the model are set. The final rent of a shop unit can be assumed to be in pounds per year:

- If a group is complete within an area, all shops of that group and area are awarded a 20% synergy bonus. Hence *B(1) = 12* and *B(0) = 10*. Note that in exceptional circumstances a shop belonging to two groups, which are both complete in an area, might have a synergy bonus of *B(2) = 14.4*.
- The attractiveness factor of an area $r_k$ is in the range [5...25].
- The size efficiency factor of shops depending on the number of shops of this type in the area is *E(1) = 10* for one small shop, *E(2) = 11.5* for one medium shop, *E(3) = 13* for one large shop, *E(4) = (3 * 13 + 10) / 4 = 12.25* for a small and a large shop together, *E(5) = (3 * 13 + 2 * 11.5) / 5 = 12.4* for a medium and a large shop together etc.
- The shop count efficiency factor is calculated as $I_j = \min(10 - |t_j - c_j|^+;\ 0)$. Thus should the actual number of shops of type *j* be ten or more away from the ideal number, the solution is regarded as so bad that its objective function value is 0.
- The fixed rent per shop type and area $r_{ik}$ is in the range [1000...3000].

For clarification, the following are examples on calculating a year's revenue dependant rent per location unit, i.e. the total rent for a medium shop occupying two location units would be twice as much etc. To obtain the total rent for the location, the fixed rent per shop type and area the location is in must be added to this.

A small sized shop, located in the least attractive area, whose group is not complete and whose shop count efficiency indicates that five shops less than the ideal number are present: 10 [size] * 5 [area] * 10 [group] * 5 [count] = 2500 (pounds per year).

A medium sized shop, located in an area of average attractiveness, whose group is incomplete and whose shop count efficiency indicates that one more shop than ideal is present: 11 [size] * 15 [area] * 10 [group] * 9 [count] = 14850 (pounds per year).

A large sized shop, located in a prime area, with a complete group present and whose shop count efficiency is ideal: 13 [size] * 25 [area] * 12 [group] * 10 [count] = 39000 (pounds per year).

### 7.1.5 Data Sets

Thoroughly testing our algorithms for robustness was the main aim when creating our data. The Mall Problem was chosen because of its similar multiple-choice character, as this enables us to try similar approaches as before with a similar encoding to the nurse scheduling problem. However, when choosing the format of the data, it was implemented in a way that one major difference would emerge. The intention was to create an overall much larger solution space, due to both larger problem dimensions and less tight constraints. This is expected to lead to a situation where all algorithms will find feasible solutions more frequently than for the nurse scheduling. Therefore, it will be interesting to see how the various approaches of the past chapters will perform when the ‘cost’ of solutions is the major factor rather than ‘feasibility’. With this in mind, seven data sets comprising of ten files each were set up.

Table 7-1 shows a summary of the specifications and dimensions used for generating the data. Furthermore, we created the files whereby the following rules were fulfilled:

- Each location is randomly assigned one area with at least 5 locations and at most 30 locations in any one area.
- Each shop is either assigned none, one or two groups, with a strong bias towards one group. However, groups must have at least three and at most ten members.
- A 'loose' shop size constraint means that any number of small, medium and large shops is allowed. A 'tight' constraint stipulates that at most 6 small, 17 medium sized and 22 large sized shops are allowed. Note that this leaves only six units slack.
- A 'loose' shop count constraint indicates that no minimum number of shops for any shop type is set. With an 'average' count constraint, the sum of the minimum numbers for all shop types equals between 60% and 80% of all locations. With a 'tight' constraint, it equals respectively between 95% and 98%.
- The minimum, ideal and maximum number of each shop type is set between zero and ten, with the total subject to the above shop count constraint and obviously such that minimum ≤ ideal ≤ maximum. In addition, the sum of the maxima for all shops must be greater or equal to the number of locations in the mall.
- The attractiveness of an area and the fixed shop type per area rent is set up in the ranges as given in section 7.1.4. Areas with a higher index have a slight bias towards being more attractive and also carry a higher fixed rent.

| Data Set | Locations | Areas | Shops | Groups | Shop Size | Shop Count |
|---|---|---|---|---|---|---|
| **1 (01-10)** | 20 | 2 | 12 | 3 | Loose | Average |
| **2 (11-20)** | 50 | 4 | 16 | 4 | Loose | Average |
| **3 (21-30)** | 100 | 5 | 50 | 8 | Loose | Loose |
| **4 (31-40)** | 100 | 5 | 20 | 5 | Loose | Average |
| **5 (41-50)** | 100 | 5 | 20 | 5 | Loose | Tight |
| **6 (51-60)** | 100 | 5 | 20 | 5 | Tight | Average |
| **7 (61-70)** | 100 | 5 | 20 | 5 | Tight | Tight |

Table 7-1: Specifications of Mall Problem data sets.

Data sets 1 and 2 were only used for preliminary experiments, hence their reduced size. No further results are reported on these. Data set 3 represents a situation before the mall is built. Therefore there are no constraints on sizes and numbers of shops and a variety of possible shop types exists. In data set 4, the 50 candidates of set 3 have been reduced to the twenty 'most promising', with some estimates of how many of each shop type are required.

Sets 5 and 6 represent a situation where the mall is built. In set 5, it is largely decided which shops will be in the mall, but it is not yet known in what sizes. In set 6, the mall is divided into empty 'shops', but it is not yet known of which type. Finally, set 7 combines the constraints of sets 5 and 6. In set 7, it is known how many shops of each type and how many of each size category there will be. However, it is unknown which shop will be of what size and where they will be placed in the mall. Examples of data can be found in D.3.

To allow for further comparisons, one file in each of data sets 4-7 uses the same group memberships, ideal number of shops, attractiveness ratings and fixed rents. This guarantees that solutions to set 7 are always achievable with sets 4-6 and solutions to set 5 and 6 could be found with set 4. For instance, the best solution found in file 65 should be worse than the ones found for files 35, 45 and 55, whilst the best solutions found with files 45 and 55 should be worse than those found with 35.

Overall, the dimensions of the problem are approximately two to five times as big as for the nurse scheduling (25 nurses multiplied by 40 shift patterns each compared with 100 locations multiplied by 20 or 50 shop types). The difference between the solution space sizes is even bigger. It is $40^{25} \approx 10^{40}$ for the nurse scheduling and $20^{100} \approx 10^{130}$ or $50^{100} \approx 10^{170}$ for the Mall Problem. Although many of these solutions will be infeasible as there is a maximum of ten shops of one type allowed in the mall, overall the constraints have been chosen such that they are not as tight as for the nurse problem.

It is conjectured that the larger solution space combined with the slacker constraints will increase the number of feasible solutions as compared to the nurse scheduling. In the nurse scheduling problem, as detailed in section 2.1.2, a pre-switched knapsack routine made the problem extremely tight. Here, even in the tightest case, there is some slackness in most constraints. Thus, we expect the genetic algorithm to have fewer problems with feasibility and more with the now non-linear solution ‘cost’ than before. As mentioned earlier, this is the desired effect.

## 7.2 Simple Direct Genetic Algorithm Approach

### 7.2.1 Description of Experiments

The experiments for the Mall Problem were carried out in a similar manner to those for the nurse scheduling problem, as described in section 4.2. For each of the 50 data files 20 runs were made on the same computer equipment as before. The results will be presented in the same feasibility and ‘cost’ format as before, too. However, in contrast to the nurse scheduling problem, we are now maximising, so a higher solution rent value is better than a lower one. The values for the rent displayed in the graphs are in thousands of pounds. Solution times are only reported if significantly different from those found during parameter tests.

As before, feasibility denotes the probability of finding a feasible solution. Rent refers to the target function value of the best feasible solution for each data set averaged over the number of data sets for which at least one feasible solution was found. Should the algorithm fail to find a single feasible solution for all 20 runs on one data file, a censored observation of zero is made instead. As we are maximising the rent, this is equivalent to a very poor solution.

Unlike for the nurse scheduling problem, the optimal solutions to the Mall problem are unknown. Thus to get a rough idea as to the quality of solutions found by the algorithms, the following theoretical upper bound can be used. The best we can hope for is that all shops are large, in a group and with an ideal size count. In addition, on average over all data files the area attractiveness factor should be close to 15 and the fixed area per shop type rent cannot exceed 3000. This is obviously too optimistic, as some files do not allow large shops only and many of the other constraints will prevent an ideal shop count for all shop types to be achieved. For instance, there is no rule saying that the sum of the ideal shop counts must be equal to the number of locations in the mall.

In this ideal case, the rent for a mall with 100 locations would be (in thousands):
100(locations) * [12(group) * 13(large) * 10(ideal) * 15(attract) + 3000(fixed)] = 2640.

### 7.2.2 Encoding

When choosing the encoding, we have to keep in mind that for a meaningful comparison with the results of the nurse scheduling problem, there must not be too many differences. For the nurse scheduling, each nurse had to be assigned a shift pattern. This was reflected in the encoding by gene *i* representing the shift pattern worked by nurse *i*. The equivalent for the Mall Problem would be gene *i* representing the type of shop built in location *i*. Thus, for a true comparison with the nurse scheduling results, this is the required encoding.

However, not to dismiss other alternatives immediately, the following two possibilities were considered. In the first, the string has as many genes as the number of shop types multiplied by the number of areas. Each gene then denotes the number of shops of a particular type built in a specific area. The second possibility is a string with as many genes as the number of locations in the mall. Here, the value of a gene would show the type of shop built in a specific location. This, as explained above, is the equivalent to the direct nurse scheduling encoding.

With the first encoding, there are a number of problems, which led us to discard the idea. The biggest problem is that after any type of standard crossover, we will almost certainly end up with infeasible children in respect to the total number of shops used. Thus, an intensive repair operator would be required. Furthermore, for some data files we would end up with very long strings of up to 250 genes. This could lead to problems regarding the formation of successful building blocks. An advantage of this encoding would be the quicker calculation of the objective function value, as the $n_{jk}$ values (number of shops of one type per area) are already known. With the second encoding, the $n_{jk}$ values have to be deducted first.

As mentioned before, the second encoding is very similar to the one used for the direct nurse scheduling. It was successful then, and again it offers advantages over the other type for the Mall Problem. First of all, crossover operations will never generate infeasible children regarding the total number of shops. Solutions can still be infeasible with respect to the shop size and shop count constraints, but this is the same with the first encoding. Secondly, strings are no longer than 100 genes for any of the data files. Thus, for its general superiority and simplicity the second encoding was chosen. Note that in order to stay in line with sorting the nurses in grade order, the locations of the string are sorted in area order.

### 7.2.3 Genetic Algorithm Set Up

Since only the multiple-choice constraints are implicitly taken care of by the chosen encoding, the remaining constraints have to be included with a penalty function. These constraints are the minimum and maximum number of shops of each type allowed in the mall and the maximum number of small, medium and large sized shops. To arrive at the penalty $p_s$ of a solution $s$, we will measure and sum the violations of all these constraints and then multiply this with a penalty weight $w_{penalty}$. If a constraint is violated, the violation is measured as the difference between the actual number and the maximum number allowed. Thus, $p_s$ is as follows:

$$p_s = w_{penalty} \left\{ \begin{array}{l} \sum_{j=1}^{S} \left[ \max(0; a_j - \sum_{i=1}^{N} x_{ij}) + \max(0; \sum_{i=1}^{N} x_{ij} - b_j) \right] \\ + \max(0; \sum_{j=1}^{S} n_{jL} - m_L) + \max(0; \sum_{j=1}^{S} n_{jM} - m_M) + \max(0; \sum_{j=1}^{S} n_{jS} - m_S) \end{array} \right\}$$

The raw fitness of an individual can now be calculated as its objective function value minus the above penalty term. With this in place, the same canonical genetic algorithm as before and as summarised in appendix A.2 can be used in the first instance. Hence, those parameter values that have been found to work best with the nurse scheduling in section 4.3 were chosen for initial experiments. Thus, the genetic algorithm uses the parameters and strategies as summarised in Table 7-2. These might no longer be ideal, since we are trying to optimise a new problem. Hence, the next section deals with some limited parameter tests.

| Parameter / Strategy | Initial Setting |
|---|---|
| Population Size | **1000** |
| Population Type | **Generational** |
| Initialisation | **Random** |
| Selection | **Rank Based** |
| Uniform Crossover | **Parameterised with *p = 0.8*** |
| Parents and Children per Crossover | **4** |
| Per Bit Mutation Probability | **1.5%** |
| Replacement Strategy | **Keep 10% Best** |
| Stopping Criteria | **No improvement for 30 generations** |
| Penalty Weight | **20** |

Table 7-2: Initial parameters and strategies of the direct genetic algorithm.

### 7.2.4 Parameter Tests

The starting point for the parameter tests is the genetic algorithm as described in Table 7-2. From previous experience with parameter tests (see sections 4.3 and 6.4), we conjecture that for most parameters there will only be a slight difference in the quality of the results, as long as parameters are chosen within a sensible range. As mentioned before, this is due to the well known robustness of genetic algorithms to a wide range of parameter settings. Therefore, we concentrate on a selected number of experiments here.

Firstly, we experiment with the population size, because the increase in the solution space leads to unacceptably long running times for the previous setting. The parameter that is most likely to be 'wrong' is the penalty weight, as the Mall Problem has a different objective function and different constraints. Finally, we will experiment with mutation rate and crossover operators to determine if the previous settings are still best for the problem. All other parameters and strategies will remain unchanged from those used before.

The results of varying the population size are shown in Figure 7-2 and in Figure 7-3. The latter shows a roughly linear increase of solution time with a bigger population size, which is expected for an efficient algorithm. Note that due to the prolonged running times, a population of size 1000 was not tested. Although the algorithm is efficient, it is about ten times slower than for the nurse scheduling. This gives an idea as to how much the solution space and / or problem dimensions have increased in comparison to the nurse scheduling.

On the other hand, as the experiments show, a smaller population than for the nurse scheduling is sufficient to solve the problem. However, reducing the population size too far runs the risk of producing low quality solutions. This is illustrated by the poor quality of solutions for a population of size 10. The graph in Figure 7-2 also shows that a population of size 50 and above gives much better results with a good feasibility. So overall in comparison to the nurse scheduling problem, the genetic algorithm converges more slowly as the solution space is bigger but finds it 'easier' to solve the Mall Problem. These results are not surprising, as this is exactly how the data files were set up. The solution space is bigger than before but all constraints contain some slackness.

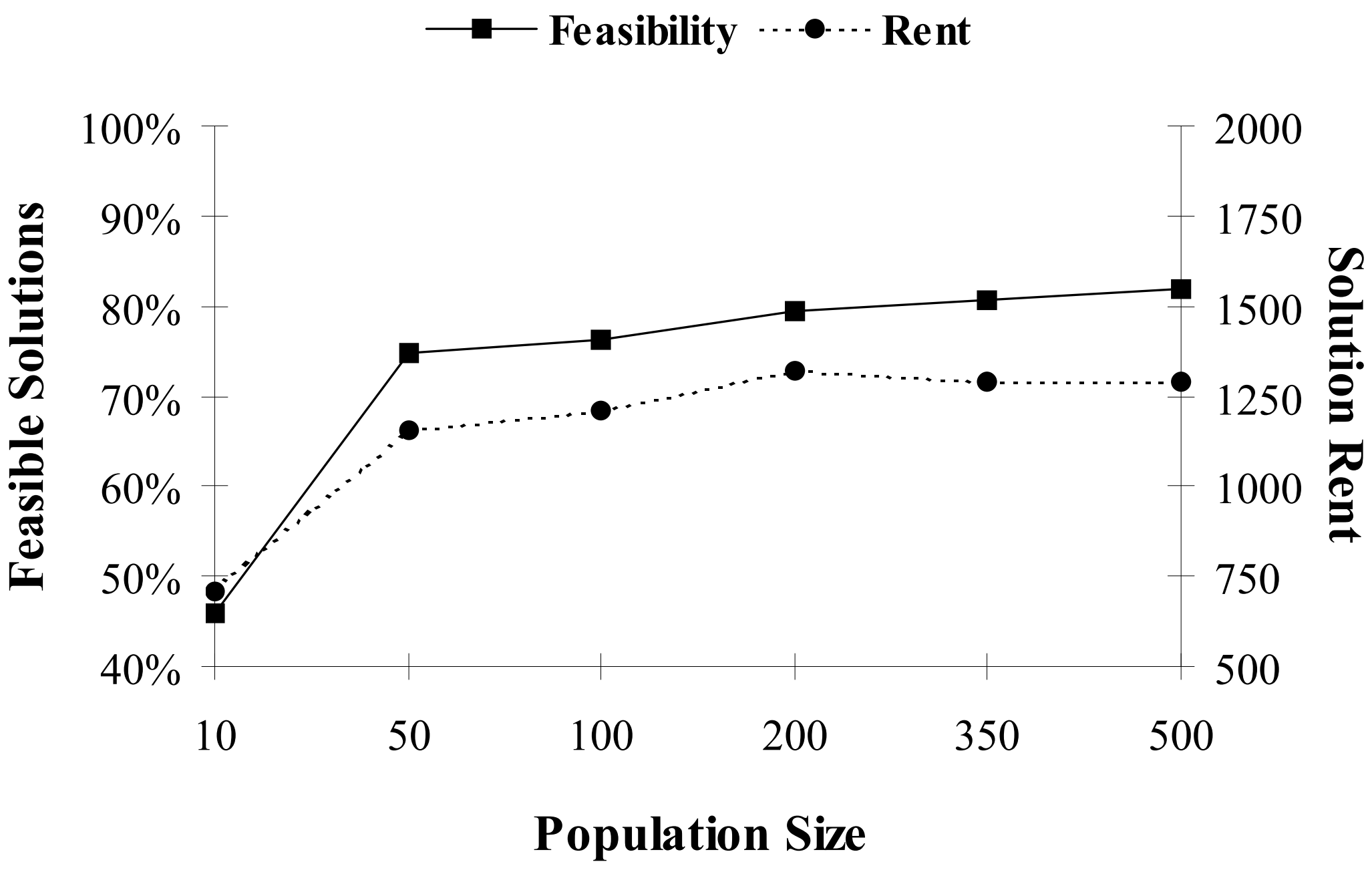

Figure 7-2: Population size versus feasibility and rent.

From the results displayed in Figure 7-2, one can see that in line with previous results, the bigger the population size the higher the rent and feasibility. The best rent is achieved with a population of size 200. This is attributed to the choice of stopping criterion, which was set to 30 generations without improvement. Presumably, for this criterion a population of size 200 is ideal, as larger populations do not have enough time to converge. Thus for other stopping criteria the 'optimal' population size might be different. Nevertheless, we decided to use a population of size 200, which appears to give a good trade-off between solution quality and speed.

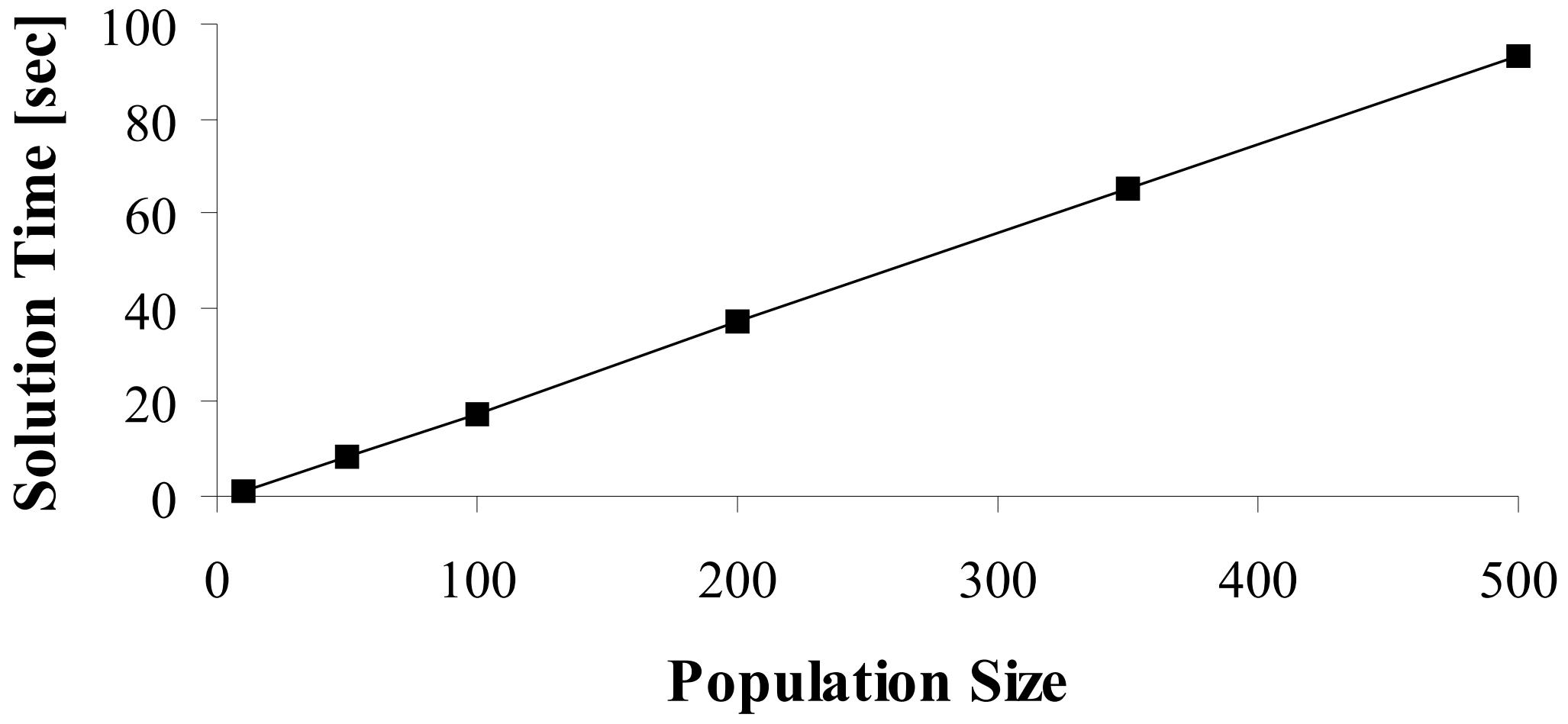

Figure 7-3: Population size versus average solution time.

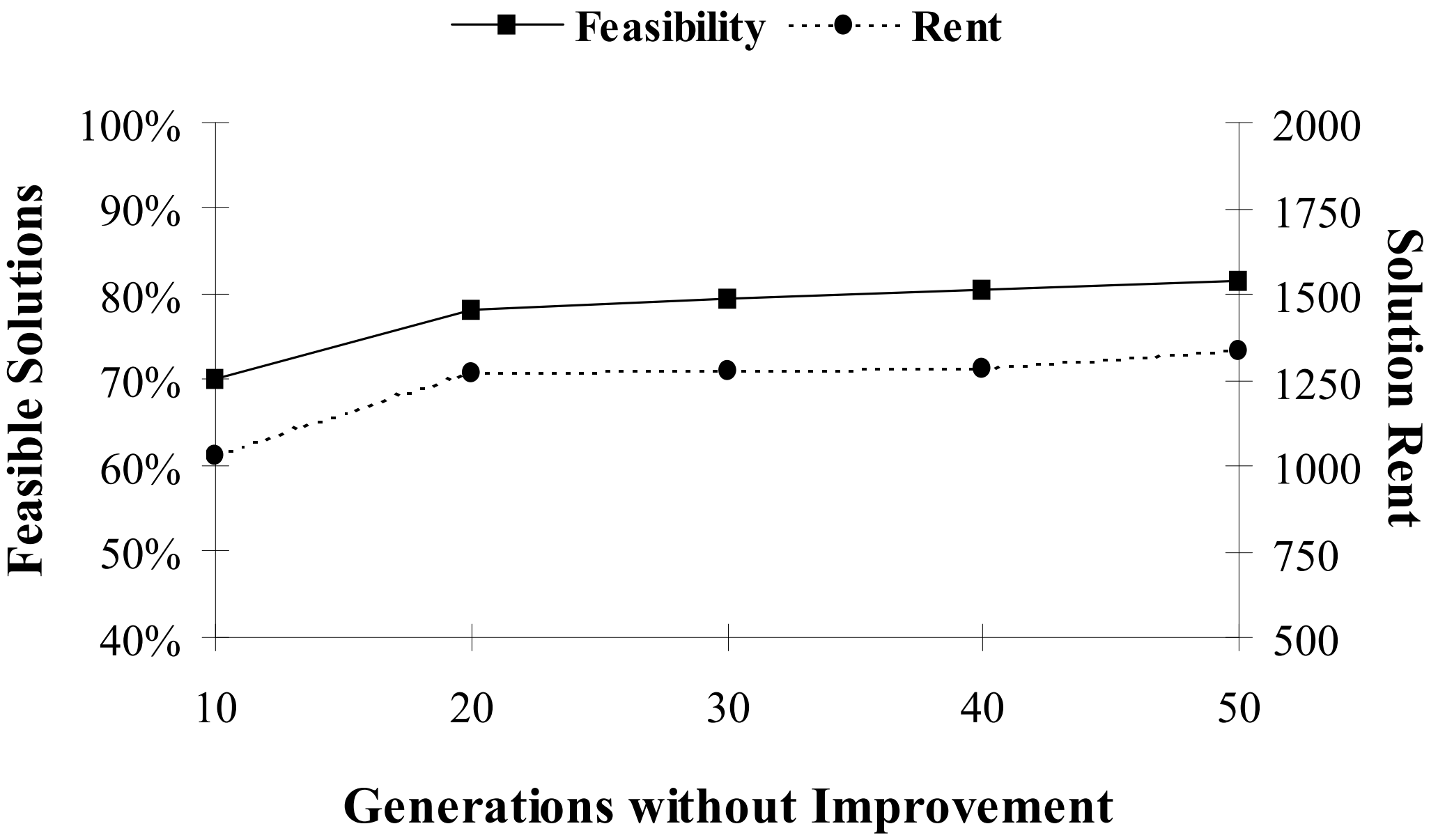

Figure 7-4: Stopping criteria versus feasibility and rent.

When experimenting with the stopping criteria, the number of generations without improvement was kept as the criterion. This measure is easy to calculate and meaningful from an optimisation point of view. The results of varying the number of generations between 10 and 50 are displayed in Figure 7-4 and in Figure 7-5. Unsurprisingly, the more generations the better the results but the slower the speed of convergence. For a population of size 200, 30 generations without improvement is a sensible stopping criteria, with a good trade-off between solution quality and speed. This setting will be used for future experiments.

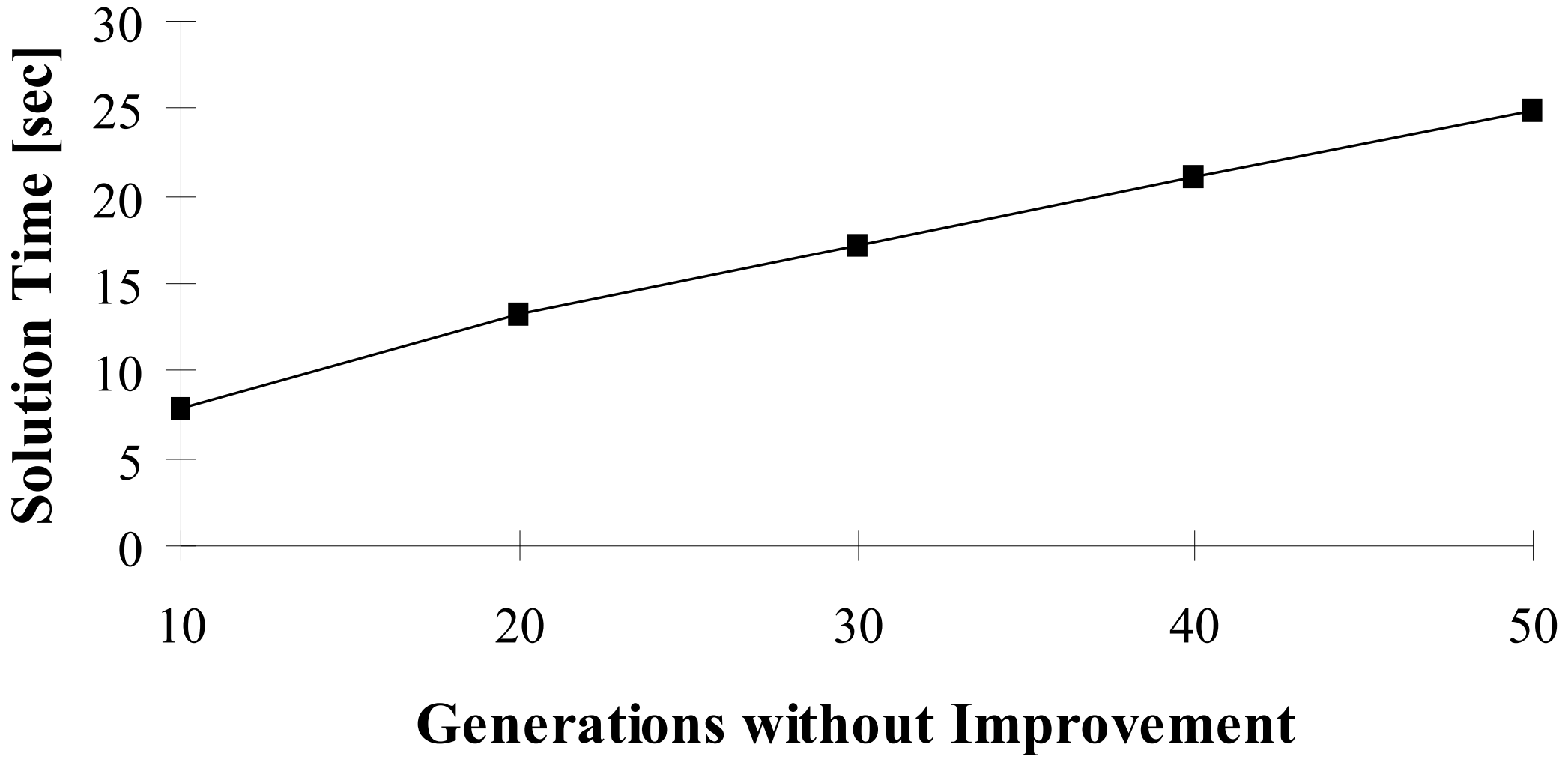

Figure 7-5: Stopping criteria versus average solution time.

As the results so far are already quite good in terms of feasibility, the pre-set penalty weight of 20 can be assumed a reasonable choice. The results of further tests on the penalty weight are displayed in Figure 7-6 and show that a penalty weight of 30 performs best. Two things are of interest here in comparison to the experiments on the penalty weight for the nurse scheduling problem in section 4.3.3. Firstly, the overall shape of the graph is similar. This shows again how a penalty weight that is set too high is detrimental to solution quality, because it restricts the search too much.

Secondly, if one compares the absolute value of the best penalty weight for the nurse problem, which was 20 with the best value here, then there is little difference. At first, this might look surprising, since the target function values that have to be counterbalanced against the penalty function are much higher here. Therefore, one would expect a higher penalty weight for the Mall Problem. However, this argument ignores the fact that the Mall Problem has less stringent constraints and penalty weights should reflect the difficulty in meeting the constraints. Taking both these facts into account, a similar penalty weight is optimal.

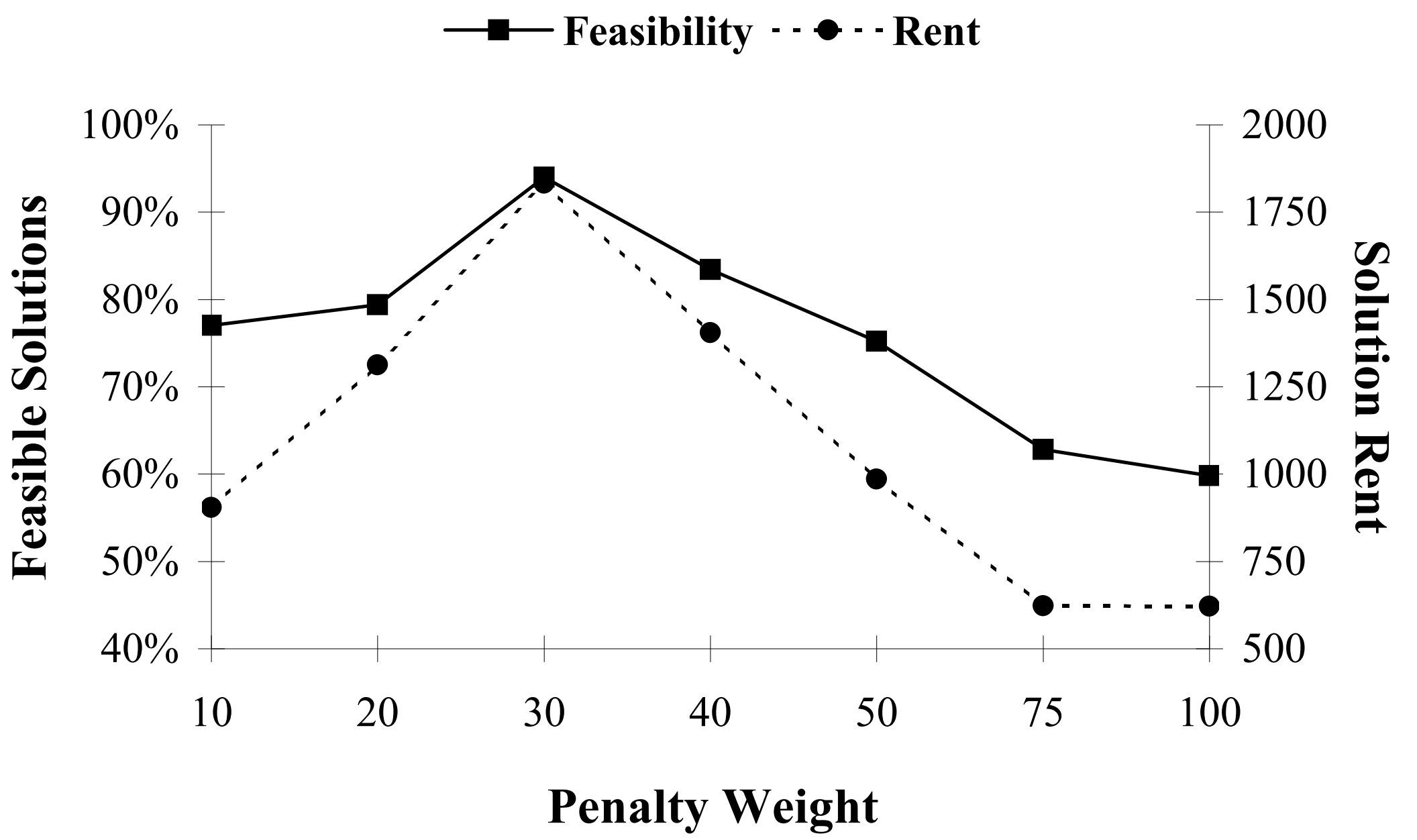

Figure 7-6: Penalty weight versus rent and feasibility.

Figure 7-7 shows the results of experimenting with the single bit mutation probability $p_M$. There is no significant difference in the results for the range between $0.5\% \leq p_M \leq 2\%$. To keep in line with the direct genetic algorithm for the nurse scheduling problem, the same mutation probability of $p_M = 1.5\%$ was chosen for all future experiments.

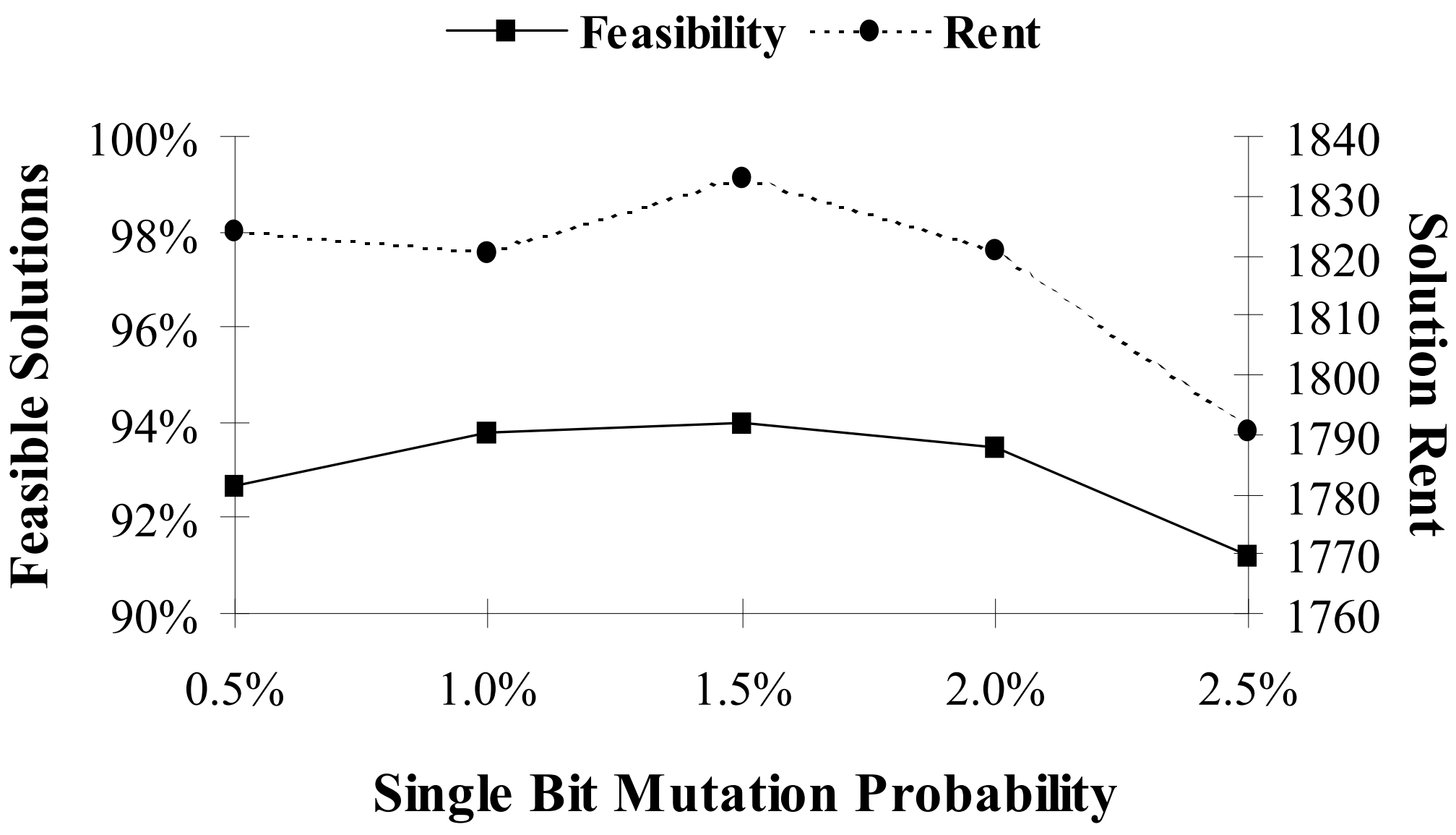

Figure 7-7: Single bit mutation probability versus feasibility and rent.

With all other parameters and strategies in place, the following final experiments tested the effect of different crossover operators. The results are pictured in Figure 7-8 with the labels indicating the crossover operator, respectively the setting of *p* for the parameterised uniform crossover. As with the nurse scheduling, parameterised uniform crossover gave the best performance. Interestingly in line with the nurse scheduling results, parameterised uniform crossover with a higher *p* was more likely to produce feasible solutions, whereas with a smaller *p* solutions were of higher rent. This may be explained by the amount of disruptiveness present in those operators.

With a *p* close to *0.5*, the disruptiveness is maximal, but so is the flexibility in creating new solutions. More disruptiveness leads to fewer large chunks being exchanged between parents, which in turn reduces the chance of a feasible offspring. On the other hand, the disruptiveness offers the chance of creating the widest possible range of children, sometimes creating offspring which more conservative crossovers would not. Hence, this results in higher rent but less feasibility. In contrast, a higher *p* offers less disruptiveness and therefore more feasibility, with the trade-off of finding lower rent solutions. In addition, two-point crossover was shown once more to be better than one-point crossover. For all further experiments parameterised uniform crossover with $p = 0.66$ was chosen.

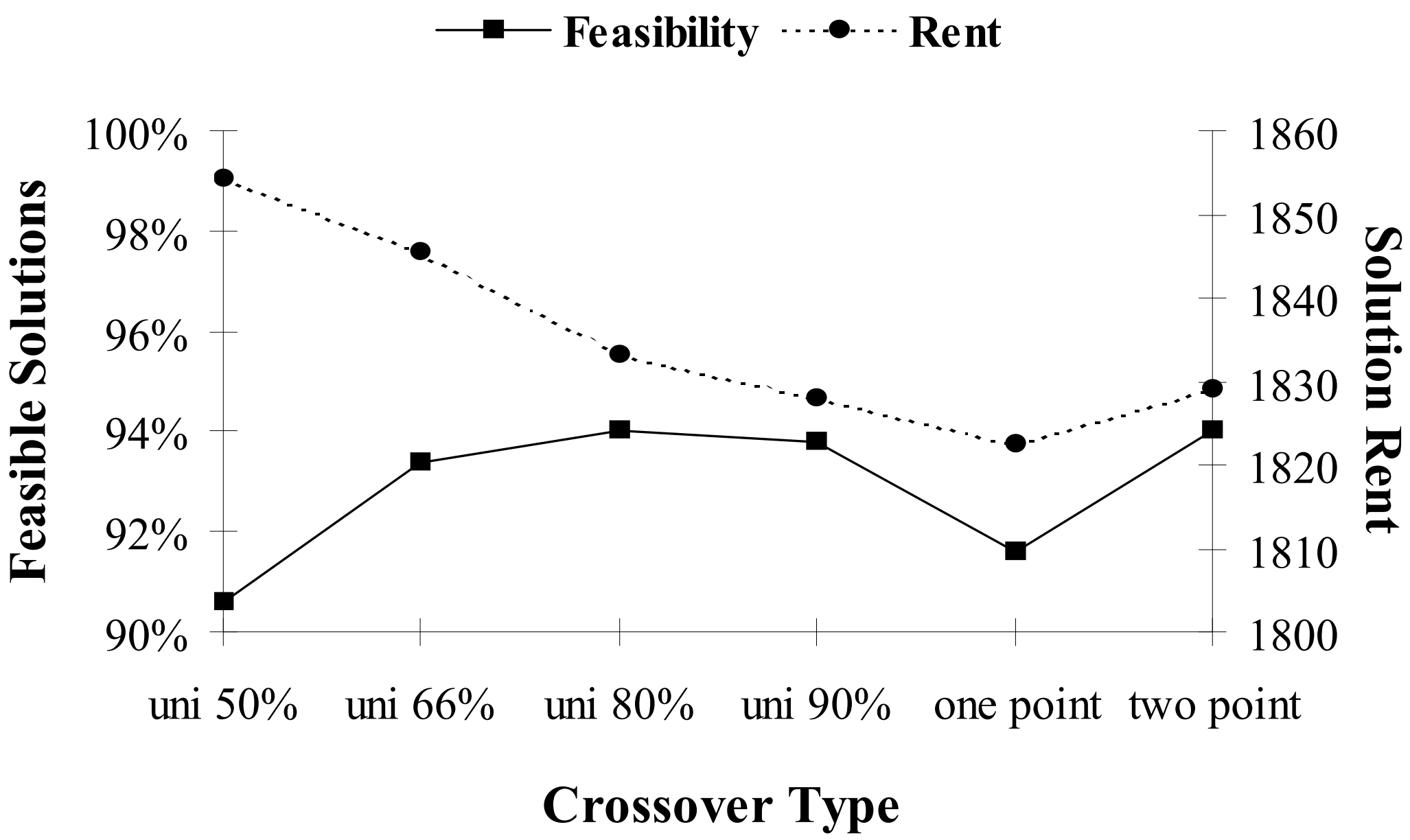

Figure 7-8: Crossover operators versus rent and feasibility.

## 7.3 Enhanced Direct Genetic Algorithm Approach

### 7.3.1 Co-Operative Co-Evolution

The results found so far with the direct genetic algorithm are of reasonable quality. Feasibility is over 90% and the rent values only improved slightly over the past experiments whilst being within 30% of the theoretical bound. This makes it an interesting situation in which to apply the co-operative co-evolutionary approach with hierarchical sub-populations, as detailed in section 5.2. There, this approach was very successful at improving the number of feasible solutions by overcoming epistasis.

Here, feasibility is not critical and hence it remains to be seen if the co-evolutionary approach is still capable of improving solutions.

The way in which to apply the sub-population structure and the 'grade-based' crossover is straightforward: In line with splitting the string into partitions with nurses of the same grade, the string is now split into the areas of the mall. Thus, we will have sub-strings with all the shops in one area belonging to them. These can then be recombined to create larger 'parts' of the mall and finally full solutions. However, the question arises how to calculate the pseudo fitness measure of the partial strings.

With the nurse scheduling, the objective function value of a partial solution was obtained by summing the $p_{ij}$ values of the nurses and shift patterns involved. Furthermore, we were able to define relatively meaningful sub-fitness scores by exploiting the 'cumulative' nature of the covering constraints due to the grade structure. Although the new pseudo covering constraints we deducted were not a perfect match to the original covering constraints, the pseudo fitness scores calculated with them allowed for an effective recombination of partial solutions for the nurse scheduling problem.

With the Mall Problem, the situation is more complicated since a large part of the objective function is a source of epistasis. Most of a solution's objective function value is 'area dependent', i.e. the size of shops (small, medium or large), whether a shop's group is complete or not, the area attractiveness factor and the fixed rent per area and shop type. The only exception is the shop count efficiency factor, which depends on the total number of shops of a particular type in the mall.

Apart from the fixed rent per area and shop type, all other objective function components add to the epistasis present in the problem. Another way to look at this is to see these objectives as 'soft' constraints similar to the use of penalty functions. Thus, the proposed partitioning of the string will not eliminate the objective function based epistasis fully.

The constraints are a second source for epistasis. In contrast to the objective function, these depend largely on the whole string, for instance the total number of shops of a particular size allowed. Only by adding up the shops and sizes for all areas does one know if a solution is feasible or not. Separating the constraints as before makes no sense now, because they are no longer of a 'cumulative' nature.

For instance, if there is a total of ten shops of one type allowed in the mall and there are five areas, then it would make no sense setting a maximum of two shops of this type for each area. This would clearly result in sub-optimal solutions, as for example no large shops could be built of this type. Equally, it does not make sense to divide the total number of small, medium and large shops allowed in total in the mall amongst the areas.

As explained above, the pseudo fitness of individuals in sub-populations can only be based on area dependent criteria. Therefore, neither the Shop Count Efficiency Factor nor any of the constraints can be taken into account. Thus, for each location in the area the following formula is used and the results are then summed up over all locations in the area to arrive at the pseudo fitness of a partial string:

$$\begin{bmatrix}\text{Attraction Factor of Area} \times \text{Total Group Bonus of Shop} \\ \times \text{Shop Size Efficiency Factor}\end{bmatrix} + \text{Fixed Rent Shop/Area}$$

Due to the complexity of the fitness calculations and the limited overall population size, we refrained from using several levels in the hierarchical design as we did with the nurse scheduling. Instead a simpler two level hierarchy is used: Five sub-populations optimising the five areas separately and one main population optimising the original problem. A special crossover then selects one solution from each sub-population and pastes them together to form a full solution.

Alternatively, we could have used a 'stair-case' design of hierarchical sub-populations, slowly building up to full solutions. Counting all possible combinations of five low-level sub-strings this would have led to ten pairs, ten triplets, five quadruplets and one full solution. Altogether, this equals 31 possible sub-strings respectively sub-populations. With a total population size of 200, implementing this would lead to far too few individuals in each sub-population. Thus this idea is not pursued any further in this thesis but may be a possibility for future research in combination with a larger population.

As for the nurse scheduling problem, within the sub-populations parameterised uniform crossover is performed to maximise diversity. The main population uses three types of crossovers. Some individuals are created from the five sub-populations, some are built from one sub-population and the main population and the remainder are made using parameterised uniform crossover within the main population. In detail, the model looks as follows:

- Five sub-populations with 25 individuals each, one main population with 75 individuals (Total population of 200 as before).
- Sub-population *u* optimises the rent for area *u* only (the remainder of the string is redundant). The above pseudo fitness function is used for this purpose.
- The main population uses the original objective.
- Crossover within the sub-populations is parameterised uniform with *p=0.66*.
- New individuals of the main population are created in equal parts as:
    - Assembled from the five sub-populations with a four-point 'grade-based' crossover.
    - Using a fixed point crossover between a member of the main population and an individual from a sub-population (taken with equal probability from any sub-population).
    - Using parameterised uniform crossover with *p=0.66* with individuals from the main population.

The results of this approach are displayed in Figure 7-9 under the 'Sub-Pops' label. For this experiment we used the above model with the remaining genetic parameters and strategies (rank-based selection, stopping criteria, random initialisation etc.) set as before. The results found are clearly disappointing, as they are worse than those achieved with the standard direct genetic algorithm (label 'Standard').

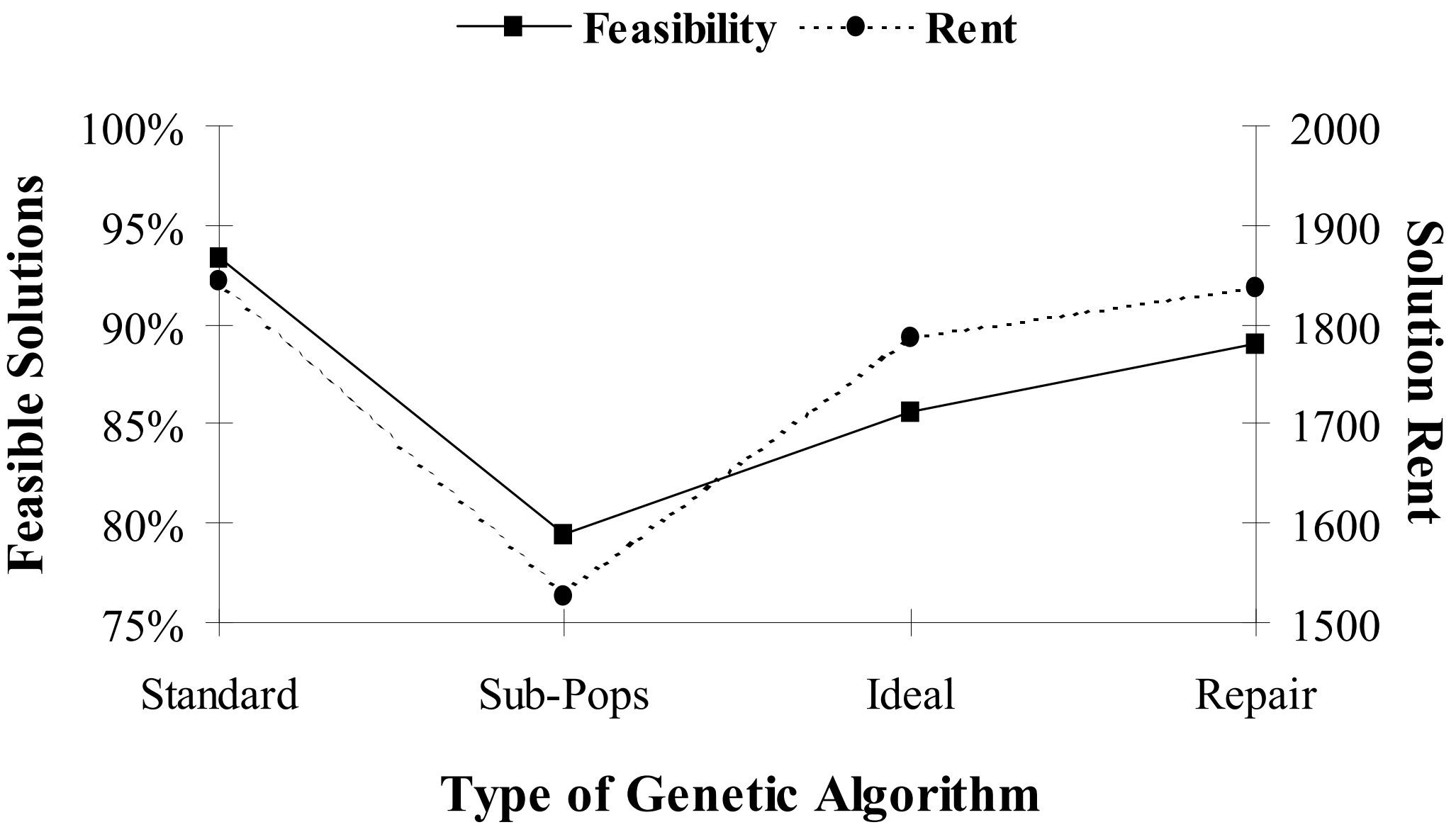

Figure 7-9: Comparison of various types of direct genetic algorithms.

However, watching the optimisation runs closely, it was quickly established why there was such a poor performance. Solutions of the sub-populations were extremely unlikely to be feasible for the overall problem, as they covered only one fifth of the string, with the remainder staying as originally initialised. It was equally unlikely for that third of solutions in the main population, which was formed from the five sub-populations alone, to be feasible. Although these solutions were of very high rent, because the sub-populations ignored the main constraints, their combination was unlikely to produce an overall feasible solution.

The situation was only slightly better with those solutions formed by a member of the sub-populations and a member of the main population. Usually, even if the member of the main population was feasible, the children were not. Again, even though the partial string from the sub-population member was of high rent, it was usually incompatible with the rest of the string, resulting in too many or too few shops of some types.

This mainly leaves it to the last third of the main population, the part using parameterised uniform crossover within the main population, to create feasible solutions. Hence, using this scheme of co-operative co-evolution is in fact similar to running a standard genetic algorithm with a population of size 25. This explains the poor results. In the next section, we will try to improve on this by combining partial strings more intelligently.

### 7.3.2 Mating

So far, the use of sub-populations has resulted in solutions of worse quality. As explained before, this is mainly due to recombining partial solutions that ignore the overall constraints and thus resulting in infeasible solutions. Here we propose an enhancement on the previous method for the middle third of the main population. Instead of combining any two parents, we let one parent select its ‘mate’ from a selection of parents. Thus, the first parent is chosen according to its fitness, whilst the second is chosen to be compatible with the first.

This approach was inspired by an idea presented by Ronald [139]. He solves Royal Roads and multi-objective optimisation problems using a genetic algorithm where the first parent is chosen following standard rules, i.e. proportional to its fitness. However, the second parent is not chosen according to its fitness, but depending on its ‘attractiveness’ to the first parent, which is measured on a different scale.

Our approach will be slightly different and only affect the second third of solutions created within the main population. This third was chosen, since it seemed to be the most promising part of the population to apply the following mating rules. The first parent is still chosen according to its rank from the main population. Then it is randomly decided from which sub-population the second parent should come. However, rather than picking one solution from this sub-population as parent, ten candidates are chosen, leaving the decision which one of these ten will become the second parent to a set of rules.

The rules are that the candidate combining best with the first parent regarding the ideal number of shops of each type is picked. To find it, the number of shops of each type in the first parent is added to those in the candidate and the difference of each shop type to its ideal count is summed up. The candidate with the lowest sum, or in the case of a tie, the first one with the lowest sum is picked. This type of rule rather than a full fitness calculation for all ten parent-candidate combinations was chosen because of the large computational cost of the latter. Furthermore, by having shops at their 'ideal' level, they are by definition above their minimum and below their maximum level, so those constraints should be taken care of implicitly.

Figure 7-9 shows the results of this mating approach under the 'Ideal' label. The results are clearly improved on from the previous ones, but still not as good as those of the standard direct genetic algorithm approach. Further investigations led us to look more closely at the solutions created by the mating. We found that although, as postulated above, the ideal level aimed for is between the minimum and maximum for each shop type, for many crossovers none of the ten mates was able to provide a feasible solution. This was due to two reasons: Either the shop size constraints were violated or even though an ideal number was aimed for, this was not achievable for all shop types and hence some shop count constraints were still violated.

As a final modification to the co-evolutionary approach, we tried to improve upon these results by repairing some solutions with regard to the second problem. In order to do this, a simple repair routine is introduced and applied randomly to 50% of all solutions created via mating. The repair scans the solution for shop types of which too few are present. It then substitutes a shop type of which at least the minimum number plus one is present, with a shop type whose count is below the minimum. If possible, this is done in an area where there is already a shop of this type, to avoid creating small and ineffective shops. As can be seen from the graph, the use of the repair algorithm improves results further. However, they still fall short both for rent and for feasibility of the results found by the standard direct genetic algorithm.

### 7.3.3 Summary of Direct Approaches

Overall, the standard genetic algorithm was able to solve the Mall Problem reasonably well, once a quick parameter optimisation was performed. Although we do not know the optimal solutions, an average feasibility of above 90% and only small improvements for rent for a large number of experiments seem to indicate that we have found solutions of at least reasonable quality. Furthermore, the solutions found were within 30% of a very optimistic upper bound. Overall the constraints were less tight than for the nurse scheduling and hence the genetic algorithm found it much easier to reach feasible solutions. Since this was already achieved with a simple fixed penalty weight, no dynamic penalties were used for the Mall Problem.

The results of the co-operative co-evolutionary approach were very disappointing, but easily explained. In contrast to the nurse scheduling problem, the constraints present in the Mall Problem are not decomposable according to the structure of the sub-populations. Hence, we were unable to calculate meaningful pseudo fitness scores. This led the sub-populations to optimise the rent of their area in isolation of what was happening in the other areas. So unsurprisingly, a combination of these partial solutions was unsuccessful because it usually violated the overall constraints.

However, one has to be cautious not to dismiss the idea of using co-operative co-evolution for the Mall Problem altogether. Because of the population size limit of 200, the sub-populations used were smaller than in the nurse scheduling case. It is well-known that once the population falls below a certain (problem-specific) size, genetic algorithms have great difficulty in solving the problem well. It is possible that the sub-populations were too small in this instance. At this stage, no answers to this question are known and the minimum acceptable size of such sub-populations is an interesting area for future work.

Furthermore, due to the limited overall population size, a more gradual build-up of sub-populations was not feasible. This would have led to even more and hence smaller sub-populations. However, this more gradual approach might have enabled the algorithm to find good feasible solutions by slowly joining together promising building-blocks. This is in contrast to the relatively harsh two-level design where building blocks had to 'succeed' immediately. Exploring the exact benefits of a gradual build-up of sub-solutions would make for another challenging area of possible future research.

Subsequently, the disappointing co-evolutionary results were improved with a mating and repair scheme. Although, results were still worse than for the standard direct approach, both ideas have their merits. Mating in combination with co-evolution is a very interesting concept and as shown is capable of improving results significantly. It is therefore quite possible that the failure was due to the underlying problem of the missing gradual build-up rather than a fault of the mating idea itself. This question can be answered once the above mentioned further research is carried out.

Of course, a similar argument applies to the results of the repair scheme. However, in comparison to the nurse scheduling approaches there is a further difference. There we were able to identify promising search states, i.e. *balanced* solutions, and then develop an effective repair mechanism. Additionally, unpromising (*unbalanced*) solutions were penalised and then quickly dropped from the population. For the Mall Problem, the situation is more complex due to the non-linear objective. Although some repairable features could be identified, repairing these in isolation does not lead to perfect solutions. What would be required is a much more comprehensive and sophisticated repair algorithm, which would possibly result in a stand-alone heuristic to solve the problem.

In conclusion, the simple direct genetic algorithm provided results of reasonable quality without the need for dynamic penalty weights. On the other hand, the co-operative co-evolutionary approach failed to achieve good results. This failure shows that the underlying structure in the nurse scheduling problem, i.e. the cumulative constraints and the day / night split, was an important factor in the success of the co-evolutionary schemes. However, some of the problems can possibly be attributed to the simpler two-level hierarchy and smaller sizes of the sub-populations used, but further research is necessary to confirm this. In the remainder of this chapter, we will investigate if an indirect genetic algorithm coupled with a decoder performs better than the direct genetic algorithms and conduct further work into the underlying structure of the Mall Problem.

## 7.4 Indirect Genetic Algorithm Approach

### 7.4.1 Encoding and Genetic Algorithm Set Up

In this section, we will describe the type of encoding used for an indirect genetic algorithm approach coupled with a decoder. As for the indirect genetic algorithm approach for the nurse scheduling problem in chapter 6, it is our intention to let the genetic algorithm find an optimal permutation of objects which is then fed into a deterministic decoder. This decoder builds the actual solution, in this case the layout of the mall, from the given permutations.

When deciding what the genotype of permutations should be, there are four choices: Shop types, shop groups, locations and areas. The match for the direct encoding would be a permutation of locations. However, before choosing this encoding, consideration must be given to the others, as they are potentially more effective encodings. Using shop groups or areas would be too general, like using grades in the nurse scheduling case. Thus we can rule shop groups out immediately, as that would leave far too many decisions for the decoder to make, i.e. choosing which shop type from the group to place it into which area. Furthermore, we would need extra rules to cope with those shop types that are in more than one or in no group.

We also decided against using a permutation of areas. With this encoding, the situation is slightly better than in the permutations of shop groups case, because all locations within one area are the same in our model. Thus, the decoder must only decide which shop type to place. However, with this permutation, all locations in one area need to be assigned shops before another area is taken into account. This would require a lot of forward planning, which again is outside the scope of a relatively simple decoder.

This leaves the decision between a permutation of shop types and a permutation of locations. For the indirect nurse scheduling, an encoding based on the permutation of shifts (equivalent to shops) would have resulted in a much more complicated decoder than one based on the permutation of nurses (equivalent to locations). The same is true here. An encoding based on a permutation of shop types has the advantage of being shorter (length is equal to maximum number of shop types = 50) than the alternative (length is equal to the maximum number of locations = 100). However, with a permutation of shop types, the decoder would have to decide how many shops of a given type to use and into which areas to place them. A permutation of locations only leaves the decision which shop type to place for the decoder.

In the former case, the decoder would have to choose from up to $\frac{15!}{10!\,5!} = 3003$ possibilities (five areas, maximum ten shops of one type in whole mall, i.e. five ways to partition 15 objects). In the latter, there are only up to 50 respectively 20 possible decisions (maximum number of shop types). This is a significant difference much simplifying and speeding up the second decoder in comparison to the first. Hence, the second type of decoder was chosen.

Now that we have decided on an encoding, the remainder of our indirect genetic algorithm can be set up. This will be along the same lines as for the direct genetic algorithm for the Mall Problem in section 7.2 and the indirect approach for the nurse scheduling in chapter 6. In other words, we use those strategies and parameters that in general proved best for the indirect genetic algorithm in chapter 6, which are PUX with *p=0.66* and standard swap mutation. All problem-specific parameters and strategies used, such as the penalty weight and penalty function, are those that worked best in section 7.2.4. Full details are shown in Table 7-3. No further parameter testing was performed.

| Parameter / Strategy | Setting |
|---|---|
| Population Size | **100** |
| Population Type | **Generational** |
| Initialisation | **Random** |
| Selection | **Rank Based** |
| Crossover | **PUX with *p = 0.66*** |
| Parents and Children per Crossover | **2** |
| Per Bit Mutation Probability | **1.5%** |
| Replacement Strategy | **Keep 10% Best** |
| Stopping Criteria | **No improvement for 30 generations** |

Table 7-3: Parameters used for the indirect genetic algorithm.

### 7.4.2 Decoder

As outlined earlier, the task of the decoder is to assign shop types to locations, which is equivalent to the nurse scheduling decoder where shift patterns were assigned to nurses. To decide which shop is best placed into the location currently under consideration, the decoder will cycle through all possible shop types. Obviously, those shop types for which the maximum number of shops allowed in the mall are already present will not be considered. Each candidate shop type is temporarily placed into the location and the following points are then taken into account to calculate its suitability:

- What is the fixed shop type per area rent for this shop type and location?
- How many shops of this type are already present in the mall and how does this compare to the minimum, ideal and maximum number allowed for this shop type?
- Would placing this shop (help) complete a group or is its group already complete?
- What shop size would be created by placing the shop and how would this affect the total number of small, medium and large shops allowed?

The decoder then assigns a score for each candidate according to its performance on the above points. The candidate with the highest score, or in the case of a tie, the first one with the highest score is placed into the location. Note that ties are very unlikely due to the randomness and large range of the fixed rent and therefore no special search orders are necessary for this decoder. Formally, the score $s_{ij}$ of a shop of type $i$ going into location $j$ is calculated as follows, where $w_1$-$w_6$ are appropriate weights:

$$s_{ij} = \begin{bmatrix} w_1(\text{medium bonus}) + w_2(\text{large bonus}) + w_3(\text{slackness in size constraint}) \\ + w_4(\text{ideal/total number of shop type}) + w_5(\text{new member}) \\ + w_6(\text{group complete}) + \text{fixed area/shop rent} \end{bmatrix}$$

Where the meaning of the terms in brackets is as follows:

- The 'medium bonus' and 'large bonus' are set to 1 if the shop would create the respective shop size, otherwise they are 0.
- The 'slackness in size constraint' is measured as the number of shops of the size, which would be created, that are still allowed in the mall. For instance, if a small shop was created and there is a total of 5 small shops allowed in the mall, with 3 small shops already present, then the 'slackness in size' would be 5 - 3 - 1 = 1. Note that this can lead to negative values for unsuitable shop sizes.
- The 'ideal / total number of shop type' is equal to the difference between the ideal and the total number for shops of this type if a shop of this type was placed. This encourages shops of those types to be placed, which are still below their ideal level, with the further below they are the higher the encouragement. Note that this should force the shop count for each type to be above the minimum and help it to remain below the maximum (due to minimum ≤ ideal ≤ maximum).
- The 'new member' is 0 if a shop of this type is already present in the area. Otherwise, it is calculated as (10 – total members of group the shop type is in + members already present in area). This encourages shops of those types whose group is more complete than others to be placed. Ten was chosen, as this is the largest possible group size. Thus, smaller groups are at an advantage, which is intentional as they are more likely to be able to reach completion than large groups. If a shop type belongs in more than one group, two scores are possible. If a shop type is in no group, the 'new member' score is zero.

- The 'group complete' is set to 1 if the group the shop type belongs to is complete in the area, otherwise it is set to 0. For shop types that belong into two groups, a score of 2 is possible. If the shop type is in no group, group complete is set to 0.
- The 'fixed area/shop rent' is equal to $r_{ik}$ of the shop type $i$ and area $k$ the location is in.

As for the nurse scheduling decoders used previously, let us compare this decoder with the rules set up by Palmer and Kershenbaum [125] and detailed in section 3.7. Rule one stipulates that we should be able to create every solution in the original solution space with the decoder. Analogous to the nurse scheduling decoders (refer to section 6.3.3), for a fixed set of decoder weights, there will always be a counter example. This is because the first shop type to be scheduled will always be the same for a specific location.

For the mall decoder and most weight settings, it will probably be the shop with the highest fixed area per shop type rent, as all other conditions are the same for most shop types. Therefore, the decoder will be unable to construct solutions such that all shop types are in areas where they do not have the highest fixed area per shop type rent. Although very unlikely as we are maximising the rent, this could prevent us from finding the optimal solution.

Rule two asks for all decoded solutions to be feasible. As mentioned previously, this distinguishes our decoder from others, as it is not always possible to meet all the covering constraints. For instance, there is a conflict between the shop count and the shop size constraints. Thus, we have to make use of the same penalty function based fitness as for the direct genetic algorithm in section 7.2.3.

As with the nurse scheduling, it is in the nature of the decoder to bias towards good and feasible solutions. As long as sensible decoder weights are chosen, all 'desirable' solutions of the original space are represented by decoded solutions. However, this does not mean that the whole solution space is covered evenly and hence rule three is violated. It remains to investigate sensible weights for this problem. Transformation between encoded and decoded solutions is reasonably fast as requested by rule four. As for the nurse scheduling, small changes in the encoded solution may or may not lead to small changes in the decoded solution (rule five), as domino effects are possible. Thus, as for the nurse scheduling decoder, most rules are violated. However, as outlined there, the violations are either intentional or unavoidable and hence do not render the decoder useless.

Although a powerful scoring mechanism taking all the problem details into account has been created, we are faced with one dilemma: How are the weights going to be set? In a first attempt, we tried the three sets of weights (Low, Medium, High) as detailed in Table 7-4, which are based on intuition, previous experience and the scatter graphs shown in Appendix F. Note that the labels refer to the relative 'magnitude' of the weights.

| Weight | Low | Medium | High |
|---|---|---|---|
| $W_{medium}$ | 500 | 500 | 500 |
| $W_{large}$ | 1000 | 1000 | 1000 |
| $W_{size}$ | 100 | 250 | 1000 |
| $W_{ideal}$ | 200 | 500 | 2000 |
| $W_{member}$ | 200 | 200 | 200 |
| $W_{group}$ | 2000 | 2000 | 2000 |

Table 7-4: Three types of weight settings for the Mall Problem decoder.

The results for these three weight sets can be seen in Figure 7-10. For the 'low' setting they are about as good as the direct genetic algorithm results, for the 'medium' setting, they are better and for the 'high' setting, they are far worse. Clearly, there is potential in the decoder, if only one could find the 'best' weights. Unlike for the nurse scheduling decoder weights, there are far too many possibilities for these decoder weights to conduct empirical tests. Thus, an 'automatic' approach will be introduced in the next section where the genetic algorithm sets the weights simultaneously whilst optimising the problem. Note that due to the difficulty of finding good manual weights for the decoder, no stand-alone results for the decoder without a genetic algorithm are presented.

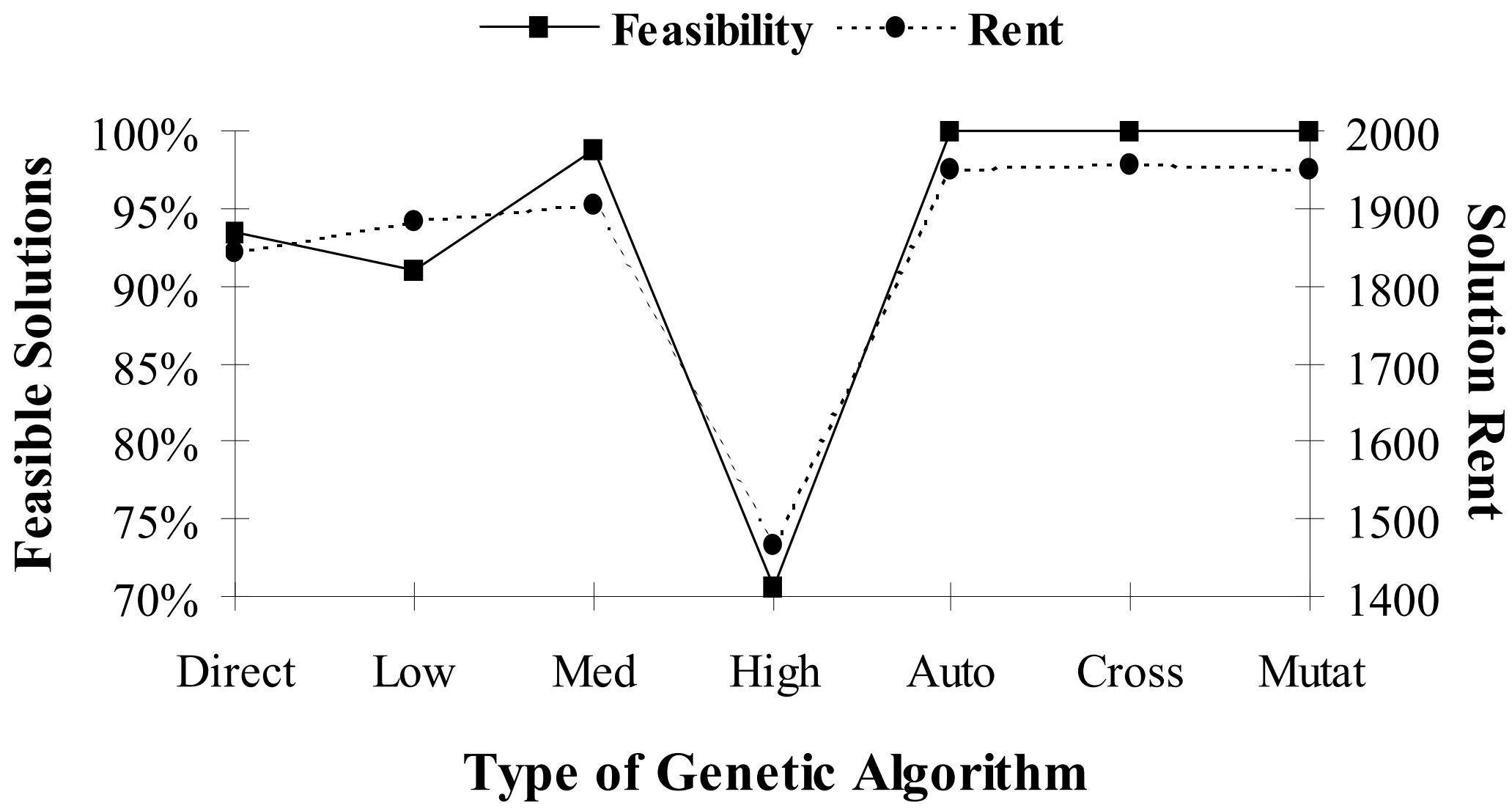

Figure 7-10: Various indirect genetic algorithms compared with the direct approach.

## 7.5 Further Decoder Enhancements

### 7.5.1 Automated Decoder Weights

The ability of genetic algorithms to set their own penalty parameters has already been investigated with adaptive penalty weights in section 4.4. Because of the higher number of weights, a slightly different approach has to be taken here. When initialised, each individual has a random set of weights $w_1 ... w_6$ assigned as six extra genes. Crossover is applied to these six extra genes so that weights producing better results have a better chance of being passed on. Similarly, mutation is used to avoid convergence and to re-introduce random values.

An additional advantage of these flexible weights is that the probability of not being able to create every solution in the solution space becomes even smaller than for a set of fixed weights. This is because different individuals will have different weights producing a greater variety of solutions. Thus, it is even less likely that the decoder will miss any valuable solutions.

The idea shares some similarities with the evolving schedule builder as suggested by Hart et al. [89], although our method is not quite as complex. Hart et al., who also solve a scheduling problem, extend the standard permutation and schedule builder approach by evolving the schedule builder itself. They do this by 'building' a unique schedule builder for each of their data sets, tailored to the characteristics of the problem.

This is achieved by having two additional sets of heuristics: One set determines how to split the task to be scheduled into parts and the other set how to assign those parts. For each task, there are about ten heuristics available. Their string then consists of multiple parts: First the usual permutation then a list of which heuristics to apply to this particular string. The heuristics are then evolved with standard crossover. This results in a robust and flexible system producing high quality solutions.

To execute our idea, a further six genes were attached to each individual which represent the six decoder weights for this particular individual. However, these extra genes are not treated as part of the string but have their own crossover and mutation operator. The values for the weights were randomly initialised in the range between 0 and 10000. This large range was chosen, as it was unknown in which region good values were lying. PUX was applied to the permutation part of the string as before. For the weights part of the string the following three crossover strategies for the weights of the children were tried:

- Taking the weights of a random parent.

- Taking the weights as rank-weighted averages of both parents
- Setting the weights at random in the range between the two weights of the parents.

Experiments showed that the last two methods performed well with the second method being slightly better and converging more quickly. The results for the second method are shown in Figure 7-10 under the 'Auto' label. They were the best found so far with 100% feasibility and improved rent. The first method did not produce as good results as the others, although still better than those found by the direct algorithm. We assume this was because this method was restricted to keeping to the weight values it was initialised at, which makes it less flexible than the other two methods.

We also wanted to make sure that the choice of 10000 as the upper initialisation limit did not have a negative effect on the optimisation. This could occur because the fixed area / shop rent component of the score has a fixed weight of one assigned to it. Therefore, the algorithm was rerun with an initialisation limit of 50000 and a different random number stream to avoid mere scaling. The results were of the same quality as before. Taking a closer look at the average weights of the final generations, some interesting similarities were discovered. In Figure 7-11, these weights are pictured as summarised by data sets.

As can be seen from the graphs, the weights behave in a very similar fashion for both choices of initialisation ranges. This indicates that the fixed area / shop rent has little to no effect on the decoder assignments. Furthermore, in 'easier' data sets, relatively higher weights were put on rent-increasing rules rather than on rules concerned with constraints. For data sets that are more difficult, the picture is different. Now more emphasis, or relatively higher weights, is put on the slackness and ideal rules, which both deal with the constraints. These results are very encouraging, as they show that the decoder behaves as intended, striking a good balance between feasibility and quality of results.

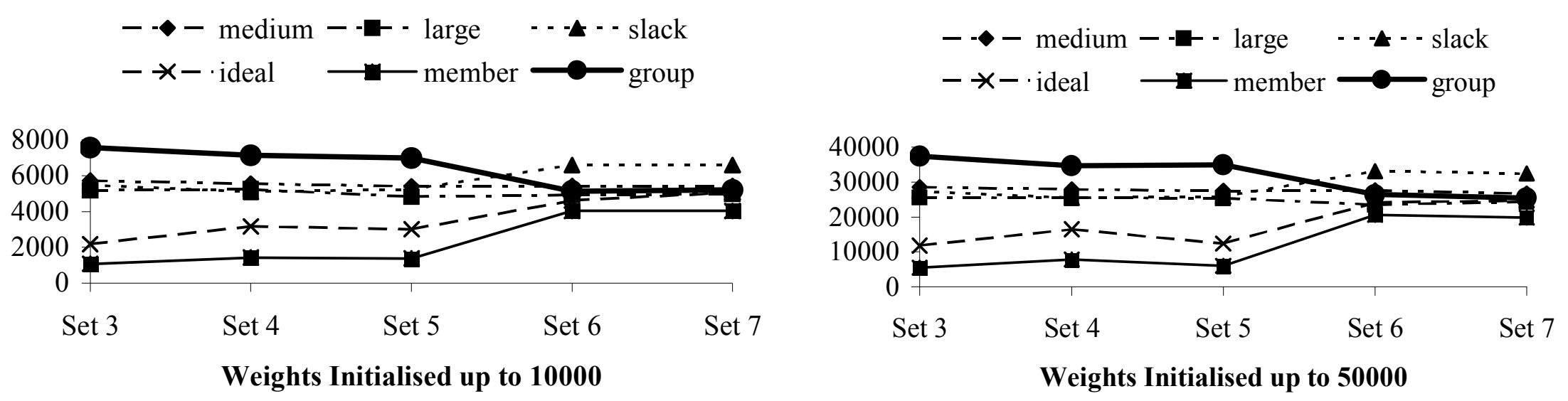

Figure 7-11: Final average weights for two initialisation ranges and different data sets.

### 7.5.2 Adapting Parameters and Strategies

After the success of the self adjusting decoder weights, to go one step further self adjusting crossover strategies and mutation rates are introduced. Such genetic algorithms that set their own parameters 'online' are not a new concept. For instance, Tuson and Ross [166] and Tuson and Ross [167] experiment with co-evolving operator settings in genetic algorithms. Further examples of adapting parameter ideas are presented in Davis [46] and Yeralan and Lin [182].

The system Tuson and Ross use is called COBRA, which is short for cost operator-based rate adaptation. COBRA is a learning rule method measuring benefit (as the increase of children's fitness over their parents' fitness) and cost (as computational effort to evaluate a child) of various operator settings. Those operators with a high benefit to cost ratio are assigned higher probabilities of being selected. After testing COBRA on a number of problems, the authors arrive at mixed results. Tuson and Ross conclude that adaptation is not necessarily a good thing and its success is problem dependent.

Here, we will follow a much simpler approach. Each individual receives two additional genes. The first is set to 1, 2 or 3 which indicates that respectively a C1, PMX or PUX with *p=0.66* crossover is performed. The second gene is a real number giving the swap mutation rate applied to the individual. The crossover tag is initialised at random, such that there is an equal probability for each crossover operator. The swap mutation rate is set randomly between 0% and 5%. When a crossover is performed, the child takes its crossover tag from the parent with the higher rank and the mutation probability of the child is set to the rank-weighted averages of the parents.

Figure 7-10 shows the results for adaptive crossover alone (label ‘Cross’) and for adaptive crossover and mutation (label ‘Mutat’). As can be seen, the use of adaptive crossover improves results, whilst the addition of adaptive mutation makes them worse. Whilst the former is investigated in the remainder of this section, the latter can be explained as follows.

Although a good setting for the mutation rate does help the algorithm as a whole, on an individual basis the mutation rate does not directly influence a solution such that those with an ‘optimal’ setting have necessarily a higher fitness. Therefore, what will happen is that the mutation rate converges to the average of the initialisation range, in our case 2.5%. This was observed to be happening in the experiments. However, for our particular problem a mutation rate of 1.5% yields better results than a rate of 2.5%. Thus, the overall worse results with adaptive mutation are due to the mutation probability being too high.

In Figure 7-12 we have a closer look at the performance of the various crossover operators. The adaptive setting (label ‘Adapt’) performs better than any of the three crossovers on their own. Similar to the results found for indirect genetic algorithm solving the nurse scheduling problem, PUX performs better than PMX and C1 is worst. We also ran a set of experiments where all three crossovers were used with equal probability throughout the optimisation (label ‘3 Cross’). This was to show that the gain in performance of the adaptive crossover was not simply based on the fact that more than one crossover operator was used at the same time. The graph shows that using three crossover operators together is better than using any in isolation, however the adaptive setting performed even better.

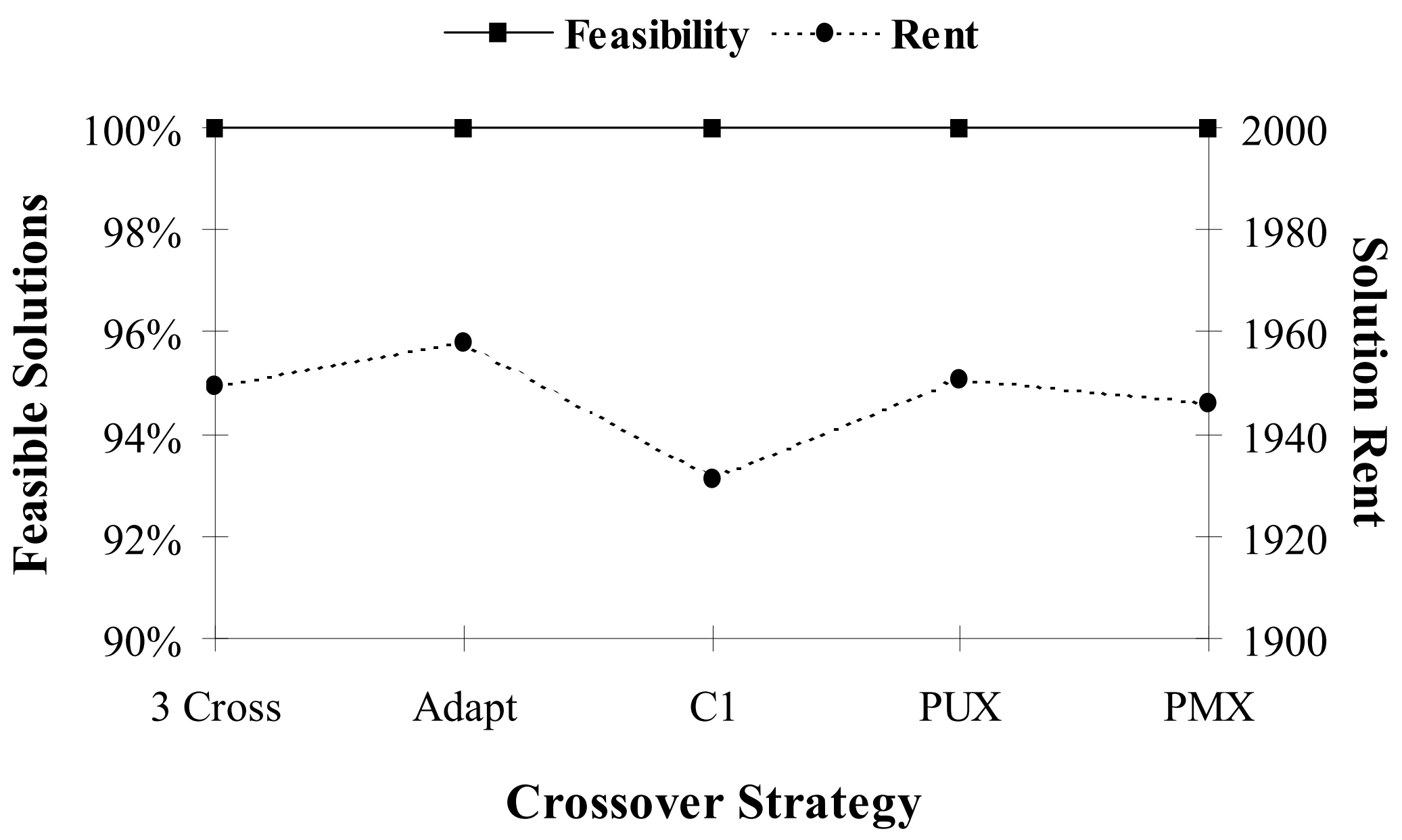

Figure 7-12: Different crossover strategies for the indirect genetic algorithm.

To further investigate what might have caused the better performance of the adaptive crossover we looked at a single run of three data files, one from set 4, one from set 5 and the last one from set 7. In Figure 7-13, Figure 7-14 and Figure 7-15 respectively, the average probability of a crossover type against the current generation is plotted. This reveals some interesting facts and indicates why the strategy works. Towards the end of the search for all three files, C1 crossover is preferred. Presumably, this is because the search is close to converging and a less disruptive crossover, which leaves big chunks of individuals intact, is performing best.

Furthermore, there is a clear tendency in the graphs that the more difficult the file is to optimise, in our case because of tighter constraints, the longer the more disruptive and more flexible PMX and PUX operators were used. Particularly for the file of set 7, the C1 operator hardly features at all during the first half of the search because its use would be too 'conservative'. The graphs also show that generally PUX is preferred over PMX and that 'harder' problems require more generations to be solved.

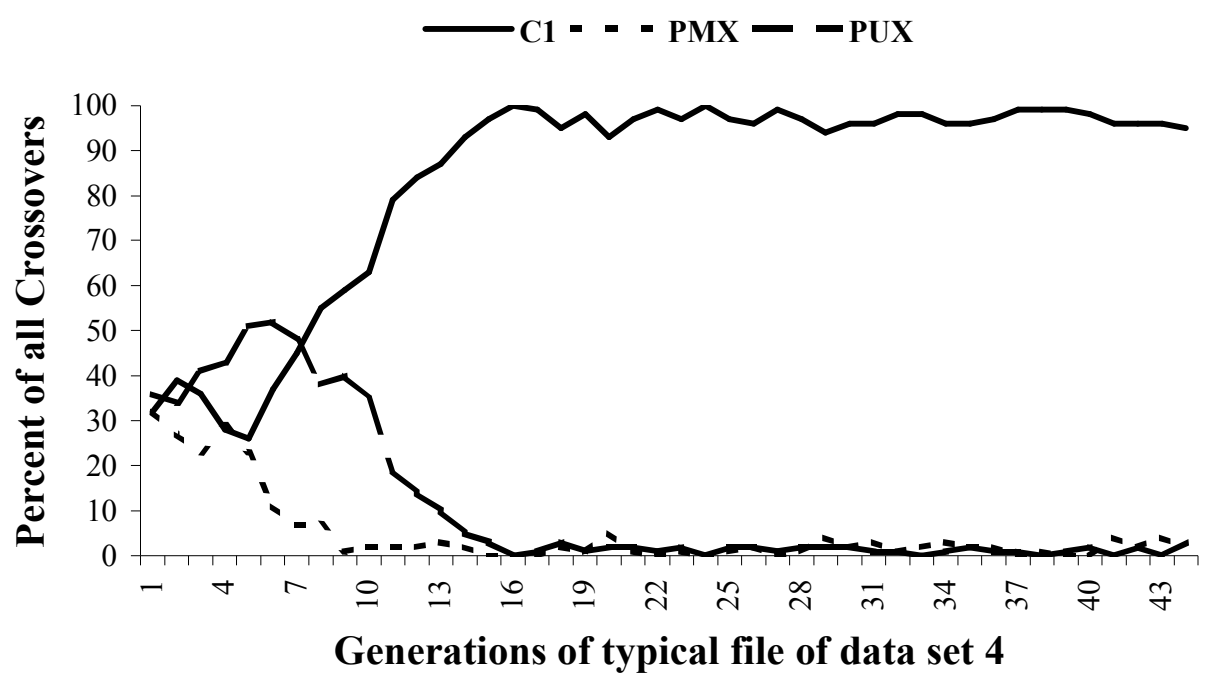

Figure 7-13: Crossover rates for adaptive crossover and a ‘relaxed’ file.

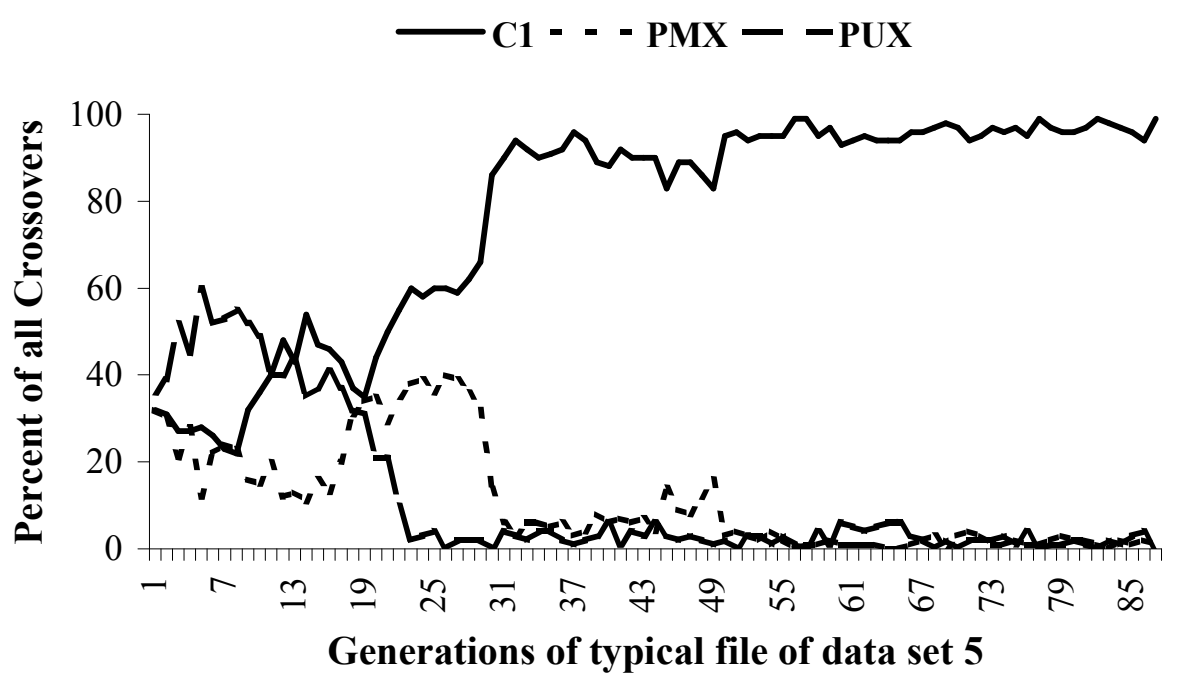

Figure 7-14: Crossover rates for adaptive crossover and a file that is ‘tight’ on the shop count.

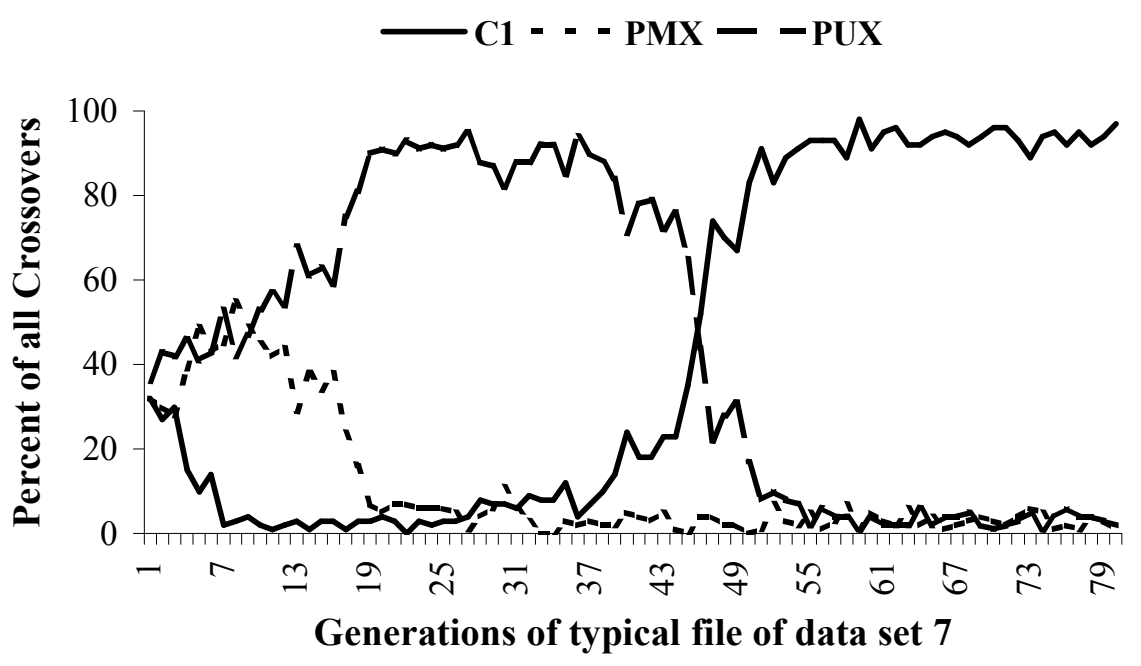

Figure 7-15: Crossover rates for adaptive crossover and a ‘tight’ file.

## 7.6 Nurse Scheduling Revisited

The success of the adaptive approaches for the Mall Problem motivated their application to the indirect genetic algorithm used for the nurse scheduling problem. To do this, the decoder remained unchanged, as described in section 6.5.1, apart from the decoder weights. The weight for the preferences $w_p$ was set to one and the weights $w_s$ of covering a shift of grade $s$ were initialised at random for each individual in the range between 0 and 100.

As before, two ways of crossing over the weights were tried. One was to set the weight of a child equal to the rank-weighted average (label 'Rank') of both parents' weights. The other method set it at random in the range between the two parents' weights (label 'Range'). We also tried the self-setting crossover approach (label 'Cross'). As for the Mall Problem, individuals were initialised with a random crossover operator (C1, PMX or PUX) and the parent with a higher rank decided which crossover to use and passed this on to the child. Again, mutating a string also mutates the decoder weights and the crossover type by re-initialising them in the full ranges.

The results of the experiments with these new strategies are displayed in Figure 7-16. When compared with the original results with fixed weights and crossover type (label 'Fixed'), the outcome is very interesting. Although not clearly dominating the original results, the new results are at least of the same quality and this was achieved without having to perform lengthy parameter setting experiments. As for the Mall Problem, there is little difference between the two ways in which the weights of the children are calculated. When both self-setting crossover and weights are combined (labels 'C+Rank' and 'C+Range'), the results are clearly better than those found by the original approach. These are extremely encouraging results, indicating the general suitability of the indirect genetic algorithm with self-setting decoder weights and crossover rates for the optimisation of multiple-choice problems.

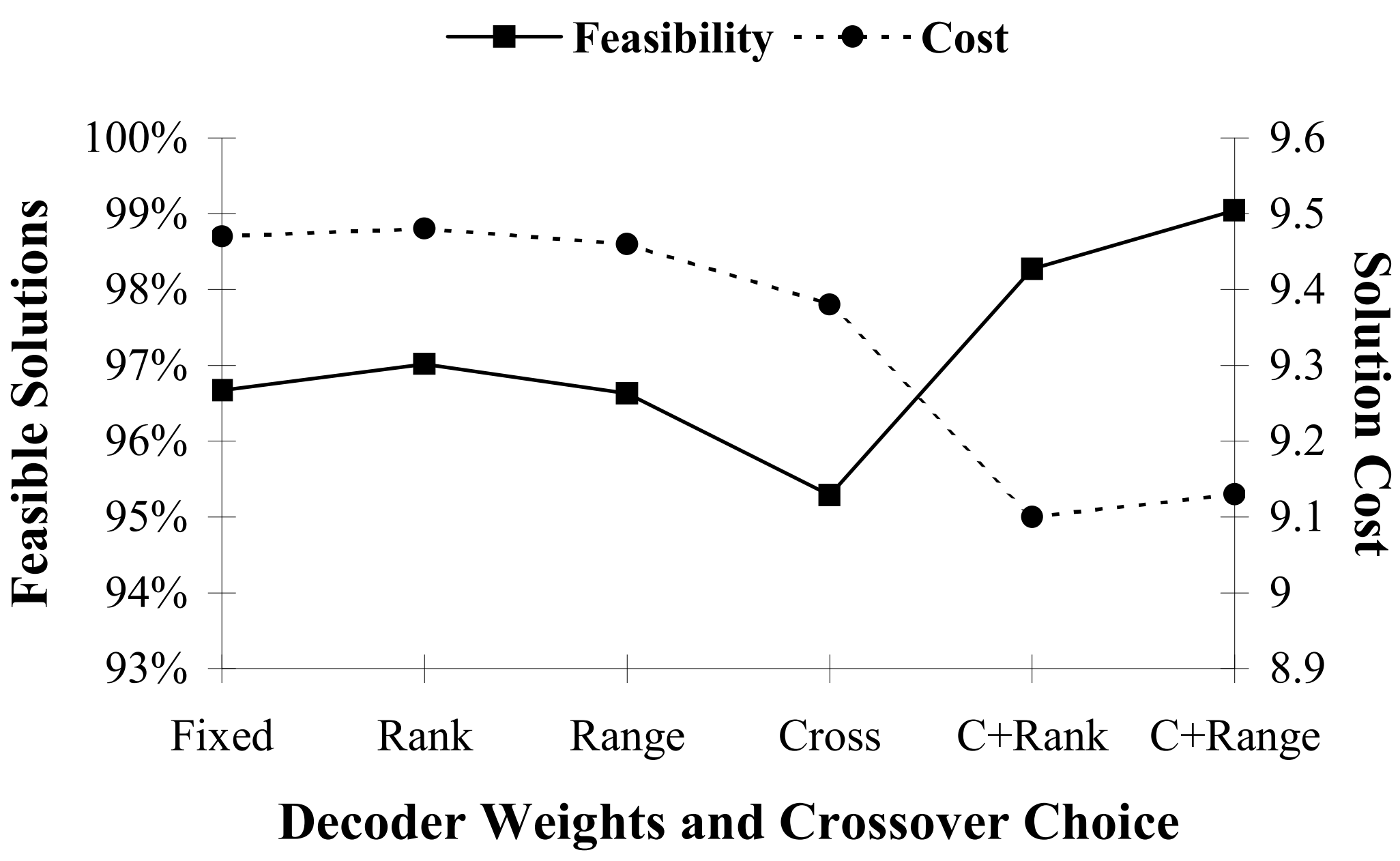

Figure 7-16: Variations of the decoder weights and crossover strategies for the indirect genetic algorithm solving the nurse scheduling problem.

## 7.7 Conclusions

As for the nurse scheduling problem, the indirect approach proved to be superior to the direct genetic algorithm optimisation. However, the differences were less severe due to the way the data was created with an emphasis on less tight constraints. Nevertheless, one might argue that the indirect algorithm and in particular the decoder was supplied with far more problem-specific information than the direct genetic algorithm. And indeed, this is true. However, this was done because the indirect approach lends itself more easily to the inclusion of such information than the direct algorithms. This is an important lesson to be learnt from this research. A final summary comparing the Mall Problem's results is shown in Figure 7-17.

Problem-specific information was included in the direct case, but the attempts at using co-operative co-evolution were ill-fated. As explained in section 7.3.3, it would be premature to discard the co-evolutionary idea as useless for the Mall Problem. The most likely reason for the failure lies in the population size limit, which denied a gradual build-up of sub-solutions. Alternatively, it could be that further research into the problem structure might yield a more promising way of defining the sub-populations. In any way, further research is necessary to establish the exact relationship between the hierarchical structure of sub-populations, the build-up of sub-solutions and the total population size.

Injecting further problem-specific information, in the way of mating, helped, but the naive direct approach was still superior. A final attempt to improve solutions was made with a repair operator. As with the genetic algorithm for the nurse scheduling problem, the operator was aimed at improving the feasibility of solutions. However, unlike for the nurse scheduling problem, no equivalent *balanced* or *unbalanced* situations could be exploited. Thus, the impact of the repair algorithm was only limited.

With the indirect genetic algorithm and decoder, it was relatively easy to incorporate problem-specific information, particularly in the form of making active use of constraints rather than as ‘dumb’ background penalties only. This again showed the advantages of this type of optimisation in comparison to the traditional genetic algorithms. The only problem faced was due to the sheer number of problem-specific features that were included in the decoder. Therefore, suitable weights needed to be found.

This was achieved by having some extra genes representing the weights and leaving the choice to the genetic algorithm via some special operators. This idea was extended by allowing the genetic algorithm to also choose the type of crossover and mutation rate employed. The results found were of excellent quality with 100% feasibility and further improved rent, which was within 15% of a very optimistically set upper bound. Another advantage is that these results were achieved without the need for lengthy and possibly complicated parameter tests to find ‘ideal’ penalty weights and crossover rates. Moreover, these dynamic parameters have the benefit to be able to adjust with the search as has been demonstrated for the crossover rates.

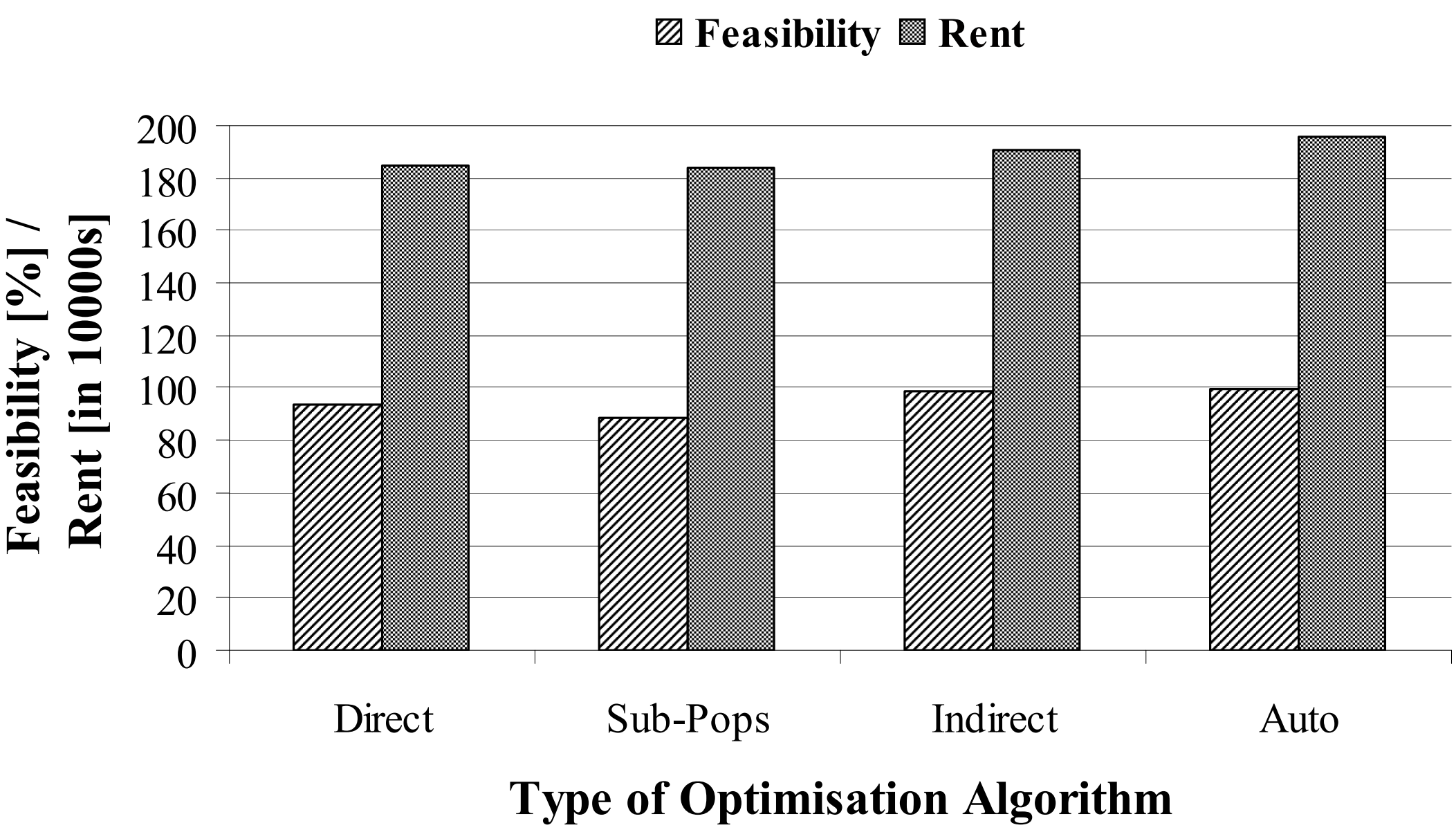

Figure 7-17: Various genetic algorithms for the Mall Problem.

To prove the general suitability of these ideas, they were re-applied to the indirect nurse scheduling approach. Again, the results were of better quality than before without any parameter tests being required. This indicates that there is potential in using this type of score assigning decoder plus a permutation based genetic algorithm for other scheduling and similar problems. The principles used in this decoder are widely applicable: Schedule one candidate at a time, measure the remaining slackness of constraints, identify positive and negative contributions to the objective and finally assign a weighted score. Of course, finding good weights will be difficult, but as shown in this research it can be left to the genetic algorithm to sort the weights out. This clearly is a very promising area for further work.

# 8 Conclusions and Future Research

## 8.1 Conclusions

The work described in this thesis has demonstrated that with help, genetic algorithms are capable of solving constrained multiple-choice problems. In particular, two major avenues for providing such help by balancing feasibility with solution quality have been outlined: A direct and problem-specific approach exploiting problem structure and a more generic indirect scheme. The former took the form of special operators for the direct genetic algorithm. In particular a co-operative co-evolutionary algorithm with hierarchical sub-populations was used. The latter used the problem-specific information contained in a set of heuristic rules as part of an indirect genetic algorithm with a self-adjusting schedule builder. For both the nurse scheduling and the mall layout problems, the indirect genetic algorithm was easier to implement and more effective. It is conjectured that this is because this approach does not rely on the problem structure. Indeed some evidence points to the fact that doing so can also be counter productive, as seen for seeding or the co-evolutionary approach to the mall layout problem.

At the heart of the direct approaches lies the co-operative co-evolutionary scheme with hierarchical sub-populations. This approach was chosen after the failure of the canonical genetic algorithm to solve the nurse scheduling problem adequately. Even with optimised parameters (as detailed in chapter 4), it was unable to provide good results to the problem. Particularly disappointing was the poor performance of the penalty function with a fixed penalty weight. The introduction of more sophisticated truly adaptive penalty weights improved results somewhat, but nowhere near to those achieved by tabu search.

The poor results of the genetic algorithm so far were explained with the epistasis created by the penalty function approach, which is in conflict with the building block hypothesis. In fact, the epistasis is twofold: One level is due to the covering constraints, whilst the second level is created by higher graded nurses being allowed to cover for nurses of lower grades. It is this second level of epistasis that the co-operative co-evolutionary algorithm intended to reduce in chapter 5.

Thus, in this approach lower level sub-populations solve the scheduling problem for one grade band only, whilst higher level sub-populations solve the problem for a combination of grades. To make this scheme viable, new sub-fitness functions were set up based on the original constraints, which had been split up into sub-constraints. This approach was chosen over the 'compatibility' based sub-string fitness suggested by Potter and De Jong [129] and used for instance by Hernandez and Corne [94]. The drawback of their method is that either a large number of computationally expensive fitness calculations have to be made or an unreliable sub-fitness is obtained due to sampling errors.

To overcome these difficulties we proposed the hierarchical approach: Computationally fast sub-fitness scores are calculated directly and a slow build-up to more and more complete solutions improves the compatibility of sub-strings over time. Thus, information is conveyed from lower level sub-populations to higher level sub-populations by means of a special 'grade-based' crossover operator. To promote further information exchange between sub-populations, migration of individuals between sub-populations was introduced. The results of this approach were very encouraging, with good solutions for most data sets.

Another conclusion from the direct approaches is that repair and mutation operators are worth trying, but some intelligence is necessary to make them successful. In particular the incentive / disincentive approach targeting repair at worthwhile candidates fared well. This was implemented as a two-way strategy to overcome the 'lack of killer instinct' effect. Firstly, the genetic algorithm was led away from solutions that would most likely become a dead-end. Secondly and simultaneously to the first operation, promising *balanced* solutions were encouraged and exploited. With these enhancements in place, the genetic algorithm achieved results that were comparable in quality to those of tabu search.

In chapter 7, we applied the same direct techniques to the non-linear mall layout and tenant selection problem. The data and problem details were set up in a way that it was easier to reach feasibility but more difficult to find an optimal solution than for the nurse scheduling problem. Predictably, the canonical direct genetic algorithm found the Mall Problem easier to solve, with results being of reasonable quality. We then turned our attention to using a co-operative co-evolutionary approach for the Mall Problem.

Again, sub-populations were defined such that they optimised only parts of the problem, in this case only one area of the mall. As with the nurse scheduling, a higher level main sub-population gathered the information found by the lower level sub-populations. Unfortunately, the problem structure did not allow the constraints to be differentiated as easily as in the nurse scheduling case. Thus, the sub-fitness scores were less meaningful than for the nurse scheduling problem. More importantly, due to the limited population size, only a two-level hierarchy of sub-populations was feasible without making the sub-populations too small. Thus, the approach failed to deliver good results. Further enhancements of the scheme, namely mating and repair, improved results somewhat, but not to a satisfactory level.

The failure of the co-operative co-evolutionary genetic algorithm teaches some important lessons. Although it is a very powerful approach and was the key to success for the nurse scheduling problem, it is imperative that meaningful sub-fitness scores are used and a gradual build-up of sub-solutions is allowed. Intuitively, one can be substituted for the other, i.e. if good sub-fitness scores are available less hierarchical levels are required and vice versa.

To find those sub-fitness scores, it is not simply enough to divide the problem into sub-problems. Furthermore, one must also be able to ‘divide’ the constraints into ‘sub-constraints’ such that their ‘reunion’ is similar to those found in the original problem. Hence reducing the number of sub-population levels at the same time as having worse sub-fitness scores was the downfall of this approach for the Mall Problem. However, more research is necessary to establish the precise relationship between the accuracy of the sub-fitness scores and the number of sub-population levels required.

The second major avenue we followed to solve multiple-choice problems was using indirect genetic algorithms. For the nurse scheduling, the genetic algorithm was set up to find permutations of the nurses which where then fed into a decoder that builds the roster from this list. Following on, three decoders employing deterministic scheduling rules were proposed in chapter 6. The more sophisticated combined decoder worked best and, after some parameter tests to find good weights for the scheduling rules, provided better results to the nurse scheduling problem than any of the direct genetic algorithm approaches.

These better results are directly attributed to the advantages of the 'two phased' indirect approach. With the direct genetic algorithm, it was very difficult to include problem-specific knowledge. For instance, constraints could only be used passively in the background to penalise solutions. Furthermore, a lot of effort was involved when repair operators and other local improvement heuristics were implemented as not to upset the general balance between exploitation and exploration. This resulted in long developing times and numerous parameter tests.

With the indirect approach, the situation is different, since the genetic algorithm can be left almost unchanged from the canonical version. Problem-specific knowledge can be included easily within the heuristic rules of the decoder. Thus, constraints can play an active role in guiding the search rather than just being in the background.

When using an indirect approach with a separate decoder, it is important to note that unlike the relaxed approaches described in the literature, in our case there is no guarantee that feasible solutions will be found. Hence, this scheme still needs to balance feasibility and solution quality. Although this and other characteristics of the decoders violate the decoder rules of Palmer and Kershenbaum [125], we found no conflict in doing so. In fact, some violations of these rules, for instance biasing towards promising regions of the solution space, seemed indeed helpful.

Finally, the more complicated the problem, the more scheduling rules or decoder weights will be needed. This has been demonstrated with the Mall Problem in chapter

7. There at least six decoder weights were required, making exhaustive parameter tests very cumbersome. This issue was confronted by proposing self-adjusting decoder weights. These work in a similar manner to adaptive penalty weights, i.e. the control of the decoder weights was handed over to the genetic algorithm. Each individual had its own set of weights attached and by using an appropriate crossover strategy we ensured that better individuals and hence better sets of weights contributed more to future decoder weights.

This approach proved highly effective, eliminating the need for parameter tests and yet producing better results than ever before for the Mall Problem. As we have shown, a particular benefit of these self-setting weights is that the weights can adapt with the differences in the data sets. Spurred on by this success, we extended the self-regulation of the genetic algorithm to crossover and mutation. Although no improvements were made in the mutation case, letting the algorithm decide which crossover to apply further improved results.

In a final set of experiments, we re-applied the idea of self-setting decoder weights and self-regulating crossover to the indirect genetic algorithm for the nurse scheduling problem. Again, the results were better than any others found previously by the genetic algorithm. This clearly shows the potential of this self-regulating indirect genetic algorithm for the optimisation of other problems.

## 8.2 Future Research

There are many avenues of future work arising from this thesis. For the problems already solved in this thesis, the genetic algorithms used could be further extended by the introduction of an intelligent mutation operator or a steady-state reproduction scheme. Although we do not anticipate a significant improvement in solution quality by these measures, it would be interesting to see if there are any conflicts between them and the enhancements already detailed in this thesis. However, of more interest is future work into the new schemes of co-operative co-evolution with hierarchical sub-populations and self-regulating indirect genetic algorithms.

The results so far for the hierarchical sub-population approaches are of mixed quality. The idea worked well for the nurse scheduling, but it failed for the Mall Problem. As explained before this was due to the differences of the objective functions and constraints of the two problems and the limit on the number of sub-populations. Nevertheless, it would be interesting to apply this approach to other scheduling or related problems with the aim of setting up guidelines under which conditions this type of algorithm can be successful. In particular the relationship between the number of sub-population levels required and the 'accuracy' of sub-fitness scores needs to be investigated further.

Areas of other possible future enhancement of this scheme include: A smart migration, for instance swapping particularly useful information; an extension of the mating idea presented combining those individuals that are most suitable; coupling Delta Coding with the use of sub-population based hypercubes and an increase in the number of 'main' populations, possibly with different characteristics.

The self-regulating indirect genetic algorithm approach has probably even greater potential because it relies less on a problem's structure. It can be used for any type of constrained or unconstrained optimisation problem. A generic way of applying it would be to use a permutation based encoding and then 'schedule' one object at a time, as we did for the mall layout and nurse scheduling problems.

The schedule builder would then measure the suitability of each object choosing the best for each slot. This measurement can be based on a mixture of slackness left in the constraints if the object was scheduled and its contribution to various target function terms. All these will have a weight assigned that will be regulated by the genetic algorithm itself in the way described earlier. Additionally, the control over the crossover operator can be left to the algorithm, too. This 'generic' indirect algorithm approach surely has considerable potential to be exploited.
Whilst experimenting with various permutation based crossover operators, a new operator named PUX was invented. PUX is a combination of standard parameterised uniform crossover and order based crossover. In this thesis, we have shown this operator to perform better than its closest rivals uniform order-based crossover and PMX. Although intuitive reasons were presented for this, more work is needed to establish the precise behaviour of this operator.

Thus, it would be beneficial to apply PUX to other problems and compare it to other permutation based operators, as well as perform a forma analysis along the lines of Cotta and Troya [43]. Additionally, one could imagine a different kind of PUX where $p$ varies with the generations or is even controlled by the genetic algorithm itself. For instance, if $p$ would slowly increase from 0.5 to 1 over the generations this would simulate the observations made for the genetic algorithm controlled crossover rates.

Another idea is to combine constraint logic programming with the indirect genetic algorithm plus a schedule builder. Barnier and Brisset [14], Bruns [29] and others have done this with good results for the direct standard genetic algorithm. Thus, it seems there would be good potential for including some constraint programming ideas within the decoder. For instance, these could be used for an intelligent look-ahead device or to schedule the last few or some nasty items.

Finally, there is the still unexplored idea of oscillating fitness functions. Although their straightforward implementation proved unsuccessful during the pilot study to this thesis, a more sophisticated approach could yet lead to success. This could for instance take a form similar to the hierarchical structure of the sub-populations in the co-evolutionary approach. Rather than just two functions as in the pilot study, there could be a 'main' function and some 'sub'-functions. The genetic algorithm could then oscillate between the main function and one sub-function at a time, possibly concentrating on certain functions depending on the course of the optimisation.

# Appendix A  Genetic Algorithm Tutorial

## A.1 What is a Genetic Algorithm?

This appendix is an introduction to genetic algorithms for a newcomer to the field and is loosely based on Whitley [174] and Davis [48]. Genetic algorithms are a class of meta-heuristics, which are modelled on the idea of evolution in nature. Their main distinctions from other methods are that

- No problem-specific information other than the objective function is needed.
- The genetic algorithm works with encoded variables, not with the actual variables themselves.
- A number of solutions are processed in parallel at the same time.
- Most operators used are stochastic not deterministic.

Before presenting a genetic algorithm in more detail in the next section, we would like to explain some of the terminology used, which is mainly derived from biology and evolutionary studies.

A variable is referred to as a *gene* or *allele*, whilst a solution is known as an *individual*, a *string* or a *chromosome*. The position of a gene is called *loci*. Usually, solutions need to be *encoded* in order that genetic operators can be applied. This will become clearer with the example in the next section. An encoded solution is then called *genotype* whilst a decoded solution is called *phenotype*. To measure the quality of a solution the term *fitness* is used, i.e. a fitter solution is a better solution. All solutions together make up a *population* and one iteration of a genetic algorithm is known as a *generation*.

# A.2 How Does a Genetic Algorithm Work?

## A.2.1 Introduction

This section will describe a canonical or original genetic algorithm. In particular, the main genetic operators of selection, crossover, mutation and replacement will be presented. After giving the following overview of the canonical genetic algorithm scheme, all steps will be explained in detail:

(a) Encoding of solutions and random initialisation of first population.
(b) Fitness evaluation of all solutions in the population.
(c) Selection of the parent solutions according to their fitness.
(d) Crossover between parents to form new solutions.
(e) Mutation of a small proportion of these new solutions.
(f) Fitness evaluation of new solutions.
(g) Replacement of old solutions by new ones keeping some of the (best) old solutions.
(h) Returning to step (c) if stopping criteria is unfulfilled.

## A.2.2 Encoding of Solutions

It is necessary to encode solutions into a format such that the genetic operators, described in the following paragraphs, can be applied. The technique for encoding solutions into chromosomes will vary from problem to problem and from genetic algorithm to genetic algorithm. Originally, encoding was carried out using concatenated bit strings of all variables. However, many other types have been used and it is probably fair to say that no one technique is best for all problems.

Hence, choosing an encoding that is right for one's problem is often considered to involve a certain amount of art and experience. Two things to bear in mind are the calculation of the fitness via an evaluation function and the inner workings of crossover and mutation. A good encoding may also be able to include some of the constraints implicitly (see section 3.3 for a discussion of this). Examples of (not necessarily good) encodings are:

- A list of the cities in the order visited for a travelling salesman problem.
- A concatenated string of all solution variables, either kept in their original format or transformed into the binary system.
- For the nurse scheduling problem, a binary list of the size (number of nurses) × (number of days) × (shifts per day), with a 1 indicating a nurse works the shift and a 0 otherwise. For further details on the encodings of the nurse scheduling problem, refer to sections 4.1 and 6.3.1.
- For the shopping mall problem, a list of size equal to the number of shop types, each element indicating the total number (measured in small units) of shops of that type present in the mall. For more information on encodings for the shopping mall problem, see sections 7.2.2 and 7.4.1.

### A.2.3 Fitness Evaluation

An evaluation of a solution's fitness is necessary for the selection stage, where fitter individuals will have a better chance of being selected as parents and can therefore pass their genes on. The raw fitness of an individual is usually equivalent to its objective function value, sometimes modified due to constraint violation via penalty functions. More information about penalty functions can be found in section 3.4.

However, some modification of the raw fitness is necessary to avoid the problem of domination and lack of selection pressure. Domination can happen if the initial population contains one or more 'super individuals' with much higher fitness than the rest of the population. If left unchecked this would lead to premature convergence to a region very close to this individual, which might be undesirable. Lack of selection pressure can happen towards the end of the search when many individuals have similar raw fitness values. In this situation it would be preferred if the better individuals were still picked more often than average ones, but because of the similar raw fitness values this is unlikely to happen.

The answer to these problems is to use some form of fitness scaling or ranking. However, fitness scaling has the disadvantage that a problem-specific scaling function has to be found. Assigning ranks is more problem independent and also more 'naturally' inspired, but there is the drawback of having to sort the population first which can be computationally time intensive. Nevertheless, our preferred method is to assign ranks.

### A.2.4 Selection of Parents

Once individuals have been assigned an adjusted fitness or in our case a rank, selection can take place. Selection is necessary to create new solutions via crossover. Depending on the crossover operator used (more about this in the next section) the appropriate number of parents needs to be chosen from amongst the whole population. To further the idea of the 'survival of the fittest', the fitter an individual, the more likely it is to be picked as parent.

The most common method of achieving this is via 'roulette wheel' selection. This can be imagined as spinning a big roulette wheel with as many slots as members in the population. The wheel is spun once for each parent required. The width of the slot and thus the chance of an individual being chosen for parenthood is proportional to its fitness. In the case of ranking, the slot would be proportional to the individual's rank. For instance, the best individual out of a population of 100 would have a rank and slot width of 100, the second best would have a rank and slot width of 99 etc. An example for five individuals, with the label indicating the rank is shown in Figure I.

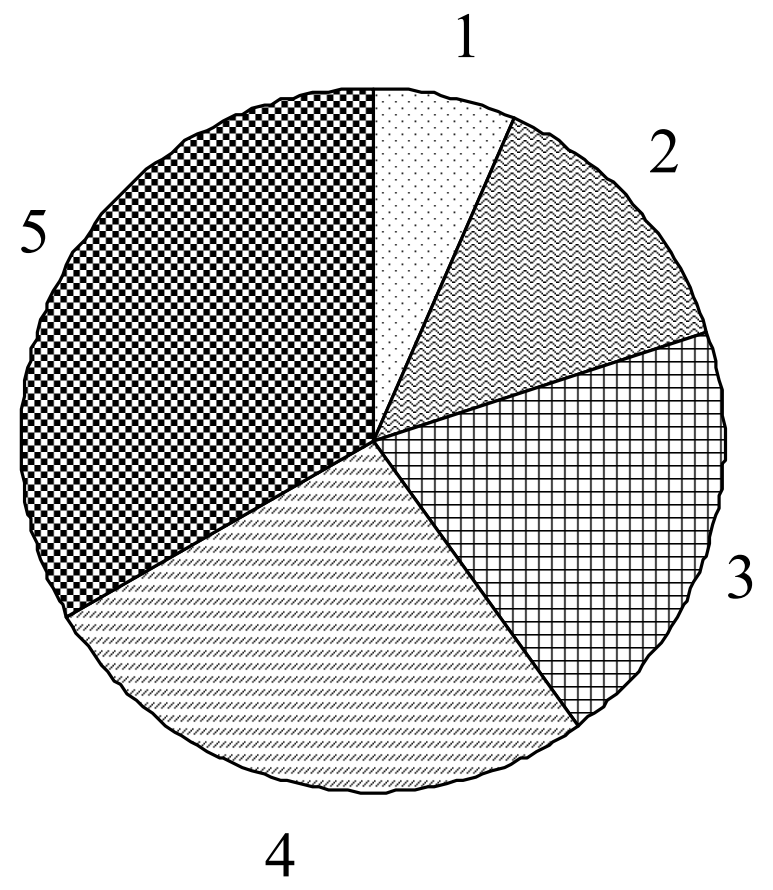

Figure I: Roulette wheel for five individuals.

As the best individual is assigned the highest rank $N$ (for a population consisting of $N$ individuals), this leads to the following number of children per individual. Note that these figures are on average and are based on two parents being required for one child:

$$2 \cdot total\ number\ of\ children\ \ to\ be\ created \cdot \frac{rank\ \ of\ \ individual}{sum\ \ of\ \ ranks\ \ of\ \ all\ \ individuals}$$

$$= 2 \cdot N \cdot \frac{rank(i)}{(N+1)N/2} = \frac{4 \cdot rank(i)}{N+1}$$

I.e. for the best individual (rank = N): $\frac{4N}{N+1} \approx 4\ \ children.$

I.e. for an average individual (rank = N/2): $\frac{2N}{N+1} \approx 2\ \ children.$

### A.2.5 Crossover

The crossover operator is central to the genetic algorithm as it creates new solutions or children. It usually works as a cut and splice operator, i.e. two (or possibly more) parents are taken and cut into pieces. These pieces are then pasted together to form a new solution. There are a vast number of different crossover operators, differing in precisely how they cut up the parents and which parts are chosen to be spliced back together again.

One can now see the close interaction between crossover and encoding. If they work against each other, for example if the cutting points are in inconvenient or even infeasible positions, then clearly one has made a bad choice for either the encoding or for the crossover operator. Although there is a vast number of different crossover strategies, of which only very few examples will be given here, none can be said to be generally superior. It very much depends on the interaction between encoding and the operator and often a successful combination can only be found experimentally or due to experience. For more information about crossover, also refer to sections 4.3.5 and 6.2.

The simplest crossover operator is the so-called one-point crossover. It uses two parents (say A and B) and cuts them both at the same point, thus creating four parts (say A1, A2, B1 and B2). Two new children are now formed by pasting A1 with B2 and B1 with A2. One can imagine a similar crossover operator with two and more crossover points. Figure II below shows an example with four crossover points.

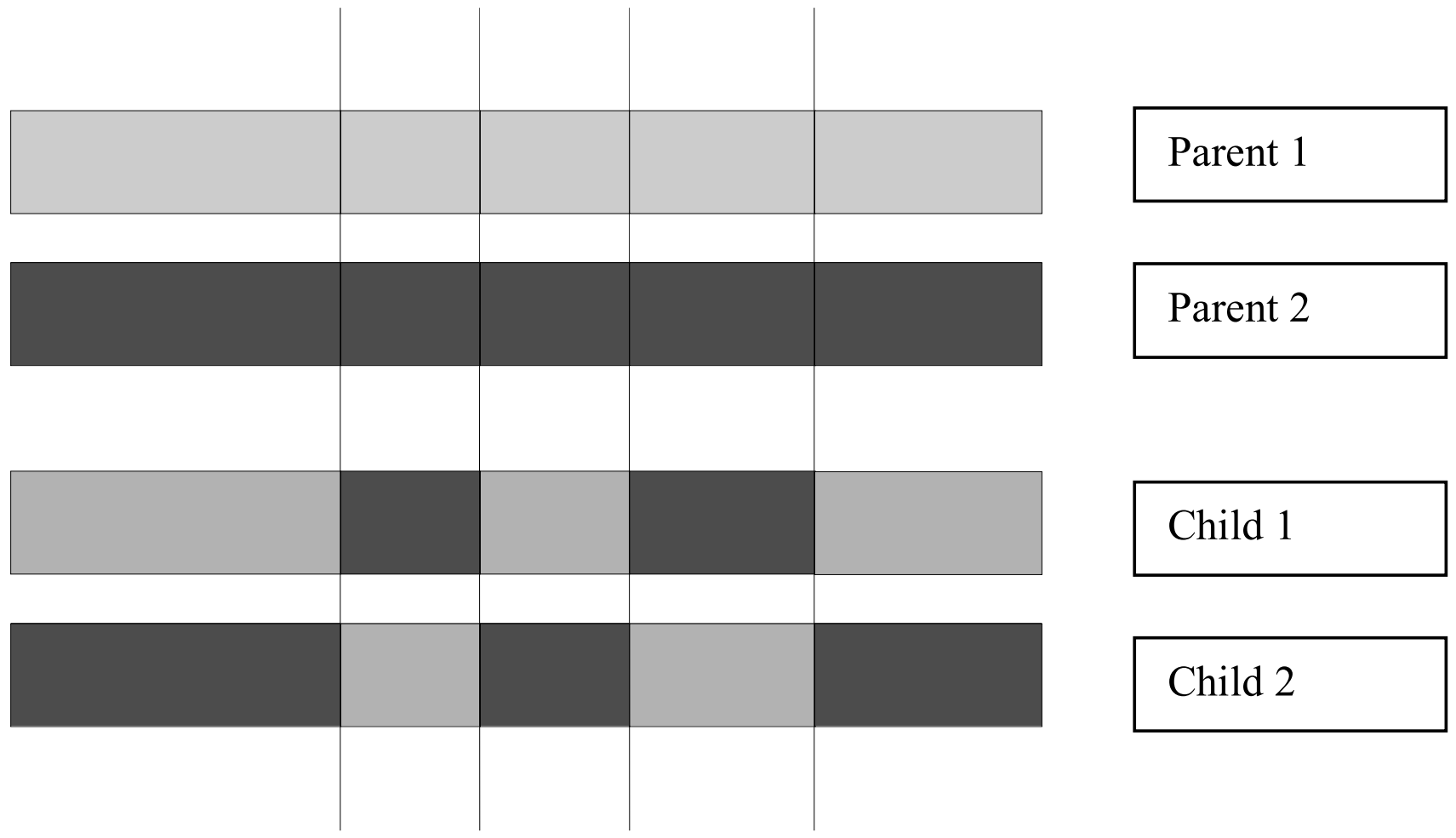

Figure II: Schematic four-point crossover.

Note that if there is more than one crossover point, the children need not necessarily be pasted together by taking pieces alternatively from the parents. This is shown in Figure II purely for illustrative purposes and is in practice uncommon. A relation of *n*-point crossover, with *n* being the number of genes minus one, is uniform crossover. Here both parents are split up into all their genes and these are passed on such that for each of the child's gene there is a 50 percent chance for it to come from either parent. Sometimes other percentages than 50 are used to bias the selection; such a strategy is then called parameterised uniform crossover.

Completely different crossover operators are necessary for strings encoded as permutations. This is because above crossovers would often lead to solutions which featured one value more than once and others not at all. This can be seen from the one-point crossover example in Figure III. The children created are invalid as operations 1 and 2 appear twice in child 1 and operations 3 and 4 are missing completely (vice versa with child 2).

| | | | | | | |
|---|---|---|---|---|---|---|
| 1 | 2 | 3 | 4 | 5 | 6 | Parent 1 |
| 3 | 4 | 2 | 1 | 6 | 5 | Parent 2 |
| | ^ | ^ | | | | Crossover-Point |
| 1 | 2 | 2 | 1 | 6 | 5 | Child 1 |
| 3 | 4 | 3 | 4 | 5 | 6 | Child 2 |

Figure III: Example of one-point crossover going wrong for permutation encoded problems.

To solve the above dilemma, special permutation crossover operators have been invented. Three of these operators are presented in section 6.2, another is C1-crossover. C1-crossover is similar to one-point crossover, in so far as it splits both parents using one crossover point. Furthermore and as before, the first half of the new children is taken identically from the respective parent. However, the other half is then constructed by filling in the missing genes only, in the order they appear in the other parent. For the above example, the result would be as shown in Figure IV.

| | | | | | | |
|---|---|---|---|---|---|---|
| 1 | 2 | 3 | 4 | 5 | 6 | Parent 1 |
| 3 | 4 | 2 | 1 | 6 | 5 | Parent 2 |
| | ^ | ^ | | | | Crossover-Point |
| 1 | 2 | 3 | 4 | 6 | 5 | Child 1 |
| 3 | 4 | 1 | 2 | 5 | 6 | Child 2 |

Figure IV: Example of C1 order based crossover.

## A.2.6 Mutation

The mutation operator is modelled after mutation in nature. It consists of making small random changes to one or a few genes in a chromosome. Translated into genetic algorithm terms, this means that there is a small chance that the value of a particular gene or variable is randomly modified. Originally, with binary encoding, a zero would be changed into a one and vice versa. With alphabets of higher cardinality, there are more options and changes can be made either at random or following a set of rules. The latter is often referred to as intelligent mutation. As with other genetic algorithm parameters, there is no optimal mutation rate for all applications, however an estimated (per gene) mutation rate of [1 / (length of string)] is often a good starting point.

The purpose of mutation is twofold: Due to the limitation of any given population size, some variables might have been initialised such that not all instances of the variable's alphabet are present. This can prove fatal, as the optimal value of a variable might be missing. Hence, it would never be reached by the genetic algorithm. Mutation offers a chance of rectifying this. Similarly, due to an unfortunate choice of parents or operators, certain variable values might be 'lost' throughout the population. Hopefully, these will mainly be those variable instances that lead to low fitness values, but situations are possible where good or even optimal values are lost. Again, mutation can re-introduce those values.

The second purpose of mutation is to gain the capability to make 'small' changes within a genetic algorithm framework. By design, the crossover operator is disruptive and is therefore very unlikely to produce children that are very similar to either parent. However, in the later stages of the search when some convergence towards good solutions has already occurred, small changes might prove more efficient than larger ones. It is here in particular where an intelligent mutation can lead to success rather than random standard mutation.

### A.2.7 Replacement

Once enough children have been created, they replace the parent generation. Many replacement strategies are possible. One of the major distinctions is whether the whole (or a large part) of the population is replaced at once or whether individuals are replaced one by one. The former is called a 'generational' approach, whilst the latter is known as 'steady-state' reproduction.

When a generational approach is used, it is common not to replace the whole population, but to keep the $x\%$ of the best parents. This method is known as elitism and prevents the algorithm from losing the best solutions found so far. Another replacement strategy is tournament replacement, where the best $k$ out of $n$ individuals survive. A variant of this is to bid children versus their own parents so that the victor proceeds into the next generation. Further ideas include rules based on similarities between individuals, to keep a good spread. Many other rules are possible. With a steady-state approach, the most common replacement rules are either tournament replacement or replacing the worst individual with a certain probability (and if that is not replaced then the second worst and so on).

### A.2.8 Stopping Criteria

The topic of stopping criteria is fairly neglected in genetic algorithm literature. Various and mainly problem-specific methods are used with none being reported to be superior for all problems. Examples of criteria are:

- Exceeding a certain number of generations or amount of computation time.
- Not improving the best solution found so far for a certain number of generations or amount of computation time.
- Convergence of the population, which is often based on a large percentage of the population being identical and therefore indicates that the search has terminated naturally.

## A.3 Why Does a Genetic Algorithm Work?

It seems intuitively obvious why genetic algorithms can be used to optimise functions and many real-life examples can be found in support of this, for instance Bäck [8] and others. These examples confirm that genetic algorithms are a robust and generic optimisation method. Nevertheless, not much theory exists to back this up, apart from work based on the original ideas of Holland [96]. Parts of his work, namely the Schema

Theorem, the Building Block Hypothesis and Implicit Parallelism, will be described in the remainder of this section.

Firstly, the term *schema* needs to be explained. A schema can be understood as a string with some of its elements having been replaced by a '#' or 'don't care' sign. For instance, in the case of a binary encoding, the # would stand for either 0 or 1. For a binary encoding of length five, some possible schemata therefore are (100#1), (##1#0) and (####1). In general, a string with $L$ positions and an alphabet of cardinality $N$ contains $(N^L-1)$ schemata and the whole search space is $[(N+1)^L-1]$. $(N+1)$ is used as # extends the cardinality of the employed alphabet by one, however one is deducted from the total as the string consisting of # only is not counted. Holland concluded that genetic algorithms manipulate schemata when they solve a problem, which is described in Holland's Schema Theorem.

How can schemata be used to explain the inner workings of a genetic algorithm? Another way of looking at a schema is to view it as being one hyperplane of the solution space. In other words, each individual is a sample point of those hyperplanes it crosses. By evaluating a population of solutions, the number of hyperplanes sampled and evaluated is far higher than the number of individuals evaluated. This effect is known as implicit parallelism and explains how a genetic algorithm can sample large areas of the solution space quickly.

Combining the hyperplane sampling idea with the fact that individuals reproduce proportionally more often with higher fitness leads to the building block hypothesis. Because fitter individuals are reproduced more often, so are their fitter schemata. However, mutation and crossover tend to disrupt individuals or more precisely long schemata. Hence, short and fit schemata will tend to occur more and more frequently in a population. Schemata of this type are known as building blocks, following the idea that the genetic algorithm can build a good solution by suitably combining them.

Another analogy sometimes used to describe the hyperplane sampling of genetic algorithms is that of a *k*-armed bandit. To understand this, imagine the following: In a supermarket, there are *k* checkout tills. At first people will queue evenly at them, but as we all know, some move faster than others do. This will prompt some people from slower queues to join faster moving ones. In the first instance, probably too many will change, making the slower queues a better option. Hence, some people will change back and so on, until the checkout time at all queues is roughly the same.

Now imagine some people playing a *k*-armed bandit: A person can only pull one arm and each arm has a certain pay-off if it is pulled. If more than one person pulls a specific arm, the pay-off is shared. This leads to a situation that is analogous to the supermarket, those arms with a higher pay-off will be pulled by proportionally more people. This is exactly what the genetic algorithm does when sampling a binary string. It will pull those arms (that is set the value of a bit to one) more often that have a higher pay-off. Of course, initially, some bits will be set to one too often and others not often enough, just like the queuing in the supermarket. However, after some time, an equilibrium will be reached with each bit being set to one proportional to its pay-off or contribution to fitness.

## A.4 Common Genetic Algorithm Enhancements

A canonical genetic algorithm as presented in appendix A.2 is often not suitable to solve a problem to optimality, or as De Jong [51] put it: 'Genetic Algorithms are NOT function optimisers' because 'they lack the killer instinct'. However, as the literature shows, modified genetic algorithms can successfully be used as function optimisers. This section will give a very brief overview of the most common modifications.

Often cited as the foremost drawback of the genetic algorithm is the lack of constraint support. Possible solutions for this are the inclusion of the constraints into the encoding, special crossover and mutation operators, penalty functions and repair operators. As these methods are discussed in detail throughout chapter 3, no further details are given here.

Many researchers have found that larger population sizes give better results. Two reasons for this are that simply more solutions are sampled and that it is less likely for the algorithm to lose valuable information due to the higher number of individuals and schemata at any time. However, a larger population means more computation time per generation and usually more overall execution time of the optimisation. Using more than one computer in parallel would speed things up and because of their population based nature, genetic algorithms are perfectly suited for parallel computing. Hence, the idea of parallel sub-populations was born.

Sub-populations can also be used for a very different reason on a single computer. Rather than just being a part of one big population and following the same rules, each sub-population could follow different strategies and therefore offer more variety. This is known as co-evolution. Many researchers have found that this helps combat one of the genetic algorithm's main drawbacks, the problem of premature convergence. We also make use of sub-populations and the reader is referred to section 5.2 for more details.

Similarly, the concepts of niching and crowding try to fight the same problem. Here, individuals are constrained such that a certain area of the solution space can only 'support' a fixed number of individuals. Hence, if there are too many similar individuals some are 'crowded out'. The result of this is the formation of various niches around local optima. This has the additional benefit of providing more than one good solution from a single run.

Another problem of genetic algorithms is due to their lack of being able to make small changes to an individual. However, this is often required in the later stages of the search, when solutions are close to an optimum. Crossover is too disruptive to achieve this, whilst mutation is too random to consistently make the right changes at the right time. A possible solution to this is the hybridisation with a local hillclimber. However, such a hillclimber will almost always need problem-specific knowledge to succeed. This leads to the loss of the general applicability of this then specifically tailored genetic algorithm. An example of a hillclimber can be found in section 5.4.

# Appendix B Summary of Diplom Thesis

This appendix will provide a short summary of the German Diplom Thesis [4]. This thesis was written at the University of Mannheim (Germany) under the supervision of Professor Schneeweis from September to December 1996. Its objective was to carry out a pilot study into solving the nurse scheduling problem with a genetic algorithm. Only six different data sets relating to one ward in six consecutive weeks were available and nurses were restricted to work either days or nights in week, i.e. there were only 218 rather than 411 possible shift patterns. The objectives were to show the general applicability of a genetic algorithm to the problem, the difficulties encountered and to gauge the possible research potential for further work into this area. A summary of the thesis follows.

The first chapter introduced the problem of nurse scheduling and presented the actual problem at a British hospital. A general introduction into genetic algorithms and their main operators followed in the second chapter. These were then put into concrete terms in chapter three with the actual nurse scheduling problem at hand. Chapter four presented the optimal solutions to the six data sets available as found by CPLEX and compared these to the results of a standard genetic algorithm with intuitively chosen parameters and strategies. In the remainder of chapter four some limited parameter and strategy tests were conducted. The results were slightly improved, however they were still far inferior to the CPLEX solutions. The optimal parameters found are those used in section 4.3 as starting values for further parameter tests.

In chapter five, six possible enhancements to the basic genetic algorithm were presented and their potential examined. The six enhancements were a variety-based replacement, ‘intelligent’ mutation, local optimisation, two oscillating target functions, dynamic penalty weights and grade based crossover.

The variety-based replacement policy tried to promote the survival of less similar individuals. However, the rules used were too strong and only very different individuals were allowed to survive resulting in worse solutions than before.

The idea of 'intelligent' mutation was based on mutating worse individuals more often than better ones. Additionally, the mutation rate was 'adapting' with the search, i.e. the better the solutions found, the smaller the mutation rate. Small improvements were made with this approach. However, this was largely due to 'lucky strikes' because of the generally much higher mutation probability.

Local optimisation tried to improve the solution value by swapping the shift patterns worked between two nurses. Thus, it would not improve feasibility by its definition. It succeeded in improving solution quality slightly, however feasibility was still very low.

Another approach to improve feasibility was to have two 'oscillating' fitness functions. One would try to improve cover, i.e. feasibility, whilst the other one was focused on the nurses' requests, i.e. cost. Only one would be used at a time, with a swap if no further improvements had been made for a number of generations. If no improvements were made for either then the search would terminate. Various ratios of one function to the other were tried, but all failed to improve results. The reasons for this were not fully clear. Possibly, the difference between the two functions was too big such that good solutions to one were unfit for the other and vice versa.

Adaptive penalty weights were defined as behaving proportionally to the generation index. Both increasing and decreasing weights and various 'cooling' and 'increasing' schedules were tried. Some of these did improve both solution cost and feasibility.

However, overall feasibility was still very low. The final idea of grade-based crossover was specifically addressing this issue. Rather than using uniform crossover all the time and hence splitting up some (partially) good schedules, a new fixed point crossover was introduced. This crossover would take place in roughly 20% of all crossovers and would divide the parents along the grade boundaries and recombine accordingly.

Beforehand, solutions were sorted by the nurses' grades from high to low. Feasibility was almost doubled to about 60% following this approach.

The final chapter of the thesis presented possible future enhancements. These included an expansion of the actual model (inclusion of all 411 shift patterns and others) and the introduction of a new fitness score based on the grade-segments of the string. Moreover, this could possibly be further expanded by the use of sub-populations based on each segment. The latter idea was further developed during the present research and is presented in section 5.2. The thesis concluded that there was a good research potential for a PhD.

# Appendix C  Nurse Scheduling Data

## C.1 Demand

This appendix presents some of the data used for the nurse scheduling problem for illustration purposes.  In this first section Table I shows the typical shifts to be covered for one ward.  The shifts are given grouped by day / night and grade of nurses required, where

- Su - Sa are the seven days respectively nights of a week.
- Q1, Q2 and Q3 are the number of nurses needed per grade.  Higher qualified ones can substitute nurses with a lower grade.

| | Day Shifts | | | | | | | Night Shifts | | | | | | |
|---|---|---|---|---|---|---|---|---|---|---|---|---|---|---|
| | Su | Mo | Tu | We | Th | Fr | Sa | Su | Mo | Tu | We | Th | Fr | Sa |
| Q1 | 2 | 2 | 2 | 2 | 2 | 2 | 2 | 1 | 1 | 1 | 1 | 1 | 1 | 1 |
| Q2 | 2 | 2 | 2 | 2 | 2 | 2 | 2 | 1 | 1 | 1 | 1 | 1 | 1 | 1 |
| Q3 | 5 | 5 | 5 | 5 | 5 | 5 | 5 | 1 | 1 | 1 | 1 | 1 | 1 | 1 |

Table I: Example of a week's demand for nurses.

## C.2 Qualifications, Working Hours and General Preferences

Table II shows the typical qualifications, working hours and general day / night preferences of the nurses on one ward.  In the table,

- i is the nurse index.
- q(i) is the grade of nurse i.
- w(i) is the weekly working hours of nurse i, where 1 = 100%, 2 = 80%, 3 = 60%, 4 = 50%, 5 = 40%, 6 = 20% and 7 = 0% of a full time nurse.

- p(i) is the general day respectively night preference with 1 days preferred, 2 days important, 3 nights preferred, 4 nights important, 5 days only, 6 nights only and 0 indifference between days and nights.

| i | 1 | 2 | 3 | 4 | 5 | 6 | 7 | 8 | 9 | 10 | 11 | 12 | 13 | 14 | 15 | 16 | 17 | 18 | 19 | 20 | 21 |
|---|---|---|---|---|---|---|---|---|---|---|---|---|---|---|---|---|---|---|---|---|---|
| Q(i) | 1 | 1 | 1 | 1 | 1 | 1 | 1 | 1 | 1 | 2 | 2 | 2 | 2 | 2 | 3 | 3 | 3 | 3 | 3 | 3 | 3 |
| W(i) | 7 | 1 | 2 | 1 | 2 | 3 | 2 | 3 | 2 | 1 | 1 | 1 | 1 | 7 | 1 | 1 | 1 | 1 | 1 | 4 | 1 |
| P(i) | 1 | 1 | 0 | 0 | 0 | 2 | 3 | 0 | 0 | 0 | 2 | 0 | 0 | 0 | 0 | 0 | 0 | 0 | 0 | 0 | 1 |

Table II: Example of nurses' general preferences and qualifications.

## C.3 Preferences

The last information needed to calculate the $p_{ij}$ values, following the rules outlined in chapter 2.1.3, are the weekly requests of the nurses. Examples of these are given in Table III, where

- i is the nurse index.
- Su - Sa are the seven days respectively nights of a week. The higher the number the stronger the preference not to work that shift. For a detailed description of the scale used, see chapter 2.1.3.

| | Day Shifts | | | | | | | Night Shifts | | | | | | |
|---|---|---|---|---|---|---|---|---|---|---|---|---|---|---|
| i | Su | Mo | Tu | We | Th | Fr | Sa | Su | Mo | Tu | We | Th | Fr | Sa |
| 1 | 0 | 0 | 0 | 0 | 0 | 2 | 1 | 0 | 0 | 0 | 0 | 0 | 0 | 0 |
| 2 | 0 | 1 | 0 | 0 | 0 | 0 | 0 | 0 | 0 | 0 | 0 | 0 | 0 | 0 |
| 3 | 0 | 0 | 0 | 0 | 0 | 0 | 0 | 0 | 0 | 0 | 0 | 0 | 0 | 0 |
| 4 | 0 | 0 | 0 | 1 | 0 | 0 | 0 | 2 | 0 | 0 | 0 | 0 | 0 | 0 |
| 5 | 0 | 0 | 0 | 0 | 0 | 0 | 0 | 0 | 0 | 0 | 0 | 0 | 0 | 2 |
| 6 | 0 | 0 | 0 | 0 | 0 | 0 | 0 | 0 | 0 | 1 | 0 | 0 | 0 | 0 |
| 7 | 3 | 2 | 0 | 0 | 0 | 0 | 2 | 0 | 0 | 0 | 0 | 0 | 0 | 0 |
| 8 | 0 | 0 | 0 | 0 | 0 | 0 | 1 | 0 | 0 | 0 | 0 | 0 | 0 | 0 |
| 9 | 0 | 0 | 0 | 1 | 0 | 0 | 0 | 0 | 0 | 0 | 0 | 1 | 0 | 0 |
| 10 | 0 | 0 | 0 | 0 | 0 | 0 | 0 | 0 | 0 | 0 | 0 | 0 | 0 | 0 |
| 11 | 2 | 0 | 1 | 0 | 0 | 2 | 3 | 2 | 0 | 0 | 0 | 0 | 0 | 0 |
| 12 | 0 | 0 | 0 | 0 | 0 | 0 | 0 | 0 | 0 | 0 | 0 | 0 | 0 | 0 |
| 13 | 0 | 0 | 0 | 1 | 0 | 0 | 0 | 0 | 0 | 0 | 0 | 1 | 0 | 0 |
| 14 | 0 | 0 | 0 | 0 | 2 | 0 | 0 | 0 | 0 | 0 | 0 | 0 | 0 | 0 |
| 15 | 2 | 2 | 0 | 0 | 0 | 0 | 0 | 0 | 0 | 1 | 0 | 0 | 0 | 0 |
| 16 | 0 | 0 | 0 | 0 | 0 | 0 | 0 | 0 | 0 | 0 | 0 | 0 | 0 | 0 |
| 17 | 0 | 0 | 0 | 0 | 0 | 0 | 0 | 4 | 0 | 0 | 0 | 0 | 0 | 0 |
| 18 | 0 | 0 | 1 | 0 | 0 | 3 | 3 | 3 | 0 | 0 | 0 | 0 | 0 | 0 |
| 19 | 2 | 0 | 0 | 0 | 0 | 0 | 0 | 0 | 0 | 0 | 0 | 0 | 0 | 0 |
| 20 | 0 | 0 | 0 | 0 | 0 | 0 | 1 | 0 | 0 | 0 | 0 | 0 | 0 | 0 |
| 21 | 0 | 0 | 0 | 0 | 0 | 0 | 0 | 0 | 0 | 0 | 0 | 0 | 0 | 0 |

Table III: Example of nurses' weekly preferences.

## C.4 Table of Shift Patterns

Depending on the working hours and requested days off, a nurse can only work a certain number of shift patterns. Table IV shows all possible shift patterns. Altogether, there are 411 possible shift patterns. In the table,

- *i* is the shift pattern index.
- *S* - *S* is 1, if shift pattern *i* covers that shift on this particular day, else it is 0.
- *v(i)* is the general value of shift pattern *i*. The lower the value the 'nicer' the shift pattern. For more details about this see chapter 2.1.3.
- *e(i)* is the first, *l(i)* the last free day of shift pattern *i*. These are required for maximum work-stretch constraints.

| | Day Shifts | | | | | | | Night Shifts | | | | | | | | | |
|---|---|---|---|---|---|---|---|---|---|---|---|---|---|---|---|---|---|
| i | S | M | T | W | T | F | S | S | M | T | W | T | F | S | v | e | l |
| 1 | 1 | 1 | 1 | 1 | 1 | 0 | 0 | 0 | 0 | 0 | 0 | 0 | 0 | 0 | 1 | 0 | 5 |
| 2 | 1 | 1 | 1 | 1 | 0 | 1 | 0 | 0 | 0 | 0 | 0 | 0 | 0 | 0 | 3 | 0 | 4 |
| 3 | 1 | 1 | 1 | 1 | 0 | 0 | 1 | 0 | 0 | 0 | 0 | 0 | 0 | 0 | 1 | 1 | 4 |
| 4 | 1 | 1 | 1 | 0 | 1 | 1 | 0 | 0 | 0 | 0 | 0 | 0 | 0 | 0 | 2 | 0 | 3 |
| 5 | 1 | 1 | 1 | 0 | 1 | 0 | 1 | 0 | 0 | 0 | 0 | 0 | 0 | 0 | 3 | 1 | 3 |
| 6 | 1 | 1 | 1 | 0 | 0 | 1 | 1 | 0 | 0 | 0 | 0 | 0 | 0 | 0 | 1 | 2 | 3 |
| 7 | 1 | 1 | 0 | 1 | 1 | 1 | 0 | 0 | 0 | 0 | 0 | 0 | 0 | 0 | 2 | 0 | 2 |
| 8 | 1 | 1 | 0 | 1 | 1 | 0 | 1 | 0 | 0 | 0 | 0 | 0 | 0 | 0 | 2 | 1 | 2 |
| 9 | 1 | 1 | 0 | 1 | 0 | 1 | 1 | 0 | 0 | 0 | 0 | 0 | 0 | 0 | 3 | 2 | 2 |
| 10 | 1 | 1 | 0 | 0 | 1 | 1 | 1 | 0 | 0 | 0 | 0 | 0 | 0 | 0 | 1 | 3 | 2 |
| 11 | 1 | 0 | 1 | 1 | 1 | 1 | 0 | 0 | 0 | 0 | 0 | 0 | 0 | 0 | 2 | 0 | 1 |
| 12 | 1 | 0 | 1 | 1 | 1 | 0 | 1 | 0 | 0 | 0 | 0 | 0 | 0 | 0 | 2 | 1 | 1 |
| 13 | 1 | 0 | 1 | 1 | 0 | 1 | 1 | 0 | 0 | 0 | 0 | 0 | 0 | 0 | 2 | 2 | 1 |
| 14 | 1 | 0 | 1 | 0 | 1 | 1 | 1 | 0 | 0 | 0 | 0 | 0 | 0 | 0 | 3 | 3 | 1 |
| 15 | 1 | 0 | 0 | 1 | 1 | 1 | 1 | 0 | 0 | 0 | 0 | 0 | 0 | 0 | 1 | 4 | 1 |
| 16 | 0 | 1 | 1 | 1 | 1 | 1 | 0 | 0 | 0 | 0 | 0 | 0 | 0 | 0 | 1 | 0 | 0 |
| 17 | 0 | 1 | 1 | 1 | 1 | 0 | 1 | 0 | 0 | 0 | 0 | 0 | 0 | 0 | 2 | 1 | 0 |
| 18 | 0 | 1 | 1 | 1 | 0 | 1 | 1 | 0 | 0 | 0 | 0 | 0 | 0 | 0 | 2 | 2 | 0 |
| 19 | 0 | 1 | 1 | 0 | 1 | 1 | 1 | 0 | 0 | 0 | 0 | 0 | 0 | 0 | 2 | 3 | 0 |
| 20 | 0 | 1 | 0 | 1 | 1 | 1 | 1 | 0 | 0 | 0 | 0 | 0 | 0 | 0 | 3 | 4 | 0 |
| 21 | 0 | 0 | 1 | 1 | 1 | 1 | 1 | 0 | 0 | 0 | 0 | 0 | 0 | 0 | 1 | 5 | 0 |
| 22 | 0 | 0 | 0 | 0 | 0 | 0 | 0 | 1 | 1 | 1 | 1 | 0 | 0 | 0 | 1 | 0 | 4 |
| 23 | 0 | 0 | 0 | 0 | 0 | 0 | 0 | 1 | 1 | 1 | 0 | 1 | 0 | 0 | 3 | 0 | 3 |
| 24 | 0 | 0 | 0 | 0 | 0 | 0 | 0 | 1 | 1 | 1 | 0 | 0 | 1 | 0 | 3 | 0 | 3 |
| 25 | 0 | 0 | 0 | 0 | 0 | 0 | 0 | 1 | 1 | 1 | 0 | 0 | 0 | 1 | 3 | 1 | 3 |
| 26 | 0 | 0 | 0 | 0 | 0 | 0 | 0 | 1 | 1 | 0 | 1 | 1 | 0 | 0 | 2 | 0 | 2 |
| 27 | 0 | 0 | 0 | 0 | 0 | 0 | 0 | 1 | 1 | 0 | 1 | 0 | 1 | 0 | 3 | 0 | 2 |
| 28 | 0 | 0 | 0 | 0 | 0 | 0 | 0 | 1 | 1 | 0 | 1 | 0 | 0 | 1 | 3 | 1 | 2 |
| 29 | 0 | 0 | 0 | 0 | 0 | 0 | 0 | 1 | 1 | 0 | 0 | 1 | 1 | 0 | 2 | 0 | 2 |
| 30 | 0 | 0 | 0 | 0 | 0 | 0 | 0 | 1 | 1 | 0 | 0 | 1 | 0 | 1 | 3 | 1 | 2 |
| 31 | 0 | 0 | 0 | 0 | 0 | 0 | 0 | 1 | 1 | 0 | 0 | 0 | 1 | 1 | 2 | 2 | 2 |
| 32 | 0 | 0 | 0 | 0 | 0 | 0 | 0 | 1 | 0 | 1 | 1 | 1 | 0 | 0 | 3 | 0 | 1 |
| 33 | 0 | 0 | 0 | 0 | 0 | 0 | 0 | 1 | 0 | 1 | 1 | 0 | 1 | 0 | 3 | 0 | 1 |
| 34 | 0 | 0 | 0 | 0 | 0 | 0 | 0 | 1 | 0 | 1 | 1 | 0 | 0 | 1 | 3 | 1 | 1 |
| 35 | 0 | 0 | 0 | 0 | 0 | 0 | 0 | 1 | 0 | 1 | 0 | 1 | 1 | 0 | 3 | 0 | 1 |
| 36 | 0 | 0 | 0 | 0 | 0 | 0 | 0 | 1 | 0 | 1 | 0 | 1 | 0 | 1 | 4 | 1 | 1 |
| 37 | 0 | 0 | 0 | 0 | 0 | 0 | 0 | 1 | 0 | 1 | 0 | 0 | 1 | 1 | 3 | 2 | 1 |
| 38 | 0 | 0 | 0 | 0 | 0 | 0 | 0 | 1 | 0 | 0 | 1 | 1 | 1 | 0 | 2 | 0 | 1 |
| 39 | 0 | 0 | 0 | 0 | 0 | 0 | 0 | 1 | 0 | 0 | 1 | 1 | 0 | 1 | 3 | 1 | 1 |
| 40 | 0 | 0 | 0 | 0 | 0 | 0 | 0 | 1 | 0 | 0 | 1 | 0 | 1 | 1 | 3 | 2 | 1 |
| 41 | 0 | 0 | 0 | 0 | 0 | 0 | 0 | 1 | 0 | 0 | 0 | 1 | 1 | 1 | 2 | 3 | 1 |
| 42 | 0 | 0 | 0 | 0 | 0 | 0 | 0 | 0 | 1 | 1 | 1 | 1 | 0 | 0 | 1 | 0 | 0 |
| 43 | 0 | 0 | 0 | 0 | 0 | 0 | 0 | 0 | 1 | 1 | 1 | 0 | 1 | 0 | 3 | 0 | 0 |
| 44 | 0 | 0 | 0 | 0 | 0 | 0 | 0 | 0 | 1 | 1 | 1 | 0 | 0 | 1 | 3 | 1 | 0 |
| 45 | 0 | 0 | 0 | 0 | 0 | 0 | 0 | 0 | 1 | 1 | 0 | 1 | 1 | 0 | 2 | 0 | 0 |
| 46 | 0 | 0 | 0 | 0 | 0 | 0 | 0 | 0 | 1 | 1 | 0 | 1 | 0 | 1 | 3 | 1 | 0 |
| 47 | 0 | 0 | 0 | 0 | 0 | 0 | 0 | 0 | 1 | 1 | 0 | 0 | 1 | 1 | 2 | 2 | 0 |
| 48 | 0 | 0 | 0 | 0 | 0 | 0 | 0 | 0 | 1 | 0 | 1 | 1 | 1 | 0 | 3 | 0 | 0 |
| 49 | 0 | 0 | 0 | 0 | 0 | 0 | 0 | 0 | 1 | 0 | 1 | 1 | 0 | 1 | 3 | 1 | 0 |
| 50 | 0 | 0 | 0 | 0 | 0 | 0 | 0 | 0 | 1 | 0 | 1 | 0 | 1 | 1 | 3 | 2 | 0 |
| 51 | 0 | 0 | 0 | 0 | 0 | 0 | 0 | 0 | 1 | 0 | 0 | 1 | 1 | 1 | 2 | 3 | 0 |
| 52 | 0 | 0 | 0 | 0 | 0 | 0 | 0 | 0 | 0 | 1 | 1 | 1 | 1 | 0 | 1 | 0 | 0 |
| 53 | 0 | 0 | 0 | 0 | 0 | 0 | 0 | 0 | 0 | 1 | 1 | 1 | 0 | 1 | 3 | 1 | 0 |
| 54 | 0 | 0 | 0 | 0 | 0 | 0 | 0 | 0 | 0 | 1 | 1 | 0 | 1 | 1 | 2 | 2 | 0 |
| 55 | 0 | 0 | 0 | 0 | 0 | 0 | 0 | 0 | 0 | 1 | 0 | 1 | 1 | 1 | 2 | 3 | 0 |
| 56 | 0 | 0 | 0 | 0 | 0 | 0 | 0 | 0 | 0 | 0 | 1 | 1 | 1 | 1 | 1 | 4 | 0 |
| 57 | 1 | 1 | 1 | 1 | 0 | 0 | 0 | 0 | 0 | 0 | 0 | 0 | 0 | 0 | 1 | 0 | 4 |
| 58 | 1 | 1 | 1 | 0 | 1 | 0 | 0 | 0 | 0 | 0 | 0 | 0 | 0 | 0 | 3 | 0 | 3 |
| 59 | 1 | 1 | 1 | 0 | 0 | 1 | 0 | 0 | 0 | 0 | 0 | 0 | 0 | 0 | 3 | 0 | 3 |
| 60 | 1 | 1 | 1 | 0 | 0 | 0 | 1 | 0 | 0 | 0 | 0 | 0 | 0 | 0 | 2 | 1 | 3 |
| 61 | 1 | 1 | 0 | 1 | 1 | 0 | 0 | 0 | 0 | 0 | 0 | 0 | 0 | 0 | 2 | 0 | 2 |
| 62 | 1 | 1 | 0 | 1 | 0 | 1 | 0 | 0 | 0 | 0 | 0 | 0 | 0 | 0 | 3 | 0 | 2 |
| 63 | 1 | 1 | 0 | 1 | 0 | 0 | 1 | 0 | 0 | 0 | 0 | 0 | 0 | 0 | 3 | 1 | 2 |
| 64 | 1 | 1 | 0 | 0 | 1 | 1 | 0 | 0 | 0 | 0 | 0 | 0 | 0 | 0 | 2 | 0 | 2 |
| 65 | 1 | 1 | 0 | 0 | 1 | 0 | 1 | 0 | 0 | 0 | 0 | 0 | 0 | 0 | 3 | 1 | 2 |
| 66 | 1 | 1 | 0 | 0 | 0 | 1 | 1 | 0 | 0 | 0 | 0 | 0 | 0 | 0 | 2 | 1 | 2 |
| 67 | 1 | 0 | 1 | 1 | 1 | 0 | 0 | 0 | 0 | 0 | 0 | 0 | 0 | 0 | 3 | 0 | 1 |
| 68 | 1 | 0 | 1 | 1 | 0 | 1 | 0 | 0 | 0 | 0 | 0 | 0 | 0 | 0 | 3 | 0 | 1 |
| 69 | 1 | 0 | 1 | 1 | 0 | 0 | 1 | 0 | 0 | 0 | 0 | 0 | 0 | 0 | 2 | 1 | 1 |
| 70 | 1 | 0 | 1 | 0 | 1 | 1 | 0 | 0 | 0 | 0 | 0 | 0 | 0 | 0 | 3 | 0 | 1 |
| 71 | 1 | 0 | 1 | 0 | 1 | 0 | 1 | 0 | 0 | 0 | 0 | 0 | 0 | 0 | 4 | 1 | 1 |
| 72 | 1 | 0 | 1 | 0 | 0 | 1 | 1 | 0 | 0 | 0 | 0 | 0 | 0 | 0 | 3 | 2 | 1 |
| 73 | 1 | 0 | 0 | 1 | 1 | 1 | 0 | 0 | 0 | 0 | 0 | 0 | 0 | 0 | 2 | 0 | 1 |
| 74 | 1 | 0 | 0 | 1 | 1 | 0 | 1 | 0 | 0 | 0 | 0 | 0 | 0 | 0 | 2 | 1 | 1 |
| 75 | 1 | 0 | 0 | 1 | 0 | 1 | 1 | 0 | 0 | 0 | 0 | 0 | 0 | 0 | 3 | 2 | 1 |
| 76 | 1 | 0 | 0 | 0 | 1 | 1 | 1 | 0 | 0 | 0 | 0 | 0 | 0 | 0 | 2 | 3 | 1 |
| 77 | 0 | 1 | 1 | 1 | 1 | 0 | 0 | 0 | 0 | 0 | 0 | 0 | 0 | 0 | 1 | 0 | 0 |
| 78 | 0 | 1 | 1 | 1 | 0 | 1 | 0 | 0 | 0 | 0 | 0 | 0 | 0 | 0 | 3 | 0 | 0 |
| 79 | 0 | 1 | 1 | 1 | 0 | 0 | 1 | 0 | 0 | 0 | 0 | 0 | 0 | 0 | 2 | 1 | 0 |
| 80 | 0 | 1 | 1 | 0 | 1 | 1 | 0 | 0 | 0 | 0 | 0 | 0 | 0 | 0 | 2 | 0 | 0 |
| 81 | 0 | 1 | 1 | 0 | 1 | 0 | 1 | 0 | 0 | 0 | 0 | 0 | 0 | 0 | 3 | 1 | 0 |
| 82 | 0 | 1 | 1 | 0 | 0 | 1 | 1 | 0 | 0 | 0 | 0 | 0 | 0 | 0 | 2 | 2 | 0 |
| 83 | 0 | 1 | 0 | 1 | 1 | 1 | 0 | 0 | 0 | 0 | 0 | 0 | 0 | 0 | 3 | 0 | 0 |
| 84 | 0 | 1 | 0 | 1 | 1 | 0 | 1 | 0 | 0 | 0 | 0 | 0 | 0 | 0 | 3 | 1 | 0 |
| 85 | 0 | 1 | 0 | 1 | 0 | 1 | 1 | 0 | 0 | 0 | 0 | 0 | 0 | 0 | 3 | 2 | 0 |
| 86 | 0 | 1 | 0 | 0 | 1 | 1 | 1 | 0 | 0 | 0 | 0 | 0 | 0 | 0 | 3 | 3 | 0 |
| 87 | 0 | 0 | 1 | 1 | 1 | 1 | 0 | 0 | 0 | 0 | 0 | 0 | 0 | 0 | 1 | 0 | 0 |
| 88 | 0 | 0 | 1 | 1 | 1 | 0 | 1 | 0 | 0 | 0 | 0 | 0 | 0 | 0 | 2 | 1 | 0 |
| 89 | 0 | 0 | 1 | 1 | 0 | 1 | 1 | 0 | 0 | 0 | 0 | 0 | 0 | 0 | 2 | 2 | 0 |
| 90 | 0 | 0 | 1 | 0 | 1 | 1 | 1 | 0 | 0 | 0 | 0 | 0 | 0 | 0 | 2 | 3 | 0 |
| 91 | 0 | 0 | 0 | 1 | 1 | 1 | 1 | 0 | 0 | 0 | 0 | 0 | 0 | 0 | 1 | 4 | 0 |
| 92 | 0 | 0 | 0 | 0 | 0 | 0 | 0 | 0 | 0 | 0 | 0 | 1 | 1 | 1 | 1 | 3 | 0 |
| 93 | 0 | 0 | 0 | 0 | 0 | 0 | 0 | 0 | 0 | 0 | 1 | 0 | 1 | 1 | 2 | 2 | 0 |
| 94 | 0 | 0 | 0 | 0 | 0 | 0 | 0 | 0 | 0 | 0 | 1 | 1 | 0 | 1 | 2 | 1 | 0 |
| 95 | 0 | 0 | 0 | 0 | 0 | 0 | 0 | 0 | 0 | 0 | 1 | 1 | 1 | 0 | 1 | 0 | 0 |
| 96 | 0 | 0 | 0 | 0 | 0 | 0 | 0 | 0 | 0 | 1 | 0 | 0 | 1 | 1 | 3 | 2 | 0 |
| 97 | 0 | 0 | 0 | 0 | 0 | 0 | 0 | 0 | 0 | 1 | 0 | 1 | 0 | 1 | 3 | 1 | 0 |
| 98 | 0 | 0 | 0 | 0 | 0 | 0 | 0 | 0 | 0 | 1 | 0 | 1 | 1 | 0 | 2 | 0 | 0 |
| 99 | 0 | 0 | 0 | 0 | 0 | 0 | 0 | 0 | 0 | 1 | 1 | 0 | 0 | 1 | 3 | 1 | 0 |
| 100 | 0 | 0 | 0 | 0 | 0 | 0 | 0 | 0 | 0 | 1 | 1 | 0 | 1 | 0 | 2 | 0 | 0 |
| 101 | 0 | 0 | 0 | 0 | 0 | 0 | 0 | 0 | 0 | 1 | 1 | 1 | 0 | 0 | 1 | 0 | 0 |
| 102 | 0 | 0 | 0 | 0 | 0 | 0 | 0 | 0 | 1 | 0 | 0 | 0 | 1 | 1 | 3 | 2 | 0 |
| 103 | 0 | 0 | 0 | 0 | 0 | 0 | 0 | 0 | 1 | 0 | 0 | 1 | 0 | 1 | 3 | 1 | 0 |
| 104 | 0 | 0 | 0 | 0 | 0 | 0 | 0 | 0 | 1 | 0 | 0 | 1 | 1 | 0 | 3 | 0 | 0 |
| 105 | 0 | 0 | 0 | 0 | 0 | 0 | 0 | 0 | 1 | 0 | 1 | 0 | 0 | 1 | 3 | 1 | 0 |
| 106 | 0 | 0 | 0 | 0 | 0 | 0 | 0 | 0 | 1 | 0 | 1 | 0 | 1 | 0 | 3 | 0 | 0 |
| 107 | 0 | 0 | 0 | 0 | 0 | 0 | 0 | 0 | 1 | 0 | 1 | 1 | 0 | 0 | 2 | 0 | 0 |
| 108 | 0 | 0 | 0 | 0 | 0 | 0 | 0 | 0 | 1 | 1 | 0 | 0 | 0 | 1 | 3 | 1 | 0 |
| 109 | 0 | 0 | 0 | 0 | 0 | 0 | 0 | 0 | 1 | 1 | 0 | 0 | 1 | 0 | 3 | 0 | 0 |
| 110 | 0 | 0 | 0 | 0 | 0 | 0 | 0 | 0 | 1 | 1 | 0 | 1 | 0 | 0 | 3 | 0 | 0 |
| 111 | 0 | 0 | 0 | 0 | 0 | 0 | 0 | 0 | 1 | 1 | 1 | 0 | 0 | 0 | 1 | 0 | 0 |
| 112 | 0 | 0 | 0 | 0 | 0 | 0 | 0 | 1 | 0 | 0 | 0 | 0 | 1 | 1 | 4 | 2 | 1 |
| 113 | 0 | 0 | 0 | 0 | 0 | 0 | 0 | 1 | 0 | 0 | 0 | 1 | 0 | 1 | 4 | 1 | 1 |
| 114 | 0 | 0 | 0 | 0 | 0 | 0 | 0 | 1 | 0 | 0 | 0 | 1 | 1 | 0 | 3 | 0 | 1 |
| 115 | 0 | 0 | 0 | 0 | 0 | 0 | 0 | 1 | 0 | 0 | 1 | 0 | 0 | 1 | 4 | 1 | 1 |
| 116 | 0 | 0 | 0 | 0 | 0 | 0 | 0 | 1 | 0 | 0 | 1 | 0 | 1 | 0 | 4 | 0 | 1 |
| 117 | 0 | 0 | 0 | 0 | 0 | 0 | 0 | 1 | 0 | 0 | 1 | 1 | 0 | 0 | 3 | 0 | 1 |
| 118 | 0 | 0 | 0 | 0 | 0 | 0 | 0 | 1 | 0 | 1 | 0 | 0 | 0 | 1 | 4 | 1 | 1 |
| 119 | 0 | 0 | 0 | 0 | 0 | 0 | 0 | 1 | 0 | 1 | 0 | 0 | 1 | 0 | 4 | 0 | 1 |
| 120 | 0 | 0 | 0 | 0 | 0 | 0 | 0 | 1 | 0 | 1 | 0 | 1 | 0 | 0 | 4 | 0 | 1 |
| 121 | 0 | 0 | 0 | 0 | 0 | 0 | 0 | 1 | 0 | 1 | 1 | 0 | 0 | 0 | 3 | 0 | 1 |
| 122 | 0 | 0 | 0 | 0 | 0 | 0 | 0 | 1 | 1 | 0 | 0 | 0 | 0 | 1 | 4 | 1 | 2 |
| 123 | 0 | 0 | 0 | 0 | 0 | 0 | 0 | 1 | 1 | 0 | 0 | 0 | 1 | 0 | 4 | 0 | 2 |
| 124 | 0 | 0 | 0 | 0 | 0 | 0 | 0 | 1 | 1 | 0 | 0 | 1 | 0 | 0 | 3 | 0 | 2 |
| 125 | 0 | 0 | 0 | 0 | 0 | 0 | 0 | 1 | 1 | 0 | 1 | 0 | 0 | 0 | 2 | 0 | 2 |
| 126 | 0 | 0 | 0 | 0 | 0 | 0 | 0 | 1 | 1 | 1 | 0 | 0 | 0 | 0 | 1 | 0 | 3 |
| 127 | 0 | 0 | 0 | 0 | 1 | 1 | 1 | 0 | 0 | 0 | 0 | 0 | 0 | 0 | 1 | 3 | 0 |
| 128 | 0 | 0 | 0 | 1 | 0 | 1 | 1 | 0 | 0 | 0 | 0 | 0 | 0 | 0 | 2 | 2 | 0 |
| 129 | 0 | 0 | 0 | 1 | 1 | 0 | 1 | 0 | 0 | 0 | 0 | 0 | 0 | 0 | 2 | 1 | 0 |
| 130 | 0 | 0 | 0 | 1 | 1 | 1 | 0 | 0 | 0 | 0 | 0 | 0 | 0 | 0 | 1 | 0 | 0 |
| 131 | 0 | 0 | 1 | 0 | 0 | 1 | 1 | 0 | 0 | 0 | 0 | 0 | 0 | 0 | 3 | 2 | 0 |
| 132 | 0 | 0 | 1 | 0 | 1 | 0 | 1 | 0 | 0 | 0 | 0 | 0 | 0 | 0 | 3 | 1 | 0 |
| 133 | 0 | 0 | 1 | 0 | 1 | 1 | 0 | 0 | 0 | 0 | 0 | 0 | 0 | 0 | 2 | 0 | 0 |
| 134 | 0 | 0 | 1 | 1 | 0 | 0 | 1 | 0 | 0 | 0 | 0 | 0 | 0 | 0 | 3 | 1 | 0 |
| 135 | 0 | 0 | 1 | 1 | 0 | 1 | 0 | 0 | 0 | 0 | 0 | 0 | 0 | 0 | 2 | 0 | 0 |
| 136 | 0 | 0 | 1 | 1 | 1 | 0 | 0 | 0 | 0 | 0 | 0 | 0 | 0 | 0 | 1 | 0 | 0 |
| 137 | 0 | 1 | 0 | 0 | 0 | 1 | 1 | 0 | 0 | 0 | 0 | 0 | 0 | 0 | 3 | 2 | 0 |
| 138 | 0 | 1 | 0 | 0 | 1 | 0 | 1 | 0 | 0 | 0 | 0 | 0 | 0 | 0 | 4 | 1 | 0 |
| 139 | 0 | 1 | 0 | 0 | 1 | 1 | 0 | 0 | 0 | 0 | 0 | 0 | 0 | 0 | 3 | 0 | 0 |
| 140 | 0 | 1 | 0 | 1 | 0 | 0 | 1 | 0 | 0 | 0 | 0 | 0 | 0 | 0 | 4 | 1 | 0 |
| 141 | 0 | 1 | 0 | 1 | 0 | 1 | 0 | 0 | 0 | 0 | 0 | 0 | 0 | 0 | 4 | 0 | 0 |
| 142 | 0 | 1 | 0 | 1 | 1 | 0 | 0 | 0 | 0 | 0 | 0 | 0 | 0 | 0 | 2 | 0 | 0 |
| 143 | 0 | 1 | 1 | 0 | 0 | 0 | 1 | 0 | 0 | 0 | 0 | 0 | 0 | 0 | 3 | 1 | 0 |
| 144 | 0 | 1 | 1 | 0 | 0 | 1 | 0 | 0 | 0 | 0 | 0 | 0 | 0 | 0 | 3 | 0 | 0 |
| 145 | 0 | 1 | 1 | 0 | 1 | 0 | 0 | 0 | 0 | 0 | 0 | 0 | 0 | 0 | 2 | 0 | 0 |
| 146 | 0 | 1 | 1 | 1 | 0 | 0 | 0 | 0 | 0 | 0 | 0 | 0 | 0 | 0 | 1 | 0 | 0 |
| 147 | 1 | 0 | 0 | 0 | 0 | 1 | 1 | 0 | 0 | 0 | 0 | 0 | 0 | 0 | 2 | 2 | 1 |
| 148 | 1 | 0 | 0 | 0 | 1 | 0 | 1 | 0 | 0 | 0 | 0 | 0 | 0 | 0 | 3 | 1 | 1 |
| 149 | 1 | 0 | 0 | 0 | 1 | 1 | 0 | 0 | 0 | 0 | 0 | 0 | 0 | 0 | 2 | 0 | 1 |
| 150 | 1 | 0 | 0 | 1 | 0 | 0 | 1 | 0 | 0 | 0 | 0 | 0 | 0 | 0 | 3 | 1 | 1 |
| 151 | 1 | 0 | 0 | 1 | 0 | 1 | 0 | 0 | 0 | 0 | 0 | 0 | 0 | 0 | 3 | 0 | 1 |
| 152 | 1 | 0 | 0 | 1 | 1 | 0 | 0 | 0 | 0 | 0 | 0 | 0 | 0 | 0 | 3 | 0 | 1 |
| 153 | 1 | 0 | 1 | 0 | 0 | 0 | 1 | 0 | 0 | 0 | 0 | 0 | 0 | 0 | 3 | 1 | 1 |
| 154 | 1 | 0 | 1 | 0 | 0 | 1 | 0 | 0 | 0 | 0 | 0 | 0 | 0 | 0 | 3 | 0 | 1 |
| 155 | 1 | 0 | 1 | 0 | 1 | 0 | 0 | 0 | 0 | 0 | 0 | 0 | 0 | 0 | 3 | 0 | 1 |
| 156 | 1 | 0 | 1 | 1 | 0 | 0 | 0 | 0 | 0 | 0 | 0 | 0 | 0 | 0 | 2 | 0 | 1 |
| 157 | 1 | 1 | 0 | 0 | 0 | 0 | 1 | 0 | 0 | 0 | 0 | 0 | 0 | 0 | 3 | 1 | 2 |
| 158 | 1 | 1 | 0 | 0 | 0 | 1 | 0 | 0 | 0 | 0 | 0 | 0 | 0 | 0 | 3 | 0 | 2 |
| 159 | 1 | 1 | 0 | 0 | 1 | 0 | 0 | 0 | 0 | 0 | 0 | 0 | 0 | 0 | 3 | 0 | 2 |
| 160 | 1 | 1 | 0 | 1 | 0 | 0 | 0 | 0 | 0 | 0 | 0 | 0 | 0 | 0 | 2 | 0 | 2 |
| 161 | 1 | 1 | 1 | 0 | 0 | 0 | 0 | 0 | 0 | 0 | 0 | 0 | 0 | 0 | 1 | 0 | 3 |
| 162 | 0 | 0 | 0 | 0 | 0 | 0 | 0 | 0 | 0 | 0 | 0 | 0 | 1 | 1 | 1 | 2 | 0 |
| 163 | 0 | 0 | 0 | 0 | 0 | 0 | 0 | 0 | 0 | 0 | 0 | 1 | 0 | 1 | 2 | 1 | 0 |
| 164 | 0 | 0 | 0 | 0 | 0 | 0 | 0 | 0 | 0 | 0 | 0 | 1 | 1 | 0 | 1 | 0 | 0 |
| 165 | 0 | 0 | 0 | 0 | 0 | 0 | 0 | 0 | 0 | 0 | 1 | 0 | 0 | 1 | 3 | 1 | 0 |
| 166 | 0 | 0 | 0 | 0 | 0 | 0 | 0 | 0 | 0 | 0 | 1 | 0 | 1 | 0 | 3 | 0 | 0 |
| 167 | 0 | 0 | 0 | 0 | 0 | 0 | 0 | 0 | 0 | 0 | 1 | 1 | 0 | 0 | 1 | 0 | 0 |
| 168 | 0 | 0 | 0 | 0 | 0 | 0 | 0 | 0 | 0 | 1 | 0 | 0 | 0 | 1 | 3 | 1 | 0 |
| 169 | 0 | 0 | 0 | 0 | 0 | 0 | 0 | 0 | 0 | 1 | 0 | 0 | 1 | 0 | 3 | 0 | 0 |
| 170 | 0 | 0 | 0 | 0 | 0 | 0 | 0 | 0 | 0 | 1 | 0 | 1 | 0 | 0 | 2 | 0 | 0 |
| 171 | 0 | 0 | 0 | 0 | 0 | 0 | 0 | 0 | 0 | 1 | 1 | 0 | 0 | 0 | 1 | 0 | 0 |
| 172 | 0 | 0 | 0 | 0 | 0 | 0 | 0 | 0 | 1 | 0 | 0 | 0 | 0 | 1 | 4 | 1 | 0 |
| 173 | 0 | 0 | 0 | 0 | 0 | 0 | 0 | 0 | 1 | 0 | 0 | 0 | 1 | 0 | 4 | 0 | 0 |
| 174 | 0 | 0 | 0 | 0 | 0 | 0 | 0 | 0 | 1 | 0 | 0 | 1 | 0 | 0 | 3 | 0 | 0 |
| 175 | 0 | 0 | 0 | 0 | 0 | 0 | 0 | 0 | 1 | 0 | 1 | 0 | 0 | 0 | 2 | 0 | 0 |
| 176 | 0 | 0 | 0 | 0 | 0 | 0 | 0 | 0 | 1 | 1 | 0 | 0 | 0 | 0 | 1 | 0 | 0 |
| 177 | 0 | 0 | 0 | 0 | 0 | 0 | 0 | 1 | 0 | 0 | 0 | 0 | 0 | 1 | 4 | 1 | 1 |
| 178 | 0 | 0 | 0 | 0 | 0 | 0 | 0 | 1 | 0 | 0 | 0 | 0 | 1 | 0 | 4 | 0 | 1 |
| 179 | 0 | 0 | 0 | 0 | 0 | 0 | 0 | 1 | 0 | 0 | 0 | 1 | 0 | 0 | 4 | 0 | 1 |
| 180 | 0 | 0 | 0 | 0 | 0 | 0 | 0 | 1 | 0 | 0 | 1 | 0 | 0 | 0 | 3 | 0 | 1 |
| 181 | 0 | 0 | 0 | 0 | 0 | 0 | 0 | 1 | 0 | 1 | 0 | 0 | 0 | 0 | 2 | 0 | 1 |
| 182 | 0 | 0 | 0 | 0 | 0 | 0 | 0 | 1 | 1 | 0 | 0 | 0 | 0 | 0 | 1 | 0 | 2 |
| 183 | 0 | 0 | 0 | 0 | 0 | 1 | 1 | 0 | 0 | 0 | 0 | 0 | 0 | 0 | 1 | 2 | 0 |
| 184 | 0 | 0 | 0 | 0 | 1 | 0 | 1 | 0 | 0 | 0 | 0 | 0 | 0 | 0 | 2 | 1 | 0 |
| 185 | 0 | 0 | 0 | 0 | 1 | 1 | 0 | 0 | 0 | 0 | 0 | 0 | 0 | 0 | 1 | 0 | 0 |
| 186 | 0 | 0 | 0 | 1 | 0 | 0 | 1 | 0 | 0 | 0 | 0 | 0 | 0 | 0 | 3 | 1 | 0 |
| 187 | 0 | 0 | 0 | 1 | 0 | 1 | 0 | 0 | 0 | 0 | 0 | 0 | 0 | 0 | 2 | 0 | 0 |
| 188 | 0 | 0 | 0 | 1 | 1 | 0 | 0 | 0 | 0 | 0 | 0 | 0 | 0 | 0 | 1 | 0 | 0 |
| 189 | 0 | 0 | 1 | 0 | 0 | 0 | 1 | 0 | 0 | 0 | 0 | 0 | 0 | 0 | 3 | 1 | 0 |
| 190 | 0 | 0 | 1 | 0 | 0 | 1 | 0 | 0 | 0 | 0 | 0 | 0 | 0 | 0 | 3 | 0 | 0 |
| 191 | 0 | 0 | 1 | 0 | 1 | 0 | 0 | 0 | 0 | 0 | 0 | 0 | 0 | 0 | 2 | 0 | 0 |
| 192 | 0 | 0 | 1 | 1 | 0 | 0 | 0 | 0 | 0 | 0 | 0 | 0 | 0 | 0 | 1 | 0 | 0 |
| 193 | 0 | 1 | 0 | 0 | 0 | 0 | 1 | 0 | 0 | 0 | 0 | 0 | 0 | 0 | 3 | 1 | 0 |
| 194 | 0 | 1 | 0 | 0 | 0 | 1 | 0 | 0 | 0 | 0 | 0 | 0 | 0 | 0 | 3 | 0 | 0 |
| 195 | 0 | 1 | 0 | 0 | 1 | 0 | 0 | 0 | 0 | 0 | 0 | 0 | 0 | 0 | 3 | 0 | 0 |
| 196 | 0 | 1 | 0 | 1 | 0 | 0 | 0 | 0 | 0 | 0 | 0 | 0 | 0 | 0 | 2 | 0 | 0 |
| 197 | 0 | 1 | 1 | 0 | 0 | 0 | 0 | 0 | 0 | 0 | 0 | 0 | 0 | 0 | 1 | 0 | 0 |
| 198 | 1 | 0 | 0 | 0 | 0 | 0 | 1 | 0 | 0 | 0 | 0 | 0 | 0 | 0 | 3 | 1 | 1 |
| 199 | 1 | 0 | 0 | 0 | 0 | 1 | 0 | 0 | 0 | 0 | 0 | 0 | 0 | 0 | 3 | 0 | 1 |
| 200 | 1 | 0 | 0 | 0 | 1 | 0 | 0 | 0 | 0 | 0 | 0 | 0 | 0 | 0 | 3 | 0 | 1 |
| 201 | 1 | 0 | 0 | 1 | 0 | 0 | 0 | 0 | 0 | 0 | 0 | 0 | 0 | 0 | 2 | 0 | 1 |
| 202 | 1 | 0 | 1 | 0 | 0 | 0 | 0 | 0 | 0 | 0 | 0 | 0 | 0 | 0 | 2 | 0 | 1 |
| 203 | 1 | 1 | 0 | 0 | 0 | 0 | 0 | 0 | 0 | 0 | 0 | 0 | 0 | 0 | 1 | 0 | 2 |
| 204 | 0 | 0 | 0 | 0 | 0 | 0 | 0 | 1 | 0 | 0 | 0 | 0 | 0 | 0 | 1 | 0 | 1 |
| 205 | 0 | 0 | 0 | 0 | 0 | 0 | 0 | 0 | 1 | 0 | 0 | 0 | 0 | 0 | 1 | 0 | 0 |
| 206 | 0 | 0 | 0 | 0 | 0 | 0 | 0 | 0 | 0 | 1 | 0 | 0 | 0 | 0 | 1 | 0 | 0 |
| 207 | 0 | 0 | 0 | 0 | 0 | 0 | 0 | 0 | 0 | 0 | 1 | 0 | 0 | 0 | 1 | 0 | 0 |
| 208 | 0 | 0 | 0 | 0 | 0 | 0 | 0 | 0 | 0 | 0 | 0 | 1 | 0 | 0 | 1 | 0 | 0 |
| 209 | 0 | 0 | 0 | 0 | 0 | 0 | 0 | 0 | 0 | 0 | 0 | 0 | 1 | 0 | 1 | 0 | 0 |
| 210 | 0 | 0 | 0 | 0 | 0 | 0 | 0 | 0 | 0 | 0 | 0 | 0 | 0 | 1 | 1 | 1 | 0 |
| 211 | 1 | 0 | 0 | 0 | 0 | 0 | 0 | 0 | 0 | 0 | 0 | 0 | 0 | 0 | 1 | 0 | 1 |
| 212 | 0 | 1 | 0 | 0 | 0 | 0 | 0 | 0 | 0 | 0 | 0 | 0 | 0 | 0 | 1 | 0 | 0 |
| 213 | 0 | 0 | 1 | 0 | 0 | 0 | 0 | 0 | 0 | 0 | 0 | 0 | 0 | 0 | 1 | 0 | 0 |
| 214 | 0 | 0 | 0 | 1 | 0 | 0 | 0 | 0 | 0 | 0 | 0 | 0 | 0 | 0 | 1 | 0 | 0 |
| 215 | 0 | 0 | 0 | 0 | 1 | 0 | 0 | 0 | 0 | 0 | 0 | 0 | 0 | 0 | 1 | 0 | 0 |
| 216 | 0 | 0 | 0 | 0 | 0 | 1 | 0 | 0 | 0 | 0 | 0 | 0 | 0 | 0 | 1 | 0 | 0 |
| 217 | 0 | 0 | 0 | 0 | 0 | 0 | 1 | 0 | 0 | 0 | 0 | 0 | 0 | 0 | 1 | 1 | 0 |
| 218 | 0 | 0 | 0 | 0 | 0 | 0 | 0 | 0 | 0 | 0 | 0 | 0 | 0 | 0 | 1 | 0 | 0 |
| 219 | 0 | 0 | 0 | 0 | 0 | 0 | 0 | 0 | 0 | 0 | 0 | 0 | 0 | 0 | 1 | 0 | 0 |
| 220 | 1 | 1 | 1 | 1 | 1 | 1 | 0 | 0 | 0 | 0 | 0 | 0 | 0 | 0 | 1 | 0 | 0 |
| 221 | 1 | 1 | 1 | 1 | 1 | 0 | 1 | 0 | 0 | 0 | 0 | 0 | 0 | 0 | 1 | 0 | 0 |
| 222 | 1 | 1 | 1 | 1 | 0 | 1 | 1 | 0 | 0 | 0 | 0 | 0 | 0 | 0 | 1 | 0 | 0 |
| 223 | 1 | 1 | 1 | 0 | 1 | 1 | 1 | 0 | 0 | 0 | 0 | 0 | 0 | 0 | 1 | 0 | 0 |
| 224 | 1 | 1 | 0 | 1 | 1 | 1 | 1 | 0 | 0 | 0 | 0 | 0 | 0 | 0 | 1 | 0 | 0 |
| 225 | 1 | 0 | 1 | 1 | 1 | 1 | 1 | 0 | 0 | 0 | 0 | 0 | 0 | 0 | 1 | 0 | 0 |
| 226 | 0 | 1 | 1 | 1 | 1 | 1 | 1 | 0 | 0 | 0 | 0 | 0 | 0 | 0 | 1 | 0 | 0 |
| 227 | 0 | 0 | 0 | 0 | 0 | 0 | 0 | 1 | 1 | 1 | 1 | 1 | 0 | 0 | 1 | 0 | 0 |
| 228 | 0 | 0 | 0 | 0 | 0 | 0 | 0 | 1 | 1 | 1 | 1 | 0 | 1 | 0 | 1 | 0 | 0 |
| 229 | 0 | 0 | 0 | 0 | 0 | 0 | 0 | 1 | 1 | 1 | 1 | 0 | 0 | 1 | 1 | 0 | 0 |
| 230 | 0 | 0 | 0 | 0 | 0 | 0 | 0 | 1 | 1 | 1 | 0 | 1 | 1 | 0 | 1 | 0 | 0 |
| 231 | 0 | 0 | 0 | 0 | 0 | 0 | 0 | 1 | 1 | 1 | 0 | 1 | 0 | 1 | 1 | 0 | 0 |
| 232 | 0 | 0 | 0 | 0 | 0 | 0 | 0 | 1 | 1 | 1 | 0 | 0 | 1 | 1 | 1 | 0 | 0 |
| 233 | 0 | 0 | 0 | 0 | 0 | 0 | 0 | 1 | 1 | 0 | 1 | 1 | 1 | 0 | 1 | 0 | 0 |
| 234 | 0 | 0 | 0 | 0 | 0 | 0 | 0 | 1 | 1 | 0 | 1 | 1 | 0 | 1 | 1 | 0 | 0 |
| 235 | 0 | 0 | 0 | 0 | 0 | 0 | 0 | 1 | 1 | 0 | 1 | 0 | 1 | 1 | 1 | 0 | 0 |
| 236 | 0 | 0 | 0 | 0 | 0 | 0 | 0 | 1 | 1 | 0 | 0 | 1 | 1 | 1 | 1 | 0 | 0 |
| 237 | 0 | 0 | 0 | 0 | 0 | 0 | 0 | 1 | 0 | 1 | 1 | 1 | 1 | 0 | 1 | 0 | 0 |
| 238 | 0 | 0 | 0 | 0 | 0 | 0 | 0 | 1 | 0 | 1 | 1 | 1 | 0 | 1 | 1 | 0 | 0 |
| 239 | 0 | 0 | 0 | 0 | 0 | 0 | 0 | 1 | 0 | 1 | 1 | 0 | 1 | 1 | 1 | 0 | 0 |
| 240 | 0 | 0 | 0 | 0 | 0 | 0 | 0 | 1 | 0 | 1 | 0 | 1 | 1 | 1 | 1 | 0 | 0 |
| 241 | 0 | 0 | 0 | 0 | 0 | 0 | 0 | 1 | 0 | 0 | 1 | 1 | 1 | 1 | 1 | 0 | 0 |
| 242 | 0 | 0 | 0 | 0 | 0 | 0 | 0 | 0 | 1 | 1 | 1 | 1 | 1 | 0 | 1 | 0 | 0 |
| 243 | 0 | 0 | 0 | 0 | 0 | 0 | 0 | 0 | 1 | 1 | 1 | 1 | 0 | 1 | 1 | 0 | 0 |
| 244 | 0 | 0 | 0 | 0 | 0 | 0 | 0 | 0 | 1 | 1 | 1 | 0 | 1 | 1 | 1 | 0 | 0 |
| 245 | 0 | 0 | 0 | 0 | 0 | 0 | 0 | 0 | 1 | 1 | 0 | 1 | 1 | 1 | 1 | 0 | 0 |
| 246 | 0 | 0 | 0 | 0 | 0 | 0 | 0 | 0 | 1 | 0 | 1 | 1 | 1 | 1 | 1 | 0 | 0 |
| 247 | 0 | 0 | 0 | 0 | 0 | 0 | 0 | 0 | 0 | 1 | 1 | 1 | 1 | 1 | 1 | 0 | 0 |
| 248 | 1 | 1 | 1 | 1 | 1 | 1 | 1 | 0 | 0 | 0 | 0 | 0 | 0 | 0 | 1 | 0 | 0 |
| 249 | 0 | 0 | 0 | 0 | 0 | 0 | 0 | 1 | 1 | 1 | 1 | 1 | 1 | 0 | 1 | 0 | 0 |
| 250 | 0 | 0 | 0 | 0 | 0 | 0 | 0 | 1 | 1 | 1 | 1 | 1 | 0 | 1 | 1 | 0 | 0 |
| 251 | 0 | 0 | 0 | 0 | 0 | 0 | 0 | 1 | 1 | 1 | 1 | 0 | 1 | 1 | 1 | 0 | 0 |
| 252 | 0 | 0 | 0 | 0 | 0 | 0 | 0 | 1 | 1 | 1 | 0 | 1 | 1 | 1 | 1 | 0 | 0 |
| 253 | 0 | 0 | 0 | 0 | 0 | 0 | 0 | 1 | 1 | 0 | 1 | 1 | 1 | 1 | 1 | 0 | 0 |
| 254 | 0 | 0 | 0 | 0 | 0 | 0 | 0 | 1 | 0 | 1 | 1 | 1 | 1 | 1 | 1 | 0 | 0 |
| 255 | 0 | 0 | 0 | 0 | 0 | 0 | 0 | 0 | 1 | 1 | 1 | 1 | 1 | 1 | 1 | 0 | 0 |
| 256 | 0 | 0 | 0 | 0 | 0 | 0 | 0 | 1 | 1 | 1 | 1 | 1 | 1 | 1 | 1 | 0 | 0 |
| 257 | 1 | 0 | 0 | 0 | 0 | 0 | 0 | 0 | 1 | 1 | 0 | 0 | 0 | 0 | 1 | 0 | 0 |
| 258 | 1 | 0 | 0 | 0 | 0 | 0 | 0 | 0 | 1 | 0 | 1 | 0 | 0 | 0 | 3 | 0 | 0 |
| 259 | 1 | 0 | 0 | 0 | 0 | 0 | 0 | 0 | 1 | 0 | 0 | 1 | 0 | 0 | 3 | 0 | 0 |
| 260 | 1 | 0 | 0 | 0 | 0 | 0 | 0 | 0 | 1 | 0 | 0 | 0 | 1 | 0 | 3 | 0 | 0 |
| 261 | 1 | 0 | 0 | 0 | 0 | 0 | 0 | 0 | 1 | 0 | 0 | 0 | 0 | 1 | 1 | 0 | 0 |
| 262 | 1 | 0 | 0 | 0 | 0 | 0 | 0 | 0 | 0 | 1 | 1 | 0 | 0 | 0 | 2 | 0 | 0 |
| 263 | 1 | 0 | 0 | 0 | 0 | 0 | 0 | 0 | 0 | 1 | 0 | 1 | 0 | 0 | 3 | 0 | 0 |
| 264 | 1 | 0 | 0 | 0 | 0 | 0 | 0 | 0 | 0 | 1 | 0 | 0 | 1 | 0 | 3 | 0 | 0 |
| 265 | 1 | 0 | 0 | 0 | 0 | 0 | 0 | 0 | 0 | 1 | 0 | 0 | 0 | 1 | 2 | 0 | 0 |
| 266 | 1 | 0 | 0 | 0 | 0 | 0 | 0 | 0 | 0 | 0 | 1 | 1 | 0 | 0 | 1 | 0 | 0 |
| 267 | 1 | 0 | 0 | 0 | 0 | 0 | 0 | 0 | 0 | 0 | 1 | 0 | 1 | 0 | 3 | 0 | 0 |
| 268 | 1 | 0 | 0 | 0 | 0 | 0 | 0 | 0 | 0 | 0 | 1 | 0 | 0 | 1 | 2 | 0 | 0 |
| 269 | 1 | 0 | 0 | 0 | 0 | 0 | 0 | 0 | 0 | 0 | 0 | 1 | 1 | 0 | 1 | 0 | 0 |
| 270 | 1 | 0 | 0 | 0 | 0 | 0 | 0 | 0 | 0 | 0 | 0 | 1 | 0 | 1 | 2 | 0 | 0 |
| 271 | 1 | 0 | 0 | 0 | 0 | 0 | 0 | 0 | 0 | 0 | 0 | 0 | 1 | 1 | 1 | 0 | 0 |
| 272 | 0 | 1 | 0 | 0 | 0 | 0 | 0 | 0 | 0 | 1 | 1 | 0 | 0 | 0 | 1 | 0 | 0 |
| 273 | 0 | 1 | 0 | 0 | 0 | 0 | 0 | 0 | 0 | 1 | 0 | 1 | 0 | 0 | 3 | 0 | 0 |
| 274 | 0 | 1 | 0 | 0 | 0 | 0 | 0 | 0 | 0 | 1 | 0 | 0 | 1 | 0 | 3 | 0 | 0 |
| 275 | 0 | 1 | 0 | 0 | 0 | 0 | 0 | 0 | 0 | 1 | 0 | 0 | 0 | 1 | 1 | 0 | 0 |
| 276 | 0 | 1 | 0 | 0 | 0 | 0 | 0 | 0 | 0 | 0 | 1 | 1 | 0 | 0 | 2 | 0 | 0 |
| 277 | 0 | 1 | 0 | 0 | 0 | 0 | 0 | 0 | 0 | 0 | 1 | 0 | 1 | 0 | 3 | 0 | 0 |
| 278 | 0 | 1 | 0 | 0 | 0 | 0 | 0 | 0 | 0 | 0 | 1 | 0 | 0 | 1 | 2 | 0 | 0 |
| 279 | 0 | 1 | 0 | 0 | 0 | 0 | 0 | 0 | 0 | 0 | 0 | 1 | 1 | 0 | 2 | 0 | 0 |
| 280 | 0 | 1 | 0 | 0 | 0 | 0 | 0 | 0 | 0 | 0 | 0 | 1 | 0 | 1 | 2 | 0 | 0 |
| 281 | 0 | 1 | 0 | 0 | 0 | 0 | 0 | 0 | 0 | 0 | 0 | 0 | 1 | 1 | 1 | 0 | 0 |
| 282 | 0 | 0 | 1 | 0 | 0 | 0 | 0 | 1 | 0 | 0 | 1 | 0 | 0 | 0 | 3 | 0 | 0 |
| 283 | 0 | 0 | 1 | 0 | 0 | 0 | 0 | 1 | 0 | 0 | 0 | 1 | 0 | 0 | 3 | 0 | 0 |
| 284 | 0 | 0 | 1 | 0 | 0 | 0 | 0 | 1 | 0 | 0 | 0 | 0 | 1 | 0 | 3 | 0 | 0 |
| 285 | 0 | 0 | 1 | 0 | 0 | 0 | 0 | 1 | 0 | 0 | 0 | 0 | 0 | 1 | 2 | 0 | 0 |
| 286 | 0 | 0 | 1 | 0 | 0 | 0 | 0 | 0 | 0 | 0 | 1 | 1 | 0 | 0 | 2 | 0 | 0 |
| 287 | 0 | 0 | 1 | 0 | 0 | 0 | 0 | 0 | 0 | 0 | 1 | 0 | 1 | 0 | 3 | 0 | 0 |
| 288 | 0 | 0 | 1 | 0 | 0 | 0 | 0 | 0 | 0 | 0 | 1 | 0 | 0 | 1 | 3 | 0 | 0 |
| 289 | 0 | 0 | 1 | 0 | 0 | 0 | 0 | 0 | 0 | 0 | 0 | 1 | 1 | 0 | 2 | 0 | 0 |
| 290 | 0 | 0 | 1 | 0 | 0 | 0 | 0 | 0 | 0 | 0 | 0 | 1 | 0 | 1 | 2 | 0 | 0 |
| 291 | 0 | 0 | 1 | 0 | 0 | 0 | 0 | 0 | 0 | 0 | 0 | 0 | 1 | 1 | 1 | 0 | 0 |
| 292 | 0 | 0 | 0 | 1 | 0 | 0 | 0 | 1 | 1 | 0 | 0 | 0 | 0 | 0 | 1 | 0 | 0 |
| 293 | 0 | 0 | 0 | 1 | 0 | 0 | 0 | 1 | 0 | 0 | 0 | 1 | 0 | 0 | 2 | 0 | 0 |
| 294 | 0 | 0 | 0 | 1 | 0 | 0 | 0 | 1 | 0 | 0 | 0 | 0 | 1 | 0 | 3 | 0 | 0 |
| 295 | 0 | 0 | 0 | 1 | 0 | 0 | 0 | 1 | 0 | 0 | 0 | 0 | 0 | 1 | 2 | 0 | 0 |
| 296 | 0 | 0 | 0 | 1 | 0 | 0 | 0 | 0 | 1 | 0 | 0 | 1 | 0 | 0 | 3 | 0 | 0 |
| 297 | 0 | 0 | 0 | 1 | 0 | 0 | 0 | 0 | 1 | 0 | 0 | 0 | 1 | 0 | 3 | 0 | 0 |
| 298 | 0 | 0 | 0 | 1 | 0 | 0 | 0 | 0 | 1 | 0 | 0 | 0 | 0 | 1 | 2 | 0 | 0 |
| 299 | 0 | 0 | 0 | 1 | 0 | 0 | 0 | 0 | 0 | 0 | 0 | 1 | 1 | 0 | 1 | 0 | 0 |
| 300 | 0 | 0 | 0 | 1 | 0 | 0 | 0 | 0 | 0 | 0 | 0 | 1 | 0 | 1 | 2 | 0 | 0 |
| 301 | 0 | 0 | 0 | 1 | 0 | 0 | 0 | 0 | 0 | 0 | 0 | 0 | 1 | 1 | 1 | 0 | 0 |
| 302 | 0 | 0 | 0 | 0 | 1 | 0 | 0 | 1 | 1 | 0 | 0 | 0 | 0 | 0 | 1 | 0 | 0 |
| 303 | 0 | 0 | 0 | 0 | 1 | 0 | 0 | 1 | 0 | 1 | 0 | 0 | 0 | 0 | 2 | 0 | 0 |
| 304 | 0 | 0 | 0 | 0 | 1 | 0 | 0 | 1 | 0 | 0 | 0 | 0 | 1 | 0 | 3 | 0 | 0 |
| 305 | 0 | 0 | 0 | 0 | 1 | 0 | 0 | 1 | 0 | 0 | 0 | 0 | 0 | 1 | 2 | 0 | 0 |
| 306 | 0 | 0 | 0 | 0 | 1 | 0 | 0 | 0 | 1 | 1 | 0 | 0 | 0 | 0 | 1 | 0 | 0 |
| 307 | 0 | 0 | 0 | 0 | 1 | 0 | 0 | 0 | 1 | 0 | 0 | 0 | 1 | 0 | 3 | 0 | 0 |
| 308 | 0 | 0 | 0 | 0 | 1 | 0 | 0 | 0 | 1 | 0 | 0 | 0 | 0 | 1 | 3 | 0 | 0 |
| 309 | 0 | 0 | 0 | 0 | 1 | 0 | 0 | 0 | 0 | 1 | 0 | 0 | 1 | 0 | 3 | 0 | 0 |
| 310 | 0 | 0 | 0 | 0 | 1 | 0 | 0 | 0 | 0 | 1 | 0 | 0 | 0 | 1 | 2 | 0 | 0 |
| 311 | 0 | 0 | 0 | 0 | 1 | 0 | 0 | 0 | 0 | 0 | 0 | 0 | 1 | 1 | 1 | 0 | 0 |
| 312 | 0 | 0 | 0 | 0 | 0 | 1 | 0 | 1 | 1 | 0 | 0 | 0 | 0 | 0 | 1 | 0 | 0 |
| 313 | 0 | 0 | 0 | 0 | 0 | 1 | 0 | 1 | 0 | 1 | 0 | 0 | 0 | 0 | 2 | 0 | 0 |
| 314 | 0 | 0 | 0 | 0 | 0 | 1 | 0 | 1 | 0 | 0 | 1 | 0 | 0 | 0 | 2 | 0 | 0 |
| 315 | 0 | 0 | 0 | 0 | 0 | 1 | 0 | 1 | 0 | 0 | 0 | 0 | 0 | 1 | 2 | 0 | 0 |
| 316 | 0 | 0 | 0 | 0 | 0 | 1 | 0 | 0 | 1 | 1 | 0 | 0 | 0 | 0 | 2 | 0 | 0 |
| 317 | 0 | 0 | 0 | 0 | 0 | 1 | 0 | 0 | 1 | 0 | 1 | 0 | 0 | 0 | 3 | 0 | 0 |
| 318 | 0 | 0 | 0 | 0 | 0 | 1 | 0 | 0 | 1 | 0 | 0 | 0 | 0 | 1 | 3 | 0 | 0 |
| 319 | 0 | 0 | 0 | 0 | 0 | 1 | 0 | 0 | 0 | 1 | 1 | 0 | 0 | 0 | 2 | 0 | 0 |
| 320 | 0 | 0 | 0 | 0 | 0 | 1 | 0 | 0 | 0 | 1 | 0 | 0 | 0 | 1 | 3 | 0 | 0 |
| 321 | 0 | 0 | 0 | 0 | 0 | 1 | 0 | 0 | 0 | 0 | 1 | 0 | 0 | 1 | 3 | 0 | 0 |
| 322 | 0 | 0 | 0 | 0 | 0 | 0 | 1 | 1 | 1 | 0 | 0 | 0 | 0 | 0 | 1 | 0 | 0 |
| 323 | 0 | 0 | 0 | 0 | 0 | 0 | 1 | 1 | 0 | 1 | 0 | 0 | 0 | 0 | 2 | 0 | 0 |
| 324 | 0 | 0 | 0 | 0 | 0 | 0 | 1 | 1 | 0 | 0 | 1 | 0 | 0 | 0 | 2 | 0 | 0 |
| 325 | 0 | 0 | 0 | 0 | 0 | 0 | 1 | 1 | 0 | 0 | 0 | 1 | 0 | 0 | 2 | 0 | 0 |
| 326 | 0 | 0 | 0 | 0 | 0 | 0 | 1 | 0 | 1 | 1 | 0 | 0 | 0 | 0 | 2 | 0 | 0 |
| 327 | 0 | 0 | 0 | 0 | 0 | 0 | 1 | 0 | 1 | 0 | 1 | 0 | 0 | 0 | 3 | 0 | 0 |
| 328 | 0 | 0 | 0 | 0 | 0 | 0 | 1 | 0 | 1 | 0 | 0 | 1 | 0 | 0 | 3 | 0 | 0 |
| 329 | 0 | 0 | 0 | 0 | 0 | 0 | 1 | 0 | 0 | 1 | 1 | 0 | 0 | 0 | 2 | 0 | 0 |
| 330 | 0 | 0 | 0 | 0 | 0 | 0 | 1 | 0 | 0 | 1 | 0 | 1 | 0 | 0 | 3 | 0 | 0 |
| 331 | 0 | 0 | 0 | 0 | 0 | 0 | 1 | 0 | 0 | 0 | 1 | 1 | 0 | 0 | 1 | 0 | 0 |
| 332 | 1 | 0 | 0 | 0 | 0 | 0 | 0 | 0 | 1 | 1 | 1 | 0 | 0 | 0 | 1 | 0 | 4 |
| 333 | 1 | 0 | 0 | 0 | 0 | 0 | 0 | 0 | 1 | 1 | 0 | 1 | 0 | 0 | 3 | 0 | 3 |
| 334 | 1 | 0 | 0 | 0 | 0 | 0 | 0 | 0 | 1 | 1 | 0 | 0 | 1 | 0 | 3 | 0 | 3 |
| 335 | 1 | 0 | 0 | 0 | 0 | 0 | 0 | 0 | 1 | 1 | 0 | 0 | 0 | 1 | 3 | 1 | 3 |
| 336 | 1 | 0 | 0 | 0 | 0 | 0 | 0 | 0 | 1 | 0 | 1 | 1 | 0 | 0 | 3 | 0 | 2 |
| 337 | 1 | 0 | 0 | 0 | 0 | 0 | 0 | 0 | 1 | 0 | 1 | 0 | 1 | 0 | 4 | 0 | 2 |
| 338 | 1 | 0 | 0 | 0 | 0 | 0 | 0 | 0 | 1 | 0 | 1 | 0 | 0 | 1 | 4 | 1 | 2 |
| 339 | 1 | 0 | 0 | 0 | 0 | 0 | 0 | 0 | 1 | 0 | 0 | 1 | 1 | 0 | 3 | 0 | 2 |
| 340 | 1 | 0 | 0 | 0 | 0 | 0 | 0 | 0 | 1 | 0 | 0 | 1 | 0 | 1 | 4 | 1 | 2 |
| 341 | 1 | 0 | 0 | 0 | 0 | 0 | 0 | 0 | 1 | 0 | 0 | 0 | 1 | 1 | 3 | 2 | 2 |
| 342 | 1 | 0 | 0 | 0 | 0 | 0 | 0 | 0 | 0 | 1 | 1 | 1 | 0 | 0 | 1 | 0 | 1 |
| 343 | 1 | 0 | 0 | 0 | 0 | 0 | 0 | 0 | 0 | 1 | 1 | 0 | 1 | 0 | 3 | 0 | 1 |
| 344 | 1 | 0 | 0 | 0 | 0 | 0 | 0 | 0 | 0 | 1 | 1 | 0 | 0 | 1 | 3 | 1 | 1 |
| 345 | 1 | 0 | 0 | 0 | 0 | 0 | 0 | 0 | 0 | 1 | 0 | 1 | 1 | 0 | 3 | 0 | 1 |
| 346 | 1 | 0 | 0 | 0 | 0 | 0 | 0 | 0 | 0 | 1 | 0 | 1 | 0 | 1 | 4 | 0 | 1 |
| 347 | 1 | 0 | 0 | 0 | 0 | 0 | 0 | 0 | 0 | 1 | 0 | 0 | 1 | 1 | 3 | 2 | 1 |
| 348 | 1 | 0 | 0 | 0 | 0 | 0 | 0 | 0 | 0 | 0 | 1 | 1 | 1 | 0 | 1 | 0 | 1 |
| 349 | 1 | 0 | 0 | 0 | 0 | 0 | 0 | 0 | 0 | 0 | 1 | 1 | 0 | 1 | 3 | 1 | 1 |
| 350 | 1 | 0 | 0 | 0 | 0 | 0 | 0 | 0 | 0 | 0 | 1 | 0 | 1 | 1 | 3 | 2 | 1 |
| 351 | 1 | 0 | 0 | 0 | 0 | 0 | 0 | 0 | 0 | 0 | 0 | 1 | 1 | 1 | 1 | 3 | 1 |
| 352 | 0 | 1 | 0 | 0 | 0 | 0 | 0 | 0 | 0 | 1 | 1 | 1 | 0 | 0 | 1 | 0 | 0 |
| 353 | 0 | 1 | 0 | 0 | 0 | 0 | 0 | 0 | 0 | 1 | 1 | 0 | 1 | 0 | 3 | 0 | 0 |
| 354 | 0 | 1 | 0 | 0 | 0 | 0 | 0 | 0 | 0 | 1 | 1 | 0 | 0 | 1 | 3 | 1 | 0 |
| 355 | 0 | 1 | 0 | 0 | 0 | 0 | 0 | 0 | 0 | 1 | 0 | 1 | 1 | 0 | 3 | 0 | 0 |
| 356 | 0 | 1 | 0 | 0 | 0 | 0 | 0 | 0 | 0 | 1 | 0 | 1 | 0 | 1 | 4 | 1 | 0 |
| 357 | 0 | 1 | 0 | 0 | 0 | 0 | 0 | 0 | 0 | 1 | 0 | 0 | 1 | 1 | 3 | 2 | 0 |
| 358 | 0 | 1 | 0 | 0 | 0 | 0 | 0 | 0 | 0 | 0 | 1 | 1 | 1 | 0 | 1 | 0 | 0 |
| 359 | 0 | 1 | 0 | 0 | 0 | 0 | 0 | 0 | 0 | 0 | 1 | 1 | 0 | 1 | 3 | 1 | 0 |
| 360 | 0 | 1 | 0 | 0 | 0 | 0 | 0 | 0 | 0 | 0 | 1 | 0 | 1 | 1 | 3 | 2 | 0 |
| 361 | 0 | 1 | 0 | 0 | 0 | 0 | 0 | 0 | 0 | 0 | 0 | 1 | 1 | 1 | 1 | 3 | 0 |
| 362 | 0 | 0 | 1 | 0 | 0 | 0 | 0 | 1 | 0 | 0 | 1 | 1 | 0 | 0 | 18 | 0 | 1 |
| 363 | 0 | 0 | 1 | 0 | 0 | 0 | 0 | 1 | 0 | 0 | 1 | 0 | 1 | 0 | 18 | 0 | 1 |
| 364 | 0 | 0 | 1 | 0 | 0 | 0 | 0 | 1 | 0 | 0 | 1 | 0 | 0 | 1 | 18 | 1 | 1 |
| 365 | 0 | 0 | 1 | 0 | 0 | 0 | 0 | 1 | 0 | 0 | 0 | 1 | 1 | 0 | 18 | 1 | 1 |
| 366 | 0 | 0 | 1 | 0 | 0 | 0 | 0 | 1 | 0 | 0 | 0 | 1 | 0 | 1 | 18 | 2 | 1 |
| 367 | 0 | 0 | 1 | 0 | 0 | 0 | 0 | 1 | 0 | 0 | 0 | 0 | 1 | 1 | 18 | 3 | 1 |
| 368 | 0 | 0 | 1 | 0 | 0 | 0 | 0 | 0 | 0 | 0 | 1 | 1 | 1 | 0 | 1 | 0 | 0 |
| 369 | 0 | 0 | 1 | 0 | 0 | 0 | 0 | 0 | 0 | 0 | 1 | 1 | 0 | 1 | 3 | 1 | 0 |
| 370 | 0 | 0 | 1 | 0 | 0 | 0 | 0 | 0 | 0 | 0 | 1 | 0 | 1 | 1 | 3 | 2 | 0 |
| 371 | 0 | 0 | 1 | 0 | 0 | 0 | 0 | 0 | 0 | 0 | 0 | 1 | 1 | 1 | 1 | 3 | 0 |
| 372 | 0 | 0 | 0 | 1 | 0 | 0 | 0 | 1 | 1 | 0 | 0 | 1 | 0 | 0 | 18 | 0 | 2 |
| 373 | 0 | 0 | 0 | 1 | 0 | 0 | 0 | 1 | 1 | 0 | 0 | 0 | 1 | 0 | 18 | 0 | 2 |
| 374 | 0 | 0 | 0 | 1 | 0 | 0 | 0 | 1 | 1 | 0 | 0 | 0 | 0 | 1 | 18 | 1 | 2 |
| 375 | 0 | 0 | 0 | 1 | 0 | 0 | 0 | 1 | 0 | 0 | 0 | 1 | 1 | 0 | 18 | 0 | 1 |
| 376 | 0 | 0 | 0 | 1 | 0 | 0 | 0 | 1 | 0 | 0 | 0 | 1 | 0 | 1 | 18 | 1 | 1 |
| 377 | 0 | 0 | 0 | 1 | 0 | 0 | 0 | 1 | 0 | 0 | 0 | 0 | 1 | 1 | 18 | 1 | 3 |
| 378 | 0 | 0 | 0 | 1 | 0 | 0 | 0 | 0 | 1 | 0 | 0 | 1 | 1 | 0 | 18 | 0 | 0 |
| 379 | 0 | 0 | 0 | 1 | 0 | 0 | 0 | 0 | 1 | 0 | 0 | 1 | 0 | 1 | 18 | 0 | 1 |
| 380 | 0 | 0 | 0 | 1 | 0 | 0 | 0 | 0 | 1 | 0 | 0 | 0 | 1 | 1 | 18 | 0 | 2 |
| 381 | 0 | 0 | 0 | 1 | 0 | 0 | 0 | 0 | 0 | 0 | 0 | 1 | 1 | 1 | 1 | 0 | 3 |
| 382 | 0 | 0 | 0 | 0 | 1 | 0 | 0 | 1 | 1 | 1 | 0 | 0 | 0 | 0 | 1 | 2 | 0 |
| 383 | 0 | 0 | 0 | 0 | 1 | 0 | 0 | 1 | 1 | 0 | 0 | 0 | 1 | 0 | 18 | 2 | 0 |
| 384 | 0 | 0 | 0 | 0 | 1 | 0 | 0 | 1 | 1 | 0 | 0 | 0 | 0 | 1 | 18 | 2 | 1 |
| 385 | 0 | 0 | 0 | 0 | 1 | 0 | 0 | 1 | 0 | 1 | 0 | 0 | 1 | 0 | 18 | 1 | 0 |
| 386 | 0 | 0 | 0 | 0 | 1 | 0 | 0 | 1 | 0 | 1 | 0 | 0 | 0 | 1 | 18 | 1 | 1 |
| 387 | 0 | 0 | 0 | 0 | 1 | 0 | 0 | 1 | 0 | 0 | 0 | 0 | 1 | 1 | 18 | 2 | 1 |
| 388 | 0 | 0 | 0 | 0 | 1 | 0 | 0 | 0 | 1 | 1 | 0 | 0 | 1 | 0 | 18 | 0 | 0 |
| 389 | 0 | 0 | 0 | 0 | 1 | 0 | 0 | 0 | 1 | 1 | 0 | 0 | 0 | 1 | 18 | 1 | 0 |
| 390 | 0 | 0 | 0 | 0 | 1 | 0 | 0 | 0 | 1 | 0 | 0 | 0 | 1 | 1 | 18 | 2 | 0 |
| 391 | 0 | 0 | 0 | 0 | 1 | 0 | 0 | 0 | 0 | 1 | 0 | 0 | 1 | 1 | 18 | 3 | 0 |
| 392 | 0 | 0 | 0 | 0 | 0 | 1 | 0 | 1 | 1 | 1 | 0 | 0 | 0 | 0 | 1 | 0 | 3 |
| 393 | 0 | 0 | 0 | 0 | 0 | 1 | 0 | 1 | 1 | 0 | 1 | 0 | 0 | 0 | 3 | 0 | 2 |
| 394 | 0 | 0 | 0 | 0 | 0 | 1 | 0 | 1 | 1 | 0 | 0 | 0 | 0 | 1 | 18 | 1 | 2 |
| 395 | 0 | 0 | 0 | 0 | 0 | 1 | 0 | 1 | 0 | 1 | 1 | 0 | 0 | 0 | 3 | 0 | 1 |
| 396 | 0 | 0 | 0 | 0 | 0 | 1 | 0 | 1 | 0 | 1 | 0 | 0 | 0 | 1 | 18 | 1 | 1 |
| 397 | 0 | 0 | 0 | 0 | 0 | 1 | 0 | 1 | 0 | 0 | 1 | 0 | 0 | 1 | 18 | 1 | 1 |
| 398 | 0 | 0 | 0 | 0 | 0 | 1 | 0 | 0 | 1 | 1 | 1 | 0 | 0 | 0 | 1 | 0 | 0 |
| 399 | 0 | 0 | 0 | 0 | 0 | 1 | 0 | 0 | 1 | 1 | 0 | 0 | 0 | 1 | 18 | 1 | 0 |
| 400 | 0 | 0 | 0 | 0 | 0 | 1 | 0 | 0 | 1 | 0 | 1 | 0 | 0 | 1 | 18 | 1 | 0 |
| 401 | 0 | 0 | 0 | 0 | 0 | 1 | 0 | 0 | 0 | 1 | 1 | 0 | 0 | 1 | 18 | 1 | 0 |
| 402 | 0 | 0 | 0 | 0 | 0 | 0 | 1 | 1 | 1 | 1 | 0 | 0 | 0 | 0 | 1 | 1 | 3 |
| 403 | 0 | 0 | 0 | 0 | 0 | 0 | 1 | 1 | 1 | 0 | 1 | 0 | 0 | 0 | 3 | 1 | 2 |
| 404 | 0 | 0 | 0 | 0 | 0 | 0 | 1 | 1 | 1 | 0 | 0 | 1 | 0 | 0 | 3 | 1 | 2 |
| 405 | 0 | 0 | 0 | 0 | 0 | 0 | 1 | 1 | 0 | 1 | 1 | 0 | 0 | 0 | 3 | 1 | 1 |
| 406 | 0 | 0 | 0 | 0 | 0 | 0 | 1 | 1 | 0 | 1 | 0 | 1 | 0 | 0 | 4 | 1 | 1 |
| 407 | 0 | 0 | 0 | 0 | 0 | 0 | 1 | 1 | 0 | 0 | 1 | 1 | 0 | 0 | 3 | 1 | 1 |
| 408 | 0 | 0 | 0 | 0 | 0 | 0 | 1 | 0 | 1 | 1 | 1 | 0 | 0 | 0 | 1 | 1 | 0 |
| 409 | 0 | 0 | 0 | 0 | 0 | 0 | 1 | 0 | 1 | 1 | 0 | 1 | 0 | 0 | 3 | 1 | 0 |
| 410 | 0 | 0 | 0 | 0 | 0 | 0 | 1 | 0 | 1 | 0 | 1 | 1 | 0 | 0 | 3 | 1 | 0 |
| 411 | 0 | 0 | 0 | 0 | 0 | 0 | 1 | 0 | 0 | 1 | 1 | 1 | 0 | 0 | 1 | 1 | 0 |

Table IV: List of all possible shift patterns.

## C.5 Final Penalty Cost

Table V was compiled from the information given in the previous sections following the rules detailed in section 2.1.3. It shows the 'cost' $p_{ij}$ of nurse $i$ working shift pattern $j$. In the table,

- The first line is the index $i$ of the nurse.
- The second line gives the upper and lower bound, first for day then for night shifts of possible shift patterns for that nurse, i.e. these are the ranges for $j$.
- The following lines show the penalty cost $p_{ij}$ associated with nurse $i$ working a shift pattern $j$ starting with the lower day shift bound.

| | | | | | | | | | | | | | |
|---|---|---|---|---|---|---|---|---|---|---|---|---|---|
| Nurse index | 1 | | | | | | | | | | | | |
| Shift pattern range | 1 | 21 | 22 | 56 | | | | | | | | | |
| Penalty Cost | 0 | 2 | 0 | 1 | 2 | 0 | 1 | 1 | 2 | 0 | 2 | 2 | 2 |
| | 2 | 1 | 0 | 1 | 1 | 1 | 2 | 0 | 18 | 20 | 20 | 20 | 19 |
| | 19 | 20 | 20 | 19 | 20 | 19 | 20 | 20 | 20 | 20 | 21 | 20 | 19 |
| | 20 | 20 | 18 | 18 | 20 | 20 | 19 | 20 | 19 | 20 | 20 | 20 | 19 |
| | 2 | 19 | 18 | 20 | | | | | | | | | |
| Nurse index | 2 | | | | | | | | | | | | |
| Shift pattern range | 57 | 91 | 92 | 126 | | | | | | | | | |
| Penalty Cost | 16 | 10 | 10 | 9 | 9 | 10 | 10 | 9 | 10 | 9 | 10 | 10 | 9 |
| | 10 | 11 | 10 | 9 | 9 | 10 | 9 | 8 | 10 | 9 | 9 | 10 | 9 |
| | 6 | 10 | 10 | 10 | 10 | 8 | 9 | 9 | 9 | 8 | 0 | 5 | 5 |
| | 6 | 0 | 6 | 7 | 7 | 6 | 7 | 7 | 5 | 6 | 5 | 0 | 6 |
| | 6 | 6 | 6 | 6 | 5 | 6 | 6 | 6 | 0 | 7 | 7 | 7 | 6 |
| | 7 | 7 | 6 | 5 | 0 | | | | | | | | |

Table V: Examples of final shift pattern cost values.

# Appendix D Additional Nurse Scheduling Results

## D.1 Graphs of Typical Genetic Algorithm Runs

The graphs in Figure V show a single run of various types of genetic algorithms with a typical data set. The 'Average' line shows the development of the population as a whole, whilst the 'Best Feasible' line shows the best feasible member of the population. A comparison for all genetic algorithm types is shown in Figure VI and in Figure VII.

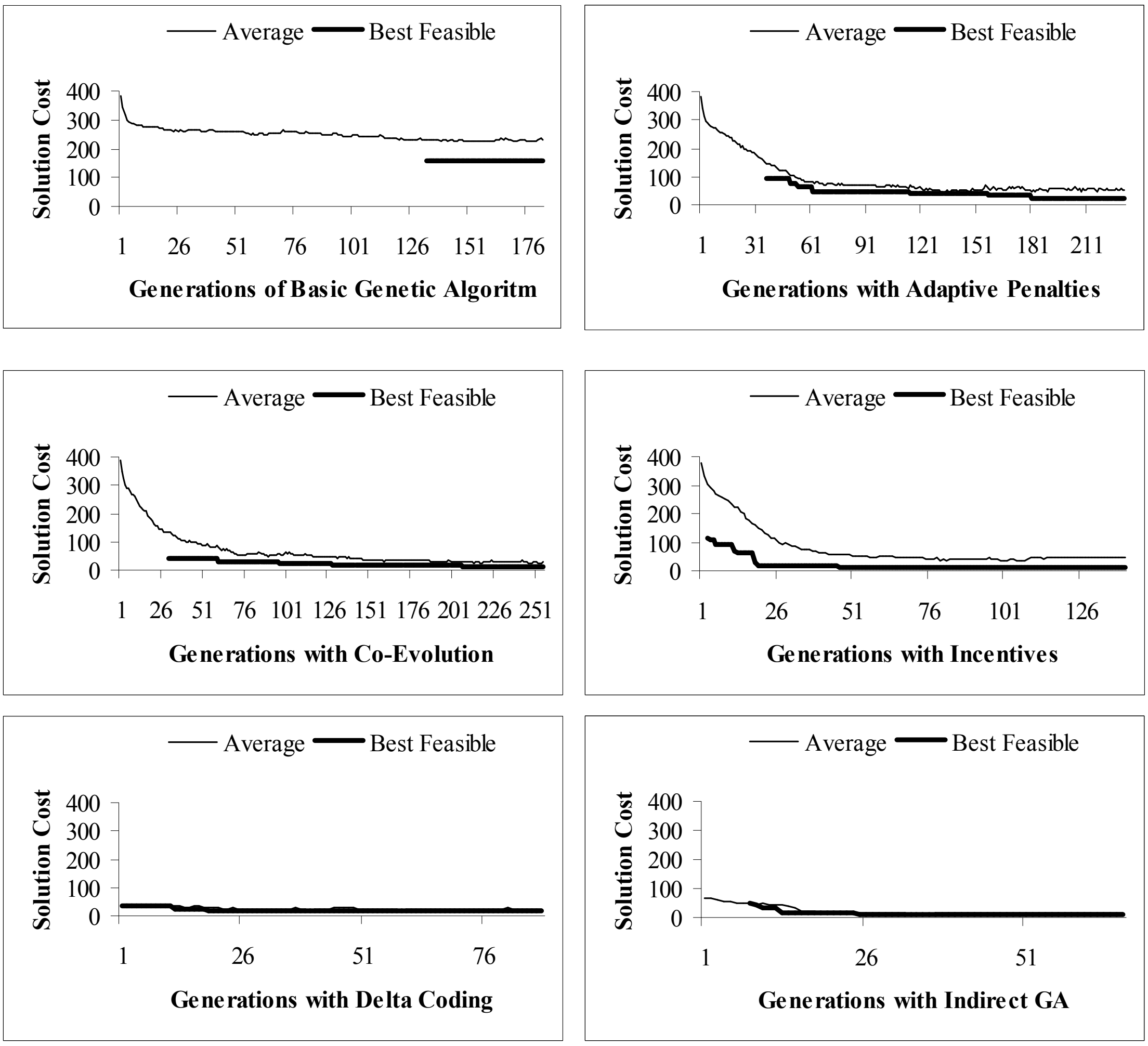

Figure V: Typical genetic algorithm runs with various strategies.

Some interesting observations can be made from the graphs. For this particular data set, introducing the dynamic weights already makes a big difference from the basic genetic algorithm. The first feasible solutions of high quality are found by the co-evolutionary approach. Once repair and incentives are introduced, the algorithm finds the first feasible solution much faster. This shows again how effective it is to add a local hillclimber to counter the ‘lack of killer instinct’ effect of genetic algorithms.

Note that the Delta Coding genetic algorithm seems to perform well, however, this is only because it is seeded with a good feasible solution from a previous run. Finally, the population average is much better for the indirect genetic algorithm than for the direct approaches (apart from delta coding for obvious reasons). This indicates that the actual solution space explored is smaller due to the bias introduced by the decoder.

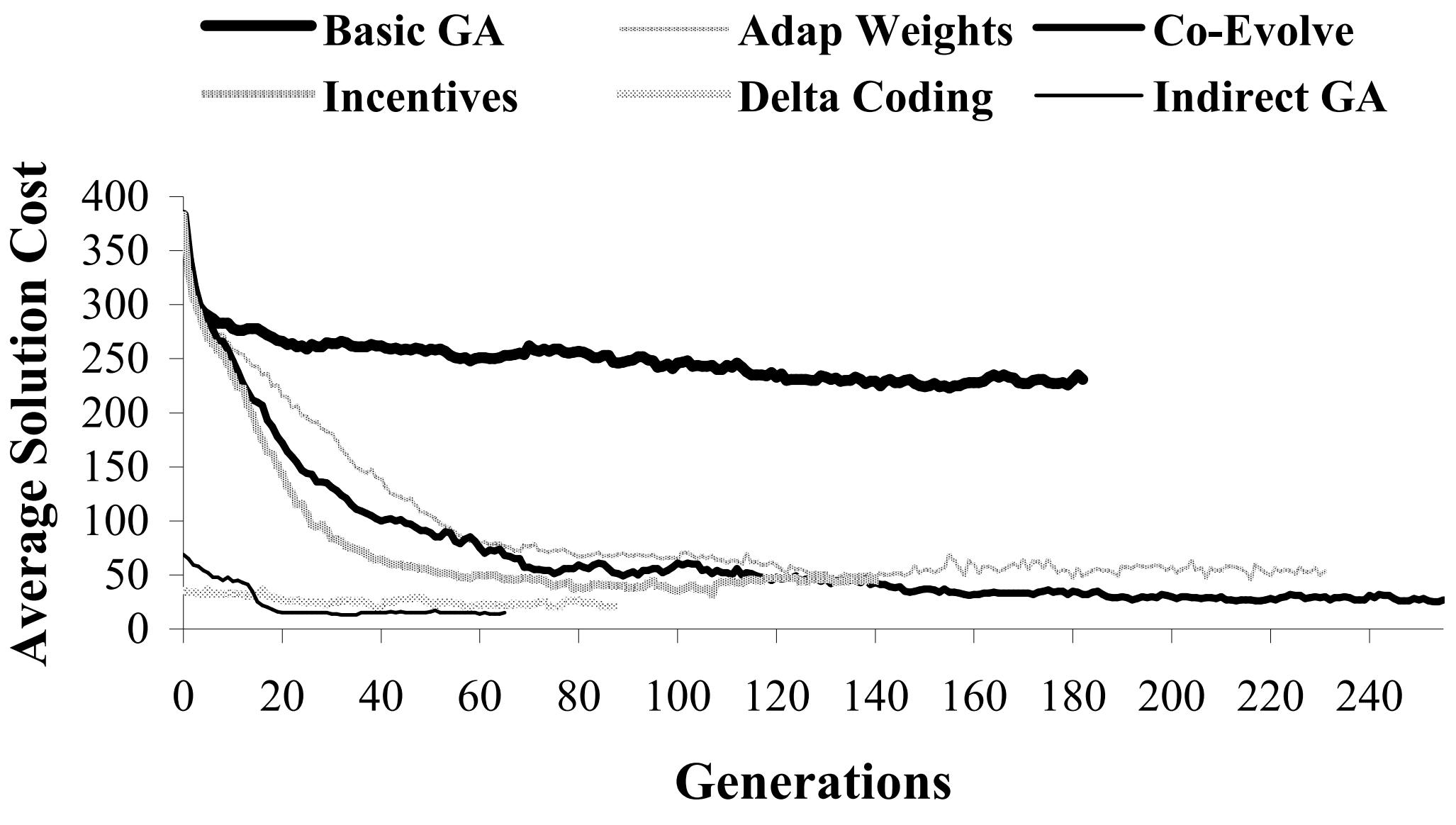

Figure VI: Comparison of average solution cost for various types of genetic algorithms and a typical data set.

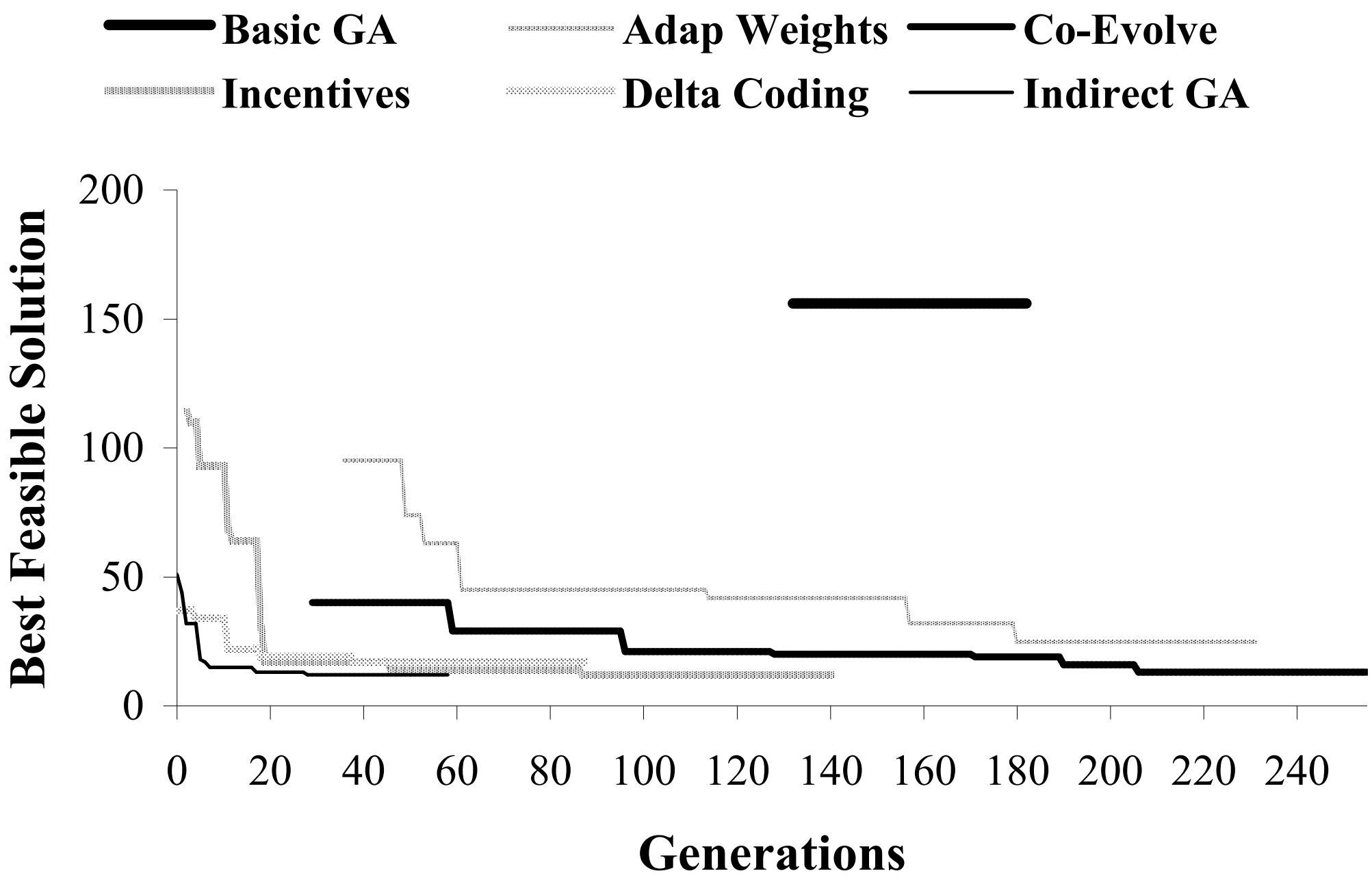

Figure VII: Comparison of best feasible solution cost for various types of genetic algorithms and a typical data set.

## D.2 Full Genetic Algorithm and Tabu Search Results

Table VI shows the feasibility and cost for all 52 data sets found with the best direct genetic algorithm, best indirect genetic algorithm, tabu search and XPRESS MP. Detailed results for other genetic algorithms are shown in Figure VIII to Figure XIII.

| | Direct GA | | Indirect GA | | Tabu Search | | IP |
|---|---|---|---|---|---|---|---|
| Week | Feasibility | Cost | Feasibility | Cost | Iterations | Cost | Cost |
| **1** | 1 | 0 | 1 | 0 | 128 | 0 | 0 |
| **2** | 1 | 12 | 1 | 12 | 67 | 11 | 11 |
| **3** | 1 | 18 | 1 | 18 | 298 | 18 | 18 |
| **4** | 1 | 0 | 1 | 0 | 180 | 0 | 0 |
| **5** | 1 | 0 | 1 | 0 | 169 | 0 | 0 |
| **6** | 1 | 1 | 1 | 1 | 203 | 1 | 1 |
| **7** | 0.5 | 13 | 1 | 11 | 127 | 11 | 11 |
| **8** | 1 | 11 | 1 | 11 | 549 | 11 | 11 |
| **9** | 0.95 | 3 | 1 | 3 | 138 | 3 | 3 |
| **10** | 1 | 1 | 1 | 2 | 173 | 1 | 1 |
| **11** | 1 | 1 | 1 | 1 | 65 | 1 | 1 |
| **12** | 1 | 0 | 1 | 0 | 141 | 0 | 0 |
| **13** | 1 | 1 | 1 | 1 | 211 | 1 | 1 |
| **14** | 1 | 3 | 1 | 3 | 373 | 3 | 3 |
| **15** | 1 | 0 | 1 | 0 | 91 | 0 | 0 |
| **16** | 0.95 | 25 | 1 | 25 | 283 | 24 | 24 |
| **17** | 1 | 4 | 1 | 4 | 347 | 4 | 4 |
| **18** | 1 | 7 | 1 | 6 | 253 | 7 | 6 |
| **19** | 1 | 1 | 1 | 1 | 89 | 1 | 1 |
| **20** | 0.95 | 5 | 1 | 4 | 261 | 4 | 4 |
| **21** | 1 | 0 | 1 | 0 | 104 | 1 | 0 |
| **22** | 1 | 1 | 1 | 1 | 274 | 1 | 1 |
| **23** | 0.95 | 0 | 1 | 0 | 88 | 0 | 0 |
| **24** | 0.75 | 1 | 1 | 1 | 122 | 1 | 1 |
| **25** | 1 | 0 | 1 | 0 | 463 | 0 | 0 |
| **26** | 0.1 | 0 | 1 | 0 | 278 | 0 | 0 |
| **27** | 1 | 2 | 1 | 3 | 76 | 2 | 2 |
| **28** | 1 | 1 | 0.95 | 1 | 123 | 1 | 1 |
| **29** | 0.35 | 3 | 1 | 1 | 498 | 2 | 1 |
| **30** | 1 | 33 | 1 | 33 | 201 | 33 | 33 |
| **31** | 0.8 | 66 | 1 | 36 | 685 | 33 | 33 |
| **32** | 1 | 21 | 1 | 21 | 404 | 20 | 20 |
| **33** | 1 | 12 | 1 | 10 | 101 | 10 | 10 |
| **34** | 1 | 17 | 1 | 16 | 169 | 15 | 15 |
| **35** | 1 | 9 | 1 | 11 | 103 | 9 | 9 |
| **36** | 1 | 7 | 1 | 6 | 209 | 6 | 6 |
| **37** | 1 | 3 | 1 | 3 | 593 | 3 | 3 |
| **38** | 1 | 3 | 1 | 0 | 70 | 0 | 0 |
| **39** | 1 | 1 | 1 | 1 | 427 | 1 | 1 |
| **40** | 1 | 5 | 1 | 4 | 299 | 4 | 4 |
| **41** | 0.95 | 27 | 1 | 27 | 594 | 27 | 27 |
| **42** | 1 | 5 | 1 | 8 | 68 | 5 | 5 |
| **43** | 0.9 | 8 | 1 | 6 | 132 | 6 | 6 |
| **44** | 0.9 | 45 | 1 | 17 | 154 | 16 | 16 |
| **45** | 1 | 0 | 1 | 0 | 103 | 0 | 0 |
| **46** | 0.7 | 6 | 1 | 4 | 301 | 3 | 3 |
| **47** | 1 | 3 | 1 | 3 | 855 | 3 | 3 |
| **48** | 1 | 4 | 1 | 4 | 71 | 4 | 4 |
| **49** | 1 | 26 | 0.7 | 25 | 236 | 24 | 24 |
| **50** | 0.35 | 38 | 0.8 | 36 | 157 | 35 | 35 |
| **51** | 0.45 | 46 | 1 | 45 | 146 | 45 | 45 |
| **52** | 0.75 | 63 | 1 | 46 | 233 | 46 | 46 |
| **Average** | 91% | 10.8 | 99% | 9.0 | 240.1 | 8.8 | 8.7 |
| **Runtime** | 15 sec | | 10 sec | | ca. 30 sec | | up to hours |

Table VI: Full genetic algorithm, tabu search and integer programming results for all data sets.

The graphs in Figure VIII to Figure XIII show detailed results for all 52 data sets and the main development stages of the genetic algorithm. The bars above the y-axis represent solution cost, with the black bars showing the number of optimal solutions, and the total bar height showing the number of solutions within three units of the optimal value. The value of three was chosen as it corresponds to the penalty cost of violating the least important level of requests in the original formulation. Thus, solutions this close to the optimum would certainly be acceptable to the hospital. The bars below the axis represent the number of times out of 20 that the run terminated without finding a single feasible solution. Thus, the less the area of the bars below the axis and the more above, the better the performance of an algorithm.

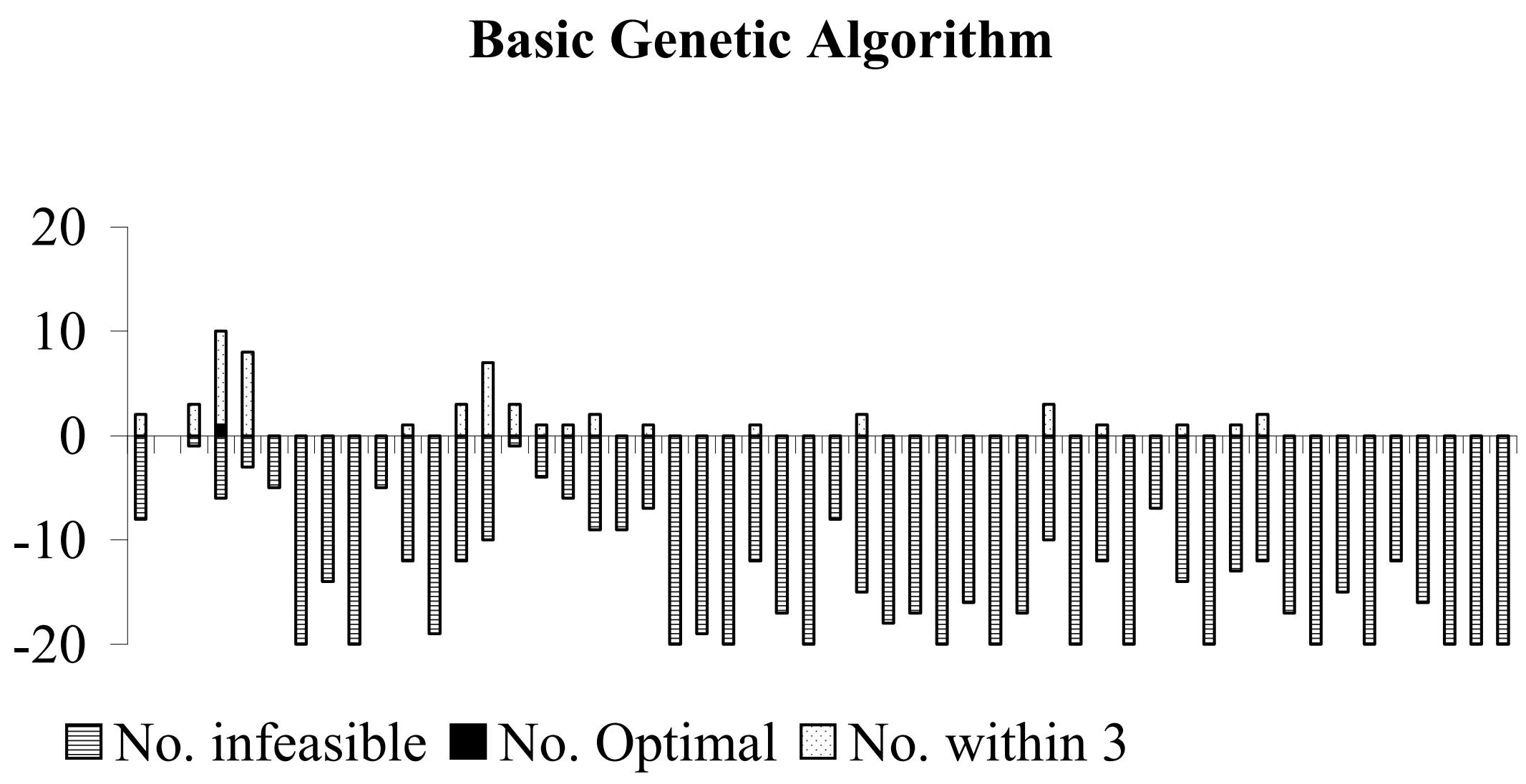

Figure VIII: Detailed results for basic genetic algorithm.

When comparing the results, one can make some interesting observations. Firstly, the impact of the co-operative co-evolutionary approach was immense, improving the results significantly from before. Furthermore, there clearly are some data sets that are more difficult to solve than others, for instance the last three data sets. This can be due to two reasons. Either those data sets are genuinely more difficult, for example due to tighter constraints etc. or the choice of genetic parameters, in particular the penalty weight, was not very suitable. To investigate this further is one area of future research as described in section 8.2. Finally, the graphs show once more the superiority of the indirect approaches, especially once the self-adjusting weights are introduced.

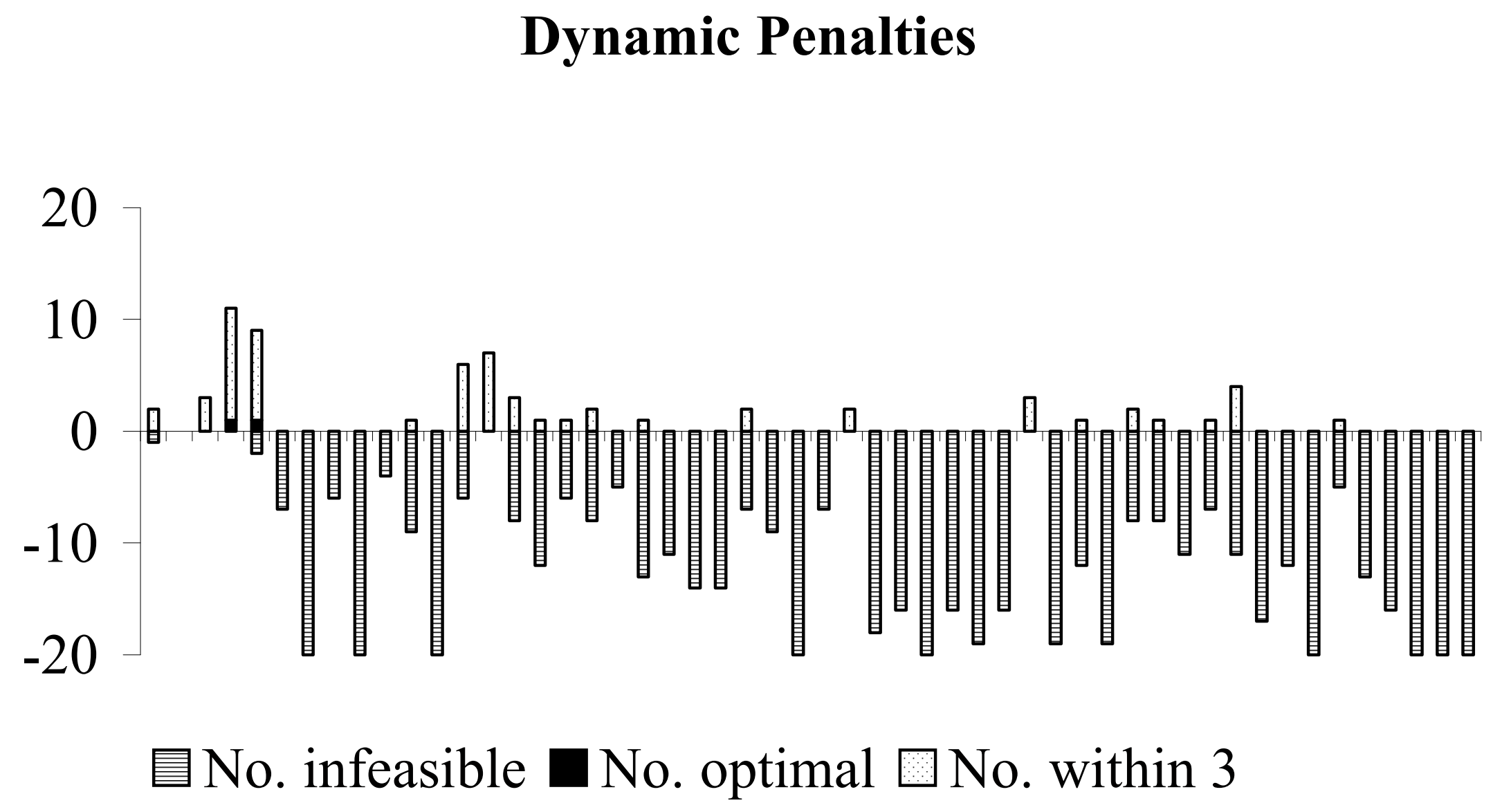

Figure IX: Detailed results for a genetic algorithm with dynamic weights and optimised parameters.

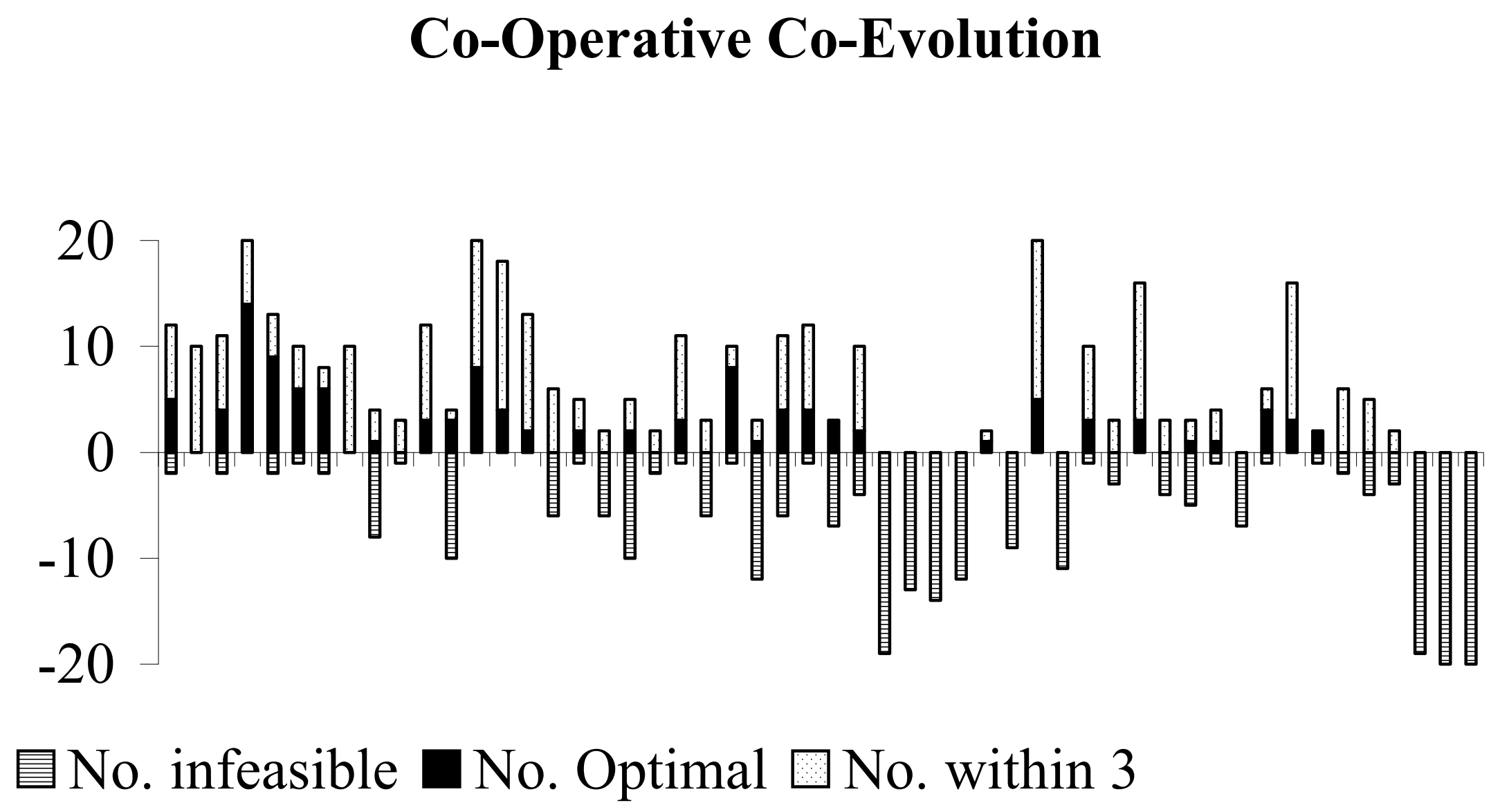

Figure X: Detailed results for a co-operative co-evolutionary approach.

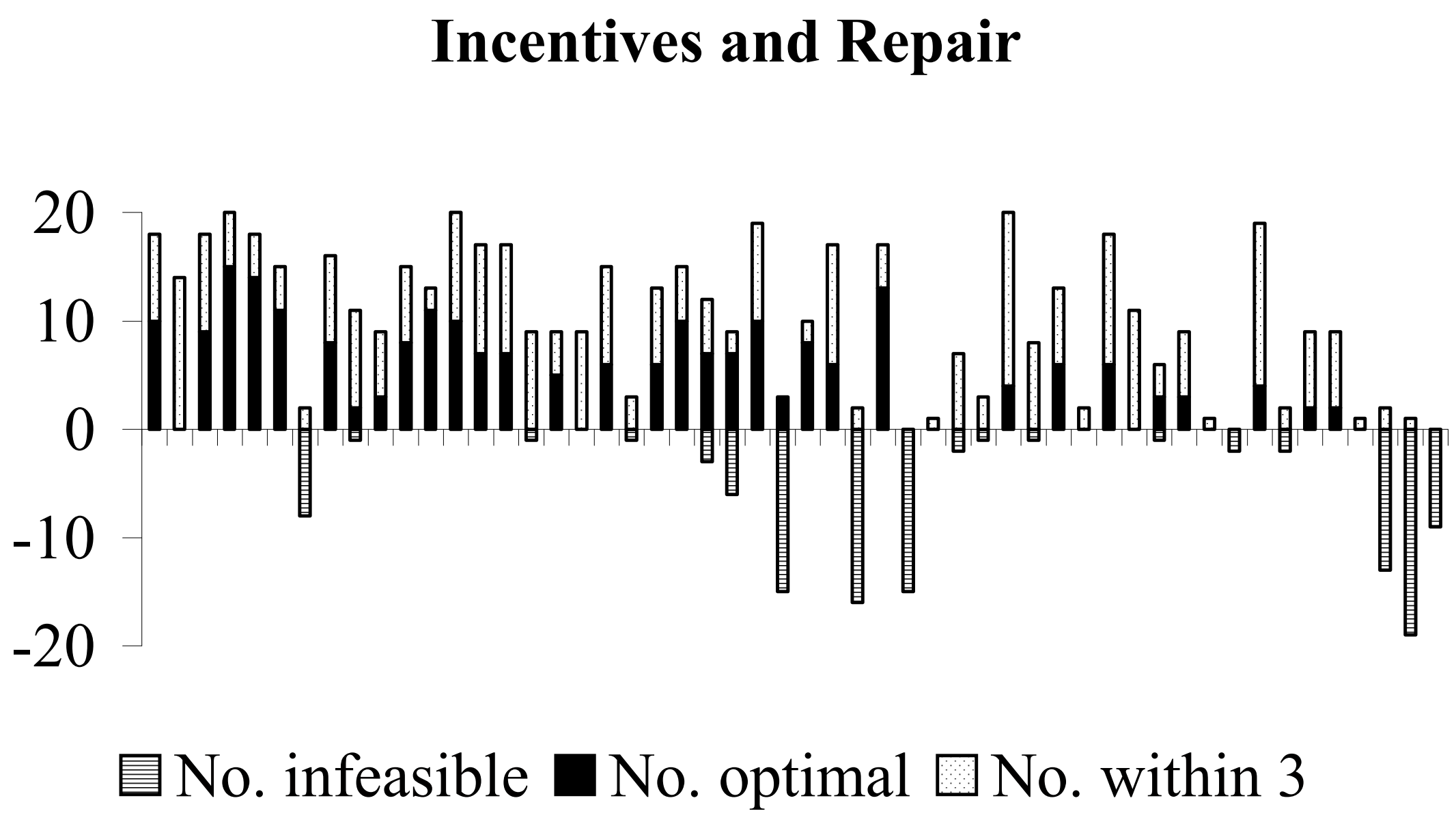

Figure XI: Detailed results for a co-operative co-evolutionary approach with repair and incentives.

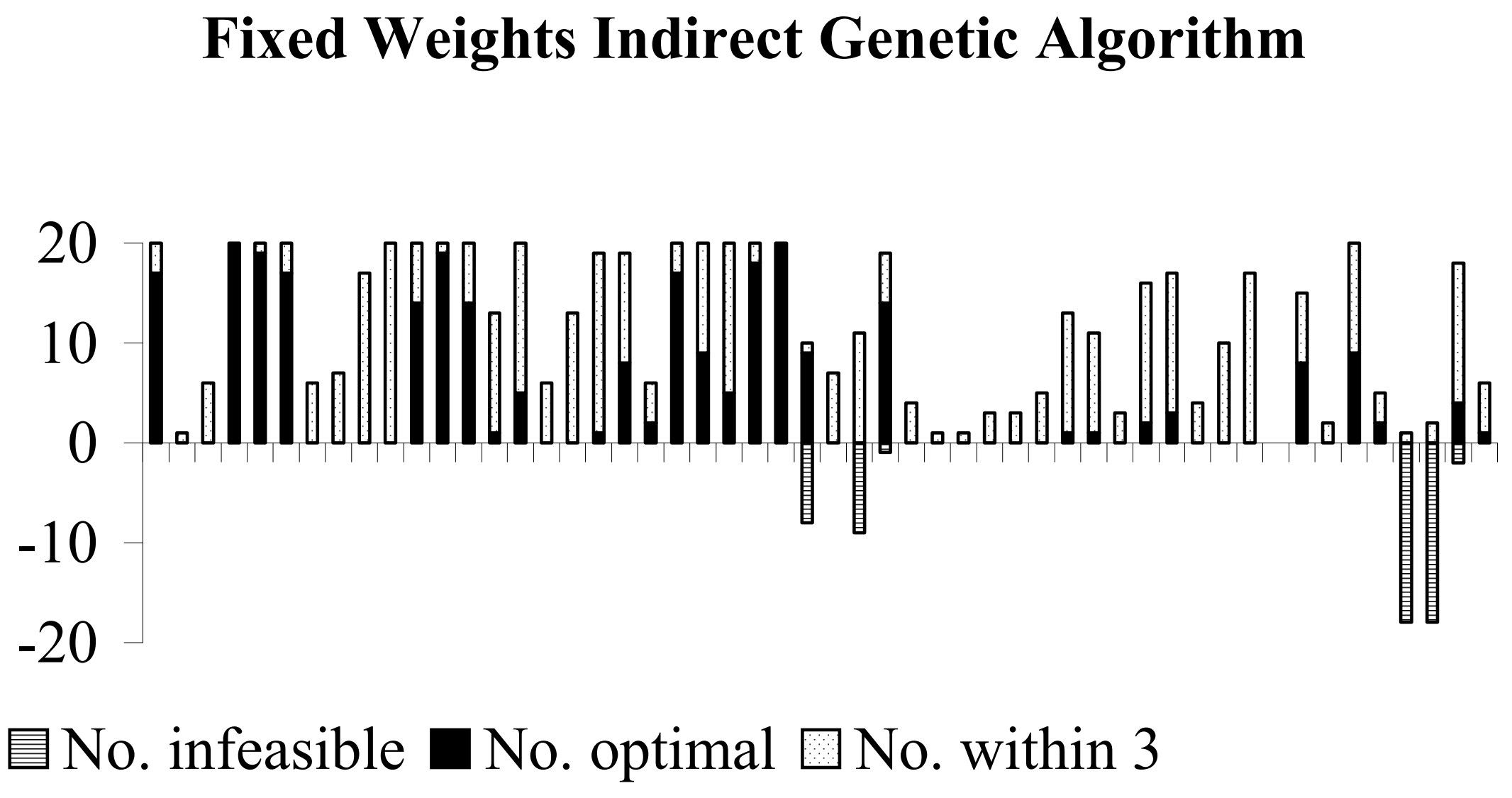

Figure XII: Detailed results for an indirect genetic algorithm with fixed decoder weights.

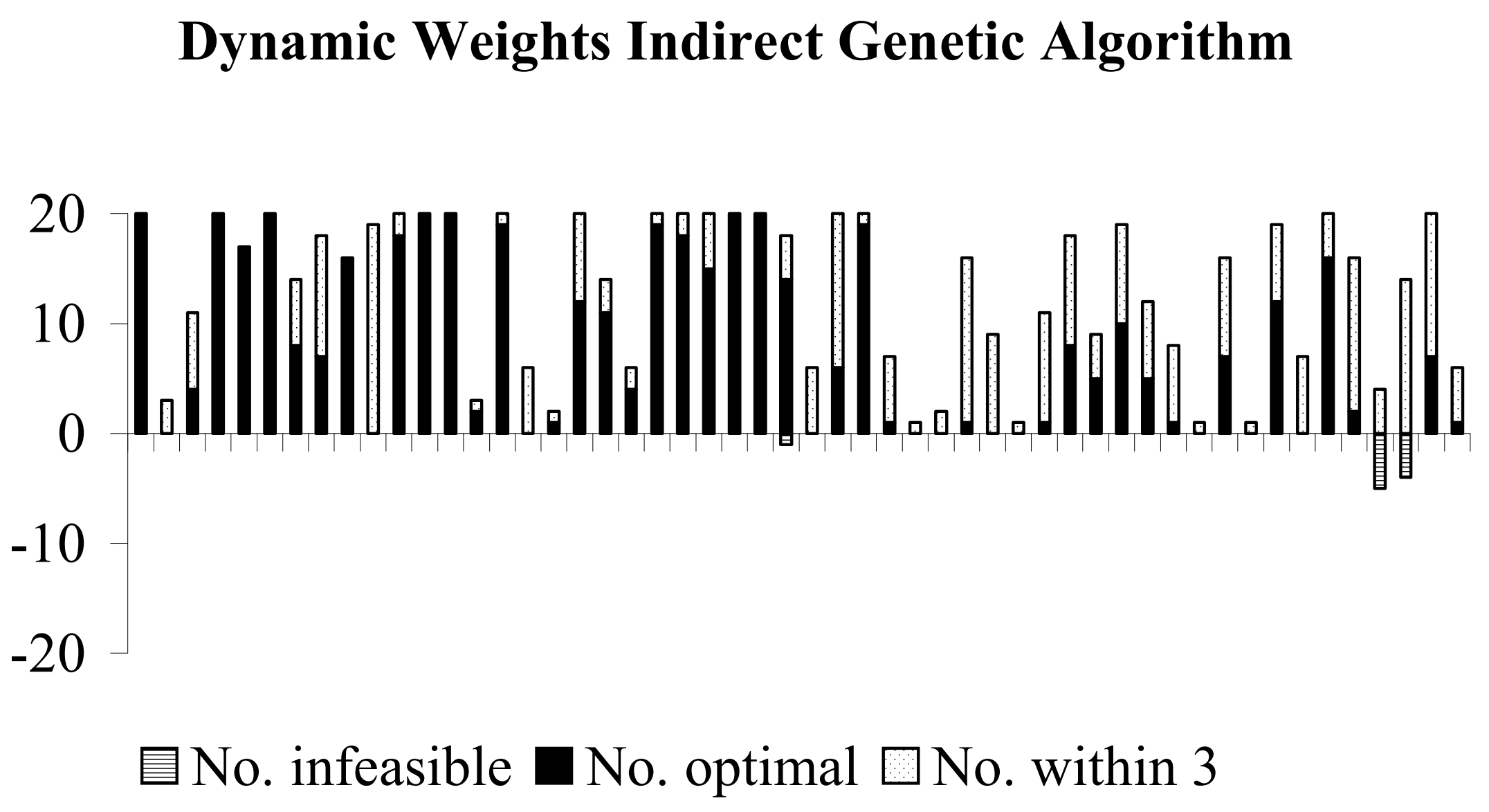

Figure XIII: Detailed results for an indirect genetic algorithm with dynamic crossover rates and decoder weights.

## D.3 Additional Results for Chapters 4 - 6

Figure XIV shows the results for using a quadratic penalty function and different penalty weights as described in section 4.3.3. The quadratic penalty is calculated by squaring the constraint violations. As mentioned in section 4.3.3, in the parameter range tested the results are significantly worse than for the linear penalty function. Thus, quadratic penalties were not used in this research.

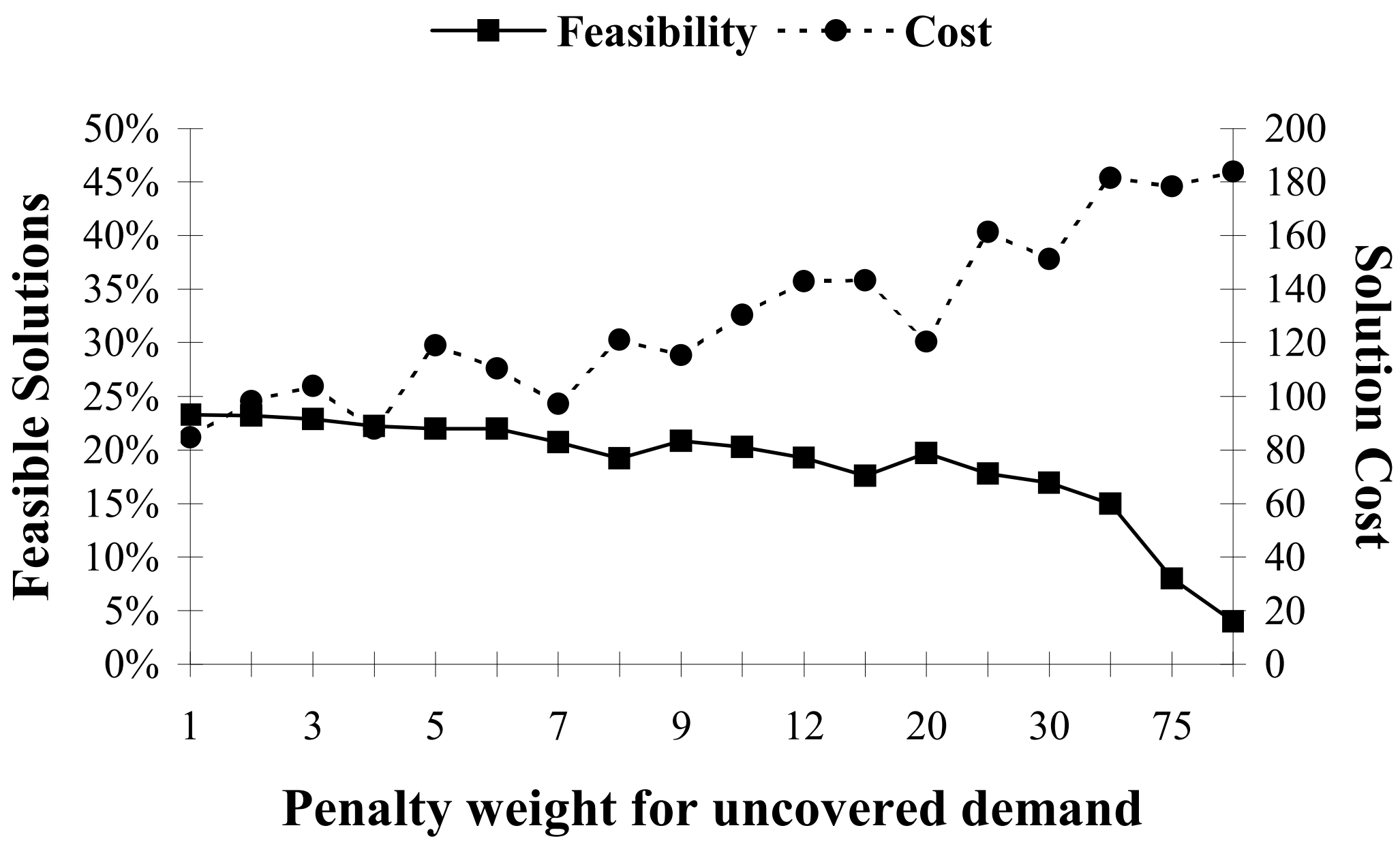

Figure XIV: Quadratic penalty weights.

Figure XV and Figure XVI show additional results for the migration operator. The graph in Figure XV shows the effect of varying the frequency of migrating the best five individuals of each sub-population into another random sub-population. Exchanging these individuals every five generations gives the best results. However, migrating random individuals performs better as pictured in Figure XVI. With random migration the best results are found for a 5% migration probability per individual and generation. As explained in section 5.2.5, the better performance of the random migration operator is attributed to the specialisation of the sub-populations.

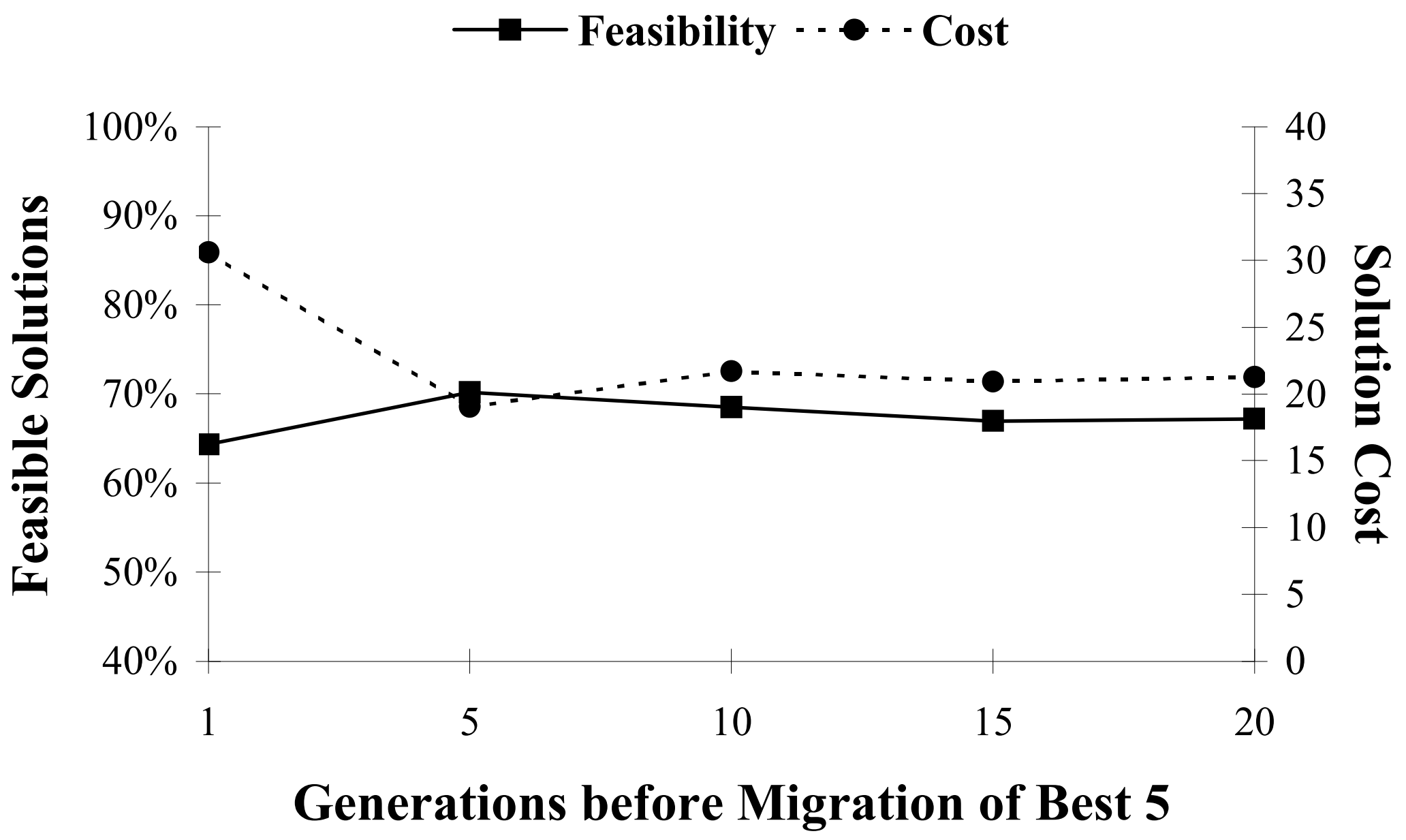

Figure XV: Migration of five best individuals of each sub-population.

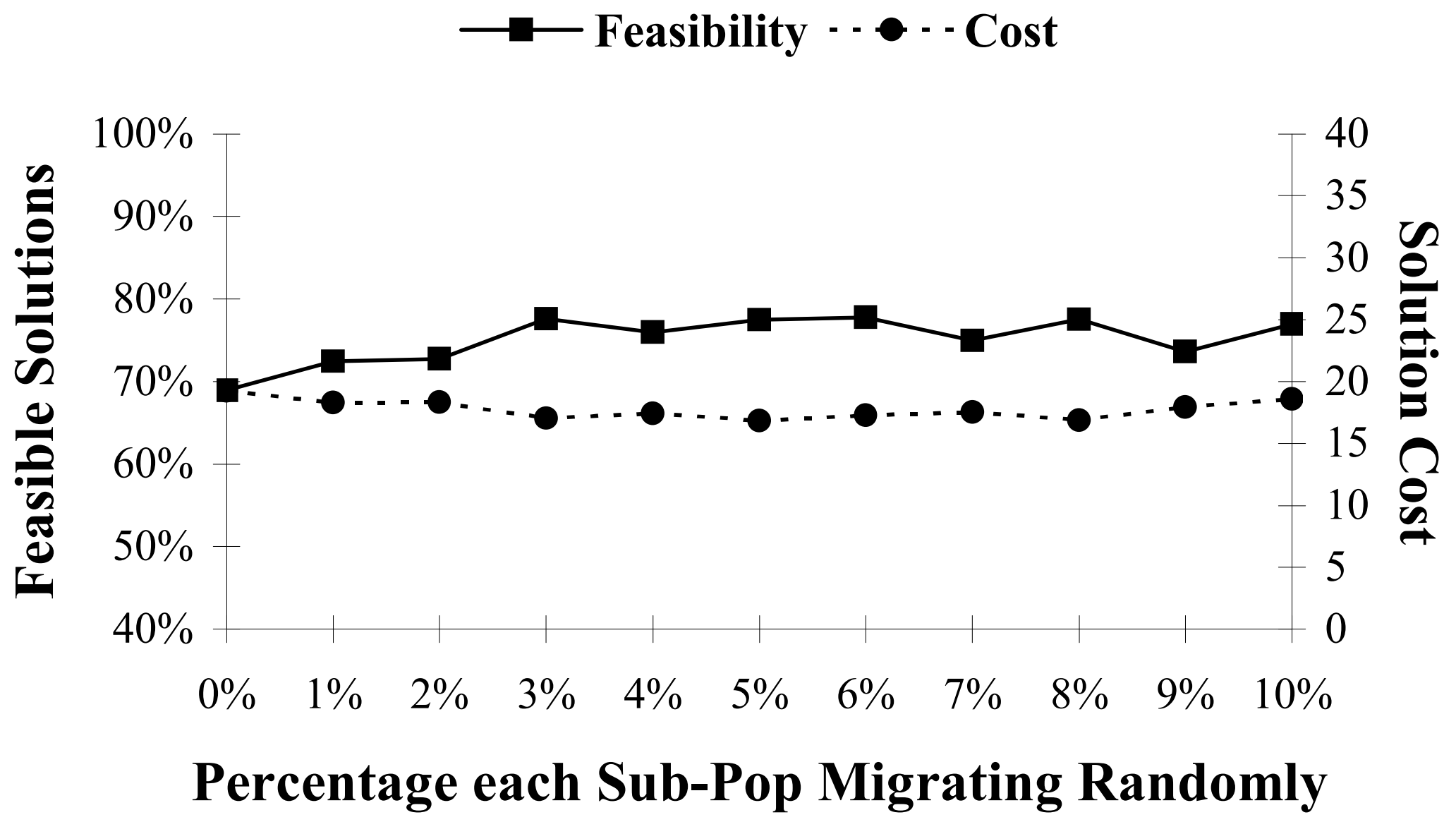

Figure XVI: Random migration between sub-populations.

# Appendix E  Mall Problem Data

This appendix details a typical data file from data set five, which is characterised by being tight on minimum, ideal and maximum number for each shop type and loose on the number of small, medium and large shops allowed. The basic size and area/location layout of the mall is defined in Table VII and Table VIII. Table IX indicates which shop type is a member of which shop group, for instance men's clothes are part of the clothing group.

| | | |
|---|---|---|
| Number of Locations; Number of Areas | 100 | 5 |
| Number of Shop Types; Number of Shop Groups | 20 | 5 |

Table VII: Problem size.

| Area | 1 | 2 | 3 | 4 | 5 |
|---|---|---|---|---|---|
| Bounds of Area | 1-18 | 19-29 | 30-57 | 58-86 | 87-100 |

Table VIII: Location distribution.

| (1 = Member; 0 = Else) | Shop Group | | | | |
|---|---|---|---|---|---|
| **Shop Type** | **1** | **2** | **3** | **4** | **5** |
| **1** | 1 | 0 | 0 | 0 | 0 |
| **2** | 0 | 0 | 0 | 0 | 1 |
| **3** | 0 | 1 | 0 | 0 | 0 |
| **4** | 1 | 0 | 0 | 0 | 1 |
| **5** | 0 | 0 | 0 | 0 | 1 |
| **6** | 1 | 0 | 0 | 1 | 0 |
| **7** | 0 | 0 | 0 | 0 | 1 |
| **8** | 0 | 1 | 0 | 0 | 1 |
| **9** | 1 | 0 | 0 | 0 | 0 |
| **10** | 1 | 0 | 0 | 0 | 0 |
| **11** | 0 | 1 | 0 | 0 | 0 |
| **12** | 0 | 1 | 0 | 1 | 0 |
| **13** | 0 | 1 | 0 | 0 | 0 |
| **14** | 0 | 0 | 0 | 1 | 0 |
| **15** | 0 | 0 | 1 | 0 | 0 |
| **16** | 0 | 1 | 0 | 0 | 0 |
| **17** | 0 | 1 | 1 | 0 | 0 |
| **18** | 0 | 0 | 1 | 0 | 0 |
| **19** | 0 | 0 | 0 | 0 | 0 |
| **20** | 0 | 0 | 1 | 0 | 0 |

Table IX: Group membership of shop types.

Table X shows the maximum number of small, medium and large shops allowed in the mall. As the example is taken from data set five, no limits are set. The efficiency factor depending on the number of shops of the same type in one area is given in Table XI. The attractiveness rating for each area is displayed in Table XII. Table XIII shows the maximum number of shops of one type allowed in the whole mall. As this data file is an example for being tight on the maximum number of shops allowed, the sum of the minima is between 95 and 98.

| Shop Size | Small | Medium | Large |
|---|---|---|---|
| Maximum Shop Number per Size | **100** | **100** | **100** |

Table X: Limits on the number of shops of one size.

| Shop Count | 0 | 1 | 2 | 3 | 4 | 5 | 6 | 7 | 8 | 9 | 10 |
|---|---|---|---|---|---|---|---|---|---|---|---|
| Efficiency Factor | **0** | **10** | **11.5** | **13** | **12.2** | **12.4** | **13** | **12.5** | **12.6** | **13** | **12.7** |

Table XI: Efficiency factor versus shop count.

| Area | 1 | 2 | 3 | 4 | 5 |
|---|---|---|---|---|---|
| Attractiveness Factor | **12** | **16** | **11** | **12** | **13** |

Table XII: Attractiveness of areas.

| Number per Shop Type in Mall | | | |
|---|---|---|---|
| Shop Type | Min | Ideal | Max |
| 1 | **6** | **6** | **7** |
| 2 | **5** | **6** | **6** |
| 3 | **4** | **5** | **8** |
| 4 | **6** | **7** | **8** |
| 5 | **5** | **9** | **9** |
| 6 | **5** | **5** | **6** |
| 7 | **4** | **4** | **7** |
| 8 | **3** | **9** | **9** |
| 9 | **6** | **6** | **6** |
| 10 | **4** | **9** | **9** |
| 11 | **6** | **6** | **9** |
| 12 | **5** | **8** | **9** |
| 13 | **4** | **7** | **9** |
| 14 | **3** | **4** | **6** |
| 15 | **5** | **7** | **7** |
| 16 | **6** | **6** | **9** |
| 17 | **5** | **9** | **9** |
| 18 | **4** | **5** | **8** |
| 19 | **6** | **7** | **8** |
| 20 | **5** | **6** | **6** |

Table XIII: Limits on the number of shops of one type.

The fixed rent of a shop unit, defined by its type and the area it is located in, is shown in Table XIV. Finally, the bonus for a complete group of shops is given in Table XV.

| | Fixed Rent per Area | | | | |
|---|---|---|---|---|---|
| Shop Type | 1 | 2 | 3 | 4 | 5 |
| 1 | **1286** | **1670** | **1761** | **2116** | **2131** |
| 2 | **1501** | **1742** | **1921** | **2222** | **2987** |
| 3 | **1616** | **1838** | **2039** | **2371** | **2612** |
| 4 | **1858** | **1437** | **2229** | **1833** | **2353** |
| 5 | **1375** | **1527** | **1726** | **2405** | **2200** |
| 6 | **1497** | **1650** | **2554** | **2113** | **2375** |
| 7 | **1758** | **2159** | **1653** | **2303** | **2687** |
| 8 | **1794** | **2339** | **1938** | **2790** | **2327** |
| 9 | **1803** | **1564** | **1891** | **2642** | **2032** |
| 10 | **1758** | **1796** | **2138** | **2589** | **2903** |
| 11 | **2055** | **1453** | **1901** | **2100** | **2689** |
| 12 | **2071** | **1999** | **2265** | **1892** | **2743** |
| 13 | **1292** | **1476** | **1923** | **1888** | **2260** |
| 14 | **1844** | **1676** | **2391** | **2156** | **2478** |
| 15 | **1671** | **1487** | **2085** | **2323** | **2439** |
| 16 | **1635** | **2075** | **1926** | **2548** | **2561** |
| 17 | **1982** | **1428** | **1738** | **1898** | **2357** |
| 18 | **1325** | **1914** | **2078** | **2627** | **2793** |
| 19 | **1979** | **1418** | **2190** | **2636** | **2728** |
| 20 | **1560** | **2302** | **2135** | **2310** | **2686** |

Table XIV: Fixed shop type and area rent.

| Group Status | Two Complete | One Complete | None Complete |
|---|---|---|---|
| Group Factor | **14.4** | **12** | **10** |

Table XV: Group bonus factors.

# Appendix F Additional Mall Problem Results

In the figures below the results of one hundred randomly created solutions to a file of the fourth data set are displayed. The graphs show the percentage of shops in small, medium and large sizes as well as the percentage of shops at an ideal count level and in groups. The line displayed is a linear approximation of the trend present. It gives us an idea of the actual workings of the mall objective function and this knowledge was used when setting the decoder weights intuitively.

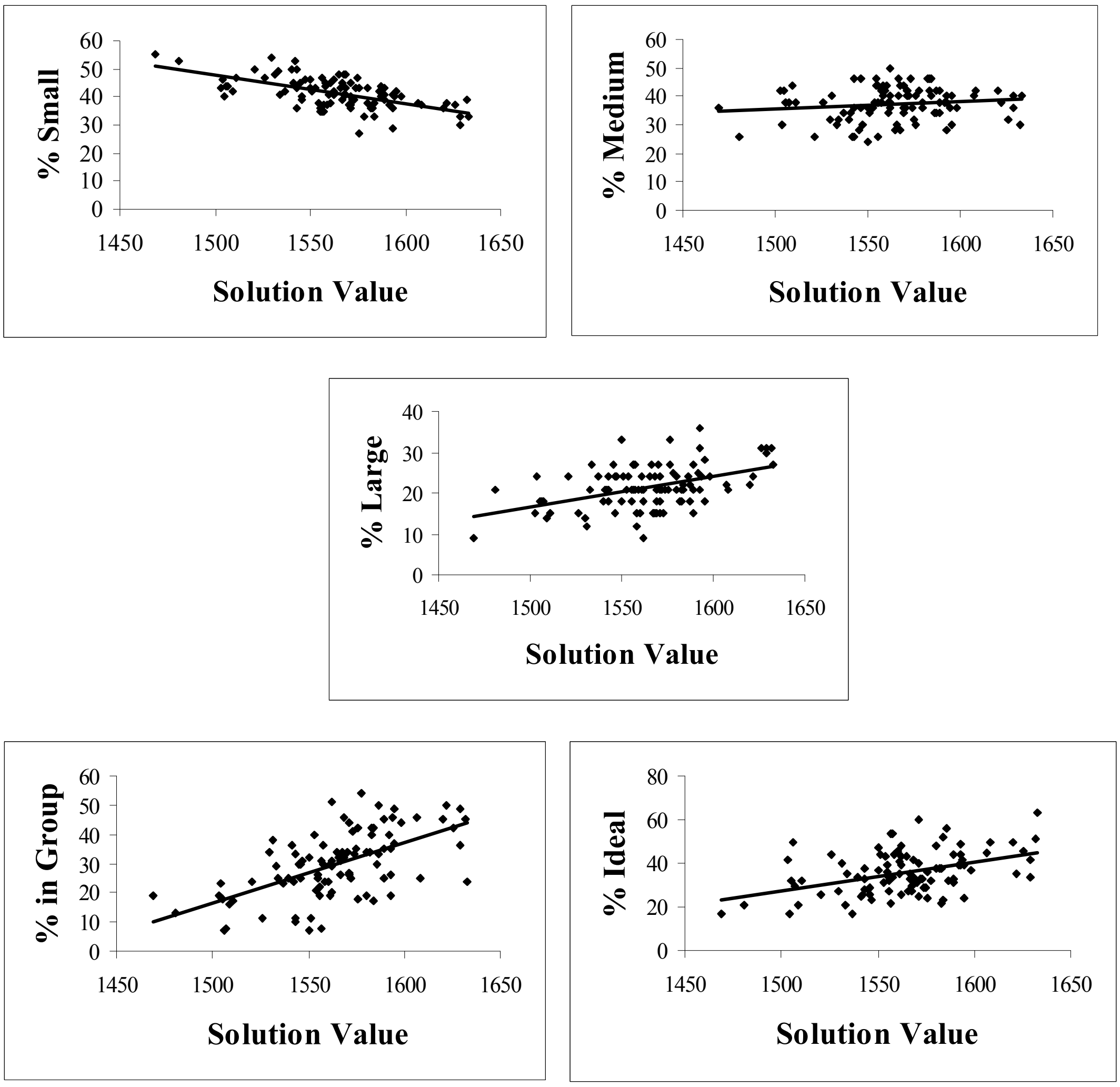

Figure XVII: Relationship between objective function value and shop sizes, shops in groups and shops with an ideal shop count.

Table XVI shows detailed results for various approaches to the Mall Problem grouped by data sets. The labels are 'Direct' for the direct genetic algorithm, 'Sub-Pop' for the simple co-evolution scheme, 'Repair' for the co-evolution with mating and repair, 'I Medium' for the indirect genetic algorithm with medium weights, 'I Auto' for the indirect approach with self setting weights and 'I Auto +' for the indirect algorithm that set its own weights and crossover rates.

In addition to the usual results of feasibility and rent, solution times and some characteristics of the solutions found are given. These characteristics are the percentage of shops in groups, the percentage at an ideal shop count level and the ratio of large, medium and small shops. These percentages clearly show the tightness of the various constraints in the data sets. For instance, files 51-60 have tight limits on the number of small, medium and large shops.

Another interesting observation can be made from the results displayed in Table XVI. If the data files do not have tight constraints, then there are clearly two ways to achieve good results: Either a high percentage of shops in groups or a high percentage at their ideal level. The direct genetic algorithm schemes take the second route. This can easily be explained. Without any further problem-specific operators, the direct approach with its encoding can maintain a full group only with difficulty, as removing one member already destroys the group. This happens frequently with most crossover operations, as different strings will work towards different groups. Thus, the children have fewer complete groups than their parents. On the other hand, maintaining the ideal shop count is easier, as all individuals work towards the same goals.

For the indirect genetic algorithm, the situation is different. Each solution is built from scratch, allocating those shops that give the biggest score. Obviously, it is not possible to have all shops at their ideal level and all groups complete. This is because there is only a limited number of locations in the mall, which are not enough too have all shop types at their ideal level. Thus, the more shop types are at their ideal level the less shops there will be of the remaining types. This makes it harder for groups to be complete as one missing member destroys the bonus. Hence in the light of this, the indirect algorithm decoder completes groups rather than maintain ideal levels, which gives better results.

| Files | Algorithm | Time [s] | % Feasible | Rent | % Group | % Ideal | % Large | % Medium | % Small |
|---|---|---|---|---|---|---|---|---|---|
| 21-30 | Direct | 43.1 | 100 | 1826.8 | 15.4 | 78.1 | 68.7 | 11.3 | 19.9 |
| | Sub-Pop | 45.3 | 100 | 1764.9 | 11.0 | 78.8 | 52.8 | 18.5 | 28.9 |
| | Repair | 59.0 | 100 | 1852.8 | 12.3 | 80.0 | 72.7 | 10.1 | 17.0 |
| | I Medium | 50.3 | 100 | 1903.0 | 73.2 | 62.3 | 61.7 | 18.1 | 20.3 |
| | I Auto | 59.1 | 100 | 1959.1 | 89.7 | 70.9 | 56.2 | 22.6 | 21.1 |
| | I Auto + | 41.7 | 100 | 1965.9 | 89.6 | 72.5 | 56.9 | 24.6 | 18.5 |
| 31-40 | Direct | 17.1 | 100 | 1973.1 | 76.1 | 82.3 | 59.0 | 15.3 | 25.9 |
| | Sub-Pop | 15.8 | 100 | 1916.3 | 70.9 | 77.8 | 50.5 | 18.7 | 30.8 |
| | Repair | 24.0 | 100 | 2012.7 | 72.5 | 73.1 | 58.7 | 19.3 | 21.9 |
| | I Medium | 25.1 | 100 | 1960.8 | 80.7 | 46.4 | 72.4 | 14.2 | 13.2 |
| | I Auto | 24.3 | 100 | 2016.7 | 89.9 | 42.3 | 65.4 | 15.3 | 19.4 |
| | I Auto + | 20.6 | 100 | 2021.1 | 90.9 | 49.4 | 65.8 | 15.6 | 18.7 |
| 41-50 | Direct | 11.3 | 100 | 1757.3 | 61.5 | 45.9 | 35.6 | 26.6 | 37.8 |
| | Sub-Pop | 12.1 | 85 | 1750.7 | 57.3 | 46.3 | 31.9 | 30.4 | 37.8 |
| | Repair | 20.6 | 100 | 1930.6 | 74.3 | 46.5 | 61.8 | 18.9 | 19.4 |
| | I Medium | 19.2 | 100 | 1911.7 | 77.1 | 44.5 | 69.1 | 16.6 | 14.1 |
| | I Auto | 22.7 | 100 | 1932.7 | 83.6 | 45.4 | 65.2 | 16.1 | 18.9 |
| | I Auto + | 18.5 | 100 | 1939.0 | 86.0 | 46.1 | 66.6 | 15.9 | 17.5 |
| 51-60 | Direct | 14.7 | 100 | 1878.9 | 32.9 | 78.2 | 62.1 | 32.9 | 5.0 |
| | Sub-Pop | 13.7 | 100 | 1855.9 | 31.7 | 65.4 | 61.7 | 33.1 | 5.2 |
| | Repair | 27.4 | 100 | 1901.8 | 42.0 | 77.0 | 62.9 | 32.0 | 5.1 |
| | I Medium | 22.1 | 100 | 1916.0 | 69.8 | 44.7 | 63.0 | 31.8 | 5.2 |
| | I Auto | 19.3 | 100 | 1951.8 | 85.7 | 48.9 | 63.1 | 31.3 | 5.6 |
| | I Auto + | 17.6 | 100 | 1958.5 | 85.7 | 51.4 | 63.0 | 32.0 | 5.0 |
| 61-70 | Direct | 13.5 | 70 | 1796.2 | 28.4 | 52.4 | 62.5 | 31.5 | 6.0 |
| | Sub-Pop | 15.3 | 12 | 1099.4 | 30.0 | 52.5 | 62.3 | 31.7 | 6.0 |
| | Repair | 15.8 | 31 | 1786.2 | 32.9 | 50.2 | 63.0 | 31.2 | 5.9 |
| | I Medium | 14.2 | 94 | 1836.6 | 44.3 | 45.2 | 65.9 | 28.3 | 5.8 |
| | I Auto | 15.5 | 100 | 1887.2 | 68.7 | 45.8 | 65.4 | 28.9 | 6.0 |
| | I Auto + | 15.7 | 100 | 1896.9 | 70.5 | 46.2 | 65.8 | 28.3 | 5.9 |
| 21-70 | Direct | 19.9 | 94 | 1846.5 | 42.9 | 67.4 | 57.6 | 23.5 | 18.9 |
| | Sub-Pop | 20.4 | 79 | 1677.4 | 40.2 | 64.2 | 51.8 | 26.5 | 21.7 |
| | Repair | 29.3 | 86 | 1896.8 | 46.8 | 65.4 | 63.8 | 22.3 | 13.9 |
| | I Medium | 26.2 | 99 | 1905.6 | 69.0 | 48.6 | 66.4 | 21.8 | 11.7 |
| | I Auto | 28.2 | 100 | 1949.5 | 83.5 | 50.7 | 63.1 | 22.8 | 14.2 |
| | I Auto + | 22.8 | 100 | 1956.3 | 84.5 | 53.1 | 63.6 | 23.3 | 13.1 |

Table XVI: Detailed Mall Problem results.